\documentclass[10pt,twocolumn,letterpaper]{article}

\usepackage{cvpr}
\definecolor{cvprblue}{rgb}{0.21,0.49,0.74}
\usepackage[pagebackref,breaklinks,colorlinks,allcolors=cvprblue]{hyperref}

\usepackage[table]{xcolor}
\usepackage[margin=1in]{geometry}
\usepackage{booktabs,multirow,array}
\usepackage{array}
\usepackage{makecell}
\usepackage{tikz}
\usepackage{graphicx}
\usepackage{arydshln} 
\usepackage{colortbl}
\usepackage{amsthm}
\usepackage{amsmath, amssymb}
\usepackage{adjustbox}
\usepackage{changepage}
\usepackage{tabularx}
\usepackage{array} 
\usepackage{lipsum}
\usepackage{paralist}
\usepackage{makecell}
\usepackage{pifont}
\usepackage{enumitem}
\usepackage[most]{tcolorbox}
\usepackage{arydshln}
\usepackage{tocloft}
\usepackage{etoc}
\tcbuselibrary{listings,breakable}
\usepackage{rotating}
\usepackage{pifont}

\newcommand{\languagelogo}{\raisebox{-0pt}{\includegraphics[width=0.9em]{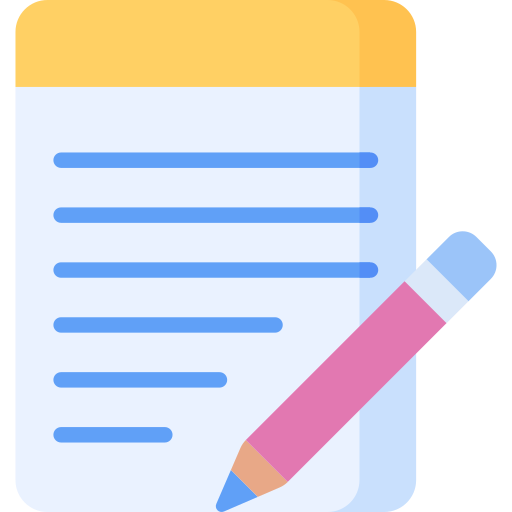}}\xspace\xspace}
\newcommand{\audiologo}{\raisebox{-0pt}{\includegraphics[width=0.9em]{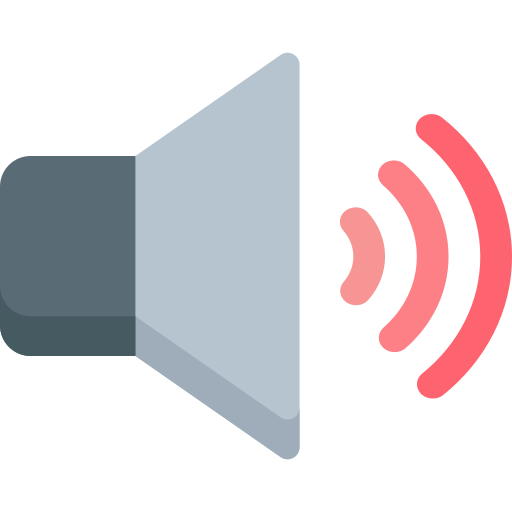}}\xspace\xspace}
\newcommand{\imagelogo}{\raisebox{-1pt}{\includegraphics[width=0.9em]{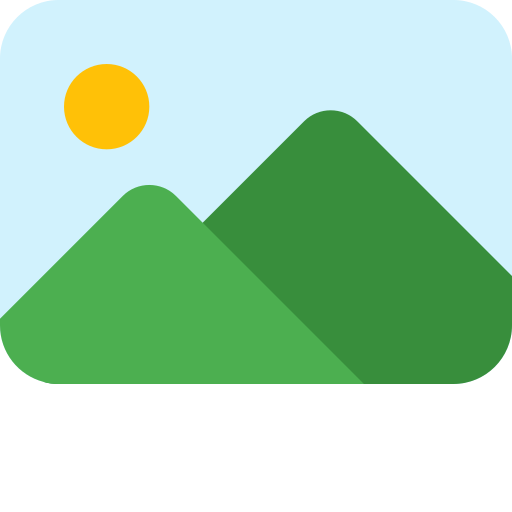}}\xspace\xspace}
\newcommand{\videologo}{\raisebox{-0.2pt}{\includegraphics[width=0.9em]{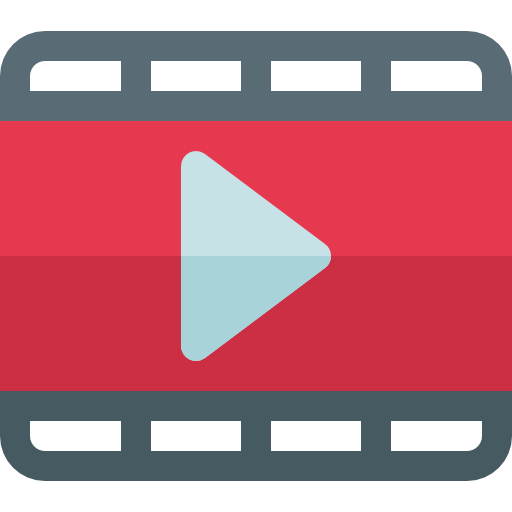}}\xspace\xspace}
\newcommand{\documentlogo}{\raisebox{-0pt}{\includegraphics[width=0.9em]{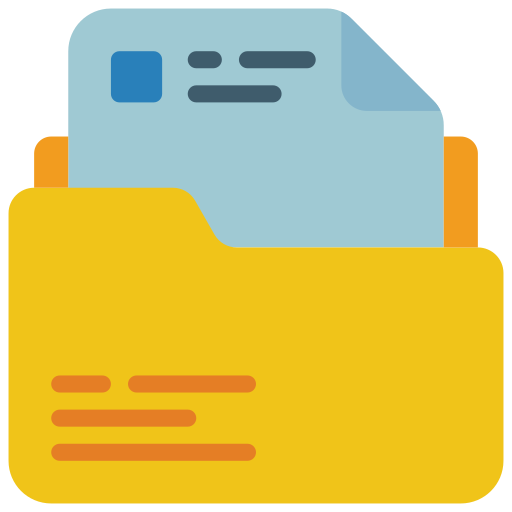}}\xspace\xspace}
\newcommand{\threedlogo}{\raisebox{-0pt}{\includegraphics[width=0.9em]{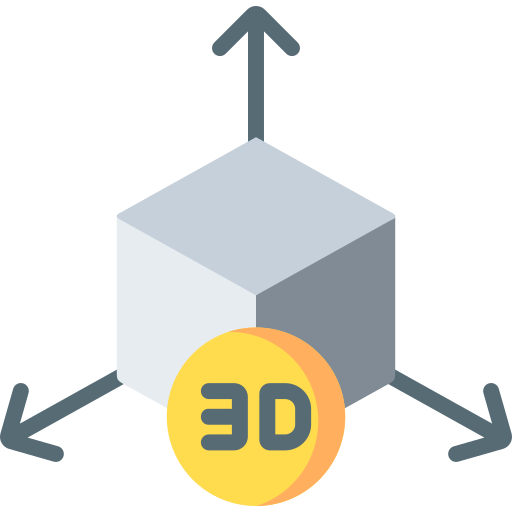}}\xspace\xspace}
\newcommand{\codelogo}{\raisebox{-0.3pt}{\includegraphics[width=0.9em]{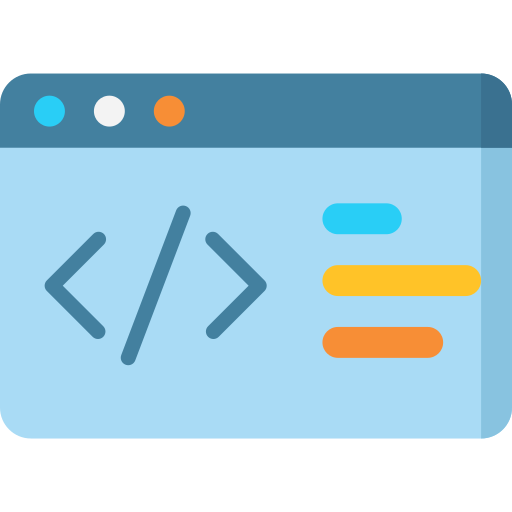}}\xspace\xspace}

\definecolor{com_green}{HTML}{67C23A}
\definecolor{com_red}{HTML}{F56C6C}
\definecolor{SCGQ_head_fill}{HTML}{E1F6FE}
\definecolor{SCGQ_head_word}{HTML}{479ED3}
\definecolor{SCGQ_table_fill}{HTML}{F2FCFC}
\definecolor{head_fill}{HTML}{DFDFDF}
\definecolor{table_fill}{HTML}{F2F2F2}
\definecolor{IC_head_fill}{HTML}{FDF3E5}
\definecolor{IC_table_fill}{HTML}{FEFAF8}
\definecolor{RSI_head_fill}{HTML}{DEFEEA}
\definecolor{RSI_table_fill}{HTML}{F0FCF7}
\definecolor{diff_gray}{HTML}{404040}
\definecolor{abl_head_fill}{HTML}{EDDCFD}
\definecolor{standard_red}{HTML}{EA3323}
\definecolor{rsi_green}{HTML}{7EC99C}
\definecolor{ics_orange}{HTML}{F4B660}
\definecolor{sqcs_blue}{HTML}{8BCEF0}

\newcommand{\tinypie}[1]{
    \begin{tikzpicture}[baseline=-0.75ex, scale=0.3]
        \fill[SCGQ_head_word] (0,0) -- (0:0.42) arc (0:#1*3.6:0.42) -- cycle;
    \end{tikzpicture}
}

\newcommand{\tinybar}[2]{
    \begin{tikzpicture}[baseline=-0.75ex]
        \fill[SCGQ_head_word] (0,-0.1) rectangle ({0.4*#1/#2},0.1);
    \end{tikzpicture}
}

\title{\textsc{UniM}: A Unified Any-to-Any Interleaved Multimodal Benchmark} 

\author{
\textbf{Yanlin Li}$^{1}$,
\textbf{Minghui Guo}$^{1}$,
\textbf{Kaiwen Zhang}$^{1}$,
\textbf{Shize Zhang}$^{1}$,
\textbf{Yiran Zhao}$^{1}$,\\
\textbf{Haodong Li}$^{2}$,
\textbf{Congyue Zhou}$^{2}$,
\textbf{Weijie Zheng}$^{3}$, 
\textbf{Yushen Yan}$^{2}$,
\textbf{Shengqiong Wu}$^{1}$, \\
\textbf{Wei Ji}$^{4}$,
\textbf{Lei Cui}$^{5}$,
\textbf{Furu Wei}$^{5}$,
\textbf{Hao Fei}$^{1}$\thanks{Corresponding author: Hao Fei.} ,
\textbf{Mong-Li Lee}$^{1}$,
\textbf{Wynne Hsu}$^{1}$ \\
$^{1}$NUS \qquad
$^{2}$SCUT \qquad
$^{3}$NTU \qquad
$^{4}$NJU \qquad
$^{5}$Microsoft Research \\
\texttt{yanlin.li@u.nus.edu, haofei7419@gmail.com}
}

\begin{document}

\maketitle

\begin{abstract}
In real-world multimodal applications, systems usually need to comprehend arbitrarily combined and interleaved multimodal inputs from users, while also generating outputs in any interleaved multimedia form.
This capability defines the goal of \textit{any-to-any interleaved multimodal learning} under a unified paradigm of understanding and generation, posing new challenges and opportunities for advancing Multimodal Large Language Models (MLLMs).
To foster and benchmark this capability, this paper introduces the \textbf{\textsc{UniM}} benchmark, the first \textit{Unified Any-to-Any Interleaved Multimodal} dataset.
\textsc{UniM} contains 31K high-quality instances across 30 domains and 7 representative modalities: text, image, audio, video, document, code, and 3D, each requiring multiple intertwined reasoning and generation capabilities.
We further introduce the \textbf{\textsc{UniM Evaluation Suite}}, which assesses models along three dimensions: Semantic Correctness \& Generation Quality, Response Structure Integrity, and Interleaved Coherence.
In addition, we propose \textbf{\textsc{UniMA}}, an agentic baseline model equipped with traceable reasoning for structured interleaved generation.
Comprehensive experiments demonstrate the difficulty of \textsc{UniM} and highlight key challenges and directions for advancing unified any-to-any multimodal intelligence. 
The project page is \url{https://any2any-mllm.github.io/unim}.
\end{abstract}

\vspace{-4mm}
\section{Introduction}
\label{sec:intro}
\vspace{-1mm}

\begin{figure*}
    \centering
    \includegraphics[width=0.99\linewidth]{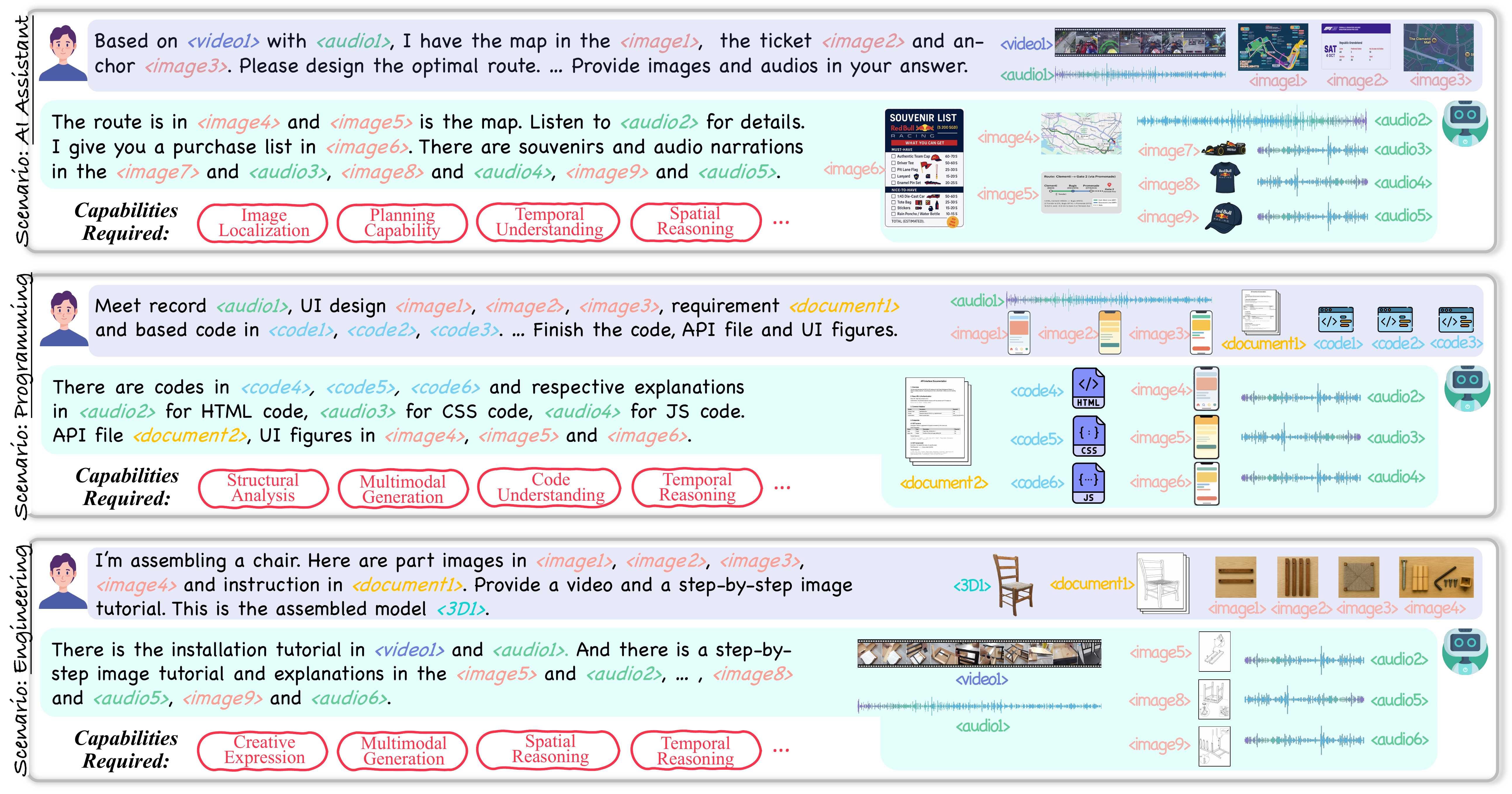}
    \vspace{-4mm}
    \caption{
    Illustration of the any-to-any interleaved multimodal paradigm with different real-world application scenarios.
    Solving any-to-any interleaved multimodal learning requires complex and combined capabilities.
    }
    \vspace{-4mm}
    \label{fig:intro}
\end{figure*}

MLLMs have rapidly progressed from early architectures centered on visual-language understanding to recent unified frameworks that jointly support both understanding and generation within a single model~\cite{fei2024vitron, xie2024showo, wu2024next, li2025onecat, ai2025ming,guo2025m2,li2025omniflow,ye2024x}.
This integration has substantially broadened the functional scope of MLLMs, enabling more comprehensive multimodal reasoning and content creation.
Yet, studies indicate that simple unification remains inadequate for achieving true general-purpose multimodal intelligence~\cite{li2024multimodal,fei2025path,mumuni2025large, wu2025janus}.
A more practical and flexible paradigm, termed \textit{interleaved multimodal learning}~\cite{lin2024learning, li2024llava, tian2024mm, wang2024cosmo, bachmann20244m, ataallah2024minigpt4, li2024textbind, wang2025contextual}, is needed, where inputs and outputs consist of arbitrarily ordered sequences of modalities (e.g., text and images).
Such interleaving better reflects real-world multimodal interactions and is key to building systems that can seamlessly perceive, reason, and respond across diverse modalities.
To advance this paradigm, several interleaved multimodal benchmarks have been introduced, including ITLVD-BENCH~\cite{liu2024holistic}, MMIE~\cite{xia2024mmie}, CoMM~\cite{chen2025comm}, ISG-Bench~\cite{chen2024interleaved}, OpenING~\cite{zhou2025opening}, which assess models on interleaved text–image understanding and generation tasks.
However, these benchmarks still exhibit notable limitations that constrain further development in the field.

The primary limitation lies in their narrow focus on only two modalities, i.e., \textbf{text} and \textbf{image}, thus failing to capture the full potential of multimodal learning.
In contrast, the MLLM landscape has rapidly advanced, with many recent models capable of understanding and generating across a broader spectrum of modalities.
Representative examples such as Unified-IO~\cite{lu2022unified}, NExT-GPT~\cite{wu2024next}, and AnyGPT~\cite{zhan2024anygpt}, etc., illustrate this growing paradigm of \textit{any-to-any multimodal} learning.
In practical applications such as AI assistants, programming copilots, and engineering design systems, users typically interact through complex interleaved multimodal inputs and expect correspondingly diverse multimodal outputs.
As shown in Fig.~\ref{fig:intro}, an AI assistant may process intertwined textual instructions, sketches, and images, and respond with a combination of textual reasoning, annotated visuals, or synthesized audio–visual content.
Such capabilities represent the next generation of MLLMs, emphasizing a unified treatment of both modality and functionality.
However, current interleaved multimodal benchmarks (e.g., MMIE~\cite{xia2024mmie}) fall short of capturing this essential \textit{any-to-any interleaved multimodal learning} paradigm within a unified framework.

Beyond the limitation in modality coverage and paradigm, existing interleaved multimodal benchmarks suffer from several additional shortcomings.
\textbf{First}, their evaluation dimensions are neither sufficiently universal nor diverse, in which each data instance typically targets a single, isolated capability, failing to reflect the composite and intertwined nature of real multimodal reasoning.
In contrast, authentic any-to-any multimodal learning usually involves multiple competencies within a single instance.
For example as in Fig.~\ref{fig:intro}, the AI assistant may simultaneously require comprehension of audio, image, and video inputs, precise image localization, and multimodal generation tasks such as producing images or audio, all demanding temporal understanding, spatial reasoning, and multi-step cognitive planning, etc.
Thus, a genuinely unified any-to-any interleaved paradigm should not only expand modality coverage but also emphasize complex reasoning and multi-stage generation.
\textbf{Second}, current benchmarks mainly concentrate on general-domain scenarios and overlook the diversity of real-world contexts.
To accurately model practical applications and rigorously evaluate MLLMs generalization, any-to-any multimodal learning should encompass a broader range of domains and tasks.

To bridge all these gaps, we propose \textbf{\textsc{UniM}}, the first \textit{Unified Any-to-Any Interleaved Multimodal} Benchmark.
We curate data from real-world sources, such as publicly open data, online social platforms, and large-scale knowledge bases such as YouTube and Wikipedia.
After rigorous manual filtering, annotation, and quality verification, we construct a dataset of 31,026 high-quality instances spanning 30 diverse domains.
\textsc{UniM} covers 7 representative modalities, i.e., text, image, audio, video, document, code, and 3D, with each instance intentionally designed to involve multiple intertwined tasks and reasoning skills.
The benchmark faithfully mirrors real-world any-to-any interleaved input–output patterns within an open-form framework.
To establish a structured evaluation protocol, all instances are divided into three difficulty levels ranging from basic to advanced.
Existing evaluation metrics (e.g., accuracy) can be largely insufficient for assessing such flexible any-to-any multimodal generation, i.e., often failing to provide objective or consistent measurement.
To remedy this, we thus develop the \textsc{UniM Evaluation Suite}, which evaluates model performance along three complementary dimensions: 
(1) \textit{Semantic Correctness \& Generation Quality}, 
(2) \textit{Response Structure Integrity}, 
and (3) \textit{Interleaved Coherence}.
Together, these criteria provide a more comprehensive and accurate assessment of a model's capability to understand and generate within a unified any-to-any interleaved paradigm.

\begin{table*}[!ht]
\centering
\vspace{-4mm}
\caption{Comparison with existing interleaved multimodal benchmarks. Inter. Comb.: Interleaved combinations of modalities. Cap. per Instance: Capability per instance. Difficulty Tax.: Difficulty taxonomy.
}
\vspace{-3mm}
\fontsize{7.5}{8.5}\selectfont
\setlength{\tabcolsep}{0.9mm}
\begin{tabular}{lccccccccc}
\toprule
\textbf{Benchmarks} & \textbf{Domains} & \textbf{Num.} & \textbf{Inter. Comb.} & \textbf{Cap. per Instance} & \textbf{Eval. Metric} & \textbf{Difficulty Tax.} & \textbf{Any-to} & \textbf{to-Any}  &\textbf{Modalities} \\
\midrule
\makecell{\scriptsize ITLVD-BENCH} 
~\cite{liu2024holistic} & 10 & 815 & 2 & Single & 5 & \textcolor{com_red}{\ding{55}} & \textcolor{com_red}{\ding{55}} & \textcolor{com_red}{\ding{55}} & \languagelogo \imagelogo \\

OpenING~\cite{zhou2025opening} & 8 & 5,400 & 4 & Single & 7 & \textcolor{com_red}{\ding{55}} & \textcolor{com_red}{\ding{55}} & \textcolor{com_red}{\ding{55}} &\languagelogo \imagelogo \\

ISG-Bench~\cite{chen2024interleaved} & 8 & 1,150 & 3 & Single & 4  & \textcolor{com_red}{\ding{55}} & \textcolor{com_red}{\ding{55}} & \textcolor{com_red}{\ding{55}} & \languagelogo \imagelogo \\

CoMM~\cite{chen2025comm} & 3 & / & 4 & Single & 3 & \textcolor{com_red}{\ding{55}} & \textcolor{com_red}{\ding{55}} & \textcolor{com_red}{\ding{55}} & \languagelogo \imagelogo \\

MMIE~\cite{xia2024mmie} & 10 & 20,103 & 3 & Single & 7 & \textcolor{com_red}{\ding{55}} & \textcolor{com_red}{\ding{55}} & \textcolor{com_red}{\ding{55}} & \languagelogo \imagelogo \\

\rowcolor{head_fill}
\textbf{\textsc{UniM} (Ours)} & \bf 30 & \bf 31,026 & \bf 41 & \bf Multiple & \textbf{13} & \textcolor{com_green}{\ding{51}} & \textcolor{com_green}{\ding{51}} & \textcolor{com_green}{\ding{51}} & \makecell{\languagelogo \imagelogo \audiologo  \videologo \documentlogo \codelogo \threedlogo} \\
\bottomrule
\end{tabular}
\label{table: comparison}
\vspace{-2mm}
\end{table*}

Benchmarking any-to-any interleaved multimodal learning requires models capable of reasoning over complex multimodal contexts and performing structured, goal-oriented generation rather than simple content synthesis.
To this end, we introduce \textbf{\textsc{UniMA}}, a \textit{Unified Any-to-Any Interleaved Multimodal Agentic} model that serves as the baseline system for \textsc{UniM}.
Built upon an agentic framework, \textsc{UniMA} integrates specialized multimodal encoders and decoders to enable coherent comprehension and generation across heterogeneous modalities.
At its core, it devises a \textit{Traceable Evidence Reasoning Module} that plans, validates, and refines intermediate reasoning steps before producing the final interleaved outputs, thereby improving response, generation quality, and overall interleaved coherence.

Extensive experiments show that current any-to-any MLLMs still struggle considerably on \textsc{UniM}, highlighting the intrinsic difficulty and challenge of this setting.
In-depth analyses further reveal the specific weaknesses of existing models and offer actionable insights for advancing unified interleaved multimodal learning.
Meanwhile, \textsc{UniMA} establishes a strong and interpretable baseline, achieving consistent improvements over prior methods.
Overall, this work presents the first comprehensive benchmark for unified any-to-any interleaved multimodal learning, offering a large-scale, high-quality dataset, a principled evaluation suite, and a robust baseline to catalyze future research in this emerging frontier.

\vspace{-1mm}
\section{Related Work}
\label{sec:related_work}
\vspace{-1.5mm}

Multimodal learning has witnessed rapid advances with the emergence of MLLMs that can integrate text and image modalities for tasks such as captioning and visual question answering~\cite{liu2023visual,li2023blip2,peng2024kosmos2,Qwen3VLHF2025,xu2025qwen3omni,wu2024deepseek}.
Recent developments ~\cite{chameleon2024,xie2024showo,emu3_2024,cui2025emu35, xiao2025omnigen,chen2025blip3,chen2024internvl} have extended these models toward interleaved comprehension and generation, where text and images appear in alternating sequences rather than isolated pairs.
Such interleaved modeling is increasingly regarded as a critical capability for next-generation multimodal systems, since it better reflects the natural patterns of human communication.
However, existing studies predominantly concentrate on the image-text scenario, and thus fall short of addressing the broader requirement of \textit{any-to-any interleaving} that involves modalities beyond vision and language, such as audio, video, documents, code, and 3D.
This paradigm represents a more realistic setting, supported by a growing number of advanced MLLMs, such as NExT-GPT~\cite{wu2024next}, AnyGPT~\cite{zhan2024anygpt}, MIO~\cite{wang2024mio}, Spider~\cite{lai2024spider}, Codi-2~\cite{tang2024codi} and ModaVerse~\cite{wang2024modaverse}.

Meanwhile, a variety of datasets and benchmarks have been proposed to support research in interleaved multimodal learning.
Relevant resources, such as MMIE~\cite{xia2024mmie} and CoMM~\cite{chen2025comm}, provide structured benchmarks for interleaved multimodal comprehension and generation.
Although these benchmarks represent an important step forward, they remain restricted to image-text scope and fail to support the evaluation of arbitrary modality combinations.
Moreover, these benchmarks only simulate relatively simple language–vision interleaving scenarios, whereas realistic any-to-any interactions are far more complex, especially when involving multiple heterogeneous modalities.
This leaves clear gaps for a unified, high-quality benchmark that can systematically assess \textit{any-to-any interleaved multimodal comprehension and generation}, which motivates the development of \textsc{UniM}.
Table~\ref{table: comparison} illustrates the comparison between \textsc{UniM} and existing interleaved benchmarks.

\section{\textsc{UniM}: Unified Any-to-Any Interleaved Multimodal Benchmark}
\vspace{-1mm}
\label{benchmark}

\textsc{UniM} takes an open-formed QA format, where input or output is a sequence of interleaved information pieces of any combined modalities, with non-textual pieces represented by placeholder tags (e.g., `$<$image1$>$', `$<$video2$>$'). 
Also, the sequences may encompass multiple modalities, and each modality contains multiple items (e.g., multiple images).
Fig. \ref{fig:intro} exemplifies the paradigm.
Appendix~\S\ref{app:task definition} gives a formal task definition.

\subsection{Data Construction}

\textbf{Process Pipeline.}
We first collect a wide range of multimodal data mainly from three sources: curated samples from public datasets, real-world multimedia content from social media (vlogs, posts), and open resources (forums, websites).
Then, we manually design interleaved combinations tailored to different modalities, providing templates for the later construction.
During the construction of QA pairs, we design task types and template instances to ensure task diversity and semantic validity.
Based on these, GPT-5-mini~\cite{OpenAI_Introducing_GPT5_2025} is employed to generate additional candidate instances for data expansion, so as to simulate more any-to-any multimodal scenarios that are hard to directly retrieve from the Internet.

\vspace{1mm}
\noindent \textbf{Quality Control.}
We adopt a two-phase quality control process.
All QA pairs are manually reviewed and revised as needed to ensure that the modality placeholder tags comply with task specifications and that the content remains logically consistent.
Then, an independent checking process is conducted, where reviewers carefully examine each completed sample to further ensure the reliability and high quality of our dataset.
Details regarding dataset construction and quality control are provided in 
Appendix~\S\ref{app: construction process} and Appendix~\S\ref{app: quality control}.

\begin{table}[t!]
\centering
\vspace{-2mm}
\caption{General statistics of \textsc{UniM} dataset.}
\vspace{-2mm}
\fontsize{7.5}{8.5}\selectfont
\setlength{\tabcolsep}{1.2mm}
\begin{tabular}{lcccccc}
\toprule
\rowcolor{head_fill}
& \multicolumn{6}{c}{\textbf{Modality}} \\
\midrule
& Image & Audio & Video & Document & Code & 3D \\
\textbf{Num.} & \makecell{22,936 \\ \tiny (73.9\%)} & \makecell{24,963 \\ \tiny (80.5\%)} & \makecell{2,336 \\ \tiny (7.5\%)} & \makecell{3,858 \\ \tiny (12.4\%)} & \makecell{807 \\ \tiny (2.6\%)} & \makecell{420 \\ \tiny (1.4\%)} \\
\midrule
\rowcolor{head_fill}
 & \multicolumn{3}{c}{\textbf{Field}} & \multicolumn{3}{c}{\textbf{Difficulty}}  \\ 
\cmidrule(lr){2-4} \cmidrule(lr){5-7}
& NS & SS & GA  & Easy & Medium & Hard \\
\textbf{Num.} & \makecell{10,624 \\ \tiny(34.2\%)} & \makecell{11,574 \\ \tiny(37.3\%)} & \makecell{8,828 \\ \tiny (28.5\%)} & \makecell{10,678 \\ \tiny (34.4\%)} & \makecell{14,088 \\ \tiny (45.4\%) } & \makecell{6,260 \\ \tiny (20.2\%)} \\
\bottomrule
\end{tabular}
\label{table: stats}
\end{table}

\begin{figure}[t!]
\vspace{-4mm}
\centering
\includegraphics[width=0.9\linewidth]{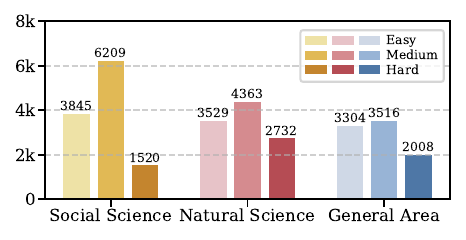}
\vspace{-4mm}
\caption{Distribution of different difficulty levels.} 
\label{fig:difficulty_level_dis}
\vspace{-4mm}
\end{figure}

\vspace{1mm}
\noindent \textbf{Data Statistics.}
The resulting \textsc{UniM} data contains a total of 31,026 instances, covering 30 real-world domains across natural science, social science and general area. Table~\ref{table: stats} presents statistics of \textsc{UniM}. 
We further define a rule-based progressive difficulty taxonomy, which categorizes instances into three levels: \textit{Easy}, \textit{Medium}, and \textit{Hard}. The distribution of difficulty levels across different fields is shown in Fig.~\ref{fig:difficulty_level_dis}.
Details regarding the classification criteria and grading process are provided in Appendix~\S\ref{app: progressive difficulty taxonomy}.

\subsection{Data Core Characteristics}

\textsc{UniM} features the following key aspects:
\noindent \textbf{Any-to-Any Interleaved Modalities.} \textsc{UniM} covers 7 modalities, supports any-to-any interleaved combinations, faithfully simulating real-world scenarios.

\noindent \textbf{Universal and Diverse Capabilities.} \textsc{UniM} evaluates the comprehensive and diverse capabilities of MLLMs. 
Appendix~\S\ref{app: capabilities to be evaluated} details definitions of each capability.

\noindent \textbf{Multi-domain Coverage.} \textsc{UniM} encompasses 30 real-world domains across different fields.

\noindent \textbf{Multiple Tasks per Instance.} Each instance in \textsc{UniM} encompasses multiple task objectives ranging from understanding to generation.
Detailed descriptions and definitions of tasks can be found in Appendix~\S 
\ref{app: task types}.

\noindent \textbf{Progressive Difficulty.} Instances in \textsc{UniM} are graded into 3-scale difficulty levels, providing comprehensive evaluation support for both simple and complex interleaved scenarios.

\noindent \textbf{Large Scale and High Quality.} \textsc{UniM} contains 31,026 instances, constructed through a rigorous pipeline for high semantic validity and logical coherence.

\vspace{-1mm}
\section{Evaluation Suite for \textsc{UniM}}
\label{sec:eval}

\vspace{-1.5mm}
Traditional metrics (e.g., accuracy) focus on single-modality or closed-form matching, which limits their applicability in complex any-to-any interleaved scenarios.
We thus rethink evaluation methods and design a systematic evaluation suite for \textsc{UniM}, with three dimensions: \textit{Semantic Correctness \& Generation Quality}, \textit{Response Structure Integrity} and \textit{Interleaved Coherence} (cf., Fig.~\ref{fig:eval_suite}).
All these metrics and algorithms details are further provided in Appendix~\S
\ref{app:details_eval_suite}.

\vspace{-1mm}
\subsection{Semantic Correctness \& Generation Quality}
\vspace{-1mm}
\textbf{Semantic Correctness (SC)} measures how well the generated output semantically aligns with the reference answer.
To ensure fair evaluation across modalities with varying instruction-following capabilities, we convert all modality outputs into comparable caption-like textual representations and employ the LLM-as-a-Judge~\cite{zheng2023judging} strategy for measurement. 
\textbf{Generation Quality (GQ)} evaluates the perceptual quality and structural coherence of generated content.
Accordingly, we design modality-specific no-reference quality assessment methods to ensure unified and comparable quality metrics across multimodal scenarios. 
Then, we compose both SC and GQ into \textbf{Semantic–Quality Coupled Score (SQCS)} to reflect the overall performance:
\setlength\abovedisplayskip{3pt}
\setlength\belowdisplayskip{3pt}
\begin{equation}
\text{SQCS} = \text{SC} \cdot \left( \eta^{\text{SQCS}}  + (1 - \eta^{\text{SQCS}}) \cdot \text{GQ} \right) \,.
\label{eq:1}
\end{equation}

\begin{figure}[!t]
\centering
\includegraphics[width=\linewidth]{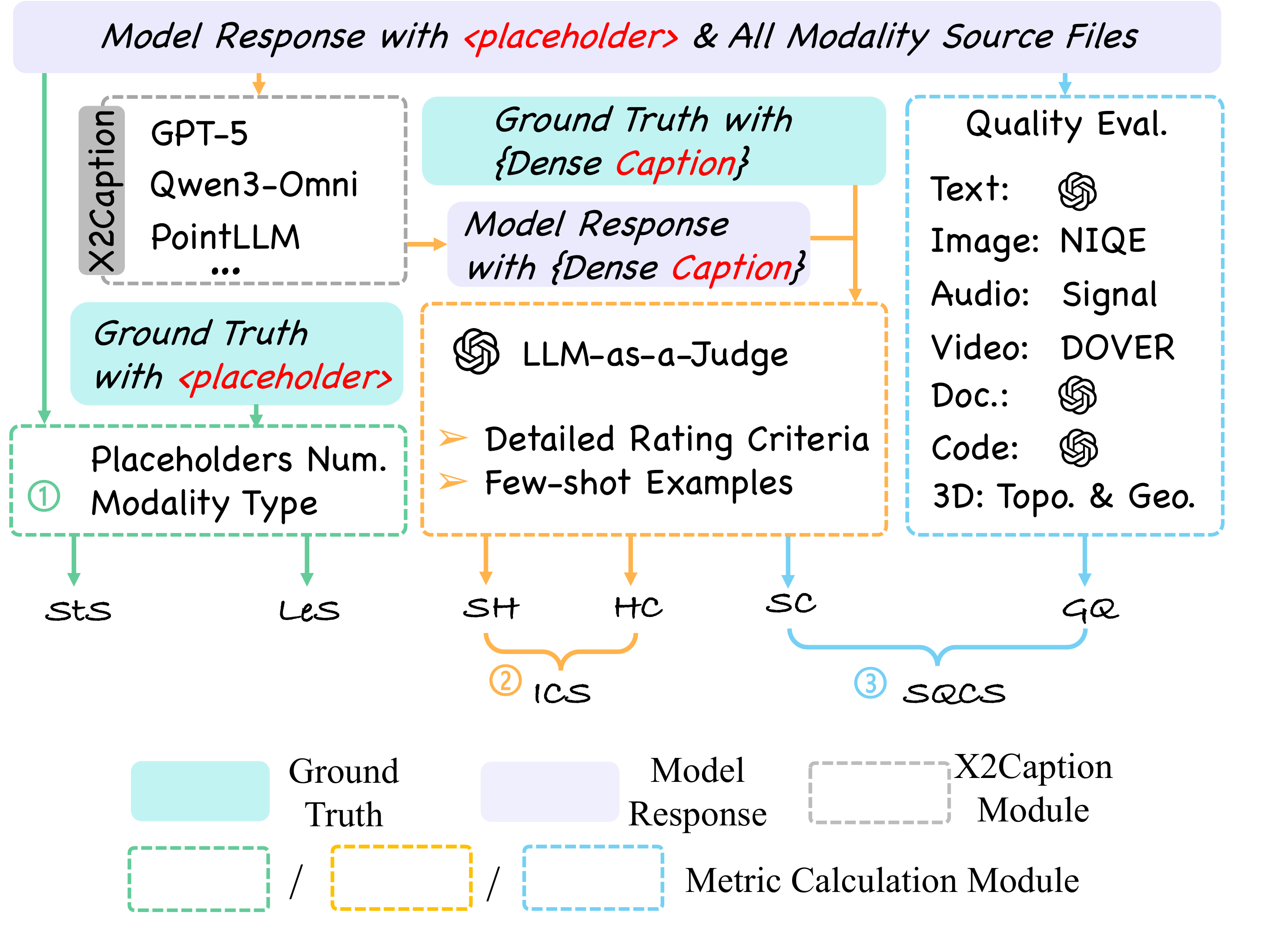}
\vspace{-8mm}
\caption{
Illustration of the \textsc{UniM} evaluation suite. \textcolor{rsi_green}{\large \ding{172}} refers to the calculation process of the StS and LeS (\S\ref{Response Structure Integrity}). \textcolor{ics_orange}{\large \ding{173}} represents the calculation process of the ICS in Eq.~\textcolor{cvprblue}{(}\ref{eq:2}\textcolor{cvprblue}{)}. 
\textcolor{sqcs_blue}{\large \ding{174}} refers to the calculation process of the SQCS; please refer to Eq.~\textcolor{cvprblue}{(}\ref{eq:1}\textcolor{cvprblue}{)}.
}
\vspace{-4mm}
\label{fig:eval_suite}
\end{figure}

\vspace{-1mm}
\subsection{Response Structure Integrity}
\label{Response Structure Integrity}

\vspace{-1mm}
We devise \textbf{Response Structure Integrity} to evaluate whether a model adheres to task-defined structural requirements regarding modality types and item quantities, regardless of semantic or logical correctness.
Technically, we break it down into two branches:

\noindent \textbf{Strict Structure Score (\textbf{StS})} evaluates the strict structural consistency of a model's output.
StS requires that the types and quantities of modalities generated in model's response precisely correspond to those in the ground truth.
Any missing or redundant modalities, or discrepancies in the number of modality placeholder tags, are explicitly penalized. 

\noindent \textbf{Lenient Structure Score (\textbf{LeS})} evaluates the degree of coverage at the modality level.
LeS assesses whether the types of modalities generated in model's response are consistent with those in the ground truth.

\vspace{-1mm}
\subsection{Interleaved Coherence}
\vspace{-1mm}

\textbf{Interleaved Coherence} is designed to evaluate a model's ability to maintain logical connectivity and expressive coordination during multimodal integration, measured by \textbf{Holistic Coherence (HC)}, which focuses on cross-modal semantic and structural consistency, and \textbf{Stylistic Harmony (SH)}, which evaluates consistency in writing style, tone, and visual aesthetics. 
We adopt the LLM-as-a-Judge~\cite{zheng2023judging} to quantify HC and SH, and ultimately use a composite metric over them: \textbf{Interleaved Coherence Score~(ICS)}:
\begin{equation}
\text{ICS} = \eta^\text{ICS} \cdot \text{HC} + (1-\eta^\text{ICS}) \cdot \text{SH} \,.
\label{eq:2}
\end{equation}

\vspace{-1mm}
\subsection{Supporting Rate}
\vspace{-1mm}
There might be the common case in \textsc{UniM}, where an MLLM may not support certain modalities and thus fail on a portion of samples (resulting in overall low metrics), yet it still can achieve top performance on those samples it supports well.
To objectively assess a model's performance, we further introduce the \textbf{Supporting Rate ($\tau$)} as a conditional modifier on top of the above three evaluation dimensions, i.e., by distinguishing the model performance between two conditions: $\mathcal{X}^{abs}$ and $\mathcal{X}^{rel}$. 
$\mathcal{X}^{abs}$ represents the model's original capability (i.e., absolute), while $\mathcal{X}^{rel}$ considers the model's supporting rate on the entire \textsc{UniM}:
\setlength\abovedisplayskip{3pt}
\setlength\belowdisplayskip{3pt}
\begin{equation}
    \mathcal{X}^{rel} =\tau \cdot \mathcal{X}^{abs} \,.
\end{equation}

\vspace{-2mm}
\section{{\textsc{UniMA}: An Agentic Model for \textsc{UniM}}}
\label{agent}

\begin{figure}[t!]
    \centering
    \includegraphics[width=0.91\linewidth]{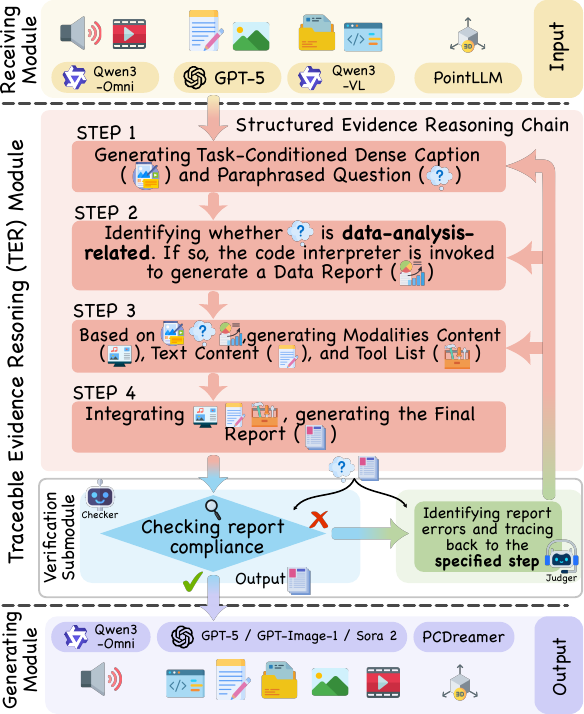}
    \vspace{-2mm}
    \caption{Overview of the \textbf{\textsc{UniMA}} architecture.}
    \vspace{-8mm}
    \label{fig:unima}
\end{figure}

\vspace{-1mm}
\noindent \textbf{Overview.}
We build \textsc{UniMA} to bridge the gap in existing MLLMs that lack strong enough any-to-any multimodal transformation across interleaved modalities.
Technically, the overall agentic pipeline operates through three coordinated modules (cf. Fig. \ref{fig:unima}). 
The \textbf{Receiving Module} converts non-text modalities into \textit{task-conditioned dense caption (TCDC)}, forming a unified text space for subsequent reasoning.
The \textbf{Traceable Evidence Reasoning (TER) Module} performs structured reasoning by generating, verifying, and refining traceable evidence to construct a logically consistent and verifiable final report that guides subsequent generation. 
Finally, the \textbf{Generating Module}, driven by the verified \textit{final report}, produces interleaved multimodal outputs, completing a loop from understanding to generation.

\vspace{-1mm}
\subsection{Traceable Evidence Reasoning Module}
\vspace{-1mm}

TER serves as the core inference module of \textsc{UniMA}, 
responsible for transforming multimodal inputs passed from the front-end into a structured and verifiable final report. 
Instead of relying on an implicit Chain-of-Thought~\cite{wei2022chain}, TER establishes an explicit evidence-grounded and traceable reasoning chain, enabling each inference step to be justified, verifiable, and revisable within the overall reasoning process. As illustrated in Fig.~\ref{fig:unima}, there are four key steps in the Structured Evidence Reasoning Chain (SERC) within TER.

\noindent \textbf{$\bullet$ Step 1:} 
TER leverages the Receiving Module to produce TCDC and a \textit{paraphrased question} from multimodal inputs, primarily enhancing the SC of the content.

\noindent \textbf{$\bullet$ Step 2:} TER determines whether the task involves data analysis; if so, the code interpreter is invoked to generate a \textit{data report}, improving SC through factual grounding.

\noindent \textbf{$\bullet$ Step 3:} The extracted information is organized into \textit{modalities content}, \textit{text content}, and a \textit{tool list}: modalities content mainly improves SQCS, text content enhances ICS, and the tool list increases StS/LeS.

\noindent \textbf{$\bullet$ Step 4:} TER integrates all evidence to produce a draft of the \textit{final report}, enhancing all evaluation dimensions to support reliable multimodal generation.

Notably, in TER, the \textit{Checker} detects factual and logical errors in the report, while the \textit{Judger} backtracks for corrective reasoning. Through iterative \textit{generation–checking–backtracking–regeneration} cycles, TER achieves traceable and reliable multimodal reasoning.
More details of \textsc{UniMA} are provided in Appendix~\S\ref{app:details_agent}.

\vspace{-1mm}
\section{Experiments}
\label{exp}

\vspace{-1mm}
We select \textbf{AnyGPT}~\cite{zhan2024anygpt}, \textbf{NExT-GPT}~\cite{wu2024next}, and \textbf{MIO}~\cite{wang2024mio} as the representative any-to-any MLLMs.\footnote{We experiment with more MLLMs, where results and analyses are shown in Appendix~\S\ref{app: exp results and analysis}.}

In our experiments, $\eta^{\text{SQCS}}=0.7, \eta^\text{ICS}=0.8$, which help to achieve the best alignment with human evaluations.
More settings are detailed in Appendix~\S\ref{app: experimental settings}.

\vspace{-2mm}
\subsection{Rationality of Evaluation Suite}
\vspace{-1mm}
Before we analyze the model performance, here, we study the rationality of our proposed evaluation suite.

\begin{figure}[t!]
\centering
\begin{minipage}[t]{0.492\linewidth}
\centering
\includegraphics[width=\linewidth]{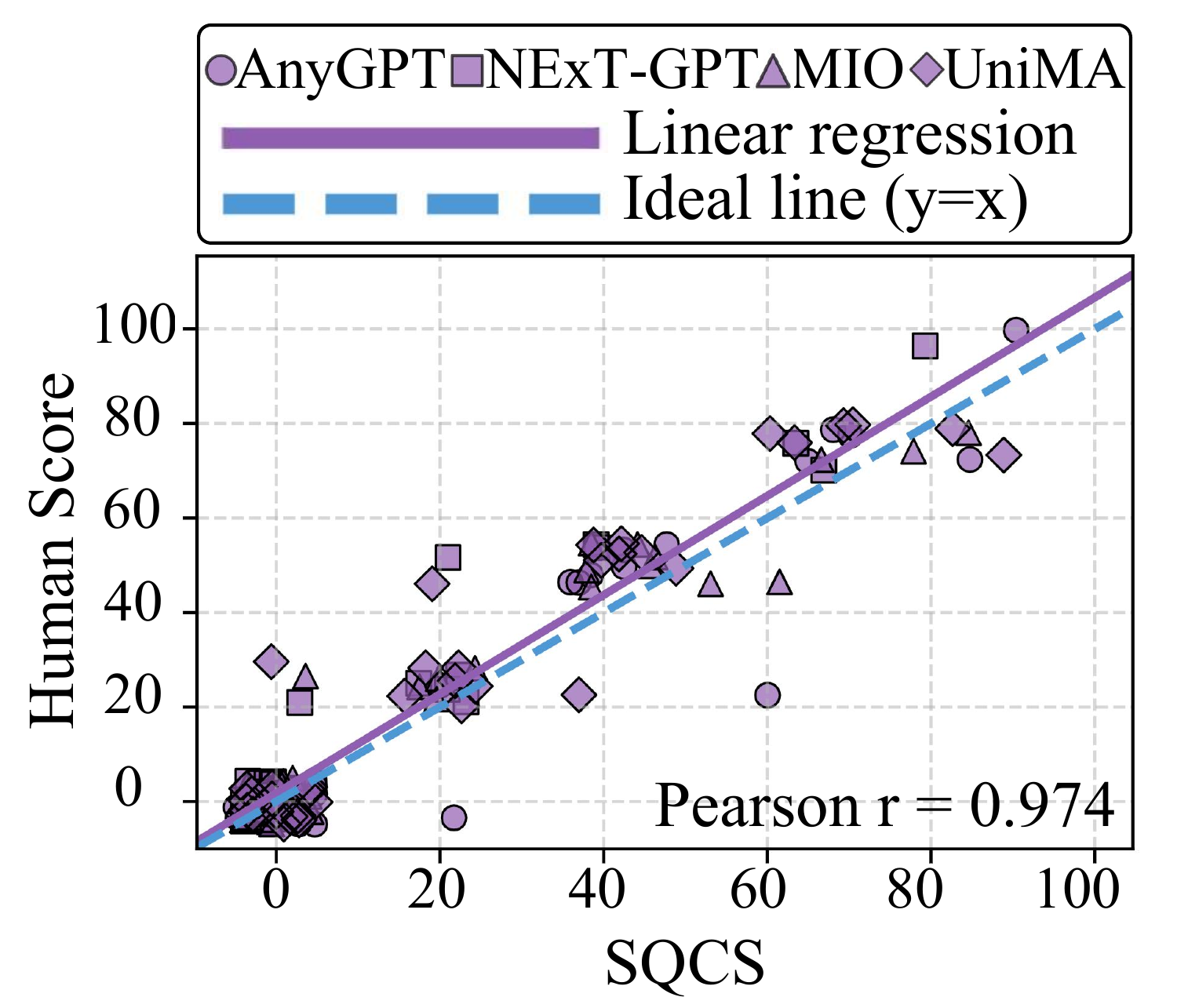}
\end{minipage}
\hfill
\begin{minipage}[t]{0.492\linewidth}
\centering
\includegraphics[width=\linewidth]{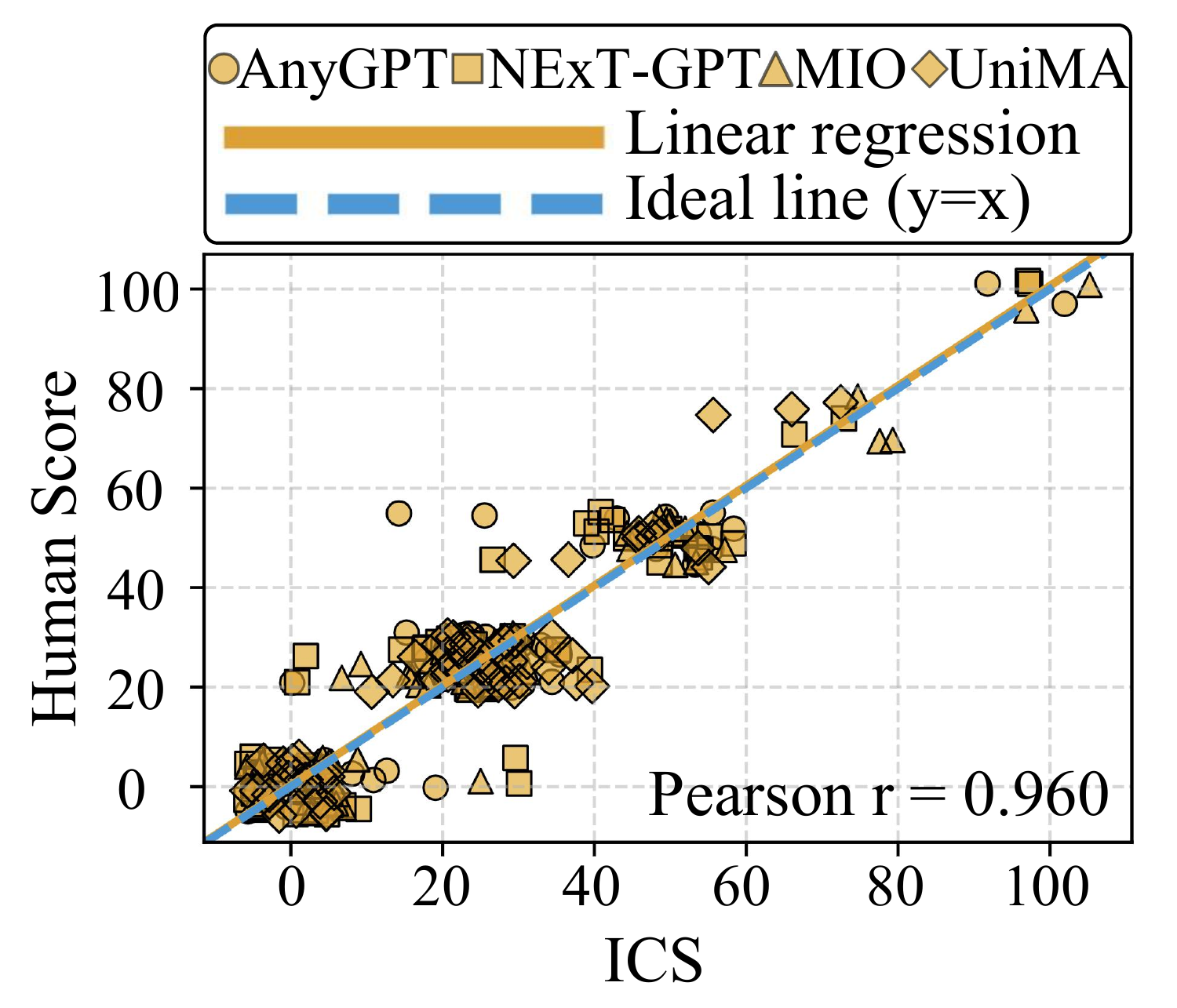}
\end{minipage}
\vspace{-7.5mm}
\caption{Results for rationality verification of SQCS and ICS.}
\vspace{-3mm}
\label{rationality_sqcs_ics}
\end{figure}

\vspace{1mm}
\noindent $\blacktriangleright$ \textbf{SQCS and ICS.}
We assess the rationality of SQCS and ICS through Pearson correlation analysis between automatic and human evaluations.
As shown in Fig.~\ref{rationality_sqcs_ics}, both metrics exhibit a remarkably high degree of linear correlation with human evaluation results, with the Pearson correlation coefficient reaching \(r=0.974\) for SQCS and \(r=0.960\) for ICS. 
These results confirm that the proposed automatic metrics can accurately reflect the performance assessed by human evaluation.

\begin{figure}[t!]
\vspace{-2mm}
\centering
\begin{minipage}[t]{0.492\linewidth}
\centering
\includegraphics[width=\linewidth]{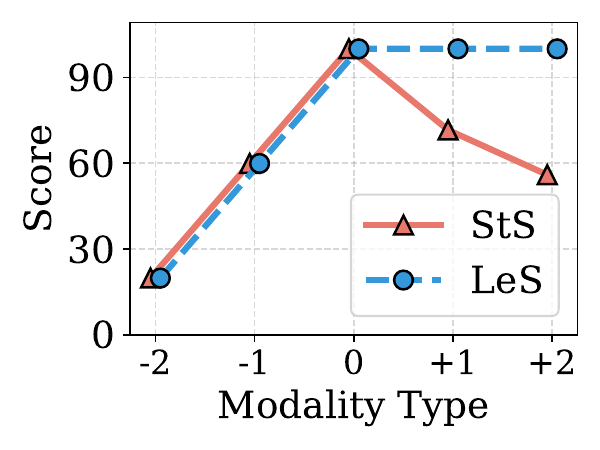}
\end{minipage}
\hfill
\begin{minipage}[t]{0.492\linewidth}
\centering
\includegraphics[width=\linewidth]{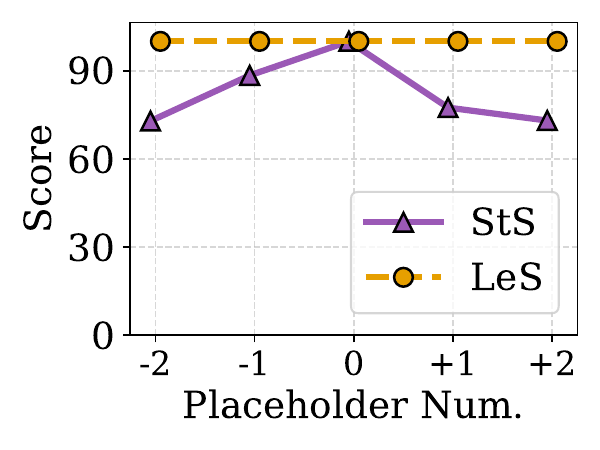}
\end{minipage}
\vspace{-8mm}
\caption{Results for rationality verification of StS and LeS.}
\vspace{-6mm}
\label{rationality_sts_les}
\end{figure}

\begin{table*}[ht!]
\centering
\vspace{-3mm}
\caption{Assessment results on Response Structure Integrity.}
\vspace{-2mm}
\fontsize{7.5}{7.5}\selectfont
\setlength{\tabcolsep}{1.8mm}
\renewcommand{\arraystretch}{1.2}
\begin{tabular}{lcccccccccccc}
\toprule
\multirow{2}{*}{\bf Model} &
\multicolumn{4}{c}{\textbf{Natural Science}} &
\multicolumn{4}{c}{\textbf{Social Science}} &
\multicolumn{4}{c}{\textbf{General Area}} \\
\cmidrule(lr){2-5} \cmidrule(lr){6-9} \cmidrule(lr){10-13}
& \bf StS$^{abs}$ & \bf LeS$^{abs}$ & \bf StS$^{rel}$ & \bf LeS$^{rel}$
& \bf StS$^{abs}$ & \bf LeS$^{abs}$ & \bf StS$^{rel}$ & \bf LeS$^{rel}$
& \bf StS$^{abs}$ & \bf LeS$^{abs}$ & \bf StS$^{rel}$ & \bf LeS$^{rel}$ \\
\midrule
\textbf{AnyGPT}~\cite{zhan2024anygpt} & 12.9 & 27.8 & 12.2 & 21.4 & 14.9 & 16.6 & 14.5 & 16.2 & 12.5 & 16.4 & 9.8 & 13.6 \\
\textbf{NExT-GPT}~\cite{wu2024next} & 2.0 & 2.2 & 1.2 & 1.3 & 1.3 & 1.7 & 1.2 & 1.5 & 2.2 & 2.5 & 1.4 & 1.7 \\
\textbf{MIO}~\cite{wang2024mio} & 1.3 & 1.9 & 0.9 & 1.3 & 4.1 & 5.2 & 4.0 & 5.1 & 3.3 & 3.8 & 2.4 & 2.9 \\
\rowcolor{RSI_head_fill}
\textbf{\textsc{UniMA}} & 50.8 & 62.2 & 50.8 & 62.2 & 58.1 & 72.9 & 58.1 & 72.9 & 71.3 & 84.3 & 71.3 & 84.3 \\
\bottomrule
\end{tabular}
\label{table: rsi}
\end{table*}

\begin{table}[t!]
\fontsize{7.5}{8.5}\selectfont
\setlength{\tabcolsep}{1.mm}
\centering
\vspace{-4mm}
\caption{Performance on Semantic Correctness \& Generation Quality and Supporting Rate ($\tau$).}
\vspace{-2mm}
\begin{tabular}{llcclrl}
\toprule
\bf Field & \bf Models & \bf SC & \bf \ \ \ GQ\ \ \ \ & \bf SQCS$^{abs}$ & \multicolumn{1}{c}{$\tau$} & \bf SQCS$^{rel}$ \\
\midrule
\multirow{3}{*}{\rotatebox{90}{\makecell{Natural\\ Science}}}
& AnyGPT~\cite{zhan2024anygpt}
& 13.7
& 37.9
& 11.1 \tinybar{0.111*2}{1.0}
& 90.4 \tinypie{90.4}
& 10.7 \tinybar{0.107*2}{1.0} \\
& NExT-GPT~\cite{wu2024next}
& 8.4
& 23.4
& 6.2\ \ \  \tinybar{0.0620*2}{1.0}
& 62.0 \tinypie{62.0}
& 2.9 \ \ \tinybar{0.029*2}{1.0}\\
& MIO~\cite{wang2024mio}
& 19.7
& 29.1
& 15.9 \tinybar{0.161*2}{1.0}
& 59.2 \tinypie{59.2}
& 10.0 \tinybar{0.10*2}{1.0} \\
\rowcolor{SCGQ_head_fill} 
& \textbf{\textsc{UniMA}}  
& 59.8
& 79.7
& 57.3 \tinybar{0.573*1.2}{1.0}
& 100 \tinypie{100}
& 57.3 \tinybar{0.573*1.2}{1.0} \\
\midrule
\multirow{3}{*}{\rotatebox{90}{\makecell{Social\\ Science}}}
& AnyGPT~\cite{zhan2024anygpt}        
& 18.0
& 23.8
& 15.5 \tinybar{0.155*2}{1.0}
& 94.7 \tinypie{94.7}
& 14.7 \tinybar{0.147*2}{1.0}\\
& NExT-GPT~\cite{wu2024next}
& 16.8
& 31.9
& 13.3 \tinybar{0.133*2}{1}
& 89.0 \tinypie{89.0}
& 10.8 \tinybar{0.108*2}{1} \\
& MIO~\cite{wang2024mio}
& 25.2
& 32.8
& 21.4 \tinybar{0.214*2}{1}
& 80.8 \tinypie{80.8}
& 16.1 \tinybar{0.161*2}{1} \\
\rowcolor{SCGQ_head_fill} 
& \textbf{\textsc{UniMA}} 
& 76.2
& 81.0
& 72.7 \tinybar{0.727*1.2}{1.0}
& 100 \tinypie{100}
& 72.7 \tinybar{0.727*1.2}{1.0} \\
\midrule
\multirow{3}{*}{\rotatebox{90}{\makecell{General\\ Area}}}
& AnyGPT~\cite{zhan2024anygpt}
& 19.0
& 30.1
& 17.9 \tinybar{0.161*2}{1}
& 90.3 \tinypie{90.3}
& 17.2 \tinybar{0.172*2}{1} \\
& NExT-GPT~\cite{wu2024next}
& 5.4
& 30.0
& 4.4 \ \ \tinybar{0.044*2}{1}
& 76.0 \tinypie{76.0}
& 3.4 \ \ \tinybar{0.034*2}{1} \\
& MIO~\cite{wang2024mio}
& 24.8 
& 37.5
& 21.2 \tinybar{0.212*2}{1}
& 71.7 \tinypie{71.7}
& 15.2 \tinybar{0.152*2}{1} \\
\rowcolor{SCGQ_head_fill} 
& \textbf{\textsc{UniMA}} 
& 64.7
& 83.6
& 62.2 \tinybar{0.622*1.2}{1.0}
& 100 \tinypie{100}
& 62.2 \tinybar{0.622*1.2}{1.0} \\
\bottomrule
\end{tabular}
\label{table: sqcs}
\end{table}

\noindent $\blacktriangleright$ \textbf{StS and LeS.} 
We introduce two types of controlled perturbations: one varying the modality types and the other varying the number of modality placeholder tags.
The perturbation set is constructed based on the ground truth. 
As shown in Fig.~\ref{rationality_sts_les}, StS decreases when modality types or placeholder tags are added or removed, whereas LeS decreases only when modality types are removed.
These results confirm that both types of perturbations induce the anticipated score deviations, validating the soundness and discriminative effectiveness of the proposed metrics.

Detailed procedures and experimental settings of rationality experiments are provided in Appendix\S~\ref{app: rationality}.

\subsection{Main Results and Observations}

$\blacktriangleright$ \textbf{Baseline models exhibit poor performance in terms of absolute metrics on \textsc{UniM}.}
First, according to the results in Table~\ref{table: sqcs}, baseline models achieve low SQCS (mostly below 20\%), indicating substantial semantic deviations between their responses and ground truth. 
Then, NExT-GPT and MIO perform quite poorly on StS and LeS (mostly below 5\%, cf. Table~\ref{table: rsi}), representing severe limitations of the baselines in matching the required modality coverage and quantity.
In addition, baselines perform slightly better on ICS than on the other metrics, but their overall performance remains low (mostly below 50\%; cf. Table~\ref{table: ic}).

\vspace{1mm}
\noindent $\blacktriangleright$ \textbf{Notably lower relative scores owing to baselines' limited support for diverse interleaved input modalities.}
All baseline models suffer notably reduced relative scores.
As shown in Table~\ref{table: rsi}, the most significant decline appears in AnyGPT in general area, where StS$^{abs}$ drops to 12.5\%, StS$^{rel}$ to 9.8\%, LeS$^{abs}$ to 16.4\%, and LeS$^{rel}$ to 13.6\%.
In Table~\ref{table: sqcs}, the largest degradation is observed for MIO in natural science, with SQCS$^{abs}$ decreasing to 15.9\% and SQCS$^{rel}$ to 10.0\%.
Similarly, in Table~\ref{table: ic}, the most substantial drop occurs for MIO in natural science, where ICS$^{abs}$ reaches 52.1\% and ICS$^{rel}$ 31.8\%.
Their restricted support for diverse interleaved input modalities (cf. $\tau$ column of Table~\ref{table: sqcs}) further indicates that these MLLMs exhibit overall inferior relative performance across tasks.

\vspace{1mm}
\noindent $\blacktriangleright$  \textbf{Model performance varies significantly across different fields and domains.}
For SQCS, models perform best in social science (SQCS$^{abs}$: 13.3\%-72.7\% \& SQCS$^{rel}$: 10.8\%-72.7\%, cf. Table~\ref{table: sqcs}), likely due to the prevalence of common concepts and descriptive reasoning patterns in pretraining data.
In contrast, natural science tasks require precise terminology and structured logic rarely seen during training, leading to weaker semantic alignment.
For ICS, the general area performs best (ICS$^{abs}$: 28.1\%-69.8\%, ICS$^{rel}$: 20.0\%-69.8\%, cf. Table~\ref{table: ic}), as open-domain data better aligns with the model’s training style for smoother coherence and tone.
Meanwhile, the higher modality and stylistic complexity of natural and social science tasks make consistent expression harder, lowering ICS performance.

\vspace{1mm}
\noindent $\blacktriangleright$  \textbf{\textsc{UniMA} shows clearly superior performance on \textsc{UniM}, serving as a reasonable baseline.} \textsc{UniMA} surpasses the baselines by a large margin, with StS/LeS 2–6$\times$ higher than AnyGPT and 15-40$\times$ higher than NExT-GPT and MIO (cf. Table~\ref{table: rsi}).
\textsc{UniMA} achieves an SQCS of around 60\% and an ICS approaching 70\%, both exceeding those of the best-performing baseline model (cf. Table~\ref{table: sqcs} and Table~\ref{table: ic}).

Overall, existing any-to-any MLLMs still face significant challenges under unified any-to-any interleaved paradigm. \textsc{UniMA} achieves acceptable performance on \textsc{UniM}, yet there is still room for further enhancement.

\begin{table}[t!]
\centering
\fontsize{7.5}{8.5}\selectfont
\setlength{\tabcolsep}{2.3mm}
\vspace{-2mm}
    \caption{Evaluation results on Interleaved Coherence.}
    \vspace{-3mm}
    \begin{tabular}{llcccc}
        \toprule
        \bf Field & \bf Models & \bf HC & \bf SH & \bf ICS$^{abs}$ & \bf ICS$^{rel}$ \\
        \midrule
        \multirow{4}{*}{\makecell{Natural\\ Science}}
        & AnyGPT~\cite{zhan2024anygpt}
        & 39.9
        & 46.3
        & 41.8
        & 38.5 \\
        & NExT-GPT~\cite{wu2024next}
        & 23.5
        & 26.1
        & 24.9
        & 16.3 \\
        & MIO~\cite{wang2024mio}
        & 49.4	 
        & 63.7
        & 52.1
        & 31.8  \\
        \rowcolor{IC_head_fill}
        & \textbf{\textsc{UniMA}} 
        & 68.4
        & 71.9
        & 69.1
        & 69.1 \\
        \midrule
        \multirow{4}{*}{\makecell{Social\\ Science}}
        & AnyGPT~\cite{zhan2024anygpt}
        & 31.3
        & 35.3
        & 32.1
        & 29.2 \\
        & NExT-GPT~\cite{wu2024next}
        & 24.5	
        & 27.1
        & 21.4
        & 19.0 \\
        & MIO~\cite{wang2024mio}
        & 46.3 
        & 55.0
        & 51.6
        & 42.0  \\
        \rowcolor{IC_head_fill}
        & \textbf{\textsc{UniMA}} 
        & 73.1
        & 76.5
        & 73.8
        & 73.8 \\
        \midrule
        \multirow{4}{*}{\makecell{General\\ Area}}
        & AnyGPT~\cite{zhan2024anygpt}
        & 36.5
        & 41.9
        & 43.6
        & 31.3 \\
        & NExT-GPT~\cite{wu2024next}
        & 27.9
        & 31.1
        & 28.1
        & 20.0 \\
        & MIO~\cite{wang2024mio}
        & 68.3 
        & 77.7
        & 60.0
        & 45.7 \\
        \rowcolor{IC_head_fill}
        & \textbf{\textsc{UniMA}} 
        & 68.7
        & 74.3
        & 69.8
        & 69.8 \\
        \bottomrule
    \end{tabular}
    \label{table: ic}
\vspace{-2mm}
\end{table}

\section{In-depth Analysis and Discussion}

\label{anal_disc}

\subsection{Performance across Various Capabilities}
\vspace{-1mm}
By evaluating all metrics across different capability dimensions, we identify the following trends.

\begin{figure}[t!]
\centering
\vspace{-6mm}
\includegraphics[width=0.9\linewidth]{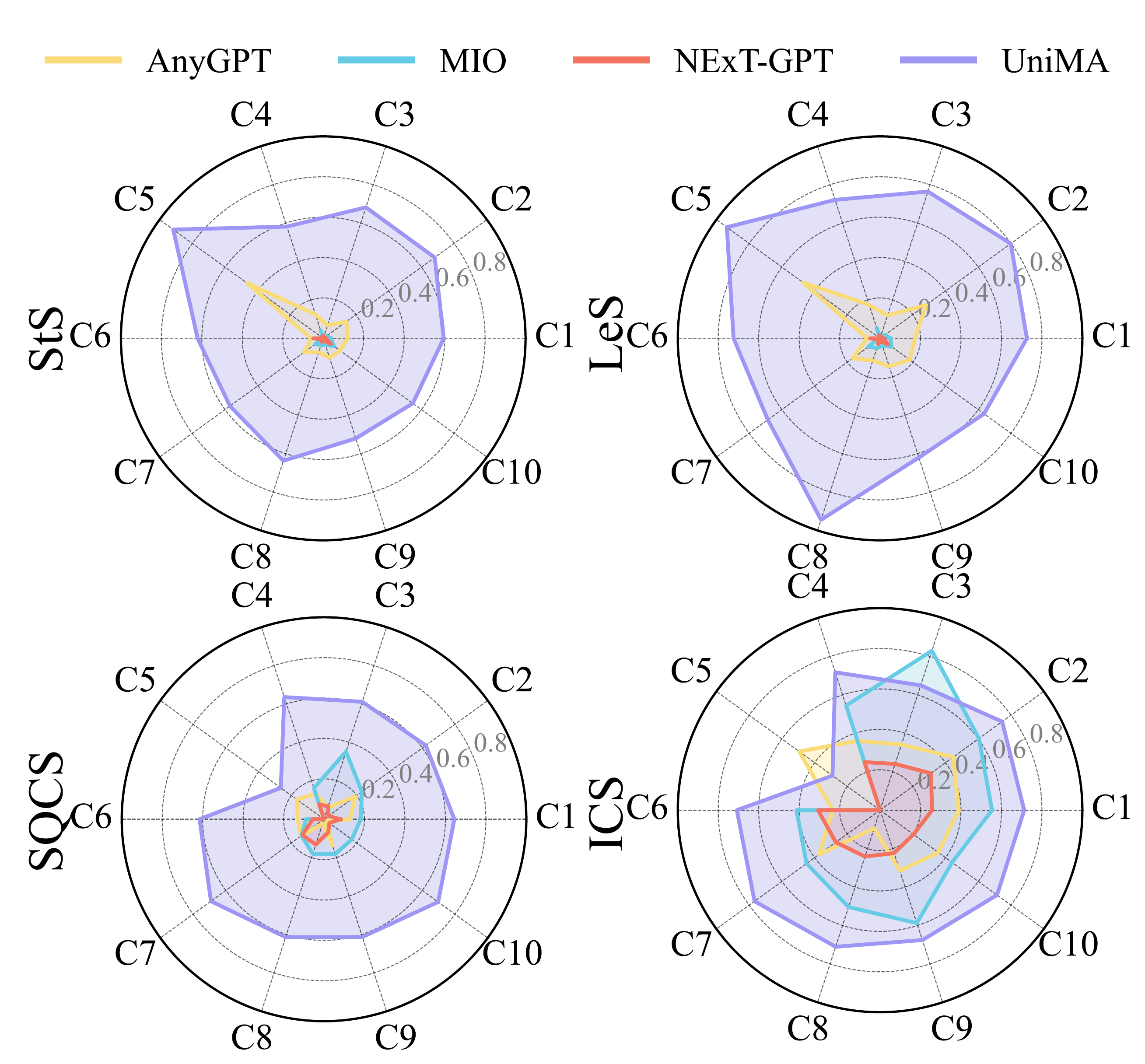}
\vspace{-3mm}
\caption{Results across 10 capabilities on \textsc{UniM}. C1: Perceptual Understanding, C2: Spatial Understanding, C3: Temporal Understanding, C4: Semantic Generation, C5: Content Editing, C6: Creative Expression, C7: Reasoning Capability, C8: Emotional Analysis, C9: Structural Analysis, and C10: Planning Capability. Refer to Appendix~\S\ref{app: capabilities to be evaluated} for details.}
\label{fig:abil}
\vspace{-4mm}
\end{figure}

\vspace{1mm}
\noindent \textit{\textbf{RQ-1: How do MLLMs perform across diverse capability dimensions?}}
As shown in Fig.~\ref{fig:abil}, \textsc{UniMA} consistently achieves the highest and most balanced performance across all 10 capability dimensions, while AnyGPT, MIO, and NExT-GPT show uneven capability distribution with sharp declines in tasks requiring complex multimodal interactions such as content editing. 
This indicates that \textsc{UniMA} attains more stable multimodal fusion and structural control, whereas the baseline models remain fragmented across capabilities, failing to sustain performance once tasks demand higher levels of cross-modal coordination.

\vspace{1mm}
\noindent\textit{\textbf{RQ-2: What trends emerge in metric behavior across different capability types?}}
Across all models, SQCS and ICS remain relatively stable in perception, spatial, and reasoning capabilities, while StS and LeS vary significantly across capabilities. 
Structurally demanding tasks such as temporal understanding, content editing, and multimodal planning show the lowest scores, with MIO and NExT-GPT both below 10\%. 
These results indicate that current MLLMs can maintain semantic alignment when input semantics are concrete but struggle to preserve modal precision and temporal synchronization in tasks with compositional or sequential dependencies, leading to failures in maintaining structural integrity, semantic correctness, and coherence during generation.

\begin{figure*}[t!]
\centering
\includegraphics[width=\textwidth]{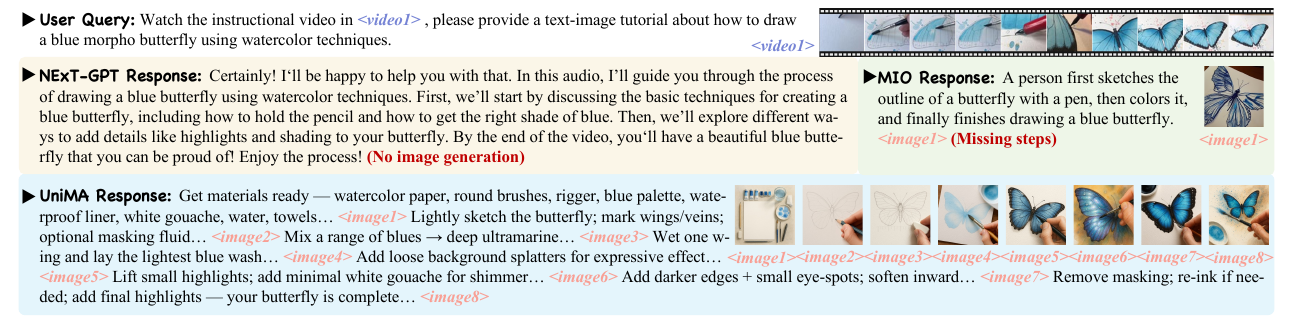}
\vspace{-5mm}
\caption{Comparison of model responses in the cross-modal painting instruction task.
NExT-GPT~\cite{wu2024next} outputs only generic text without image generation, lacking procedural structure.
MIO~\cite{wang2024mio} produces a single sketch image but omits key intermediate steps.
\textsc{UniMA} generates a coherent step-by-step text–image workflow aligned with the video, covering the entire painting process.}
\label{fig:casestudy}
\end{figure*}

\begin{figure}[t!]
\vspace{-7mm}
\centering
\includegraphics[width=\linewidth]{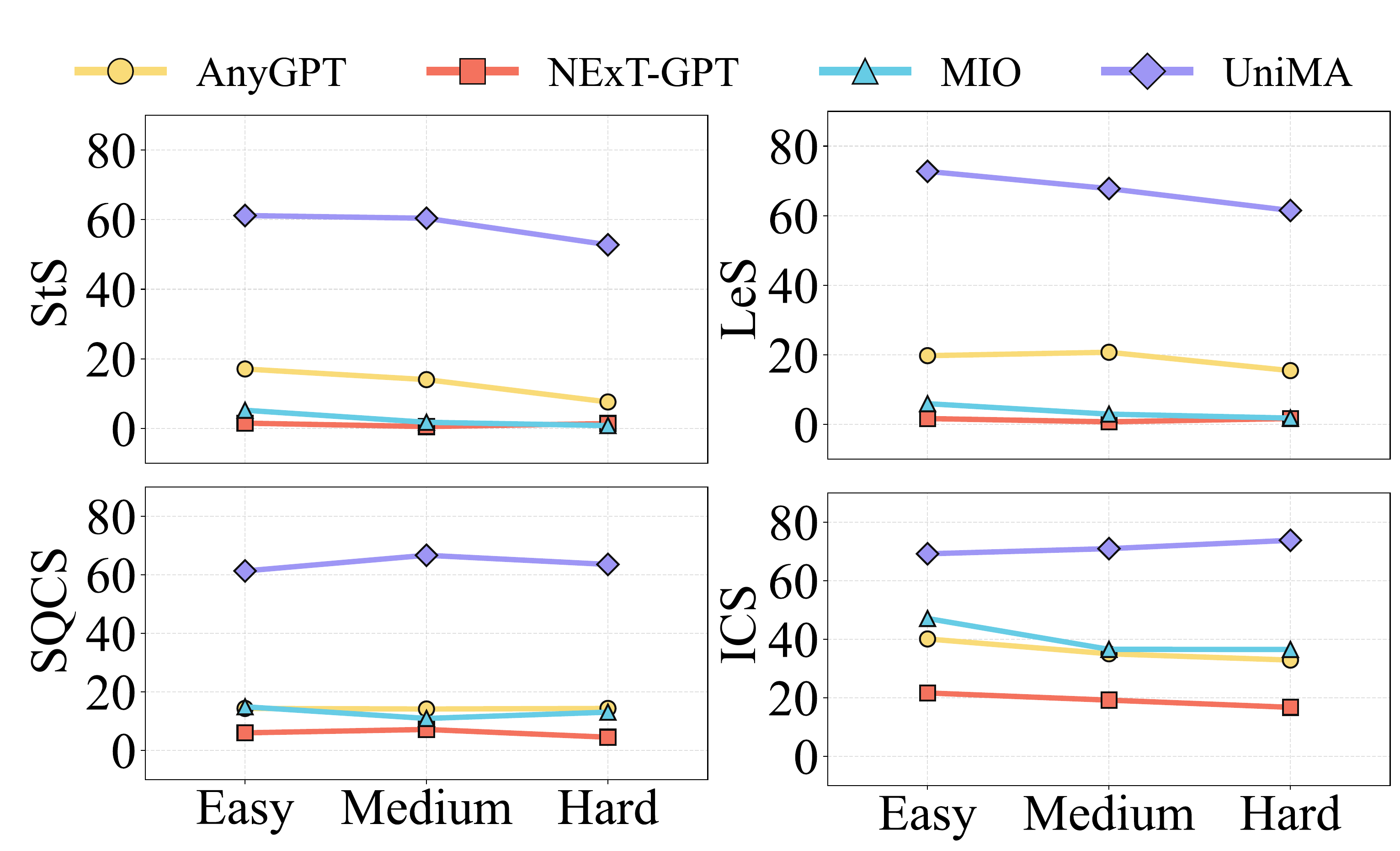}
\vspace{-5mm}
\caption{Performance comparison of MLLMs under different difficulty levels across evaluation metrics. 
Each curve illustrates the performance trend of a model as task complexity increases, where higher values indicate better performance.}
\vspace{-5mm}
\label{fig:diff}
\end{figure}

\vspace{-2mm}
\subsection{Analysis across Varying Levels of Difficulty}
\vspace{-1.5mm}
We analyze evaluated MLLMs performance across instances of varying difficulty and find the following.

\vspace{1mm}
\noindent \textbf{\textit{RQ-1: Do MLLMs exhibit performance variation consistent with task difficulty?}}  
Only \textsc{UniMA} shows a clear performance gradient aligned with increasing difficulty, whereas AnyGPT, NExT-GPT, and MIO consistently fail, even at the easiest level (cf. Fig.~\ref{fig:diff}). 
Notably, while \textsc{UniMA}'s StS and LeS decline as difficulty increases, while its SQCS and ICS remain stable, suggesting robustness in semantic reasoning but limitations in high-order modality interleaving. 

\vspace{1mm}
\noindent \textbf{\textit{RQ-2: Why do baselines perform equally poorly across all difficulty levels?}}
Most current MLLMs are insensitive to task difficulty because even easy interleaved tasks already exceed their compositional and alignment capacities. 
AnyGPT, NExT-GPT, and MIO show uniformly low scores (mostly below 20\%), regardless of difficulty, indicating that even the simplest interleaved tasks in \textsc{UniM} surpass their multimodal compositional and alignment capacities. Therefore, these models fail to distinguish task complexity.

\vspace{-1mm}
\subsection{Ablation Study on \textsc{UniMA}}
\vspace{-0.5mm}
\label{sec:ablation_unima}

Following the previous results where \textsc{UniMA} consistently outperforms baselines, we further conduct a controlled ablation to quantify each component’s contribution. We compare the full model with three variants: removing the TER module, replacing TCDC with a vanilla dense caption, and removing only the verification submodule. As shown in Table~\ref{tab:ablation}, removing TER causes the largest StS/LeS drops, confirming its central role in structural adherence and instruction-following. Replacing TCDC leads to moderate SQCS and ICS declines, indicating that task-conditioned evidence mainly enhances semantic grounding and cross-modal coherence. Removing the verification submodule produces notable degradation across all metrics, demonstrating that checking–backtracking–regeneration design is critical for reliable interleaved outputs. 
More model ablation details are provided in Appendix~\S\ref{app: ablation study}.

\begin{table}[t!]
\centering
\vspace{-4.5mm}
\caption{Ablation results on \textsc{UniMA}.}
\vspace{-2mm}
\fontsize{7.5}{8.5}\selectfont
\setlength{\tabcolsep}{1.2mm}
\begin{tabular}{lcccc}
\toprule
& \bf SQCS & \bf ICS & \bf StS & \bf LeS \\
\midrule
\rowcolor{abl_head_fill}
\textbf{\textsc{UniMA}} & \textbf{85.1} & \textbf{63.4} & \textbf{52.7} & \textbf{82.6} \\
\quad w/o \textbf{TER} & \makecell{72.9  \textcolor{standard_red}{(-12.2)}} & \makecell{56.6  \textcolor{standard_red}{(-6.8)}} & \makecell{16.4 \textcolor{standard_red}{(-36.3)}} & \makecell{21.8  \textcolor{standard_red}{(-60.8)}} \\
\quad w/o \textbf{TCDC} & \makecell{78.4 \textcolor{standard_red}{(-6.7)}} & \makecell{57.7  \textcolor{standard_red}{(-5.7)}} & \makecell{46.2 \textcolor{standard_red}{(-6.5)}} & \makecell{82.1  \textcolor{standard_red}{(-0.5)}} \\
\quad w/o \textbf{Ver.} & \makecell{72.9  \textcolor{standard_red}{(-12.2)}} & \makecell{54.7  \textcolor{standard_red}{(-8.7)}} & \makecell{38.3  \textcolor{standard_red}{(-14.4)}} & \makecell{66.8  \textcolor{standard_red}{(-15.8)}} \\
\bottomrule
\end{tabular}
\vspace{-4mm}
\label{tab:ablation}
\end{table}

\vspace{-1mm}
\subsection{Qualitative Case Study}
\vspace{-0.5mm}

Finally, to better illustrate the performance gap between existing baselines and our \textsc{UniMA}, we present a visualization of case study in Fig.~\ref{fig:casestudy}, comparing the outputs of \textsc{UniMA} with representative models (MIO, NExT-GPT). 
Compared with existing models that struggle to achieve synchronization between text and image modalities, \textsc{UniMA} constructs a complete text–image instructional workflow.
Its generated results maintain consistency in temporal order, modality coverage and stylistic coordination, enabling the model to produce semantically grounded and structurally clear instructional content, fully demonstrating \textsc{UniMA}'s advances in multimodal reasoning and coordinated generation.

\vspace{-2mm}
\section{Conclusion}
\label{conclusion}
\vspace{-1mm}

This paper presents \textbf{\textsc{UniM}}, for the first time benchmarking the \textit{Unified Any-to-Any Interleaved Multimodal Learning}.
\textsc{UniM} provides a large-scale, high-quality dataset covering 7 modalities and 30 real-world domains, involving complex task-solving capabilities, together with a principled evaluation suite for assessing comprehension and generation in complex interleaved scenarios.
Furthermore, an agentic baseline model \textbf{\textsc{UniMA}} is deliberately designed to better benchmark the task.
Our experiments show that existing MLLMs still face significant challenges under this unified any-to-any Interleaved paradigm, revealing key limitations and opportunities for future progress.

\section*{Acknowledgement}
This work is supported by the Ministry of Education, Singapore, under its MOE AcRF TIER 3 Grant (MOE-MOET32022-0001).

{
\small
\bibliographystyle{cvpr2026}
\bibliography{cvpr2026}
}

\clearpage

\etoctoccontentsline{part}{Supplementary Material}

\onecolumn

\begin{center}
    {\Large \textbf{\thetitle}} \\[0.5em]
    {\Large Supplementary Material} \\[0.3em]
\end{center}

\appendix

\renewcommand{\cftsecfont}{\bfseries \small}
\renewcommand{\cftsecpagefont}{\bfseries \small}
\renewcommand{\cftsubsecfont}{\small}
\renewcommand{\cftsubsecpagefont}{\small}
\renewcommand{\cftsubsubsecfont}{\small}
\renewcommand{\cftsubsubsecpagefont}{\small}

\setlength{\cftbeforesecskip}{0.6mm}
\cftsetindents{section}{0mm}{6mm}
\cftsetindents{subsection}{8mm}{8mm}
\cftsetindents{subsubsection}{16mm}{10mm}

\hypersetup{linkbordercolor=black,linkcolor=black}
{
\renewcommand{\contentsname}{}
\vspace{-8mm}
\localtableofcontents
}
\hypersetup{linkbordercolor=red,linkcolor=cvprblue}

\twocolumn

\newpage

\section{Potential Limitations and Future Work}
\label{app:limitation_future_work}
\subsection{Potential Limitations}
In this paper, we introduce the first large-scale unified any-to-any interleaved multimodal benchmark \textsc{UniM} and propose a unified any-to-any interleaved multimodal agentic model \textsc{UniMA} as baseline. Although this work represents a pioneering effort that significantly advances the development of interleaved multimodal learning, we acknowledge that it still has certain limitations.

\noindent \textbf{Limited Modality Combinations and Types of \textsc{UniM}.}
Although \textsc{UniM} is the first unified any-to-any interleaved multimodal benchmark that encompasses 7 modalities: text, image, audio, video, document, code, and 3D, which represents an important expansion and pioneering step beyond previous interleaved benchmarks that primarily focused on image-text scenarios, we acknowledge certain constraints. 
First, human perception of the world extends beyond these 7 modalities. 
Second, the current benchmark includes only a restricted set of modality combinations, while real-world interleaved multimodal scenarios may involve far more complex and interleaved configurations.

\noindent \textbf{Dependence of \textsc{UniMA}'s Performance on External Tools.}
The performance of \textsc{UniMA} largely depends on the multimodal understanding and generation capabilities of the external tools it invokes. 
Although we employ state-of-the-art tools, their inherent limitations may still influence the overall performance of \textsc{UniMA}.

\noindent \textbf{System Complexity and Computational Overhead of \textsc{UniMA}.}
To enable fine-grained comprehension, reasoning and generation, \textsc{UniMA} invokes up to six external tools, introducing substantial complexity and computational overhead into the system. Compared with end-to-end models, this multi-step and sequential reasoning process leads to higher system complexity.

\subsection{What To Do Next with \textbf{\textsc{UniM}} and \textbf{\textsc{UniMA}}}
Building on this pioneering work, we believe that future research on \textbf{\textit{interleaved multimodal learning}} can be further advanced along several key directions.

\noindent \textbf{From Modular to End-to-End Interleaved Multimodal Foundation Models.}
An important future direction is the development of fully end-to-end foundation models.
Current end-to-end MLLMs do not fully support flexible any-to-any multimodal interleaving, as they are unable to process arbitrary modality combinations or multiple items of the same modality. 
Therefore, an important future direction is to develop a unified encoder–decoder architecture supporting arbitrary combinations of seven modalities.

\noindent \textbf{Improving Multi-Capability Coordination in Any-to-Any MLLMs.} 
Although current any-to-any MLLMs have demonstrated certain effectiveness in several basic capabilities, their performance remains suboptimal in multi-capability coordination, particularly in complex tasks. 
Therefore, enhancing the model’s performance to integrate and coordinate multiple capabilities emerges as a research direction worthy of in-depth exploration.

\noindent \textbf{Cross-Modal Synergy and Complementarity.}
Future work may further investigate the synergistic and complementary interactions among modalities in interleaved multimodal scenarios. 
Instead of treating each modality as an independent source of information, models may benefit from explicitly modeling their dynamic interplay across tasks, contexts, and heterogeneous signal-quality conditions. 
Achieving stable and synergy-based multimodal integration is expected to clearly improve semantic consistency, generation quality, and generalization in complex interleaved multimodal settings.

\noindent \textbf{Dynamic Reasoning Mechanisms and Contextual Adaptation.}
Future work may further investigate dynamic reasoning and contextual adaptability in interleaved multimodal settings. 
Rather than assuming fixed modality contributions, models should adjust the influence of modality features based on task context, focusing on the most informative modalities at different stages. 
This adaptive strategy can enhance robustness and enable more context-aware cross-modal reasoning.

\noindent \textbf{Interleaved Rewarding.} Future work may further explore an interleaved reward mechanism that explicitly incorporates both semantic accuracy and the handling of interleaved multimodal structures into the optimization objective, thereby improving consistency and robustness in complex modality composition scenarios.

\noindent \textbf{Self-Verification and Iterative Refinement under Interleaved Settings.} Future work may incorporate self-verification and iterative refinement mechanisms, enabling models to assess and locally revise generation outputs to better handle errors arising from one-shot processing in complex interleaved multimodal tasks.

\noindent \textbf{Cognitive-Style Interleaved Modeling.} Future research may further investigate cognitive-style interleaved modeling, where models emulate human multimodal reasoning strategies by identifying primary perceptual modalities, establishing reasoning order, and leveraging complementary modalities for validation or refinement. 
This cognitive-inspired approach is expected to enhance interpretability and robustness under complex interleaved multimodal scenarios.

\begin{figure*}
\centering
\includegraphics[width=0.99\linewidth]{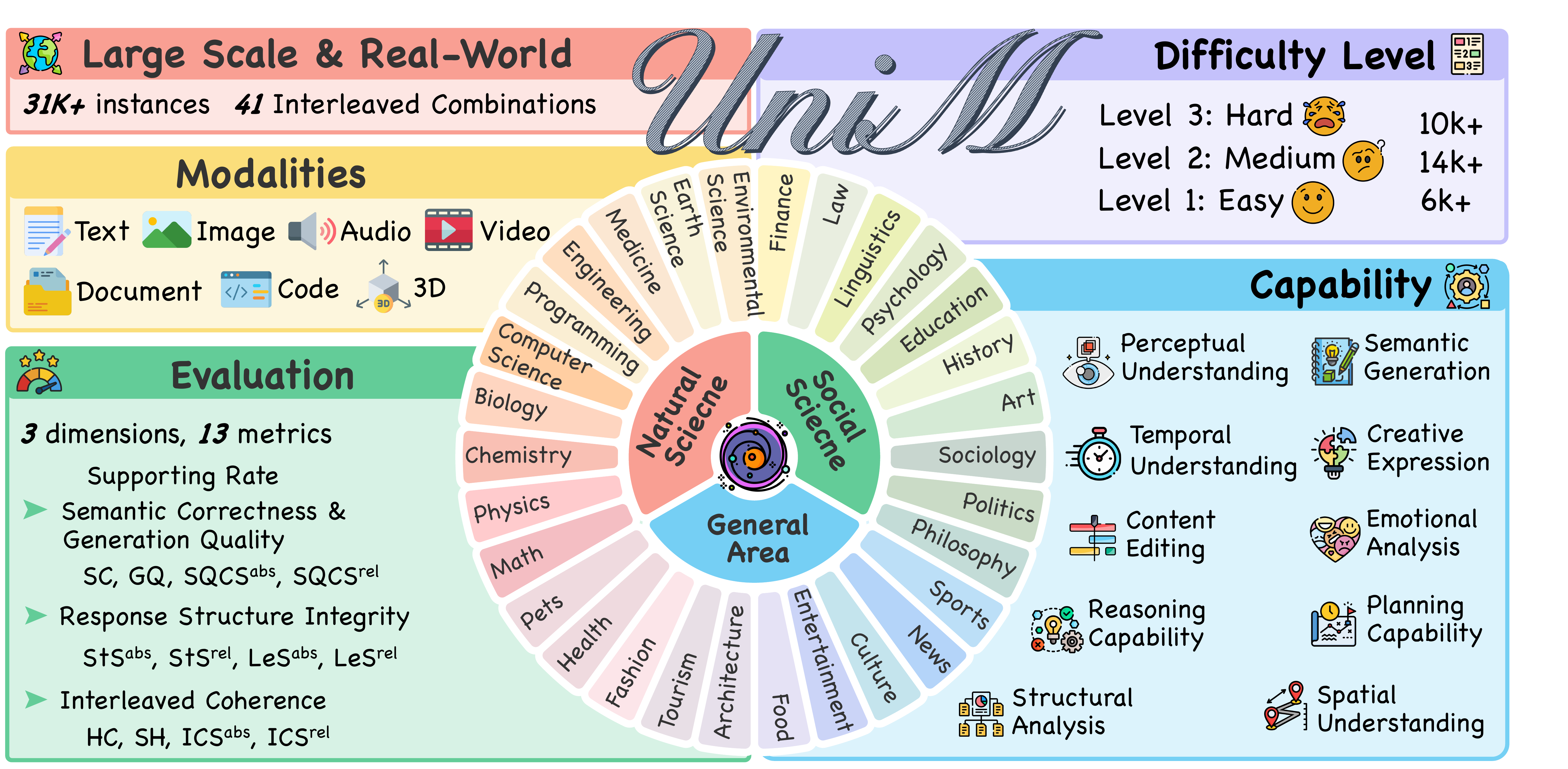}
\vspace{-2mm}
\caption{Overview of \textsc{UniM} benchmark.}
\label{fig:teaser}
\end{figure*}

\section{Task Definition}
\label{app:task definition}
The proposed task requires the model to perform comprehension and generation under the \textit{any-to-any modalities interleaved} format. 
Specifically, the input is an interleaved sequence $S_{\text{in}}$, and models aim to generate a valid interleaved output sequence $S_{\text{out}}$. 
Both $S_{\text{in}}$ and $S_{\text{out}}$ contain modality placeholder tags embedded in natural language, which denote non-text modalities.

Let $\mathcal{M}$ denote the set of modalities. 
The input sequence is defined as 
\begin{equation}
    S_{\text{in}} = [x_1, x_2, \dots, x_{L_{\text{in}}}].
\end{equation}
Each sequence element $x_j$ satisfies
\begin{equation}
x_j \in \mathcal{V} \cup \{ \langle m_k^{\text{in}} \rangle \mid m_k \in \mathcal{M} \setminus \{\text{text}\} \},
\end{equation}
where $\mathcal{V}$ denotes natural language tokens, $m_k$ represents a specific modality type, and $\langle m_k^{\text{in}} \rangle$ indicates a modality placeholder tag appearing in $S_{\text{in}}$.

For the model $\mathcal{F}$, the objective is to sequentially produce the output sequence $S_{\text{out}}$ based on the given input sequence $S_{\text{in}}$. 
Each output token $y_j$ is generated as 
\begin{equation}
y_j \sim \mathcal{F}_{\theta}(\cdot \mid S_{\text{in}}, y_1, \dots, y_{j-1}), \quad j = 1, 2, \dots, L_{\text{out}}.
\end{equation}

Similarly, the format of $y_j$ should satisfy 
\begin{equation}
y_j \in \mathcal{V} \cup \{ \langle m_k^{\text{out}} \rangle \mid m_k \in \mathcal{M} \setminus \{\text{text}\} \},
\end{equation}
where $\mathcal{V}$ denotes natural language tokens, and $\langle m_k^{\text{out}} \rangle$ represents a placeholder tag of the modality that appears in the output sequence.
Therefore, the output sequence can be expressed as 
\begin{equation}
S_{\text{out}} = [y_1, y_2, \dots, y_{L_{\text{out}}}],
\end{equation}

To ensure format consistency and the feasibility of evaluation, we impose the following constraints on the placeholders in both the input and output sequences.

\noindent \textbf{Constraint 1 (Directionality).}
Placeholders appearing in the input sequence should only occur in the input text, 
and those in the output sequence should only occur in the output text:
\begin{equation}
    \begin{aligned}
        \forall\, m_k \in \mathcal{M} \setminus \{\text{text}\}: \quad
        &\langle m_k^{\text{in}} \rangle \in S_{\text{in}} \Rightarrow \langle m_k^{\text{in}} \rangle \notin S_{\text{out}}, \\
        &\langle m_k^{\text{out}} \rangle \in S_{\text{out}} \Rightarrow \langle m_k^{\text{out}} \rangle \notin S_{\text{in}}.
    \end{aligned}
\end{equation}

\noindent \textbf{Constraint 2 (Uniqueness).}
Each placeholder appears exactly once in its corresponding sequence:
\begin{equation}
    \begin{aligned}
        \forall\, \langle m_k^{\text{in}} \rangle \in S_{\text{in}}, \quad 
        & |\langle m_k^{\text{in}} \rangle| = 1, \\
        \forall\, \langle m_k^{\text{out}} \rangle \in S_{\text{out}}, \quad 
        & |\langle m_k^{\text{out}} \rangle|=1.
    \end{aligned}
\end{equation}

\section{Details of \textsc{UniM}}

In this section, we provide an exposition of \textsc{UniM}, with Fig.~\ref{fig:teaser} serving as a comprehensive overview.

\subsection{Construction Process}
\label{app: construction process}
\begin{figure*}[!t]
    \centering
    \includegraphics[width=0.99\linewidth]{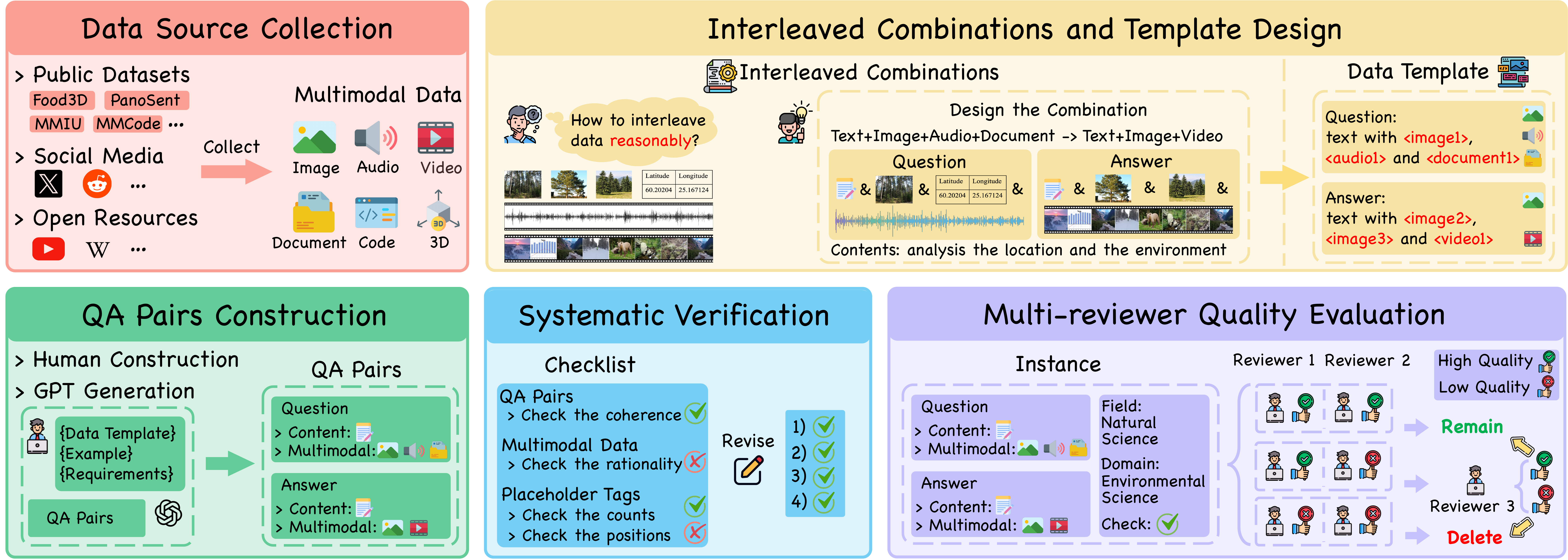}
    \vspace{-2mm}
    \caption{Overview of dataset construction.}
    \label{overview}
    \vspace{-2mm}
\end{figure*}

In this section, we provide a detailed description of its construction process, quality control protocol, and statistical characteristics.
Fig.~\ref{overview} provides a high-level overview of data construction and annotation workflow.

\subsubsection{Data Sources Collection}

For multimodal data collection, we rely mainly on three sources. 
First, we curate samples from existing high-quality public datasets. 
Second, we obtain publicly accessible real-world multimodal content from famous social media. 
Third, we supplement the multimedia materials with open resources.

\begin{table}[t!]
\centering
\caption{Public dataset sources.}
\fontsize{9}{11}\selectfont
\setlength{\tabcolsep}{1.2mm}
\vspace{-2mm}
\begin{tabular}{ll}
\toprule
\bf Modality & \textbf{Public Dataset} \\
\midrule
Image & \makecell[l]{FoodieQA~\cite{li2024foodieqa}, MuSLR~\cite{xu2025muslr}, TaxaBench-8k~\cite{sastry2025taxabind}, \\ MSEarth~\cite{zhao2025msearth}, LAB-Bench~\cite{laurent2024lab}, Chempile~\cite{mirza2025chempile}\\ OmniCorpus~\cite{li2024omnicorpus}, OmniEarth-Bench~\cite{wang2025omniearth}, \\ MMRB~\cite{cheng2025evaluating}, StateControl-Bench~\cite{wu2025see} \\ iMaterialist~\cite{guo2019imaterialist}, DeepFashion~\cite{liu2016deepfashion}}\\
\hdashline  
Audio & Panosent~\cite{luo2024panosent}, CMI-Bench~\cite{ma2025cmi}, ESC~\cite{piczak2015dataset} \\
\hdashline  
Video & \makecell[l]{Video-MME~\cite{fu2025video}, MuMa-ToM~\cite{shi2025muma}, \\ MovieChat~\cite{song2024moviechat}, Social-IQ~\cite{zadeh2019social}, PhysLab~\cite{price2019physlab}, \\ EgoTextVQA~\cite{zhou2025egotextvqa}, HowTo100M~\cite{miech2019howto100m}}\\
\hdashline  
Document & \makecell[l]{MedTrinity-25M~\cite{xie2024medtrinity}, CaseSumm~\cite{heddaya2025casesumm},\\ FDABench~\cite{wang2025fdabench}, Synfintabs~\cite{bradley2024synfintabs}}\\
\hdashline  
Code & \makecell[l]{MMCode~\cite{li2024mmcode}, ScreenSpot-Pro~\cite{li2025screenspot}, \\ CodeEditorBench~\cite{guo2024codeeditorbench}} \\
\hdashline  
3D & MetaFood3D~\cite{chen2024metafood3d}, General-Bench~\cite{fei2025path}\\
\bottomrule
\end{tabular}
\label{tab:dataset resources}
\end{table}

\noindent \textbf{Public Datasets.} We first obtain high-quality multimodal resources from existing public multimodal datasets and perform systematic filtering and structured organization. 
These datasets typically offer well-defined annotation schemes and stable data quality, providing reliable and diverse raw materials for constructing interleaved multimodal instances. 
By integrating various task types and modality forms, we ensure sufficient diversity and modality coverage in the foundational data. 
The specific data sources we use are listed in Table~\ref{tab:dataset resources}.

\noindent \textbf{Social Media.} Second, we harvest large quantities of real-world multimodal data from publicly accessible social media platforms (e.g., Twitter, Reddit), including posts and vlogs. 
Data on these platforms are highly dynamic, open-ended, and span various media types such as text, images, audio, and video, making them more representative of real-world scenarios. 
Under strict adherence to platform access and usage policies, we filter and preliminarily organize the collected multimodal data items, enabling them to serve as an important source for constructing interleaved multimodal instances.

\noindent \textbf{Open Resources.} Finally, we draw on a variety of highly accessible open resources to complement modality types that are relatively underrepresented in the previous sources, thereby strengthening the overall multimodal coverage. 
We acquire modality-specific materials from platforms that host particular types of data, for example, long-form videos from YouTube, document-hosting websites, code repositories, and libraries containing 3D models. 
After basic filtering and structured organization, these multimodal data items serve as an important supplement for constructing interleaved multimodal instances, substantially enhancing both modality richness and scenario diversity within the dataset.

\begin{table}[t!]
\centering
\caption{Multimodal files formats.}
\fontsize{9}{11}\selectfont
\setlength{\tabcolsep}{1.6mm}
\vspace{-2mm}
\begin{tabular}{ll}
\toprule
\bf Modality & \textbf{File Format} \\
\midrule
Image & PNG, JPG, JPEG\\
Audio & WAV, MP3\\
Video & MP4, WEBM\\
Document & PDF, DOC, CSV, TXT, PNG, JPG\\
Code & Markdown Block\\
3D & JPG+MTL+OBJ, OFF, PLY \\
\bottomrule
\end{tabular}
\label{tab:data file format}
\vspace{-3mm}
\end{table}

Table~\ref{tab:data file format} summarizes the data file formats associated with each modality in \textsc{UniM}.

\subsubsection{Interleaved Combinations and Template Design}

After obtaining the multimodal files, we begin by manually analyzing the characteristics of each modality and designing appropriate \textbf{interleaved combinations} based on their semantic relatedness, information density, and interaction patterns. 
Specifically, while preserving the semantic information of each original modality, we reorganize images, audio, video, documents, code, and 3D into coherent interleaved structures that can jointly express information within a single question–answer pair. 
The positions, quantities, and interleaving orders of different modalities in both input and output are semantically driven, ensuring that the combinations are diverse yet aligned with realistic interleaved multimodal comprehension and generation requirements.

Following this principle, we construct \textbf{instance templates} in which multimodal elements are inserted into text using placeholder tags such as `$<$image1$>$', `$<$audio2$>$', `$<$code3$>$', and `$<$3d4$>$'. For the multimodal files collected in \textsc{UniM}, we design a total of 41 interleaved combinations. The interleaved combinations are presented in Table~\ref{tab:interleaved combinations}.

\begin{table*}[t!]
\centering
\caption{Interleaved combinations in \textsc{UniM}. I: Image, V: Video, A: Audio, D: Document, C: Code.}
\label{tab:interleaved combinations}
\vspace{-2mm}
\fontsize{10}{14}\selectfont
\setlength{\tabcolsep}{1.2mm}

\begin{tabular}{l*{14}{c}}
\toprule
\textbf{Input}  &
D & A+D & I & I+A & V+A & V+A & I+V+A & I+D & D & A+I & A & I+A & A & D \\
\textbf{Output} &
I & A & I & I+A & I   & A   & V+A    & A   & D & A+V & I & D   & A & A \\
\hdashline
\textbf{Input}  &
V & I & I+A & I+D & I & 3D & A & V+I & A & I+A & V & A & V+A & V+A \\
\textbf{Output} &
I & A & I   & I   & D & I+D & 3D & V   & D & A   & A & V & V+A & D \\
\hdashline
\textbf{Input}  &
C & I & D & I+A+D & 3D & I+V+A & I+A+D & D+A & I+A & D & I & A+V+D & {} & {} \\
\textbf{Output} &
C & C & C & C     & 3D & A      & I      & I    & V   & I+A & V+A & A & {} & {} \\
\bottomrule
\end{tabular}
\vspace{-3mm}
\end{table*}

\subsubsection{QA Pairs Construction}

In the QA-pair construction stage, we adopt a collaborative strategy of ``human-led authoring with model-assisted expansion" to ensure both semantic quality and scalable data generation. 

First, human annotators manually compose five high-quality representative QA pairs for each template designed in the previous stage, taking into account the characteristics of different task types. 
These manually created examples clarify task intent and provide guidance for subsequent expansion, ensuring coherent reasoning, self-consistent semantics, and natural cross-modal relationships.

\begin{figure*}
\centering
\includegraphics[width=0.99\linewidth]{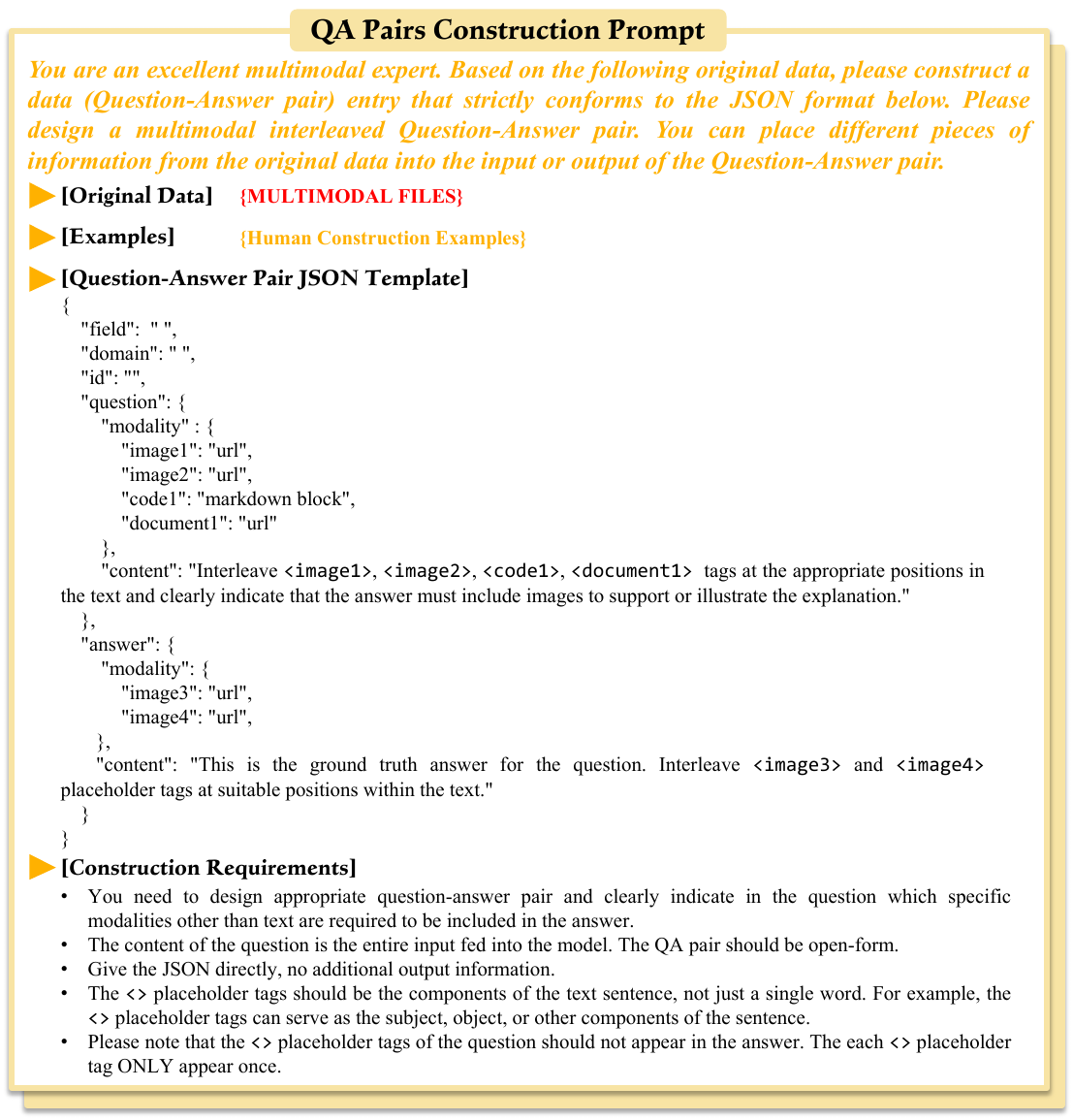}
\caption{QA pairs construction prompt example.}
\label{fig:construction prompt}
\end{figure*}

Further, we leverage GPT-5 mini~\cite{OpenAI_Introducing_GPT5_2025} to expand these human-designed examples. Annotators provide the model with the templates, exemplars, and explicit construction requirements, enabling it to generate additional candidate QA pairs under controlled structural constraints. The full prompt used is shown in Fig.~\ref{fig:construction prompt}.

\subsection{Quality Control}
\label{app: quality control}
To ensure the reliability and high quality of the dataset, we design and conduct a two-stage quality control procedure consisting of a systematic validation and multi-reviewer evaluation.

\subsubsection{Systematic Verification}

\begin{sidewaysfigure*}
\centering
\includegraphics[width=0.99\textheight]{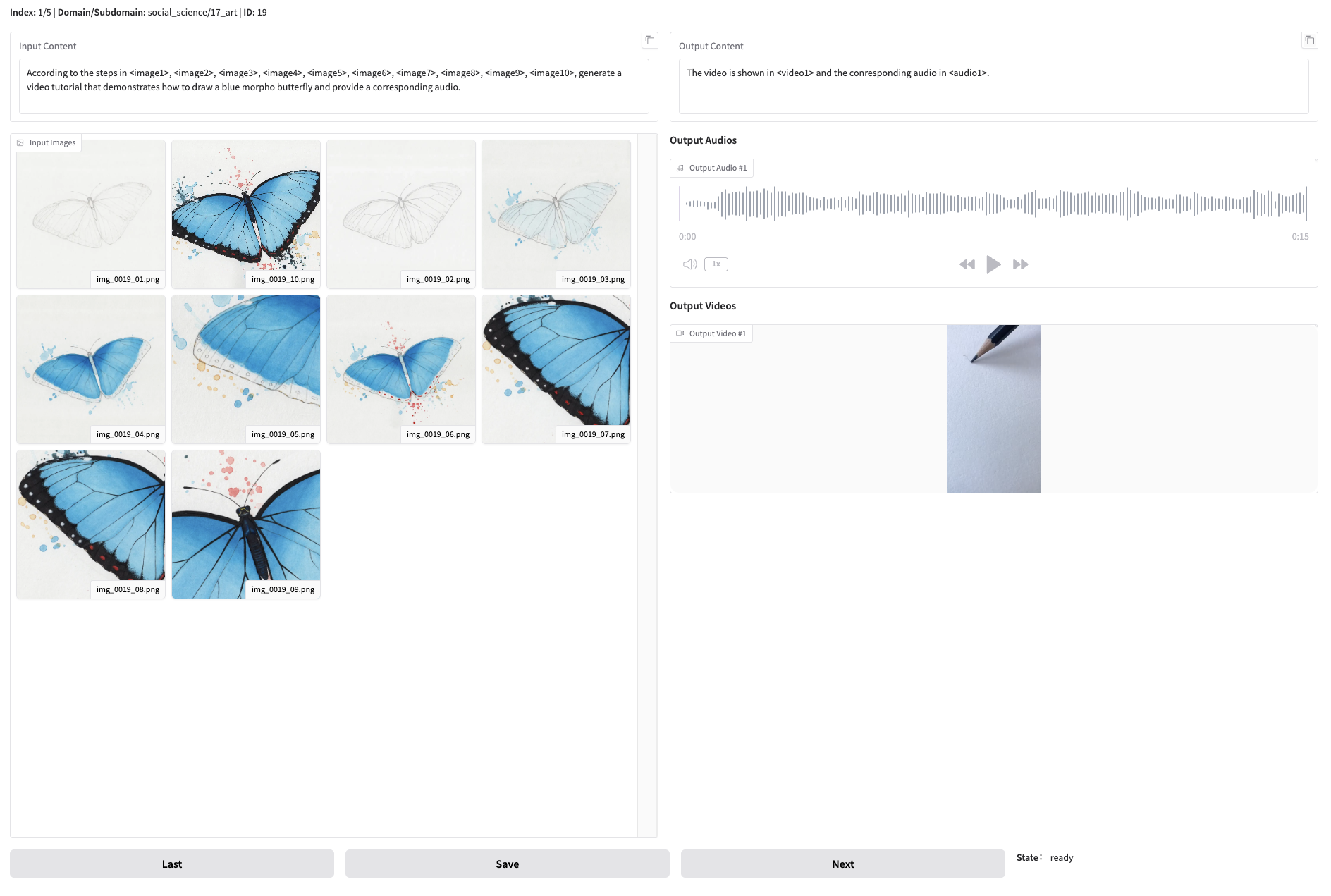}
\caption{The screenshot of systematic verification tool we developed.}
\label{fig:systematic verification}
\end{sidewaysfigure*}

First, each constructed instance undergoes a systematic verification process conducted by human reviewers. 
Reviewers examine, item by item, the semantic coherence of the QA content, the correctness of the alignment between multimodal files and textual descriptions, and the accuracy of both the number and positions of placeholder tags. 
If inconsistencies or unreasonable elements are identified, reviewers are required to revise or rewrite the sample to ensure structural clarity, logical completeness, and correct modality representation. We develop an internal tool to support this reviewing and revising process, as illustrated in Fig.~\ref{fig:systematic verification}.

\subsubsection{Multi-reviewer Quality Evaluation}

\begin{sidewaysfigure*}
\centering
\includegraphics[width=0.99\textheight]{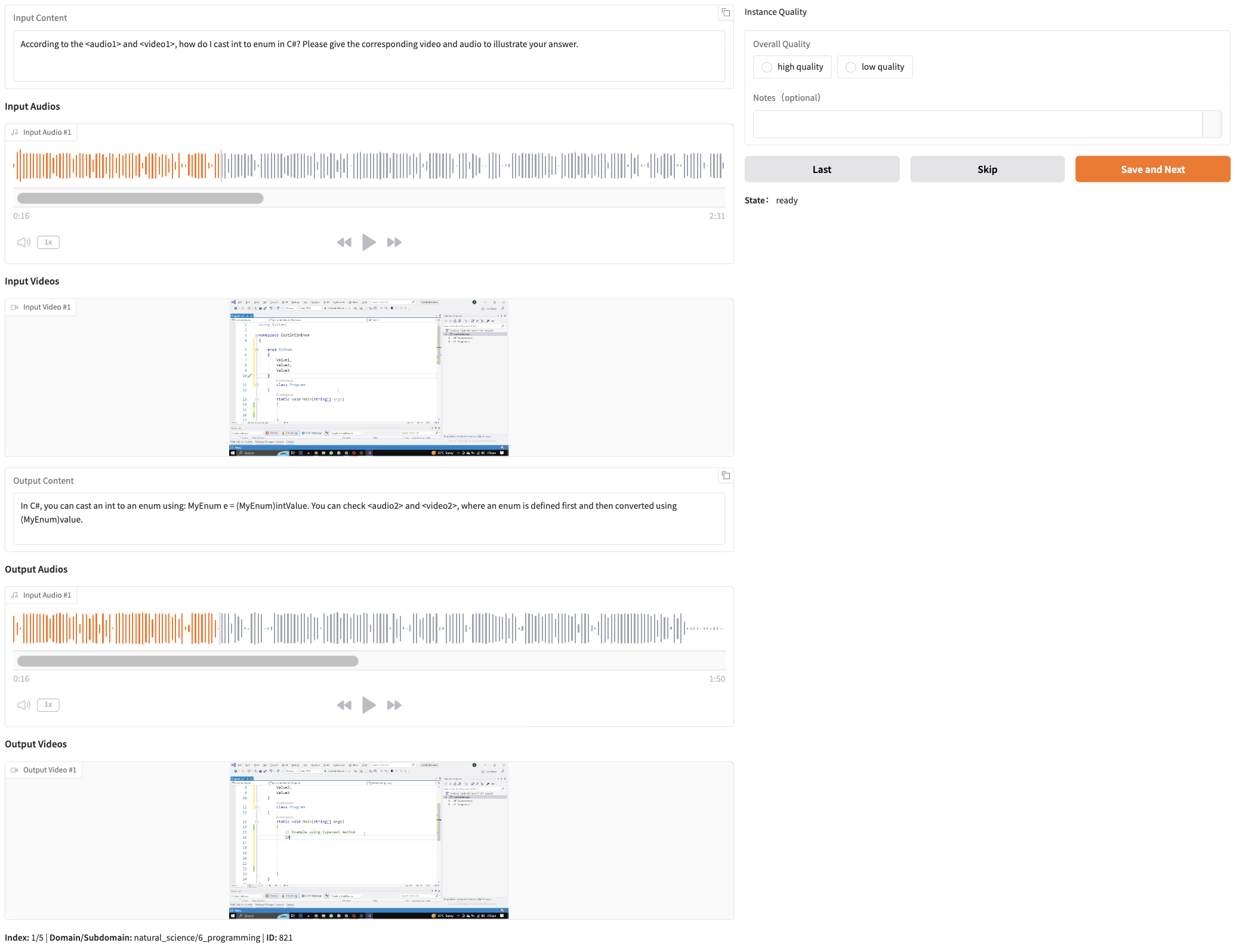}
\caption{The screenshot of multi-reviewer quality evaluation tool we developed.}
\label{fig:multi-reviewer evaluation}
\end{sidewaysfigure*}

After the initial revisions, all instances proceed to an independent multi-reviewer evaluation stage. 
Each instance is assigned to two reviewers who independently assess its quality, examining whether multimodal information is accurate, whether the task satisfies the template requirements, whether the reasoning is rigorous, and whether any ambiguities or missing content are present. 
Instances that both reviewers rate as high quality are retained directly, whereas those rated as low quality by both are removed. 
For instances with conflicting assessments, a third reviewer is introduced to provide an adjudication, ensuring the reliability and consistency of the final decision. We develop an internal tool to support this process, as illustrated in Fig.~\ref{fig:multi-reviewer evaluation}.

\subsection{Progressive Difficulty Taxonomy}
\label{app: progressive difficulty taxonomy}
To systematically characterize the complexity of \textsc{UniM} and quantitatively assess model performance on it, we establish a progressive difficulty taxonomy, which categorizes dataset instances into three hierarchical levels: Hard, Medium, and Easy.

\vspace{1mm}
\noindent \textbf{Detailed Definitions of Different Difficulty Levels.}
\vspace{0.5mm}

\begin{figure}[t!]
\centering
\begin{tcolorbox}[colback=gray!10, colframe=diff_gray, boxrule=0.8pt, arc=3pt, left=3mm, right=3mm, width=\linewidth, title={Level 3: Hard}]
\textbf{Comprehension: }Inputs contain highly complex modalities (e.g., 3D, code, video) or involve $\geq$5 intricate interleavings across modalities.\\
\textbf{Generation: }Outputs involve complex structured generation (e.g., 3D point cloud completion, video editing, code translation) or $\geq$4 interleavings across modalities.\\
\textbf{Reasoning: }Involves multi-step reasoning, deep logical analysis, or reliance on highly specialized domain knowledge.\\
\textbf{Task: }Task objectives are complex or open-ended, possibly comprising multiple interdependent subtasks.
\end{tcolorbox} 
\vspace{-2mm}
\caption{Definition of hard level.}
\label{colbox:easy definition}
\vspace{-2mm}
\end{figure}

\begin{figure}[t!]
\centering
\begin{tcolorbox}[colback=gray!10, colframe=diff_gray, boxrule=0.8pt, arc=3pt, left=3mm, right=3mm, width=\linewidth, title={Level 2: Medium}]
\textbf{Comprehension: }Inputs involve moderately complex modalities (e.g., documents, audio) or typically exhibit between three and four interleavings across modalities.\\
\textbf{Generation: }Outputs involve localized editing or limited structured generation (e.g., image editing, object detection) or typically contain between two and three interleavings across modalities.\\
\textbf{Reasoning: }Requires one to two steps of cross-modal or cross-instance reasoning, possibly involving commonsense judgment.\\
\textbf{Task: }Task objectives consist of multiple subtasks that are generally independent of one another.
\end{tcolorbox} 
\vspace{-2mm}
\caption{Definition of medium level.}
\label{colbox:medium definition}
\end{figure}

\begin{figure}[t!]
\centering
\begin{tcolorbox}[colback=gray!10, colframe=diff_gray, boxrule=0.8pt, arc=3pt, left=3mm, right=3mm, width=\linewidth, title={Level 1: Easy}]
\textbf{Comprehension: }Inputs include relatively simple modalities (e.g., text, image) or typically contain no more than two simple interleavings across modalities.\\
\textbf{Generation: }Outputs consist of straightforward, direct generation without any editing or structural modification.\\
\textbf{Reasoning: }Involves only single-step perception and alignment, without cross-modal reasoning or reliance on commonsense or external knowledge.\\
\textbf{Task: }Task objectives are singular, with short processes and no additional constraints.
\end{tcolorbox}
\vspace{-2mm}
\caption{Definition of easy level.}
\label{colbox:hard definition}
\vspace{-4mm}
\end{figure}

\noindent We begin by manually defining the difficulty criteria across four core dimensions: comprehension, generation, reasoning, and task. These four dimensions jointly describe the overall difficulty of a multimodal instance. The detailed definitions and decision criteria for each difficulty level are provided in Fig.~\ref{colbox:easy definition}, Fig.~\ref{colbox:medium definition} and Fig.~\ref{colbox:hard definition}.

\vspace{1mm}
\noindent \textbf{Top-down Classification Process from Hard to Easy.}
\vspace{0.5mm}
\noindent We employ a top-down process for difficulty classification. 
Beginning from the hard level and proceeding downward, each instance is evaluated dimension by dimension across the four criteria: comprehension, generation, reasoning, and task. An instance is classified to a given difficulty level if it satisfies at least two of the four criteria for that level. This process ensures hierarchical consistency in the classification standards while enhancing the precision and reproducibility of the difficulty categorization.

\subsection{Capabilities to Evaluate}
\label{app: capabilities to be evaluated}

\begin{table*}[t!]
\centering
\caption{Definitions of capabilities.}
\vspace{-2mm}
\fontsize{10}{12}\selectfont
\setlength{\tabcolsep}{1.5mm}
\begin{tabular}{lll}
\toprule
\bf ID & \bf Capability & \textbf{Definition} \\
\midrule
C1 & Perceptual Understanding & \makecell[l]{Identifying and interpreting content, entities, attributes, and semantic relations\\ from multimodal inputs.} \\
\hdashline
C2 & Spatial Understanding & \makecell[l]{Comprehending spatial relationships, geometric structures, 3D scene character-\\istics, and physical layouts.} \\
\hdashline
C3 & Temporal Understanding & \makecell[l]{Understanding event or action sequences, temporal dynamics, and time-depend-\\ent dependencies.} \\
\hdashline
C4 & Semantic Generation & \makecell[l]{Generating multimodal content that is semantically coherent, contextually appro-\\priate, and stylistically consistent.} \\
\hdashline
C5 & Content Editing & \makecell[l]{Modifying, refining, or transforming existing content, including stylistic adjust-\\ments and targeted edits.} \\
\hdashline
C6 & Creative Expression & Producing content with creativity, expressiveness, and aesthetic value. \\
\hdashline
C7 & Reasoning Capability & \makecell[l]{Performing mathematical, spatial, logical, and causal reasoning under multimodal\\ conditions.} \\
\hdashline
C8 & Emotional Analysis & Recognizing and interpreting emotions, intentions, mental states, and affective cues. \\
\hdashline
C9 & Structural Analysis & \makecell[l]{Understanding structured or symbolically encoded information such as documents,\\tables, and code.} \\
\hdashline
C10 & Planning Capability & \makecell[l]{Decomposing tasks and making dynamic adjustments to ultimately achieve the\\ target in complex scenarios.} \\
\bottomrule
\end{tabular}
\label{tab:capability definition}
\end{table*}

To comprehensively evaluate model performance across different interleaved multimodal scenarios, we conduct a structured analysis of the capabilities required by the task types in \textsc{UniM}, taking into account their semantic demands and cross-modal reasoning patterns. Based on this analysis, we derive ten core capability dimensions. The detailed definitions for each capability are provided in Table~\ref{tab:capability definition}. For every instance, we further annotate a set of soft labels that indicate the competencies a model must possess in order to answer that instance effectively.

\subsection{Task Types}
\label{app: task types}

\begin{table*}[t!]
\centering
\caption{Definitions of task types.}
\vspace{-2mm}
\label{tab:task definitions}
\fontsize{9}{12}\selectfont
\setlength{\tabcolsep}{2mm}
\begin{tabular}{ll}
\toprule
\bf Modality & \bf Task Type \\
\midrule
Text &  \makecell[l]{Text summarization; Text generation; Machine translation; Sentiment analysis; Stance detection;\\ Named entity recognition; Text style transfer; ...} \\
\hdashline
Image & \makecell[l]{Image captioning; Image generation; Semantic segmentation; Image editing; Visual localization;\\ Visual question answering; Image style transfer; ...}\\
\hdashline
Audio & \makecell[l]{Audio captioning; Audio generation; Audio timbre analysis; Audio sentiment analysis; Audio content analysis; ...}\\
\hdashline
Video & \makecell[l]{Video captioning; Video localization; Visual question answering; Video editing; Text-to-video generation;\\ Audio-to-video generation; Viewpoint-transformation video synthesis; ...}\\
\hdashline
Document & \makecell[l]{Document OCR; Document generation; Document description; Document summarization;\\ Document question answering; ...}\\
\hdashline
Code & \makecell[l]{Code generation; Code testing; Code debugging; Code translation; Code understanding; Code question answering;\\ Algorithm implementation; ...}\\
\hdashline
3D & \makecell[l]{3D content understanding; 3D question answering; 3D point cloud completion; 3D object recognition;\\ 3D generation; ...
} \\
\bottomrule
\end{tabular}
\vspace{-2mm}
\end{table*}

To comprehensively characterize the range of tasks covered in \textsc{UniM}, we organize the task types and categorize them by modality. Table~\ref{tab:task definitions} presents the major task types associated with each modality.

\vspace{-1mm}
\subsection{Detailed Statistics}
\vspace{-1.5mm}

\begin{table}[t!]
\centering
\caption{The number of multimodal files in \textsc{UniM}.}
\vspace{-2mm}
\label{tab:modality files}
\fontsize{9}{11}\selectfont
\setlength{\tabcolsep}{2mm}
\begin{tabular}{cccccc}
\toprule
Image & Audio & Video & Document & Code & 3D \\
68,651 & 54,329 & 2,559 & 4,728 & 1,207 & 820 \\ 
\bottomrule
\vspace{-2mm}
\end{tabular}
\vspace{-2mm}
\end{table}

We perform a detailed statistical analysis of \textsc{UniM}.
Table~\ref{tab:modality files} reports the number of each modality files contained in the dataset.
Then, we present the number of instances across the easy, medium and hard levels for each domain, as summarized in Table~\ref{tab:detailed stats}.

\begin{table*}[t!]
\centering
\caption{Detailed statistics of \textsc{UniM}.}
\label{tab:detailed stats}
\vspace{-2mm}
\fontsize{8.5}{10}\selectfont
\setlength{\tabcolsep}{1.1mm}
\begin{tabular}{llcccllcccllccc}
\toprule
\multicolumn{5}{c}{\textbf{Natural Science}} &
\multicolumn{5}{c}{\textbf{Social Science}} &
\multicolumn{5}{c}{\textbf{General Area}} \\
\cmidrule(lr){1-5} \cmidrule(lr){6-10} \cmidrule(lr){11-15}
\bf ID & \bf Name & \bf Easy & \bf Medium & \bf Hard & \bf ID & \bf Name & \bf Easy & \bf Medium & \bf Hard  & \bf ID & \bf Name & \bf Easy & \bf Medium & \bf Hard  \\
\midrule
\bf \#1 & Math & 806 & 777 & 48 & 
\bf \#11 & Finance & 739 & 1694 & 135 & 
\bf \#21 & Sports & 57 & 313 & 486  \\
\bf \#2 & Physics & 40 & 591 & 374 & 
\bf \#12 & Law & 256 & 634 & 121 & 
\bf \#22 & News & 200 & 456 & 350  \\
\bf \#3 & Chemistry & 312 & 617 & 113 & 
\bf \#13 & Linguistics & 570 & 207 & 53 &
\bf \#23 & Culture & 128 & 170 & 100 \\ 
\bf \#4 & Biology & 545	& 333 & 263 & 
\bf \#14 & Psychology & 398 & 300 & 397 & 
\bf \#24 & Entertainment & 301 & 188 & 299 \\
\bf \#5 & Compute Science & 72	& 868 & 218 & 
\bf \#15 & Education & 165 & 463 & 72 & 
\bf \#25 & Food & 256 & 624 & 296  \\
\bf \#6 & Programming & 15 & 64 & 743 & 
\bf \#16 & History & 402 & 196 & 225 & 
\bf \#26 & Architecture & 557 & 601 & 59 \\
\bf \#7 & Engineering & 299 & 310 & 549 &
\bf \#17 & Art & 747 & 1492 & 279 &
\bf \#27 & Tourism & 363 & 617 & 105 \\
\bf \#8 & Medicine & 666 & 462 & 77 & 
\bf \#18 & Sociology & 65 & 879 & 182 &
\bf \#28 & Fashion & 700 & 160 & 85 \\
\bf \#9 & Earth Science & 212 & 155 & 273 & 
\bf \#19 & Politics & 204 & 153	& 56 & 
\bf \#29 & Health & 662 & 348 & 215 \\
\bf \#10 & Env. Science & 562 & 186 & 74 & 
\bf \#20 & Philosophy & 299	& 191 & 0 & 
\bf \#30 & Pets & 80 & 39 & 13 \\
\bottomrule
\end{tabular}
\vspace{-2mm}
\end{table*}

\vspace{-1mm}
\subsection{Data Example}
\begin{figure*}[t!]
  \centering
  \begin{minipage}[b]{0.99\linewidth}
    \centering
    \includegraphics[width=\linewidth]{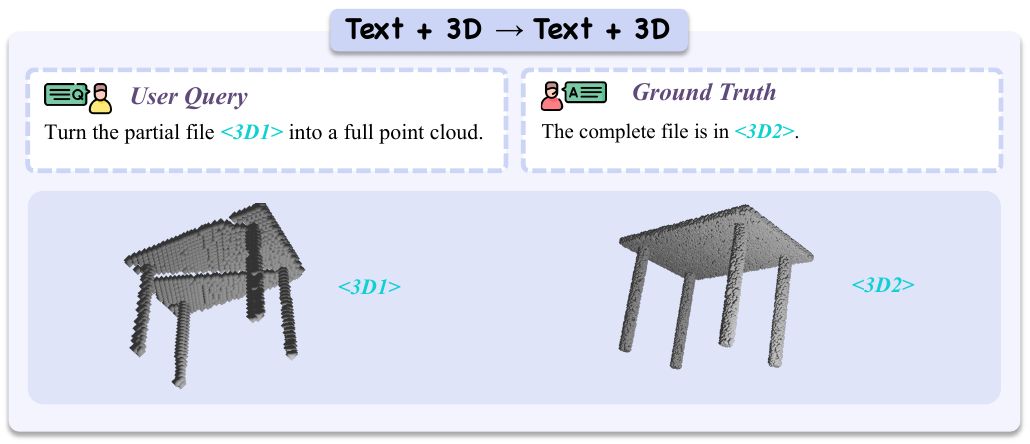}
    \caption{Data example of T + 3D to T + 3D interleaved combinations.}
    \label{fig:data1}
  \end{minipage}
  \begin{minipage}[b]{0.99\linewidth}
    \centering
    \vspace{6mm}
    \includegraphics[width=\linewidth]{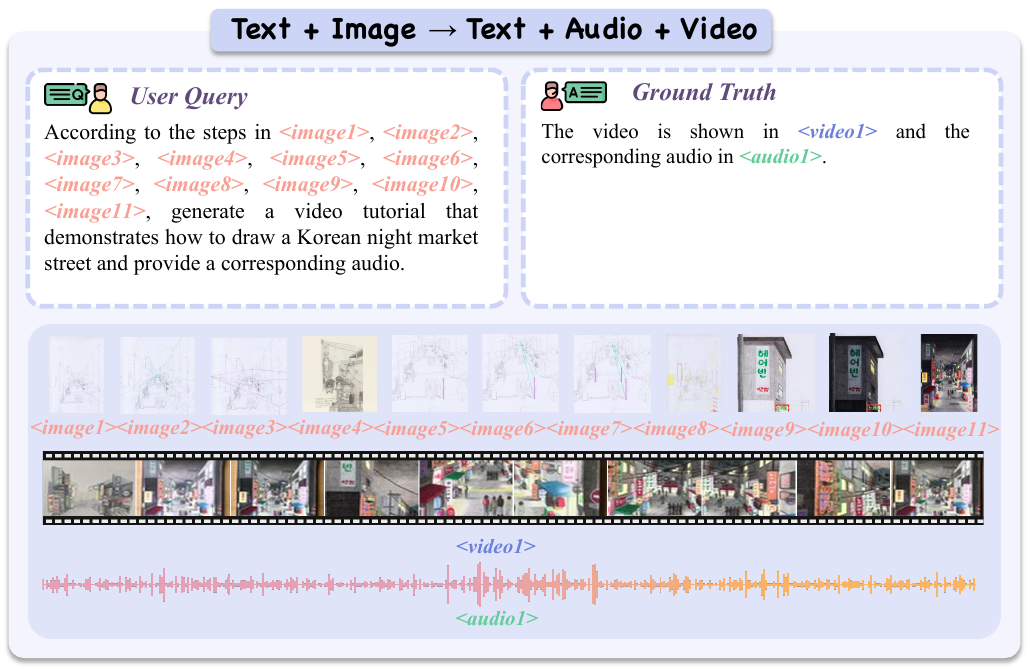}
    \caption{Data example of T + I to T + A + V interleaved combinations.}
    \label{fig:data2}
  \end{minipage}
\end{figure*}

\begin{figure*}[t!]
  \centering
  \begin{minipage}[b]{0.99\linewidth}
    \centering
    \includegraphics[width=\linewidth]{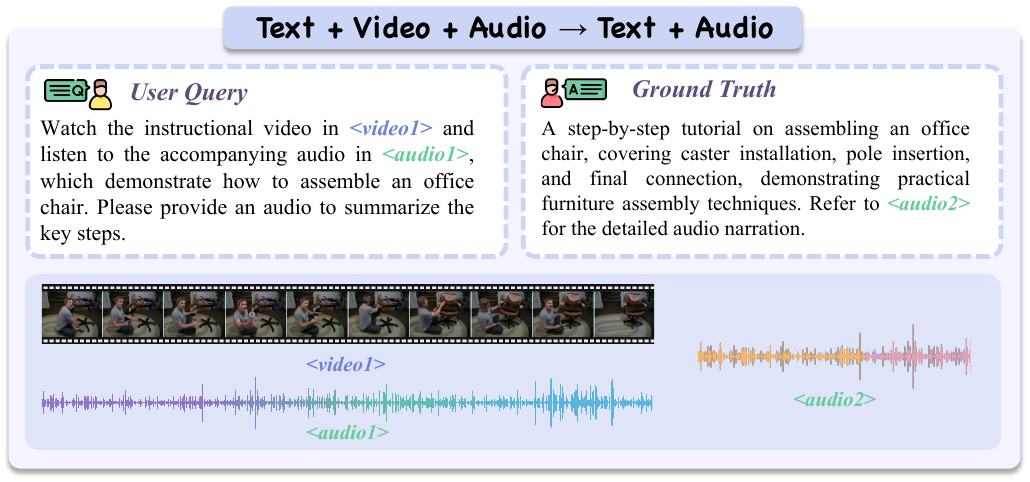}
    \caption{Data example of T + V + A to T + A interleaved combinations.}
    \label{fig:data3}
  \end{minipage}

  \begin{minipage}[b]{0.99\linewidth}
    \centering
    \vspace{6mm}
    \includegraphics[width=\linewidth]{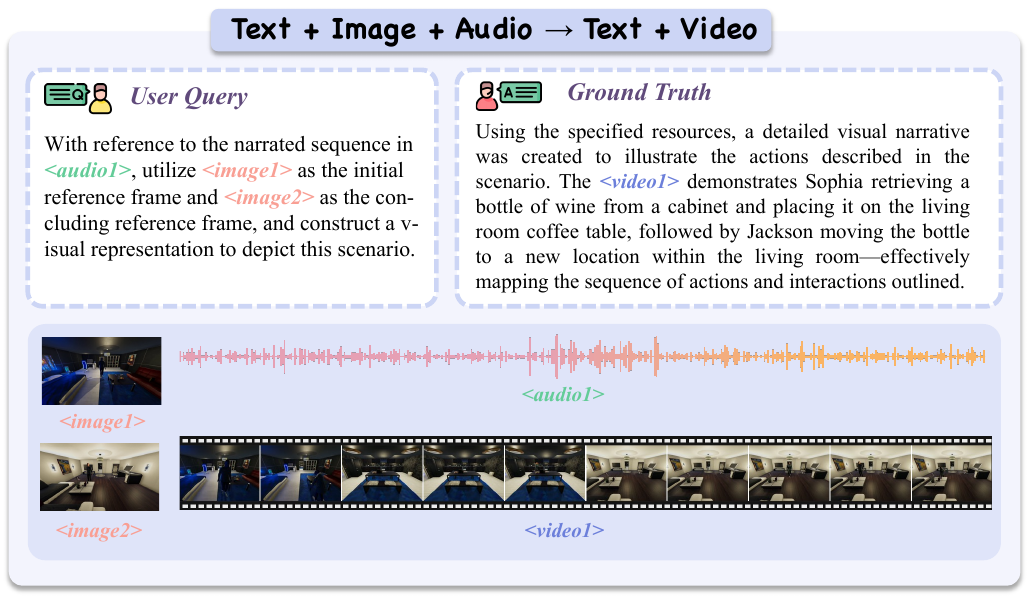}
    \caption{Data example of T + I + A to T + V interleaved combinations.}
    \label{fig:data4}
  \end{minipage}
\end{figure*}

\begin{figure*}[t!]
  \centering
  \begin{minipage}[b]{0.99\linewidth}
    \centering
    \includegraphics[width=\linewidth]{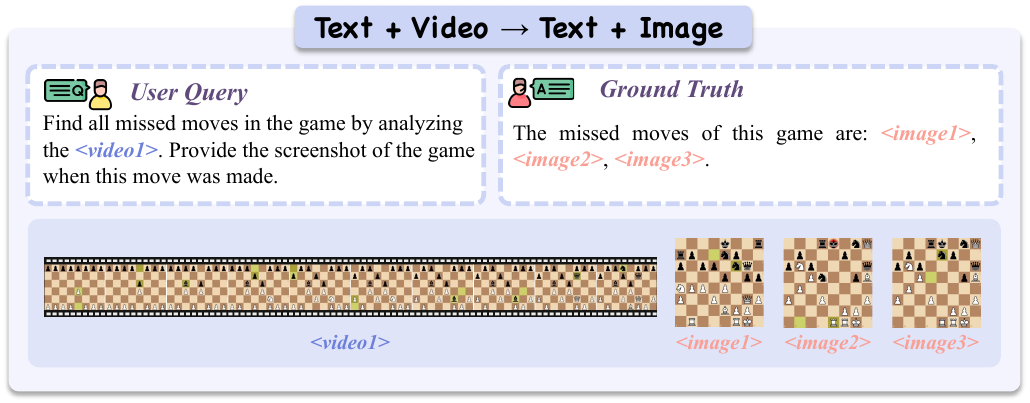}
    \caption{Data example of T + V to T + I interleaved combinations.}
    \label{fig:data5}
  \end{minipage}

  \begin{minipage}[b]{0.99\linewidth}
    \centering
    \vspace{6mm}
    \includegraphics[width=\linewidth]{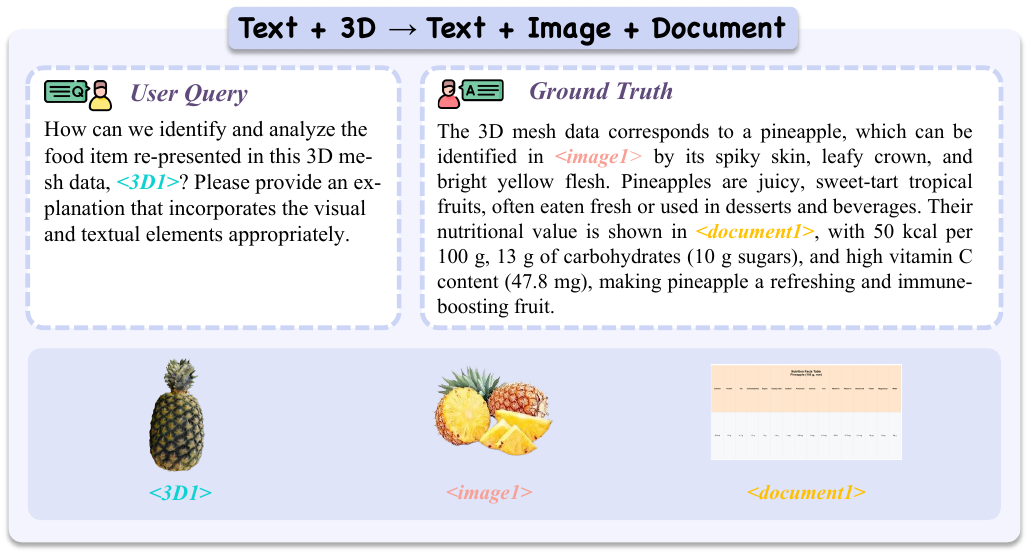}
    \caption{Data example of T + 3D to T + I + D interleaved combinations.}
    \label{fig:data6}
  \end{minipage}
\end{figure*}

\begin{figure*}[t!]
  \centering
  \begin{minipage}[b]{0.99\linewidth}
    \centering
    \includegraphics[width=\linewidth]{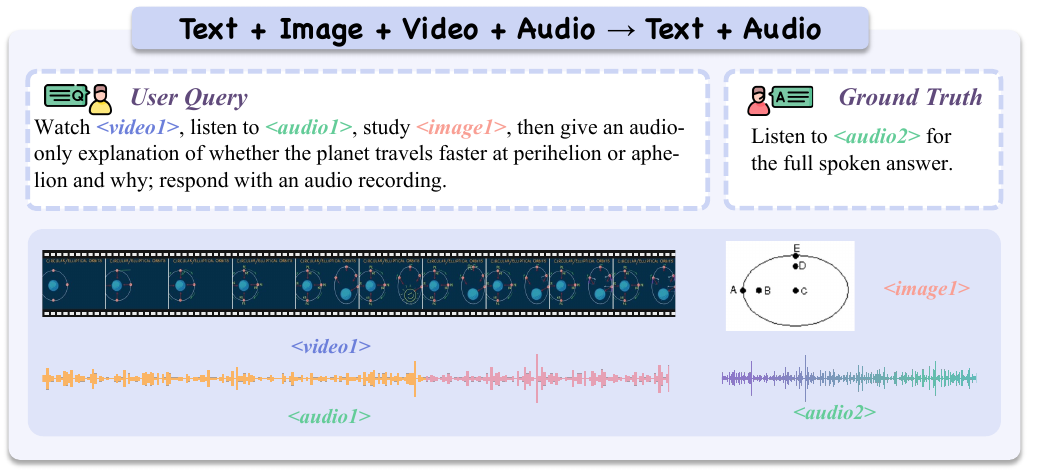}
    \caption{Data example of T + I + V + A to T + A interleaved combinations.}
    \label{fig:data7}
  \end{minipage}

  \begin{minipage}[b]{0.99\linewidth}
    \centering
    \vspace{6mm}
    \includegraphics[width=\linewidth]{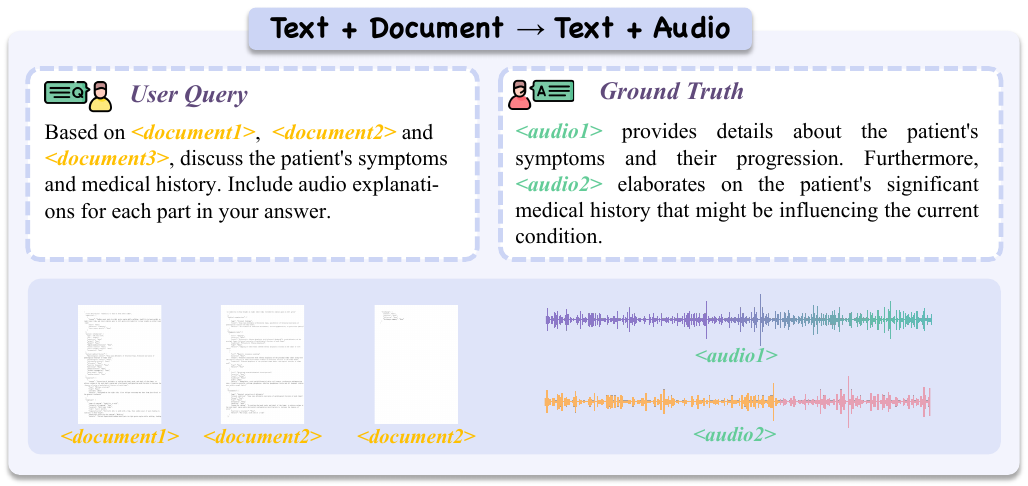}
    \caption{Data example of T + D to T + A interleaved combinations.}
    \label{fig:data8}
  \end{minipage}
\end{figure*}

\begin{figure*}[t!]
  \centering
  \begin{minipage}[b]{0.99\linewidth}
    \centering
    \includegraphics[width=\linewidth]{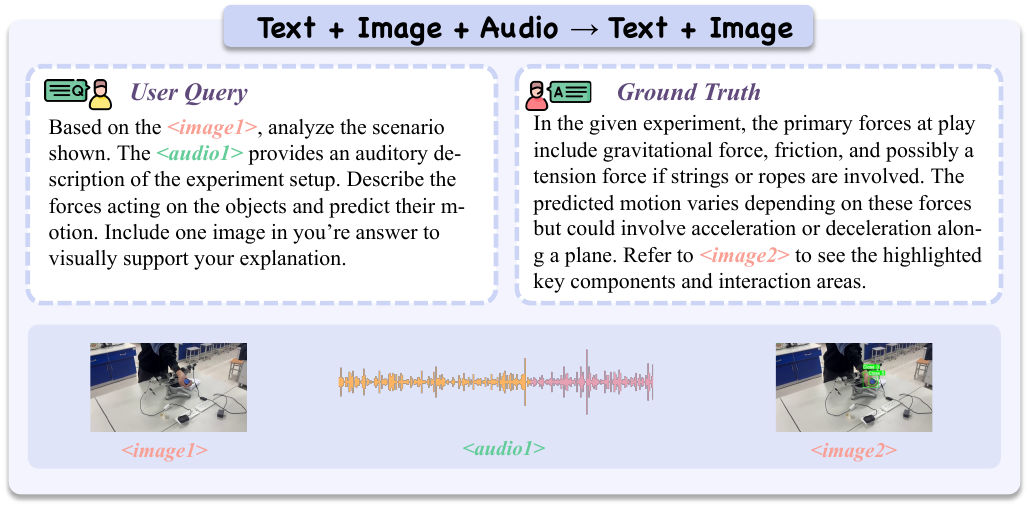}
    \caption{Data example of T + I + A to T + I interleaved combinations.}
    \label{fig:data9}
  \end{minipage}

  \begin{minipage}[b]{0.99\linewidth}
    \centering
    \vspace{6mm}
    \includegraphics[width=\linewidth]{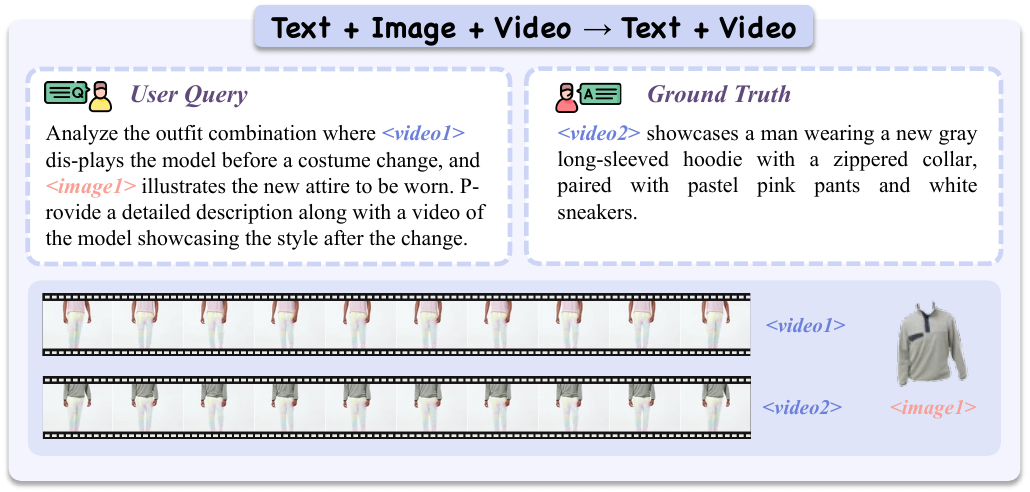}
    \caption{Data example of T + I + V to T + V interleaved combinations.}
    \label{fig:data10}
  \end{minipage}
\end{figure*}

\begin{figure*}[t!]
  \centering
  \begin{minipage}[b]{0.99\linewidth}
    \centering
    \includegraphics[width=\linewidth]{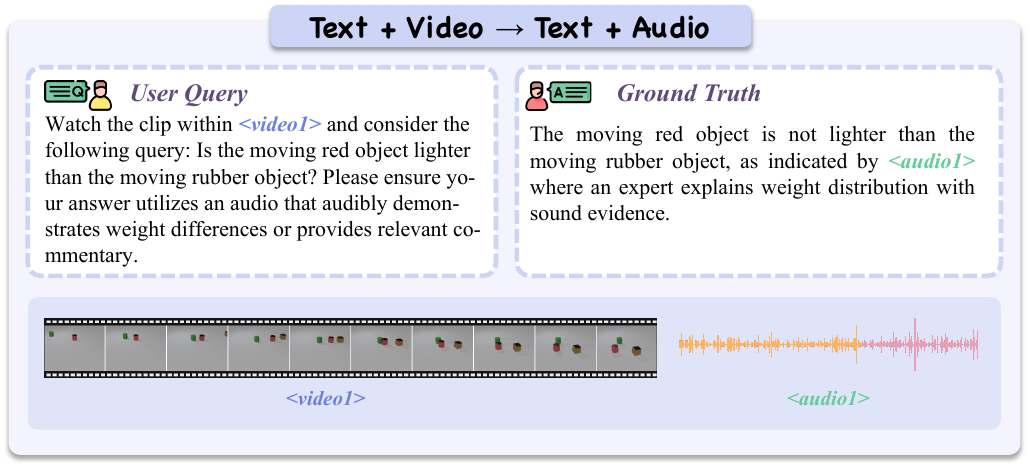}
    \caption{Data example of T + V to T + A interleaved combinations.}
    \label{fig:data11}
  \end{minipage}

  \begin{minipage}[b]{0.99\linewidth}
    \centering
    \vspace{6mm}
    \includegraphics[width=\linewidth]{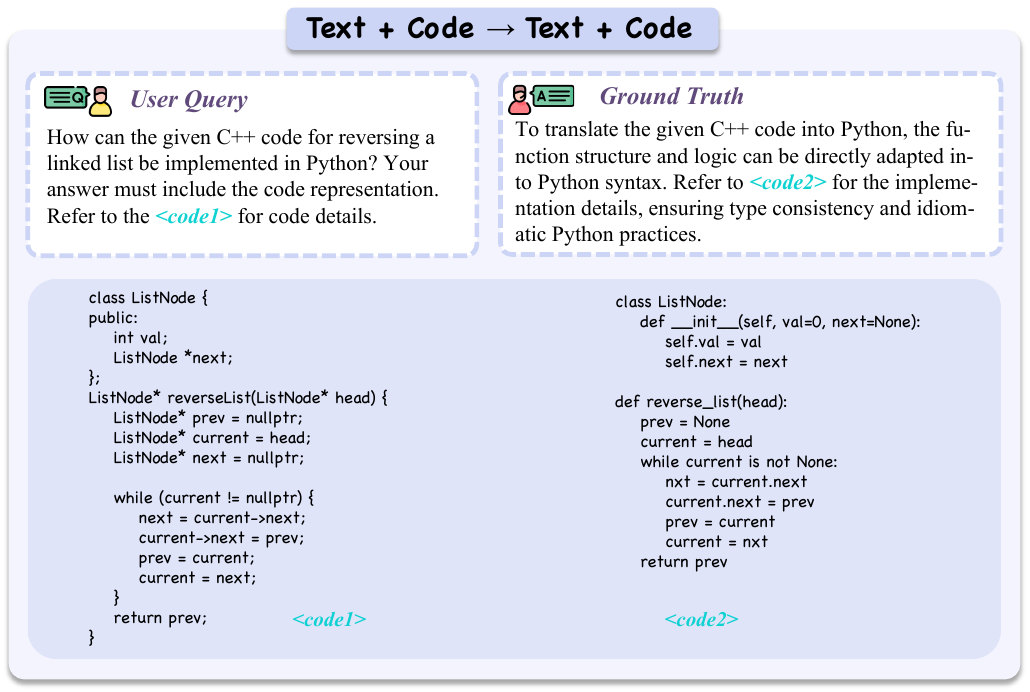}
    \caption{Data example of T + C to T + C interleaved combinations.}
    \label{fig:data12}
  \end{minipage}
\end{figure*}

\begin{figure*}[t!]
  \centering
  \begin{minipage}[b]{0.99\linewidth}
    \centering
    \includegraphics[width=\linewidth]{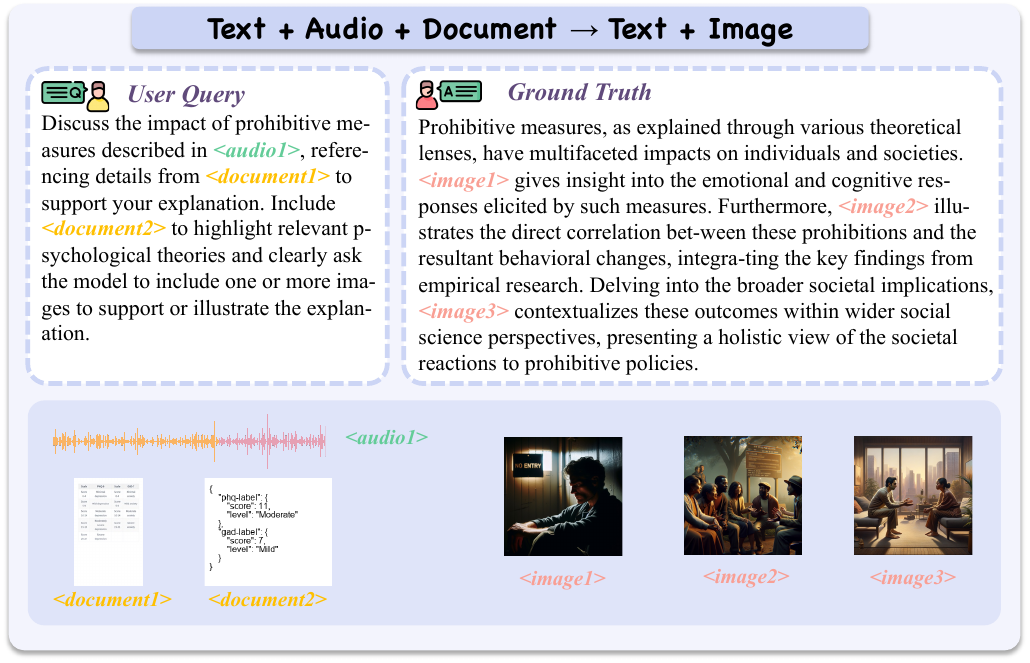}
    \caption{Data example of T + A to T + V interleaved combinations.}
    \label{fig:data13}
  \end{minipage}

  \begin{minipage}[b]{0.99\linewidth}
    \centering
    \vspace{6mm}
    \includegraphics[width=\linewidth]{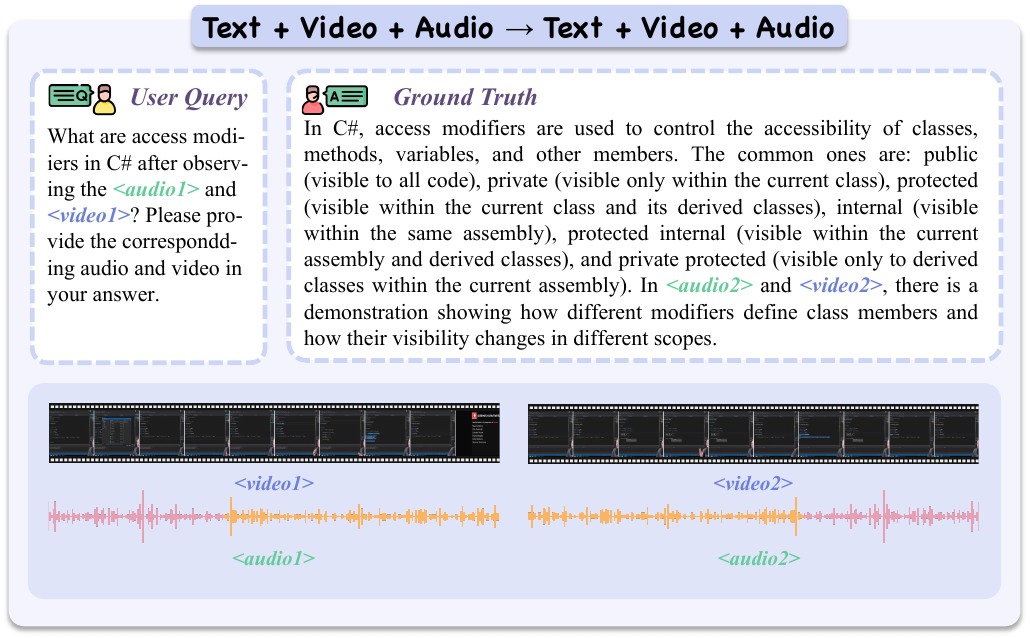}
    \caption{Data example of T + V + A to T + V + A interleaved combinations.}
    \label{fig:data14}
  \end{minipage}
\end{figure*}

\begin{figure*}[t!]
  \centering
  \begin{minipage}[b]{0.99\linewidth}
    \centering
    \includegraphics[width=\linewidth]{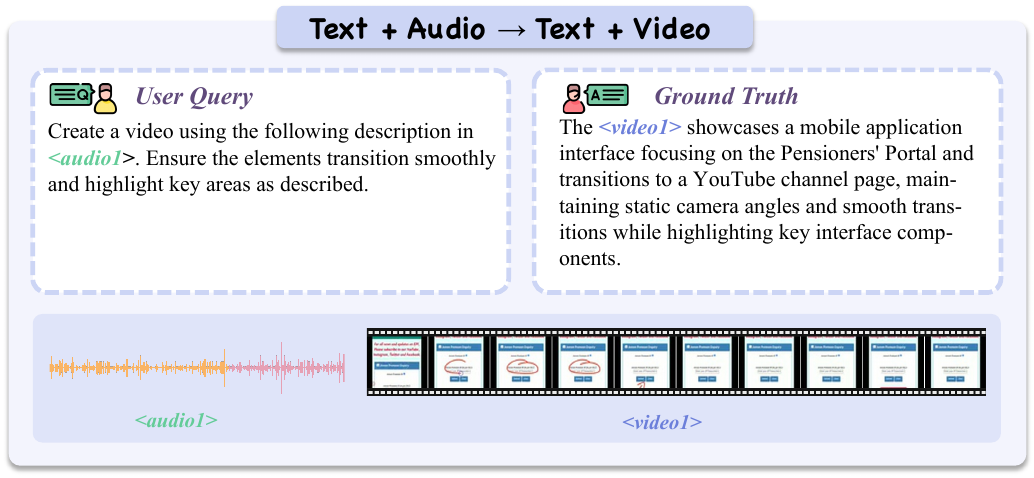}
    \caption{Data example of T + A + D to T + I interleaved combinations.}
    \label{fig:data15}
  \end{minipage}

  \begin{minipage}[b]{0.99\linewidth}
    \centering
    \vspace{6mm}
    \includegraphics[width=\linewidth]{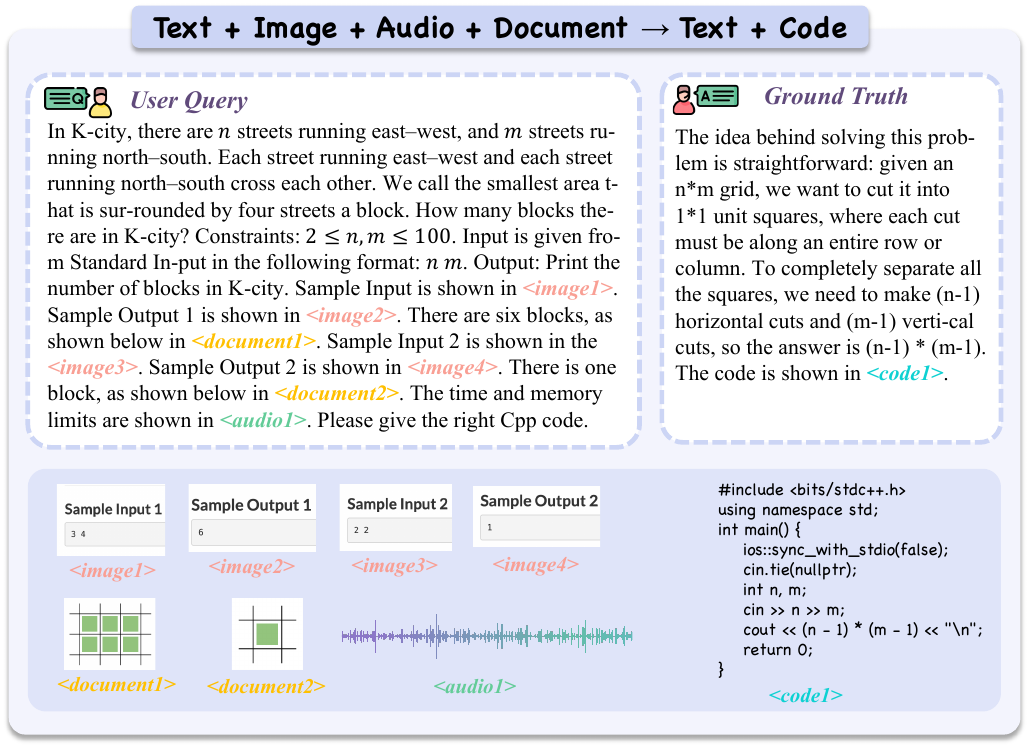}
    \caption{Data example of T + I + A + D to T + C interleaved combinations.}
    \label{fig:data16}
  \end{minipage}
\end{figure*}

\vspace{-1.5mm}
We provide some representative data examples of \textsc{UniM} as shown in Fig~\ref{fig:data1}, Fig~\ref{fig:data2}, Fig~\ref{fig:data3}, Fig~\ref{fig:data4}, Fig~\ref{fig:data5}, Fig~\ref{fig:data6}, Fig~\ref{fig:data7}, Fig~\ref{fig:data8}, Fig~\ref{fig:data9}, Fig~\ref{fig:data10}, Fig~\ref{fig:data11}, Fig~\ref{fig:data12}, Fig~\ref{fig:data13}, Fig~\ref{fig:data14}, Fig~\ref{fig:data15} and Fig~\ref{fig:data16}.

\begin{table}[t!]
\centering
\caption{Caption methods for each modality.}
\label{tab:caption_methods}
\vspace{-2mm}
\begin{tabular}{ll}
\toprule
\textbf{Modality} & \textbf{Caption Tool} \\
\midrule
Text      & Original format retained \\
Image     & GPT-5 mini \\
Video     & Qwen3-Omni \\
Audio     & Qwen3-Omni \\
Document  & GPT-5 mini \\
Code      & Original format retained \\
3D        & PointLLM \\
\bottomrule
\end{tabular}
\vspace{-2mm}
\end{table}

\vspace{-2mm}
\section{Details of Evaluation Suite}
\label{app:details_eval_suite}

\vspace{-2mm}
\subsection{Semantic Correctness \& Generation Quality}
To ensure cross-modal comparability and stable evaluation outcomes in assessing semantic correctness and generation quality, all experiments are conducted under a unified multimodal evaluation framework. This section provides the full technical details underlying the evaluation system described in the main text.

\subsubsection{Semantic Correctness}
Semantic Correctness measures the degree to which the model’s generated content is semantically aligned with the ground truth. 
In  the interleaved multimodal generation settings, a model may output text, images, audio, video, documents, code, or 3D data, but non-text modalities cannot be directly compared to the ground truth for semantic alignment. 
To address this limitation, we convert all modalities into textual representations prior to evaluation, enabling multimodal responses to be assessed within a unified semantic space using a consistent comparison protocol.

We convert non-text modalities into textual captions.
This procedure converts the entire multimodal output into a unified textual representation. The ground truth undergoes the same process during data preparation, ensuring that model response and ground truth are compared within an aligned semantic space. The captioning tools used for each modality are summarized in Table~\ref{tab:caption_methods}.

\begin{figure*}[t]
\centering
\includegraphics[width=0.99\linewidth]{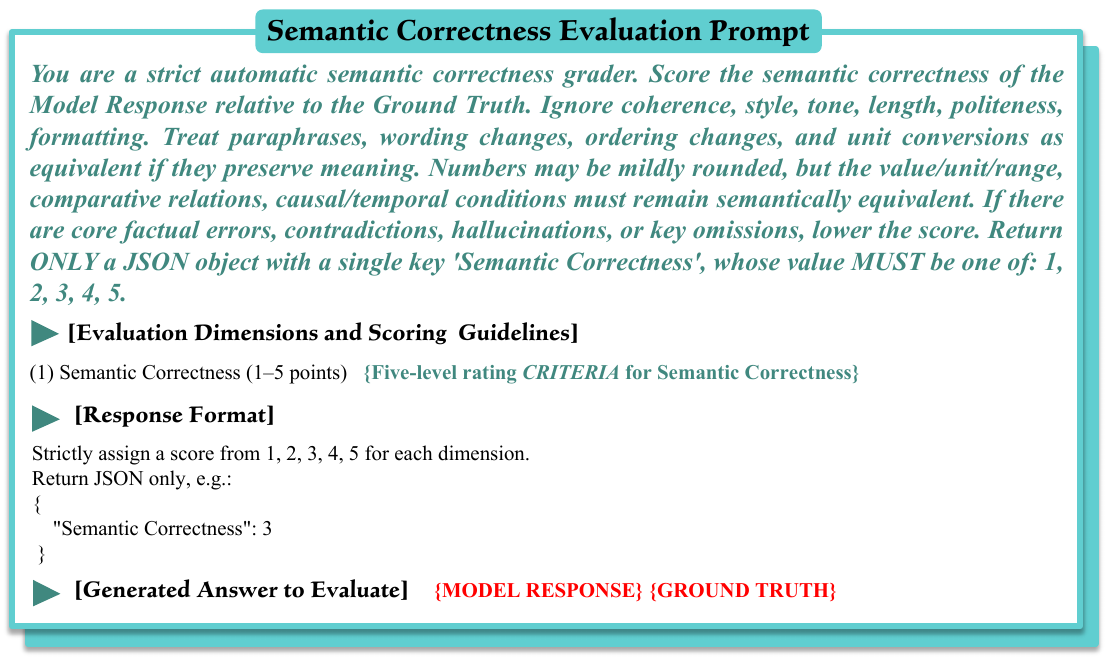}
\caption{Prompt template for \textit{Semantic Correctness}.}
\label{fig:sc_prompt}
\end{figure*}

After completing the textual alignment, we assess semantic correctness using an LLM-as-a-Judge scoring paradigm. The evaluation model takes as input the responses with captions and the ground truth, and outputs a score drawn from a predefined five-level semantic consistency criteria, with possible values $\{1, 2, 3, 4, 5\}$.
The scoring process focuses exclusively on semantic equivalence, disregarding style, tone, coherence, or surface form; variations in phrasing, ordering, or mild numerical rounding are permitted as long as the underlying meaning is preserved. 
Core factual errors, critical omissions, or contradictions result in lower scores. 
For each instance, the resulting discrete scores are further mapped onto the continuous interval 0-1 for normalization.
The scoring criteria are provided in Table~\ref{tab:sc_criteria}. The full prompt is shown in Fig.~\ref{fig:sc_prompt}.

\begin{table*}[t!]
\centering
\caption{Five-level rating criteria for \textit{Semantic Correctness}.}
\label{tab:sc_criteria}
\vspace{-2mm}
\fontsize{10}{12}\selectfont
\setlength{\tabcolsep}{2mm}
\begin{tabular}{ll}
\toprule
\textbf{Grade} & \textbf{Semantic Correctness Description} \\
\midrule
\textbf{5} &
\makecell[l]{a) Completely equivalent; \\
b) All key facts correct/covered; \\
c) No contradictions; \\
d) Units, ranges, and relational constraints remain consistent (paraphrasing, reordering, and \\ \ \ \ \ minor rounding are allowable).
}\\
\hdashline
\textbf{4} & 
\makecell[l]{a) Almost equivalent;\\
b) Most of key facts correct/covered;\\
c) No major contradiction;\\
d) Only minor omissions/ambiguity that do not affect the main conclusion.}\\
\hdashline
\textbf{3} & 
\makecell[l]{a) Partially correct;\\
b) Roughly half key facts correct;\\
c) No major contradiction but noticeable;\\
d) Notable omissions or minor misinterpretations, but the main conclusion is not fully overturned.}\\
\hdashline
\textbf{2} & 
\makecell[l]{a) Low correctness;\\
b) Less than half covered;\\
c) Important errors/contradictions/confusions (numbers/entities);\\
d) The core conclusion drifts, but still loosely on topic.}\\
\hdashline
\textbf{1} & 
\makecell[l]{a) Almost incorrect;\\
b) Nearly irrelevant;\\
c) Mostly contradictory, or hallucinated;\\
d) Mostly wrong/missing, or non-answers.}\\
\bottomrule
\end{tabular}
\vspace{-2mm}
\end{table*}

\subsubsection{Generation Quality}
Generation quality measures the clarity, stability, usability, and overall perceptual performance of the model's outputs across different modalities. In the interleaved multimodal generation tasks, semantic correctness captures only whether the content produced by the model aligns with the intended meaning, but it does not reflect the visual, auditory, structural, or readability quality of the outputs. Practical applications typically rely on both content reliability and perceptual quality; thus, generation quality must be evaluated as a dimension independent from semantic correctness.

Because different modalities exhibit substantial differences in signal form, structural properties, and degradation patterns, it is difficult to apply a single unified metric.
To address this, we design modality-specific quality assessment methods for seven distinct modalities and map all resulting scores onto 0-1 range, enabling cross-modal comparison.

\begin{figure*}[t]
\centering
\includegraphics[width=0.99\linewidth]{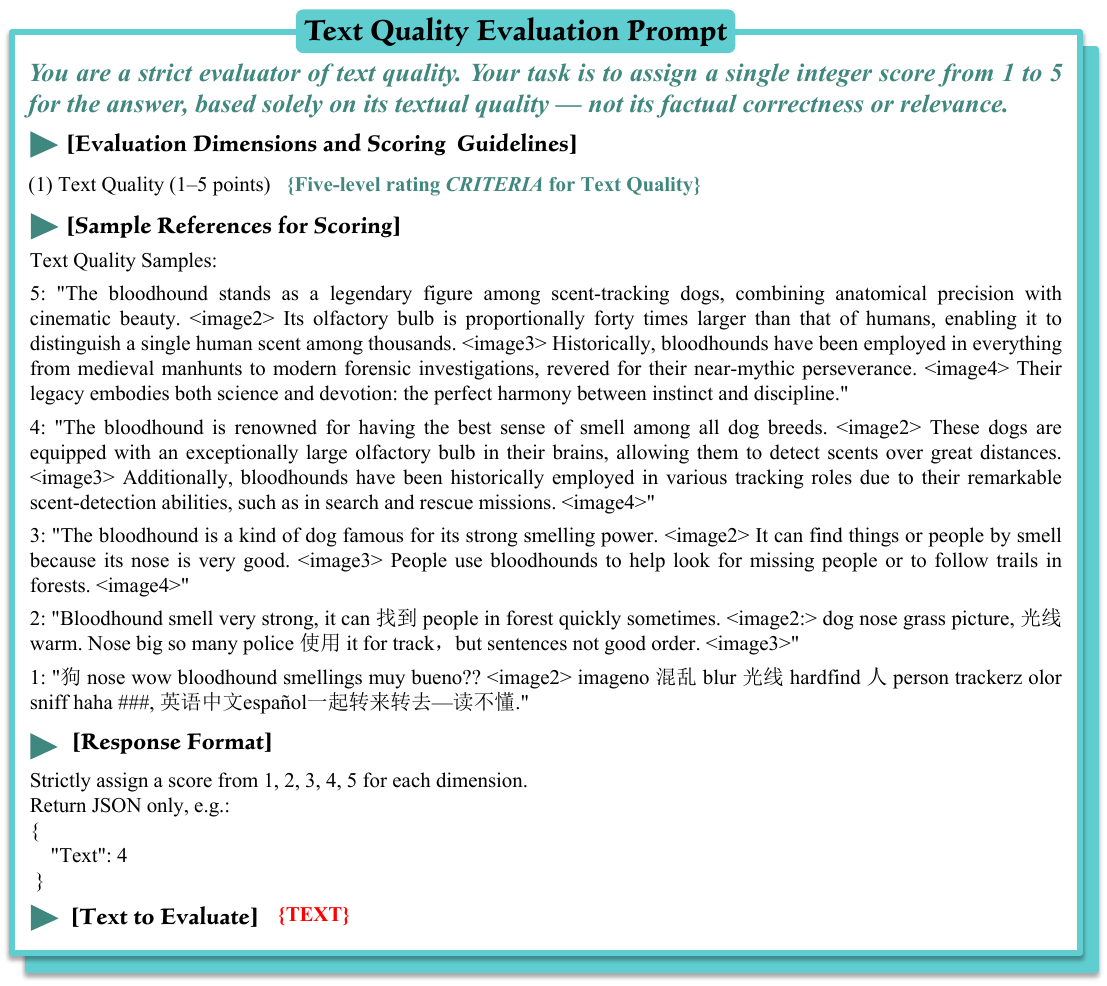}
\caption{Prompt template for text in \textit{Generation Quality}.}
\label{fig:gq_text_prompt}
\end{figure*}

\noindent \textbf{Text.} For the text modality, generation quality primarily reflects the completeness of the content, clarity of structure, fluency of language, and consistency of the output language. We adopt an LLM-as-a-Judge approach, where a GPT-5 mini evaluator constrained by a fixed prompt assesses the overall quality of the generated text. The evaluation follows a five-level quality criteria and is supported by a small set of examples to ensure consistency and comparability. The full criteria definition and prompt design are provided in Table~\ref{tab:gq_text_criteria} and Fig.~\ref{fig:gq_text_prompt}.

\begin{table*}[t!]
\centering
\caption{Five-level rating criteria for text in \textit{Generation Quality}.}
\label{tab:gq_text_criteria}
\fontsize{10}{12}\selectfont
\setlength{\tabcolsep}{2mm}
\begin{tabular}{ll}
\toprule
\textbf{Grade} & \textbf{Text Quality Description} \\
\midrule
\textbf{5} &
\makecell[l]{a) Content is rich, self-consistent, and detailed; no major omissions or vague generalities; stands\\ \ \ \ \ alone as a coherent text; \\
b) Structure is clear and well-organized; transitions are smooth (e.g., “firstly…then…therefore…”);\\ \ \ \ \ reasoning shows causal or hierarchical logic; \\
c) Language is natural and grammatically flawless; diverse sentence structures; no syntactic errors; \\
d) Entirely in one language; any foreign words appear only as necessary terminology.}\\
\hdashline
\textbf{4} & 
\makecell[l]{a) Content is generally complete with sufficient detail but slightly shallow or missing minor points;\\
b) Structure is good, with only mild jumps or awkward transitions that don’t affect comprehension;\\
c) Language is fluent with few minor grammatical or collocation issues;\\
d) Mostly consistent language, with rare short foreign terms that do not disrupt flow.}\\
\hdashline
\textbf{3} & 
\makecell[l]{a) Content covers the main idea but lacks depth or specific details;\\
b) Organization somewhat weak; order or topic shifts slightly; meaning still clear overall;\\
c) Several grammatical or spelling mistakes; simple or repetitive sentence patterns;\\
d) Minor language switching between sentences or paragraphs, noticeable but not confusing.}\\
\hdashline
\textbf{2} & 
\makecell[l]{a) Content is shallow, missing key details or explanations; very low information density;\\
b) Poor structure; sentences disjointed; reader must infer connections;\\
c) Frequent grammar errors; awkward or broken phrasing; readability low;\\
d) Frequent in-sentence language mixing that affects readability.}\\
\hdashline
\textbf{1} & 
\makecell[l]{a) Content is empty or meaningless; repetitive or irrelevant phrases; conveys no clear information;\\
b) No logical order; severe contradictions; text barely comprehensible;\\
c) Major grammatical breakdowns; unnatural or non-human syntax;\\
d) Chaotic multilingual mixing (e.g., Chinese + English + Spanish, random spelling noise)}\\
\bottomrule
\end{tabular}
\end{table*}

\noindent \textbf{Image.} For the image modality, we employ the Natural Image Quality Evaluator (NIQE) as a no-reference quality metric to measure the deviation of generated images from natural scene statistics(NSS): 
\begin{equation}
\label{eq:niqe}
\mathrm{NIQE}(I)
=
\sqrt{
(\mu_t - \mu_n)^{\mathsf{T}}
\left( \frac{\Sigma_t + \Sigma_n}{2} \right)^{-1}
(\mu_t - \mu_n)
}.
\end{equation}
Here, $\mu_{t}$ and $\Sigma_{t}$ denote the mean vector and covariance matrix of the evaluated image in the NSS feature space, while $\mu_{n}$ and $\Sigma_{n}$ represent the reference statistics estimated from a collection of high-quality natural images. The metric computes the Mahalanobis distance between these two feature distributions, reflecting the naturalness and degree of distortion of the image. Lower scores indicate that the generated image is closer to the distribution of natural images and thus exhibits higher perceptual quality.
In implementation, each image is partitioned into local patches from which we compute luminance normalization coefficients, local variances, and other NSS features. The resulting statistics are then compared with the reference distribution to estimate their deviations.

\noindent \textbf{Audio.}
The audio modality is evaluated using a fully statistics-driven, non-learning pipeline. 
Given an input waveform $x(t)$ resampled to 48\,kHz, we first apply mono conversion, 
mean removal, and energy-based trimming. The magnitude spectrogram is computed as
\begin{equation}
S(f,t) = \lvert\mathrm{STFT}(x)\rvert.
\end{equation}
The corresponding power spectrum is
\begin{equation}
P(f,t) = S(f,t)^2.
\end{equation}

A set of robust signal-to-noise ratio estimates is computed, including the original
energy ratio, subband stability, and a harmonic-percussive separation based metric. 
The effective SNR is defined as
\begin{equation}
\mathrm{SNR}_{\mathrm{eff}}
=
\max\!\bigl(
\mathrm{SNR}_{\mathrm{orig}},\,
\mathrm{SNR}_{\mathrm{sf}},\,
\mathrm{SNR}_{\mathrm{hpss}}
\bigr).
\end{equation}
It is mapped through a logistic function,
\begin{equation}
q_{\mathrm{snr}}
=
\left[1+\exp\{-k(\mathrm{SNR}_{\mathrm{eff}}-x_0)\}\right]^{-1}.
\end{equation}

Global spectral flatness is quantified as
{\begingroup
\small
\begin{equation}
\mathrm{SF}_g
=
\mathrm{median}_t
\left(
\frac{\exp\!\left(\frac{1}{F}\sum_f \ln (S(f,t)+\epsilon)\right)}
     {\frac{1}{F}\sum_f (S(f,t)+\epsilon)}
\right),
\end{equation}
\endgroup
}

and the corresponding structural score is
\begin{equation}
q_{\mathrm{struct}} = 1 - \mathrm{SF}_g.
\end{equation}

Dynamic range is extracted via percentile RMS and mapped to $q_{\mathrm{dr}}$, while 
loudness is measured using integrated LUFS and normalized to $q_{\mathrm{lufs}}$. 
Level stability, transient smoothness, and the crest factor are also included. 
The crest factor is computed as
\begin{equation}
\mathrm{crest}
=
20\log_{10}
\frac{\max_t |x(t)|}{\sqrt{\mathbb{E}[x(t)^2]}}.
\end{equation}
Its normalized score is
\begin{equation}
q_{\mathrm{crest}} = \mathrm{band\_score}(\mathrm{crest}).
\end{equation}
Bandwidth quality is derived from the 95\% energy frequency $f_{95}$ and mapped to $q_{\mathrm{bw}}$. 
Chroma stability and spectral contrast yield $q_{\mathrm{chroma}}$ and $q_{\mathrm{contrast}}$.

High-frequency hiss is penalized through the regression slope of the log-power spectrum 
in the 4--12\,kHz band, yielding $p_{\mathrm{hiss}}$. Excessive high-frequency energy produces 
a penalty $p_{\mathrm{hf}}$. Mid-section dropout or extended silence is handled by an adaptive 
coverage penalty $p_{\mathrm{gap}}$, and low-contrast noise textures result in a penalty 
$p_{\mathrm{lowC}}$.

Calibrated compensation factors are introduced to account for audio with pronounced
structural or periodic components. Structural cues contribute
\begin{equation}
b_{\mathrm{str}}
=
g_{\mathrm{struct}}\bigl(q_{\mathrm{contrast}},\, q_{\mathrm{chroma}}\bigr),
\end{equation}
while periodicity contributes
\begin{equation}
b_{\mathrm{per}}
=
g_{\mathrm{periodic}}\bigl(q_{\mathrm{periodic}}\bigr).
\end{equation}

The set of core quality terms $\{q_i\}_{i=1}^{N}$ is consolidated through 
a geometric-mean aggregation,
\begin{equation}
q_{\mathrm{base}}
=
\exp\!\left(
\frac{1}{N}\sum_{i=1}^{N} \ln q_i
\right).
\end{equation}

The final audio quality index is defined as
\begin{equation}
q_{\mathrm{audio}}
=
q_{\mathrm{base}}
\, p_{\mathrm{clip}}
\, p_{\mathrm{gap}}
\, p_{\mathrm{noise}}
\, p_{\mathrm{hiss}}
\, p_{\mathrm{hf}}
\, p_{\mathrm{lowC}}
\, b_{\mathrm{str}}
\, b_{\mathrm{per}},
\end{equation}
where $q_{\mathrm{audio}} \in [0,1]$.

\begin{figure*}[t]
    \centering
    \includegraphics[width=1\linewidth]{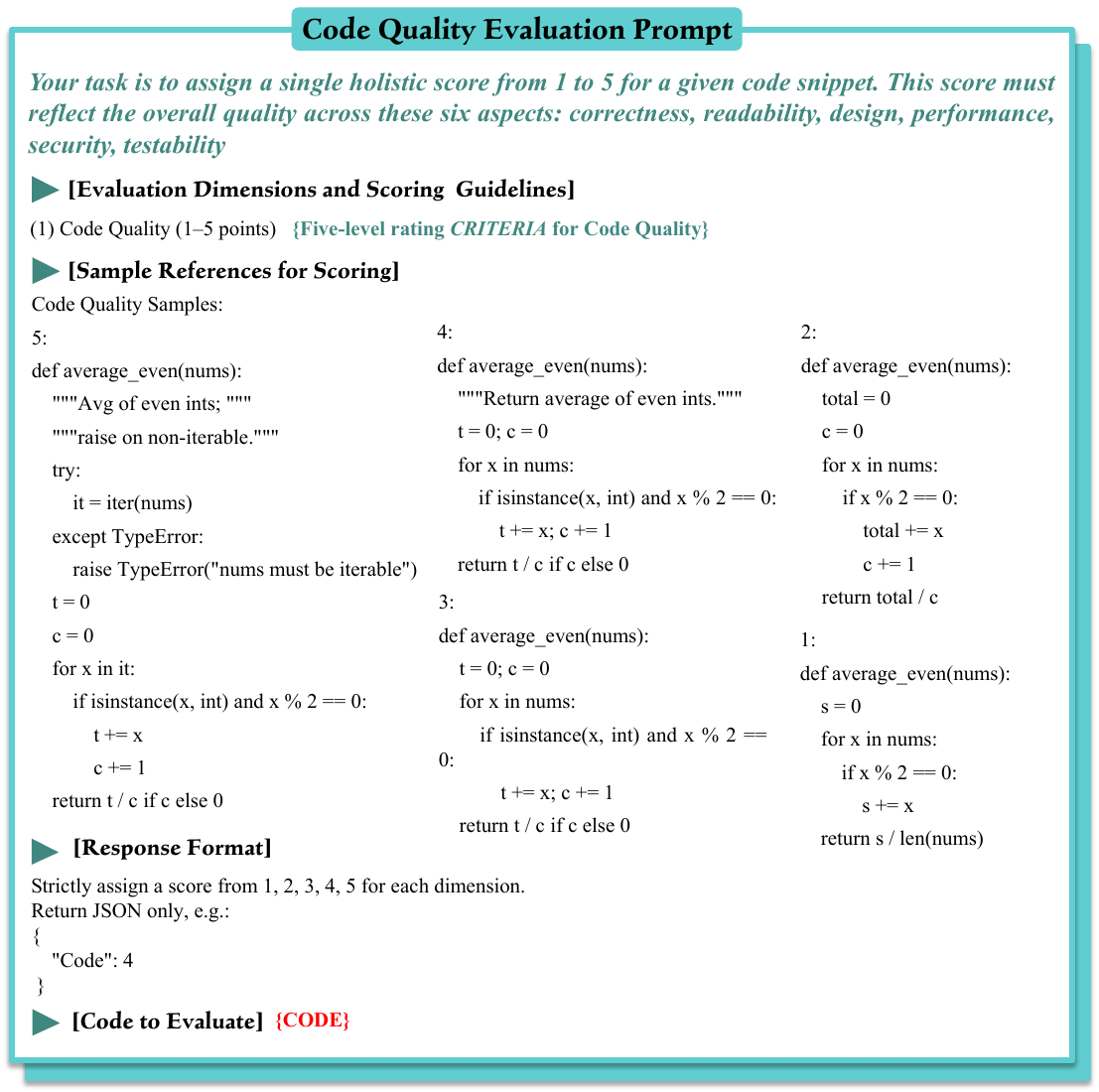}
    \caption{Prompt template for code in \textit{Generation Quality}.}
    \label{fig:gq_code_prompt}
\end{figure*}

\vspace{1mm}
\noindent \textbf{Code.} In the code modality, generation quality primarily reflects engineering quality rather than perceptual aesthetics. Accordingly, we adopt an LLM-as-a-Judge approach to perform structured quality assessment for each generated code segment. Specifically, the evaluator is configured as a language-agnostic and rigorous code reviewer, providing an overall quality judgment based on six dimensions: correctness, readability, design soundness, runtime efficiency, security, and testability. To enhance scoring consistency and interpretability, the prompt includes a five-level global quality criteria ranging from severe defects to engineering-grade excellence, along with representative few-shot examples covering different quality levels. These elements guide the LLM to make unified quality assessments across programming languages and coding styles. The complete prompt and criteria are provided in Table~\ref{tab:gq_code_criteria} and Fig.~\ref{fig:gq_code_prompt}.

\begin{table*}[t!]
\centering
\caption{Five-level rating criteria for code in \textit{Generation Quality}.}
\label{tab:gq_code_criteria}
\vspace{-2mm}
\fontsize{10}{12}\selectfont
\setlength{\tabcolsep}{2mm}
\begin{tabular}{ll}
\toprule
\textbf{Grade} & \textbf{Code Quality Description} \\
\midrule
\textbf{5} &
\makecell[l]{a) High-quality, professional-grade code with virtually no noticeable defects; \\
b) Logically rigorous, structurally well-organized, and highly robust; \\
c) Fully adheres to modern software engineering best practices, easy to maintain, and directly reusable.}\\
\hdashline
\textbf{4} & 
\makecell[l]{a) High overall quality with a reasonable structure and clear logic;\\
b) Contains only minor issues that do not affect usability;\\
c) Aligns with most software engineering best practices and is easy to maintain.}\\
\hdashline
\textbf{3} & 
\makecell[l]{a) Basically usable, with acceptable overall correctness but clear deficiencies;\\
b) Contains multiple areas for improvement that may affect maintainability or reliability;\\
c) The overall quality is functional but remains below desirable standards.}\\
\hdashline
\textbf{2} & 
\makecell[l]{a) Contains multiple evident issues, with overall quality falling below acceptable standards;\\
b) Barely runnable or only partially functional, requiring substantial repairs;\\
c) Exhibits major deficiencies in logic, structure, performance, or security.}\\
\hdashline
\textbf{1} & 
\makecell[l]{a) Contains numerous critical defects, resulting in extremely low overall quality;\\
b) Exhibits chaotic logic, unclear structure, and significant risks;\\
c) Largely non-reusable and unmaintainable, requiring a complete rewrite.}\\
\bottomrule
\end{tabular}
\vspace{-2mm}
\end{table*}

\begin{figure*}[t]
    \centering
    \includegraphics[width=1\linewidth]{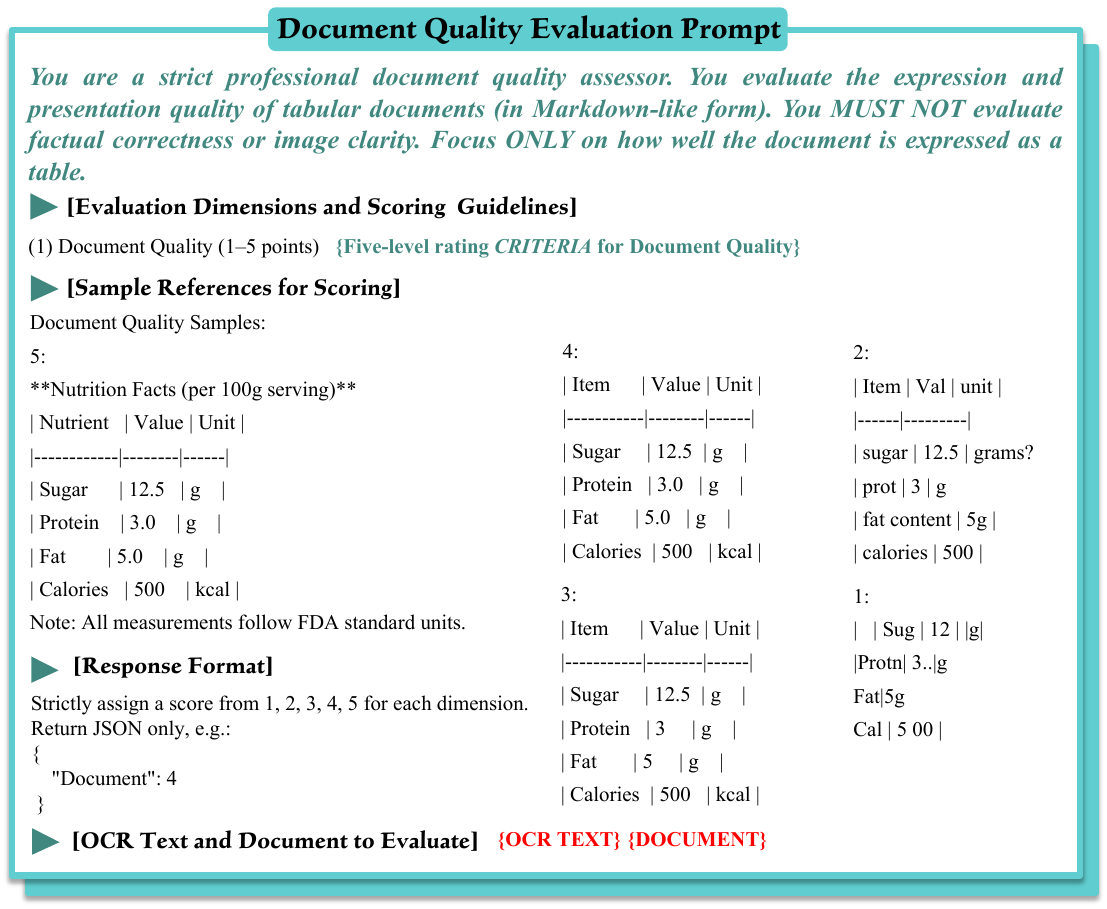}
    \caption{Prompt template for document in \textit{Generation Quality}.}
    \label{fig:gq_doc_prompt}
\end{figure*}

\vspace{1mm}
\noindent \textbf{Document.} In the document modality, the focus is not on image clarity or semantic correctness, but rather on the overall quality of tabular representation. Specifically, when a table presented as an image is recognized and transcribed into structured text, we assess whether the title and column headers are clear, whether the header and data rows are properly aligned, whether units and numerical values are consistent, and whether the table is sufficiently coherent to be interpreted and used independently.Therefore, we first apply OCR to transcribe the document image into raw text, then organize it into a Markdown-style table using simple heuristic rules. Additional formatting cues, such as consistency of decimal places and presence of measurement units, are extracted and provided to the evaluator as auxiliary hints. Building on this representation, we adopt an LLM-as-a-Judge paradigm that categorizes document quality into five levels, each specified by a corresponding set of criteria and supported by few-shot examples. This enables the model to make stable and consistent judgments regarding label clarity, structural organization, internal consistency, and overall self-containedness. The evaluation prompt and criteria are provided in Table~\ref{tab:gq_document_criteria} and Fig.~\ref{fig:gq_doc_prompt}.

\begin{table*}[t!]
\centering
\caption{Five-level rating criteria for document in \textit{Generation Quality}.}
\label{tab:gq_document_criteria}
\vspace{-2mm}
\fontsize{10}{12}\selectfont
\setlength{\tabcolsep}{2mm}
\begin{tabular}{ll}
\toprule
\textbf{Grade} & \textbf{Document Quality Description} \\
\midrule
\textbf{5} &
\makecell[l]{a) All titles, labels, and column names are clear and unambiguous; \\
b) The table structure is complete, with well-organized logical partitions and clear hierarchy; \\
c) Units, naming conventions, capitalization, and punctuation are fully consistent; \\
d) Numerical presentation is standardized, including uniform decimal places, units, and precision,\\ \ \ \ \  with proper alignment; \\
e) The surrounding context provides sufficient information for the table to be understood independently \\ \ \ \ \ without additional explanation.}\\
\hdashline
\textbf{4} & 
\makecell[l]{a) Most titles and column names are accurate, with only minor ambiguities;\\
b) The table structure is generally reasonable, though some misalignment or slightly unclear grouping \\ \ \ \ \ is present;\\
c) Overall consistency is good, with only small variations in decimal places or units;\\
d) Data presentation is mostly standardized, with mild irregularities that do not hinder usability; \\
e) The table can largely be understood independently, though some contextual information is slightly \\ \ \ \ \ lacking.}\\
\hdashline
\textbf{3} & 
\makecell[l]{a) Contains noticeably ambiguous titles or labels whose meanings require contextual inference;\\
b) Table structure is relatively disorganized, with unclear grouping, poorly defined relationships \\ \ \ \ \ between columns;\\
c) Multiple inconsistencies in formatting, naming, or units, reducing overall usability;\\
d) Numerical presentation is irregular, with large variations in decimal places and missing or mixed \\ \ \ \ \ units; \\
e) Lacks sufficient contextual explanation, making the table difficult to interpret independently.}\\
\hdashline
\textbf{2} & 
\makecell[l]{a) Titles or column names are missing or severely ambiguous, with clear potential for misinterpretation;\\
b) The table structure is incomplete, exhibiting column misalignment, unclear row logic, or missing \\ \ \ \ \ grouping;\\
c) Formatting is highly inconsistent, with disordered or entirely missing units;\\
d) Numerical presentation shows prominent issues, including irregular precision, inconsistent units, \\ \ \ \ \ and potentially misleading values; \\
e) The table is largely uninterpretable on its own and requires substantial supplementary explanation.}\\
\hdashline
\textbf{1} & 
\makecell[l]{a) Titles or column names are extensively incorrect, missing, or unrecognizable;\\
b) The table structure is entirely broken, with row–column relationships indistinguishable;\\
c) No formatting consistency is preserved, with content unordered and misaligned;\\
d) Numerical values are meaningless due to severe unit confusion, row mismatches, or erroneous \\ \ \ \ \ combinations; \\
e) The table is completely unusable on its own and requires re-OCR or full reconstruction from the \\ \ \ \ \ original source.}\\
\bottomrule
\end{tabular}
\vspace{-2mm}
\end{table*}

\noindent \textbf{3D.} For the 3D modality, we employ the no-reference quality metric NR3D-Q to assess the generated 3D objects. This metric provides a unified score by jointly evaluating topology and completeness (Topology, T), geometric fidelity (Geometry, G), and sampling uniformity (Sampling Uniformity, S) for point clouds or mesh structures, with all results average and normalized to a common scale.
The topology and completeness score (\(T\)) evaluates the global structural plausibility and integrity of a 3D object. For mesh representations, \(T\) reflects properties such as closure, watertightness, the proportion of non-manifold edges, self-intersections, and the number of connected components. For point clouds, \(T\) approximates the object’s “surface-likeness’’ and structural continuity through intrinsic dimensionality estimation, local connectivity statistics, and normal-vector consistency.
The geometry quality score (\(G\)) focuses on whether the local geometric structures are smooth, stable, and free from noticeable noise. It incorporates curvature statistics derived from local PCA, multi-scale normal stability, and neighborhood-level normal consistency.
The sampling uniformity score (\(S\)) measures whether points or surface elements are evenly distributed in space, penalizing large sparse regions or excessively dense clusters. It is computed using statistics such as the distribution of kNN distances, the proportion of outliers, and the variation in mesh face areas or vertex valence.

% 加了一句，gq是平均
Based on the above calculations, GQ is defined as the average of the scores assigned to all content items in the output of the instance, including both the text content and all multimodal content files.

\subsubsection{Semantic-Quality Coupled Score}
Evaluating SC and GQ in isolation often fails to fully capture a model’s practical utility. To address this limitation, we introduce the Semantic-Quality Coupled Score (SQCS), which is computed for each instance using the normalized SC and GQ values:
\begin{equation}
\text{SQCS} = \text{SC} \cdot \left( \eta^{\text{SQCS}}  + (1 - \eta^{\text{SQCS}}) \cdot \text{GQ} \right) \,.
\label{eq:1_1}
\end{equation}
The metric treats semantic correctness as the primary factor and incorporates generation quality as a modulation term that adjusts the score only when semantic conditions are satisfied. Specifically, when SC is low, the overall SQCS remains strongly suppressed even if GQ is high. Conversely, when SC is high, improvements in GQ drive the SQCS closer to the upper bound of SC, enabling finer discrimination of generation quality among semantically correct samples.

\subsection{Response Structure Integrity}
In this section, we provide a detailed elaboration of the two metrics under the \textit{Response Structure Integrity} evaluation dimension, Strict Structure Score (StS) and Lenient Structure Score (LeS), including their formal definitions and computational procedures.

\subsubsection{Strict Structure Score} 

The StS is designed to evaluate the strict structural consistency of a model's output.
This metric requires that the types and quantities of modalities generated in the model's response precisely correspond to those in the ground truth.
Any missing or redundant modalities, or discrepancies in the number of modality placeholder tags, are explicitly penalized. 

Specifically, we define the modality set involved in score computation as 
\begin{equation}
    \mathcal{M}^{'} = \{ m \mid g_m > 0 \vee r_m > 0 \},
\end{equation}

where $g_m$ denotes the number of placeholder tags of modality $m$ in the ground truth, and $r_m$ denotes the number of placeholder tokens of modality $m$ in the model response. 
The set $\mathcal{M}^{'}$ thus contains all modalities that appear at least once in either the ground truth or the model response. 
If a modality does not appear in either $g_m = r_m = 0$, it is considered irrelevant and excluded from evaluation process. 
For each modality, we define the number of matches $n_m = min(g_m,r_m)$. The precision $P_m$ and recall $R_m$ for modality placeholder counts as
\begin{equation}
    P_m = 
    \begin{cases}
        \dfrac{n_m}{r_m}, & r_m > 0, \\
        0, & r_m = 0,
    \end{cases}
    \qquad
    R_m = 
    \begin{cases}
        \dfrac{n_m}{g_m}, & g_m > 0, \\
        0, & g_m = 0.
    \end{cases}
\end{equation}

The F1 score for each modality is calculated as 
\begin{equation}
    F1_m = \dfrac{2 P_m R_m}{P_m + R_m}.
\end{equation}

Finally, the StS is defined as the average of $F1_m$, as shown in the following equation.
\begin{equation}
    \text{StS} = \frac{1}{|\mathcal{M}^{'}|} \sum_{m \in \mathcal{M}^{'}} F1_m.
\end{equation}

\subsubsection{Lenient Structure Score} 

The LeS focuses on evaluating the degree of coverage at the modality level.
This metric assesses whether the types of modalities generated in the response are consistent with those in the ground truth.

Let $g_t$ denote the set of modality types appearing in the ground truth, and $r_t$ denote the set of modality types appearing in the model response. 
We then define the modality overlap set as 
\begin{equation}
    \text{Overlap} = g_t \cap r_t. 
\end{equation}

The LeS is defined as the ratio of the number of overlapping modality types to the total number of modality types in the ground truth, as shown in the following equation.
\begin{equation}
    \text{LeS} = \frac{|\text{Overlap}|}{|g_t|}.
\end{equation}

\subsection{Interleaved Coherence}
Within the \textit{Interleaved Coherence} dimension, we further refine it into two complementary perspectives: Holistic Coherence and Stylistic Harmony.
The former reflects the model's ability to maintain semantic consistency and logical completeness across modalities, while the latter captures its capacity to preserve a unified narrative tone and style during cross-modal generation, serving as a key indicator of the naturalness of the generated content.

\subsubsection{Holistic Coherence}
Holistic Coherence aims to assess whether a model can maintain global consistency in semantic logic, narrative structure, and modality transitions during multimodal interleaved generation.
This perspective focuses on evaluating whether the model truly understands and integrates information across different modalities, rather than merely performing superficial modality stacking.

If the model's response demonstrates specific and accurate cross-modal references, with complementary information across modalities, natural logical flow, and well-organized interleaved labels that together produce a clear and coherent narrative, it should be rated high.
Conversely, if there are conflicts or disconnections between modalities, such as evident logical leaps, semantic contradictions, or disordered modality sequencing that hinder overall comprehension—the response should be rated low.
We define five-level rating criteria, with detailed definitions for each level shown in Table~\ref{tab:coherence-scale}.

\begin{table*}[t!]
\centering
\caption{Five-level rating criteria for \textit{Holistic Coherence}.}
\label{tab:coherence-scale}
\vspace{-2mm}
\fontsize{10}{12}\selectfont
\setlength{\tabcolsep}{2mm}
\begin{tabular}{ll}
\toprule
\textbf{Grade} & \textbf{Holistic Coherence Description} \\
\midrule
\textbf{5} &
\makecell[l]{a) Output highly matches the input, multimodal references accurate and specific; \\
b) Different modality information complements each other without noticeable contradictions; \\
c) Logic is rigorous, structure is reasonable, overall semantic/logical order is clear; \\
d) Interleaved multimodal tags naturally ordered, reading/understanding experience smooth.}\\
\hdashline
\textbf{4} & 
\makecell[l]{a) Output mostly matches the input, references generally reasonable;\\
b) Different modality information mostly complementary, minor omissions or redundancy;\\
c) Overall logic coherent, local repetitions or jumps minor, not affecting understanding;\\
d) Interleaved tags mostly ordered, minor adjustments do not affect understanding.}\\
\hdashline
\textbf{3} & 
\makecell[l]{a) Output has general relation to input, references are vague or partially unclear;\\
b) Some modality blocks repeated or missing, minor contradictions exist;\\
c) Local logic jumps or slight contradictions, requires extra reasoning;\\
d) Interleaved tags partially disordered, local understanding may be difficult.}\\
\hdashline
\textbf{2} & 
\makecell[l]{a) Output lowly matches the input, most references vague or incorrect;\\
b) Cross-modal information repetitive or conflicting;\\
c) Logic obviously chaotic, many contradictions;\\
d) Interleaved tags order clearly disordered, understanding difficult.}\\
\hdashline
\textbf{1} & 
\makecell[l]{a) Output is almost irrelevant to input, references missing or wrong;\\
b) Different modalities barely complement, major contradictions;\\
c) Logic completely collapsed, frequent contradictions;\\
d) Interleaved tags order completely disordered, hard to understand.}\\
\bottomrule
\end{tabular}
\vspace{-2mm}
\end{table*}

\subsubsection{Stylistic Harmony}
Stylistic Harmony is used to assess whether a model can maintain uniformity in register, tone, terminology, and visual style when generating interleaved multimodal content.
In multimodal narratives, consistent style enhances the naturalness and credibility of the generated output: for instance, if the textual style is formal but the visual content appears cartoonish, or if terminology and naming are inconsistent across modalities, the overall reading experience will degrade significantly.

If the model's response demonstrates high coordination between linguistic and visual styles, with consistent tone, sentence structure, and rhetoric; and if key concepts, naming conventions, and modality tags are used uniformly, resulting in a smooth and natural overall reading experience, it should be rated high.
Conversely, if the response exhibits chaotic register, inconsistent terminology, or obvious stylistic conflicts that make the content difficult to understand, it should be rated low. We define five-level rating criteria, with detailed definitions for each level shown in Table~\ref{tab:stylistic-scale}.

\begin{table*}[t!]
\centering
\caption{Five-level rating criteria for \textit{Stylistic Harmony}.}
\label{tab:stylistic-scale}
\vspace{-2mm}
\fontsize{10}{12}\selectfont
\setlength{\tabcolsep}{1.2mm}
\begin{tabular}{ll}
\toprule
\textbf{Grade} & \textbf{Stylistic Harmony Description} \\
\midrule
\textbf{5} & 
\makecell[l]{a) Style highly consistent, cross-modal narration uniform, tone and sentence structures without deviation;\\
b) Terminology fully consistent, key concepts, tags, and naming completely aligned;\\
c) Expression and visual style fully aligned, wording, sentence structures, and rhetorical/visual style\\ \ \ \ \ coordinated, overall experience smooth.}\\
\hdashline
\textbf{4} & 
\makecell[l]{a) Style mostly consistent, minor deviations;\\
b) Terminology generally consistent, key concepts, tags, and naming mostly aligned;\\
c) Expression and visual style mostly aligned, minor deviations, overall experience unaffected.}\\
\hdashline
\textbf{3} & 
\makecell[l]{a) Style partially consistent, noticeable differences in narration;\\
b) Terminology partially mixed, key concepts, tags, naming sometimes inconsistent;\\
c) Expression and visual style partially aligned, wording, sentence structures, or rhetorical/visual style\\ \ \ \ \   obviously deviate, reading/watching experience affected.}\\
\hdashline
\textbf{2} & 
\makecell[l]{a) Style inconsistent, cross-modal narration uncoordinated;\\
b) Terminology not uniform, key concepts, tags, naming frequently mixed;\\
c) Expression and visual style not aligned, wording, sentence structures, or rhetorical/visual style conflicting, \\ \ \ \ \ understanding affected.}\\
\hdashline
\textbf{1} & 
\makecell[l]{a) Style completely inconsistent, cross-modal narration chaotic;\\
b) Terminology completely wrong, key concepts, tags, naming incorrect or inconsistent;\\
c) Expression and visual style completely chaotic, wording, sentence structures, or rhetorical/visual style \\  \ \ \ \ extremely uncoordinated, almost impossible to understand.}\\
\bottomrule
\end{tabular}
\vspace{-2mm}
\end{table*}

\subsubsection{Interleaved Coherence Score}

To provide a unified and reliable metric for evaluating model performance in interleaved multimodal generation, we introduce the Interleaved Coherence Score (ICS).
Existing metrics often focus on semantic correctness or isolated stylistic quality but fail to capture how well a model maintains coherence when multiple modalities appear in an interleaved sequence. ICS addresses this gap by jointly assessing the semantic and structural alignment across modalities and the stylistic consistency of the generated output.
This unified metric enables fair comparison across diverse multimodal configurations and highlights failure cases that conventional single-modality metrics cannot reveal.

ICS is defined as a weighted combination of SH and HC, given by
\begin{equation}
\text{ICS} = \eta^\text{ICS} \cdot \text{HC} + (1-\eta^\text{ICS}) \cdot \text{SH} \,.
\label{eq:ics}
\end{equation}

To ensure the reproducibility of our evaluation process, we present the complete evaluation prompt used in the \textit{Interleaved Coherence} dimension.
This prompt guides the evaluation model to assign scores to generated content based on predefined criteria across two dimensions: Holistic Coherence and Stylistic Harmony.
To enhance the stability of model scoring, we incorporate a small number of few-shot guiding examples, which demonstrate sample outputs at different score levels along with their corresponding explanations. 

Before conducting the actual evaluation, we apply a unified text-based conversion to all non-text modalities.
Specifically, we use caption tools to automatically caption the original image, audio, video, and other non-text inputs, and replace placeholder tags in the prompt with their transcribed textual representations. This ensures that the LLM performs scoring solely based on a standardized, text-only format.
This preprocessing step guarantees comparability across different modalities and prevents biases arising from inconsistent perceptual capabilities of the evaluation model.

\begin{figure*}[t!]
    \centering
    \includegraphics[width=\linewidth]{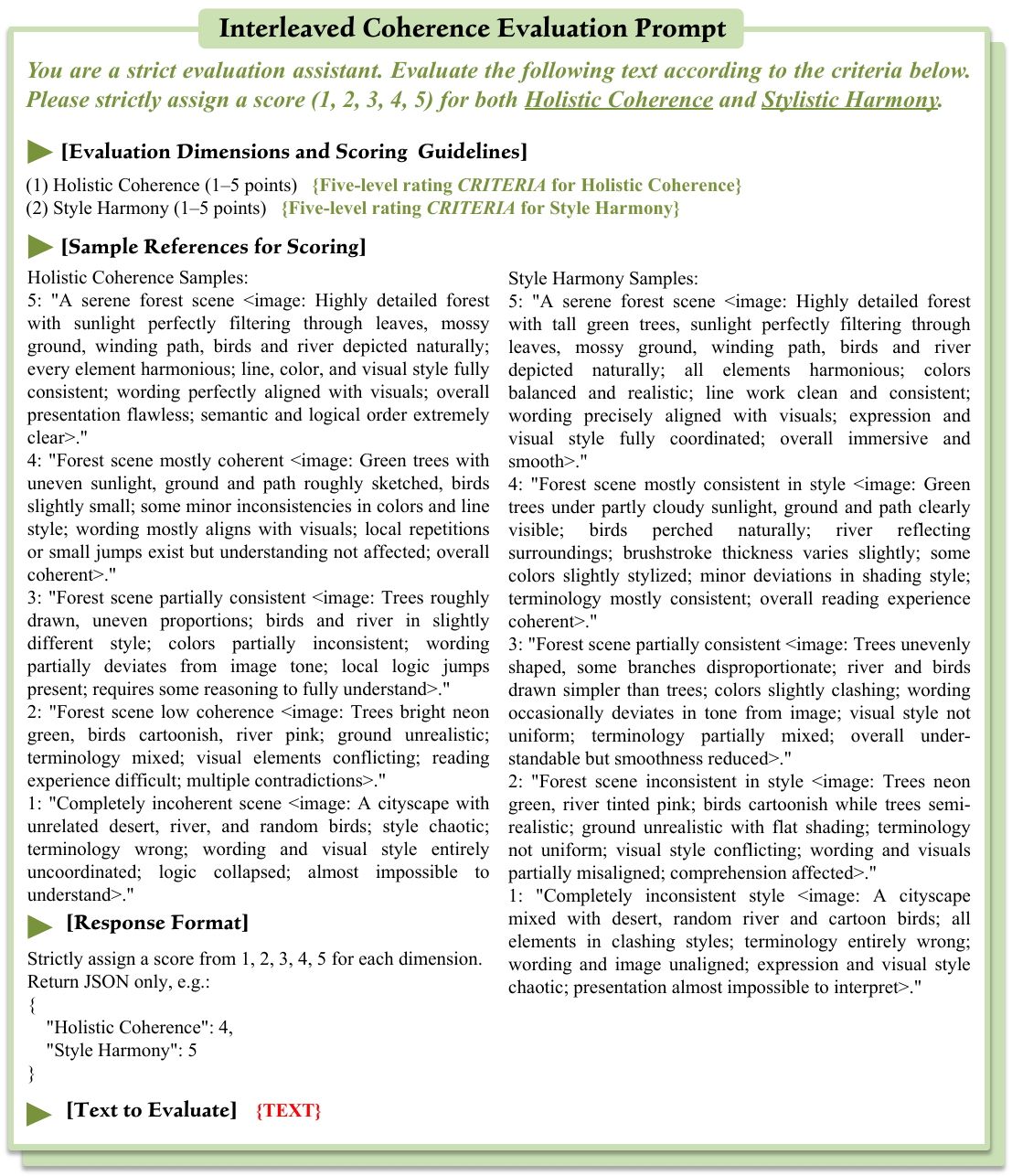}
    \caption{Interleaved coherence evaluation prompt.}
    \label{fig:ic prompt}
\end{figure*}

Finally, to standardize the evaluation scale across experiments, we linearly normalize all ICS from the original 1–5 scale to a 0–1 range. The caption tools used for each modality are listed in Table~\ref{tab:transcription_tools}. The full prompt is shown in Fig.~\ref{fig:ic prompt}.

\begin{table}[t]
\centering
\caption{Caption tools for each modality.}
\label{tab:transcription_tools}
\vspace{-2mm}
\begin{tabular}{ll}
\toprule
\textbf{Modality} & \textbf{Caption Tool} \\
\midrule
Image     & GPT-5 mini \\
Video     & Qwen3-Omni \\
Audio     & Qwen3-Omni \\
Document  & GPT-5 mini \\
Code      & Original format retained \\
3D        & Three-view rendering $\rightarrow$ GPT-5 mini \\
\bottomrule
\end{tabular}
\vspace{-1mm}
\end{table}

\subsection{Supporting Rate}
Since no current existing MLLM is perfect enough to support interleaved multimodal learning across the whole 7 modalities (text, image, audio, video, document, code, and 3D), there should be a common case in \textsc{UniM} that baseline MLLMs are likely to be unable to accept certain modality types as input.
To ensure fairness in evaluation, we emphasize full-modality visibility when conducting the assessment procedure. 
Specifically, a model must be able to receive and handle all modality information contained in an instance’s input, ensuring that its outputs are based on a complete understanding of the query rather than being affected by missing inputs or modality incompatibilities.

Therefore, we treat a model’s ability to support all input modalities of an instance as a prerequisite for evaluation: an instance is considered supported only if the model can effectively receive and process every modality present in its input. 
We then define the Supporting Rate ($\tau $) of a model as the proportion of supported instances relative to the entire \textsc{UniM} dataset. 
This metric reflects the model’s breadth of adaptability and coverage across the multimodal input space.

Building on the supporting rate as a conditional modifier, we further distinguish two types of metrics in our evaluation suite: $\mathcal{X}^{abs}$ and $\mathcal{X}^{rel}$.
$\mathcal{X}^{abs}$ denotes the average performance computed only over the subset of instances that a model can support, that is, those for which the model can fully process all input modalities. 
This reflects the model’s true capability within its valid operating range. 
$\mathcal{X}^{rel}$, in contrast, characterizes the model’s overall effectiveness on \textsc{UniM} by incorporating its coverage across the full dataset.
Together, $\mathcal{X}^{abs}$ and $\mathcal{X}^{rel}$ capture model performance from modality supportability and holistic benchmark completeness.

\subsection{Metric List}

\begin{table}[t!]
\centering
\caption{Metrics list of \textsc{UniM} evaluation suite.} 
\label{tab:metric list}
\vspace{-2mm}
\fontsize{8.5}{10}\selectfont
\setlength{\tabcolsep}{0.4mm}
\begin{tabular}{lll}
\toprule
\bf Metric & \bf Range & \bf Dimension \\
\midrule
$\tau \uparrow$ & [0, 1] & /  \\ 
SC $\uparrow$ & [0, 1] & Semantic Correctness \& Generation Quality  \\ 
GQ $\uparrow$ & [0, 1] & Semantic Correctness \& Generation Quality  \\ 
SQCS$^{abs}$ $\uparrow$ & [0, 1] & Semantic Correctness \& Generation Quality  \\ 
SQCS$^{rel}$ $\uparrow$ & [0, 1] & Semantic Correctness \& Generation Quality  \\ 
StS$^{abs}$ $\uparrow$ & [0, 1] & Response Structure Integrity \\ 
LeS$^{abs}$ $\uparrow$ & [0, 1] & Response Structure Integrity \\ 
StS$^{rel}$ $\uparrow$ & [0, 1] & Response Structure Integrity \\ 
LeS$^{rel}$ $\uparrow$ & [0, 1] & Response Structure Integrity \\ 
HC $\uparrow$ & [0, 1] & Interleaved Coherence  \\ 
SH $\uparrow$ & [0, 1] & Interleaved Coherence  \\ 
ICS$^{abs}$ $\uparrow$ & [0, 1] & Interleaved Coherence  \\ 
ICS$^{rel}$ $\uparrow$ & [0, 1] & Interleaved Coherence  \\ 
\bottomrule
\end{tabular}
\vspace{-2mm}
\end{table}

Overall, the \textsc{UniM} evaluation suite consists of 3 dimensions and 13 metrics, as illustrated in Table~\ref{tab:metric list}. It provides a comprehensive assessment of MLLMs performance in interleaved multimodal learning.

\section{Details of \textsc{UniMA}}
\label{app:details_agent}
\subsection{Receiving Module}

The Receiving Module serves as the entry point for multimodal interleaved inputs spanning seven modalities. Its primary function is to apply modality-specific expert tools to extract and convert raw inputs into dense captions for each item.

Receiving Module integrates four specialized tools:

\noindent
\textbf{GPT-5 mini} \cite{OpenAI_Introducing_GPT5_2025} handles text, image, document, and code comprehension, performing semantic parsing and symbolic reasoning to extract structured textual representations.

\noindent
\textbf{Qwen3-Omni Thinker} \cite{xu2025qwen3omni} is dedicated to audio and video understanding, capturing temporal dependencies and cross-modal correlations in continuous spatio-temporal signals.

\noindent
\textbf{Qwen3-VL} \cite{Qwen3VLHF2025} serves as the visual grounding module, extracting object-level details and bounding-box coordinates as structured visual evidence.

\noindent
\textbf{PointLLM} \cite{xu2024pointllm} processes 3D point-cloud data, enabling geometric reasoning and spatial perception within three-dimensional environments.

\subsection{Traceable Evidence Reasoning Module}

Traceable Evidence Reasoning (TER) module includes two parts: Structured Evidence Reasoning Chain and Verification Submodule.
\subsubsection{Structured Evidence Reasoning Chain}

\begin{figure*}[t!]
    \centering
    \includegraphics[width=0.94\linewidth]{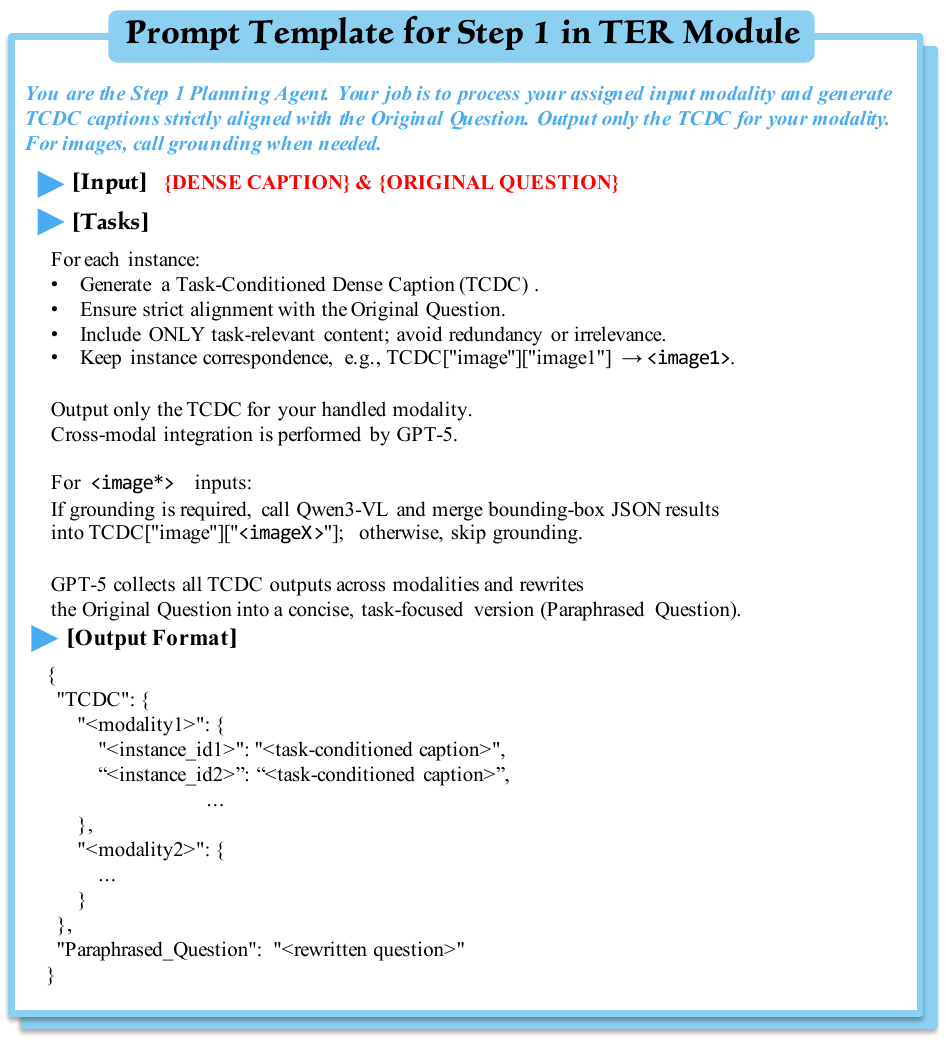}
    \vspace{-2mm}
    \caption{Prompt template for step 1 in TER module.}
    \label{fig:step1}
\end{figure*}

The main purpose of \textbf{Step~1} is to generate \textit{task-conditioned dense caption (TCDC)} and the \textit{paraphrased question} in parallel by using the output of Receiving Module and original question. 
TCDC differs from conventional dense captioning that merely provides detailed semantic descriptions of visual or video content. It explicitly incorporates the task context into the caption generation process, allowing the semantic content, granularity, and style of the caption to be adaptively constrained by the current task.
Formally, it can be expressed as
\begin{equation}
Y = f_{\text{caption}}(X \mid T),   
\end{equation}
where $X$ denotes the non-textual input modality and $T$ represents the task context.
The resulting captions automatically emphasize semantics relevant to $T$ while filtering out irrelevant details.
Paraphrased question is a rephrased version of the original question. It emphasizes the key point of the original question, making the description clearer. Then, all multimodal evidence is first normalized and reorganized into unified, task-conditioned representations that support the subsequent stages of structured reasoning. All prompt templates used across these tools are summarized in Fig.~\ref{fig:step1}.

\begin{figure*}[t!]
    \centering
    \includegraphics[width=0.96\linewidth]{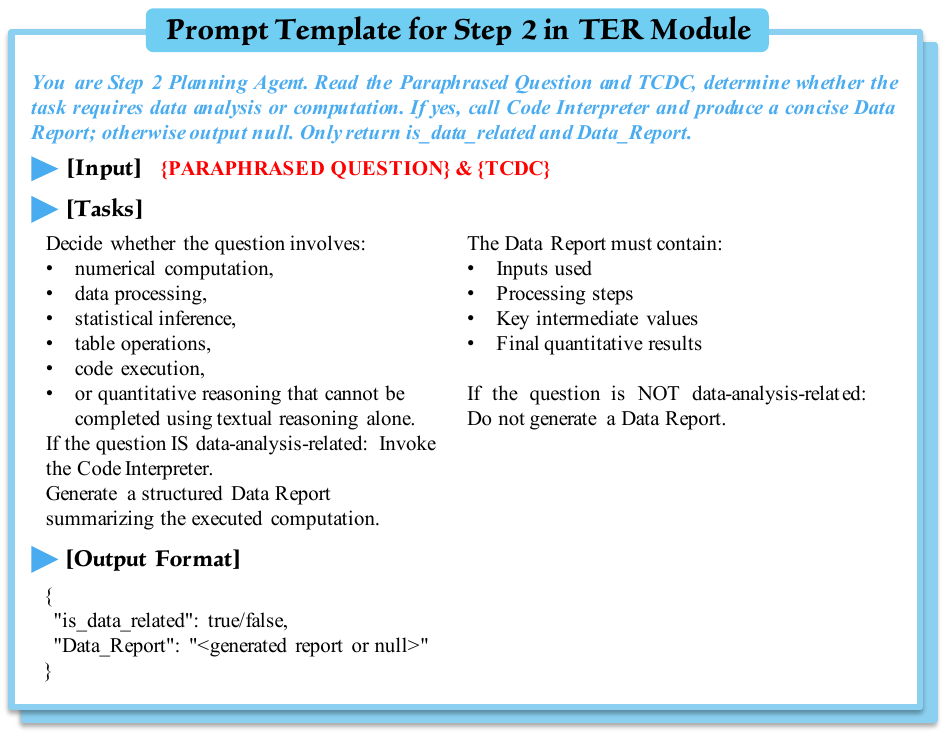}
    \vspace{-2mm}
    \caption{Prompt template for step 2 in TER module.}
    \label{fig:step2}
\end{figure*}

In \textbf{Step~2}, \textsc{UniMA} determines whether the user query requires quantitative
computation beyond pure textual reasoning.
The module reads both the paraphrased question and the cross-modal TCDC
evidence, and checks for the presence of numerical operations, data processing,
statistical inference, table manipulation, or any task that cannot be reliably
solved through language reasoning alone.
If such requirements are detected, Step 2 automatically invokes the Code
Interpreter, executes the necessary computation, and returns a concise
\textit{data report} summarizing the inputs, intermediate steps, and final results.
If not required, the \textit{data report} is omitted. Fig. \ref{fig:step2} shows the prompt template.

\begin{figure*}[t!]
    \centering
    \includegraphics[width=0.94\linewidth]{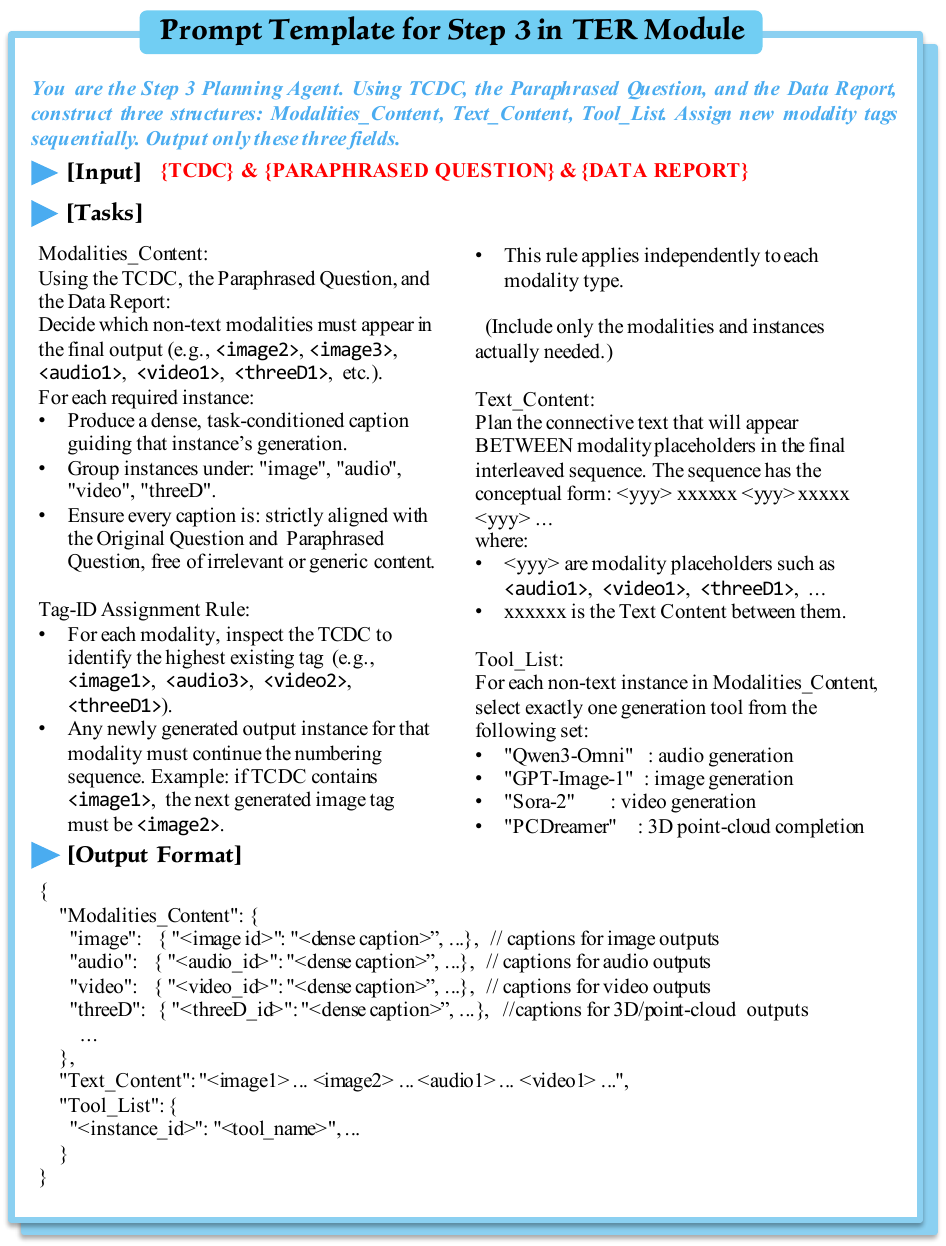}
    \vspace{-2mm}
    \caption{Prompt template for step 3 in TER module.}
    \label{fig:step3}
\end{figure*}

In \textbf{Step~3}, \textsc{UniMA} converts the intermediate outputs (TCDC, paraphrased question, and data report) from the 
Receiving and TER modules into three structured components that guide the Generating Module. \textit{modalities content} component specifies all non-text modalities that must appear in the final 
interleaved output (e.g., $<$image2$>$, $<$audio1$>$, $<$video1$>$).  
For each instance, the agent produces a concise, task-aligned dense caption derived 
from the TCDC and the paraphrased question. The result is a JSON grouping of modality 
instances by type. 
\textit{text content} defines the connective narrative that will be interleaved with non-text 
modalities. Instead of generating the final prose, the agent outputs a structured 
planning string of the form`$<$image1$>$\texttt{xx}$<$audio1$>$\texttt{xx}$<$video1$>$\texttt{xx}', where the segments between placeholders provide high-level textual transitions. The third part is \textit{tool list}, for each non-text placeholder tag in modalities content, the agent selects 
the appropriate generation tool from the system’s tool set 
(Qwen3-Omni for audio, GPT-Image-1 for images, Sora-2 for video, PCDreamer for 3D).  
The mapping is returned as a JSON dictionary linking each modality instance to its 
corresponding tool. Together, these three components provide a complete, machine-readable plan that 
specifies what to generate, how it fits into the interleaved 
sequence, and which tool will handle each modality. The prompt template of this step is shown in Fig. \ref{fig:step3}.

\begin{figure*}[t!]
    \centering
    \includegraphics[width=0.96\linewidth]{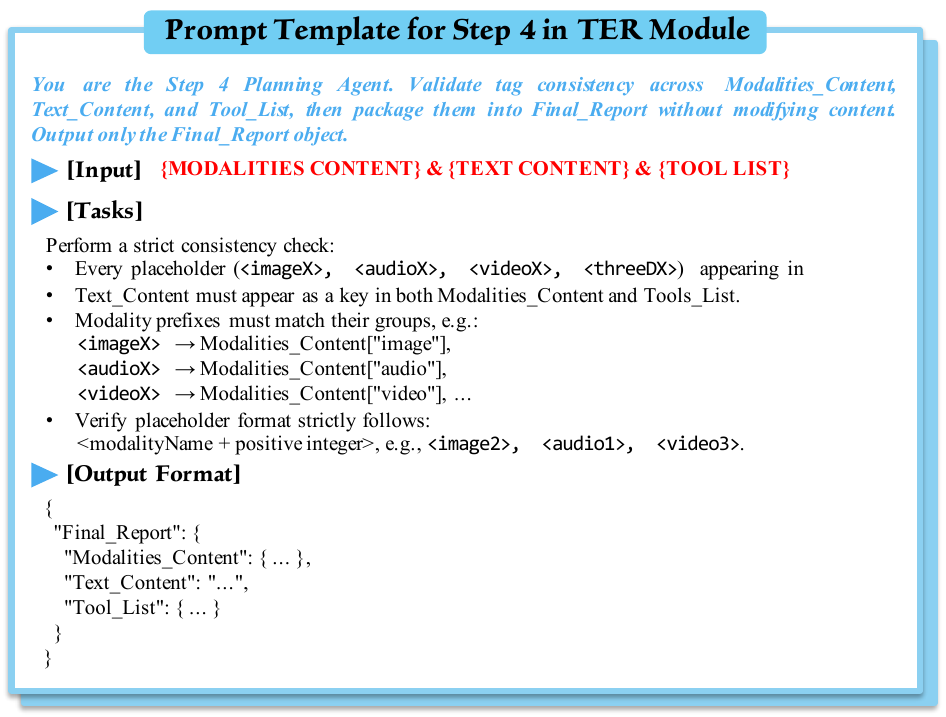}
    \vspace{-2mm}
    \caption{Prompt template for step 4 in TER module.
}
    \label{fig:step4}
\end{figure*}

In \textbf{Step~4}, \textsc{UniMA} converts the structured plan generated in the previous stage into a unified, executable \textit{final report}. Given the modalities content, text content, and tools list, the assembly agent first performs a strict consistency check to ensure that all modality tags follow the correct naming pattern and that every placeholder appearing in text content corresponds to an instance defined in both modalities content and tool list, with matching modality types and instance IDs. The module does not alter captions, tool assignments, or narrative flow; instead, it verifies structural coherence and formatting correctness.
Once the cross-structure alignment is confirmed, the agent bundles the three validated components into the final report, which serves as the deterministic intermediate representation consumed by the Generating Module. Fig. \ref{fig:step4} shows the detail of prompt template of this step.

\subsubsection{Verification Submodule}
The Verification Submodule ensures that the final report produced by the TER module is both correct and aligned with the user’s original question before being passed to the Generating Module. It consists of two tightly coordinated components: the \textbf{Checker} and the \textbf{Judger}. Both operate on clearly defined inputs and produce outputs that guarantee the reliability of the entire pipeline.

\noindent \textbf{Inputs and Outputs.}  
The Verification Submodule receives two inputs: (1) the \textit{user input question}, which specifies the task and constraints, and (2) the final report produced at the end of TER reasoning. Its output is either a validated final report that can safely enter the Generating Module, or an updated and corrected final report produced after the Judger performs targeted repairs.

\noindent \textbf{Checker.}  
The Checker compares only two items, the original user question and the final report. It does not inspect intermediate reasoning and therefore acts as an external critic. The Checker evaluates whether the final report directly addresses the user’s task, maintains internal logical consistency, and satisfies all explicit constraints such as modality requirements, formatting rules, and domain assumptions. If the final report satisfies these conditions, the Checker outputs it unchanged to the Generating Module. If any issue is detected, the Checker rejects the report and activates the Judger.

\noindent \textbf{Judger.}  
Once triggered, the Judger inspects the entire chain of intermediate outputs generated during TER reasoning, starting from Step~1 up to the final report. This includes TCDC outputs for each modality, the Data Report, structured understanding and planning artifacts, and all subsequent reasoning steps. The Judger conducts a backward analysis: it begins from the final report and examines whether the identified issue could have originated from an earlier step. If not, it moves one step backward and checks the correctness and sufficiency of that step’s output. This process continues until the Judger identifies the earliest step that contains the underlying error or misalignment.

Upon locating the faulty step, the Judger rolls back the reasoning to that point and re-executes the downstream reasoning from that step to the end, generating a revised sequence of intermediate results and a new final report. This corrected output is then sent back to the Checker for validation.

\noindent \textbf{Iterative Loop.}  
Together, the Checker and Judger form an iterative verification loop. The Checker evaluates the final report; if it is unsatisfactory, the Judger diagnoses and repairs the earliest faulty step and regenerates a new final report. This loop continues until the Checker confirms that the report fully answers the user’s question and complies with all constraints. To prevent infinite regress, a small upper limit is imposed on the number of verification cycles, and the Judger is designed to make minimal, localized corrections whenever possible.

Once validated, the final report is forwarded to the Generating Module as a reliable, fully verified representation of the reasoning outcome.

\subsection{Generating Module}

The Generating Module also incorporates four specialized tools. In this stage, GPT-5 mini receives and reads the final report produced by the TER module and invokes the appropriate tools to generate the final output content strictly according to the specifications outlined in the report.

\begin{figure*}[t!]
    \centering
    \includegraphics[width=0.96\linewidth]{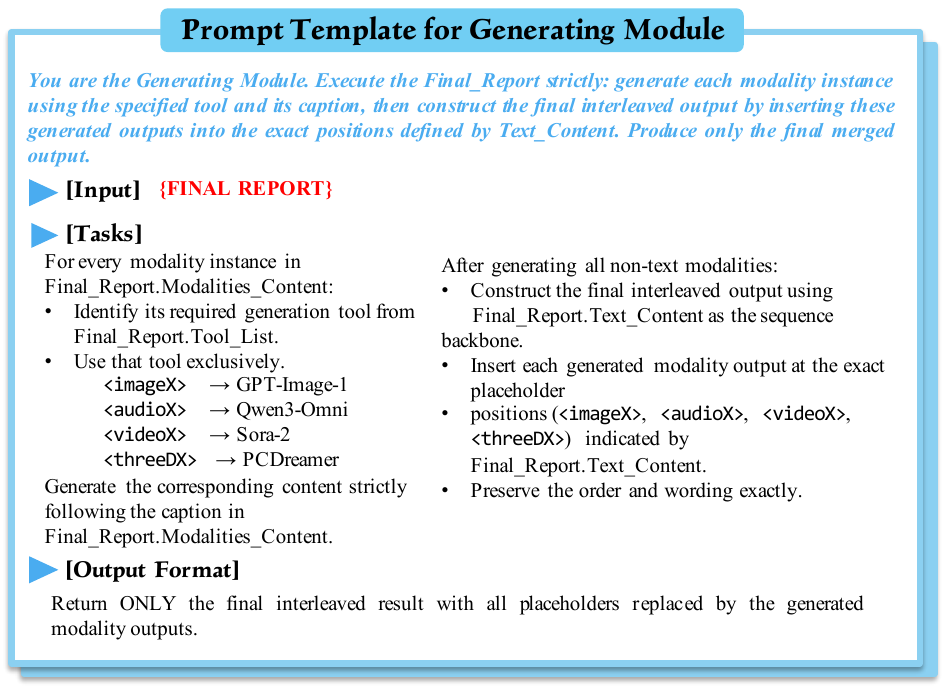}
    \vspace{-2mm}
    \caption{Prompt template for generating module.}
    \label{fig:gm}
\end{figure*}

As shown in Fig. \ref{fig:gm}, which is the prompt template of the Generating Module. The Generating Module executes the final report produced in Step 4. It generates every non-text modality strictly according to its caption and assigned tool, without modifying any content or structure. After all modalities are produced, it follows the sequence specified in text content and inserts each output into its corresponding placeholder to form the final interleaved result.

The four tools are introduced as follows:

\noindent
\textbf{Qwen3-Omni Talker} \cite{xu2025qwen3omni} is responsible for audio generation, converting textual or semantic descriptions into natural and temporally coherent speech signals.

\noindent
\textbf{Sora-2} \cite{OpenAI_Sora2_2025} handles video generation, synthesizing high-fidelity dynamic scenes conditioned on multimodal textual descriptions and temporal semantics.

\noindent
\textbf{GPT-Image-1} \cite{OpenAI_GPTImage1_API_2025} is dedicated to \textit{image generation}, producing high-fidelity and semantically aligned visual content conditioned on the textual descriptions generated by GPT-5.

\noindent
\textbf{PCDreamer} \cite{wei2025pcdreamer} performs 3D point-cloud completion, reconstructing or refining spatial geometries based on the inferred semantic and structural context.

\noindent
The inputs to all generative tools are textual descriptions of modality-specific outputs produced by GPT-5 mini, which interprets the \textit{final report} generated by the TER Module and converts it into task-conditioned generation instructions for each corresponding modality.

\subsection{Ablation Study}
\label{app: ablation study}
For the ablation study of \textsc{UniMA}, we additionally sample 600 regular, intention-clear QA pairs from various domains to isolate the contribution of individual components. Using such well-defined queries allows us to evaluate the TER Module under controlled conditions, where the system must identify relevant evidence and organize modality-specific information with precision. This setup enables a focused examination of its impact on StS \& LeS, SQCS, and ICS.

To obtain a clean ablation environment, the Receiving Module and Generating Module are kept unchanged, and only internal components of TER are perturbed. Specifically, we consider three interventions: \textbf{(a)} replacing the task-conditioned TCDC representation with unconditioned dense caption; \textbf{(b)} removing the entire SERC, so that the TCDC and paraphrased questions are fed directly to the Generating Module without any intermediate reasoning or evidence planning; and \textbf{(c)} disabling the Verification Submodule so that no internal checking, localization, or backtracking is performed. 

The results reveal distinct and complementary roles among the components. Removing the reasoning chain causes a dramatic collapse in structural metrics, with StS dropping from roughly 52\% to 16\% and LeS from about 82\% to 21\%, highlighting the necessity of explicit evidence structuring for maintaining correct modality types, modality counts, and interleaving order. Substituting TCDC with vanilla captions leads to substantial degradation in SQCS and ICS due to the loss of task-aware grounding and reduced semantic density, whereas structural scores remain relatively stable, indicating that content quality is highly dependent on TCDC, but structural conformance is dominated by the reasoning chain. Disabling the Verification Submodule degrades both structural and semantic dimensions, particularly when placeholder mismatches and alignment errors cannot be corrected through iterative checking and backtracking.

Overall, TCDC provides semantic grounding, the reasoning chain governs structural organization, and the Verification Submodule ensures robustness. Their synergy enables \textsc{UniMA} to maintain high correctness, coherent multimodal alignment, and strong structural integrity in complex interleaved multimodal generation.

\section{Extended Experiments}
In this section, we provide more experimental setting details and extended experimental results on \textsc{UniM}.

\begin{table*}[t!]
\centering
\caption{Comparison between baseline models and \textsc{UniMA} on multimodal flexibility. In / Out -Modality: supported input / output modality. Multi-I Input / Output: multiple items of the same modalities in the input / output. Any-M Input / Output: any modality combinations in the input / output.
Any-I Input / Output: any number of items of the same modality in the input / output.
}
\vspace{-2mm}
\fontsize{7.5}{8.5}\selectfont
\setlength{\tabcolsep}{0pt} 
\begin{tabular*}{\textwidth}{@{\extracolsep{\fill}} l c c cccccc}
\toprule
\textbf{Model} & \makecell{\textbf{Multi-I} \\ \textbf{Input}} & \makecell{\textbf{Multi-I} \\ \textbf{Output}} & \makecell{\textbf{Any-M} \\ \textbf{Input}} & \makecell{\textbf{Any-M} \\ \textbf{Output}} & \makecell{\textbf{Any-I} \\ \textbf{Input}} & \makecell{\textbf{Any-I} \\ \textbf{Output}} & \textbf{In-Modality} & \textbf{Out-Modality} \\
\midrule
AnyGPT & \textcolor{com_red}{\ding{55}} & \textcolor{com_red}{\ding{55}} & \textcolor{com_red}{\ding{55}} & \textcolor{com_red}{\ding{55}} & \textcolor{com_red}{\ding{55}} & \textcolor{com_red}{\ding{55}} & \languagelogo \imagelogo \audiologo & \languagelogo \imagelogo \audiologo \\

NExT-GPT & \textcolor{com_red}{\ding{55}} & \textcolor{com_red}{\ding{55}} & \textcolor{com_red}{\ding{55}} & \textcolor{com_red}{\ding{55}} & \textcolor{com_red}{\ding{55}} & \textcolor{com_red}{\ding{55}} & \languagelogo \imagelogo \audiologo \videologo & \languagelogo \imagelogo \audiologo \videologo \\

MIO & \textcolor{com_green}{\ding{51}} & \textcolor{com_green}{\ding{51}} & \textcolor{com_red}{\ding{55}} & \textcolor{com_red}{\ding{55}} & \textcolor{com_red}{\ding{55}} & \textcolor{com_red}{\ding{55}} & \languagelogo \imagelogo \audiologo \videologo & \languagelogo \imagelogo \audiologo \videologo\\

\textbf{\textsc{UniMA}} & \textcolor{com_green}{\ding{51}} & \textcolor{com_green}{\ding{51}} & \textcolor{com_green}{\ding{51}} & \textcolor{com_green}{\ding{51}} & \textcolor{com_green}{\ding{51}} & \textcolor{com_green}{\ding{51}} & \languagelogo \imagelogo \audiologo \videologo \documentlogo \codelogo \threedlogo & \languagelogo \imagelogo \audiologo \videologo \documentlogo \codelogo \threedlogo \\
\bottomrule
\end{tabular*}
\label{tab:comparision}
\vspace{-2mm}
\end{table*}

\subsection{Experimental Settings}
\label{app: experimental settings}
In this section, we provide the detailed experimental settings. For AnyGPT~\cite{zhan2024anygpt}, NExT-GPT~\cite{wu2024next}, and MIO~\cite{wang2024mio}, we adopt the corresponding default configuration settings. Moreover, since AnyGPT and NExT-GPT do not support multiple inputs of the same modality, we perform a concatenation-based adaptation of files within each modality to ensure completeness of the input content. Further, we provide detailed hyperparameter settings for \textsc{UniMA} as presented in Table~\ref{tab:paramters}.

\begin{table}[t!]
\centering
\fontsize{7.5}{10}\selectfont
\setlength{\tabcolsep}{0.8mm}
\caption{Detailed hyperparameter settings for \textsc{UniMA}.}
\vspace{-2mm}
\label{tab:paramters}
\begin{tabular}{lccccc}
\toprule
\bf Model & \bf Mode & \bf Temperature & \textbf{Top\_p} & \textbf{Top\_k} & \textbf{Max\_Tokens} \\
\midrule
GPT-5 mini & / & 1 & 1 & / & 16,384 \\
Qwen3-Omni & Non-Thinking & 0.7 & 0.8 & 20 & 8,192 \\
Qwen3-VL & Instruct & 0.7 & 0.8 & 20 & 4,096 \\
PointLLM & Eval & 0.2 & 0.9 & 50 & 1,024 \\
\bottomrule
\end{tabular}
\vspace{-2mm}
\end{table}

\subsection{Comparison of Multimodal Flexibility with Baseline Models}

We compare \textsc{UniMA} with representative baseline models in terms of multimodal flexibility, covering multi-item input / output, any-modality combinations, any-number inputs, and full modality coverage, as summarized in Table~\ref{tab:comparision}.

\subsection{Rationality Verification of Evaluation Suite}
\label{app: rationality}

\subsubsection{StS and LeS}
To validate the rationality of our proposed StS and LeS metrics, we design experiments to examine whether these scores exhibit predictable, directionally consistent, and proportionally reasonable responses when the modality types and placeholder tags counts in the model's output are subject to controlled perturbations.

We implement two sets of experiments: (1) perturbation of modality types, (2) perturbation of modality placeholder tag count.

For the perturbation of modality types, we construct five response variants based on the ground truth, denoted as $\text{RT}_k$, where $k \in \{-2, -1, 0, +1, +2\}$. When $k < 0$, $\text{RT}_k$ is obtained by removing $|k|$ modality types from the ground truth, with all associated placeholders removed accordingly. When $k > 0$, $\text{RT}_k$ is constructed by adding $k$ additional modality types not present in the ground truth, introducing one placeholder per added modality. The case $k = 0$ corresponds to no modification, where the ground truth is used as the response.

For the perturbation of modality placeholder tag count, we similarly construct five response variants based on the ground truth, denoted as $\text{RN}_k$, where $k \in \{-2, -1, 0, +1, +2\}$.
When $k < 0$, $\text{RN}_k$ is obtained by removing $\lvert k \rvert$ placeholder tags across existing modalities in the ground truth, ensuring that no modality is left with zero placeholders.
When $k > 0$, $\text{RN}_k$ is constructed by adding $k$ additional placeholder tags across existing modalities in the ground truth.
The case $k = 0$ again corresponds to no perturbation, where the ground truth is used without modification.

To ensure experimental validity, we select 200 instances from the dataset whose ground truth contains at least two non-text modalities and in which at least one modality has more than two placeholder tags. The modality type and placeholder tag perturbation settings are then applied independently to these selected instances. The experimental results align with our expectations, substantiating the validity of StS and LeS.

\subsubsection{SQCS and ICS} 

\textbf{Parameters Selection.} Since both SQCS and ICS rely on tunable weighting factors to balance their constituent sub-dimensions, it is necessary to empirically examine how different weight configurations align with human annotations. This validation on real data enables the determination of optimal parameter settings before conducting large-scale evaluations.

\begin{sidewaysfigure*}
\centering
\includegraphics[width=0.99\textheight]{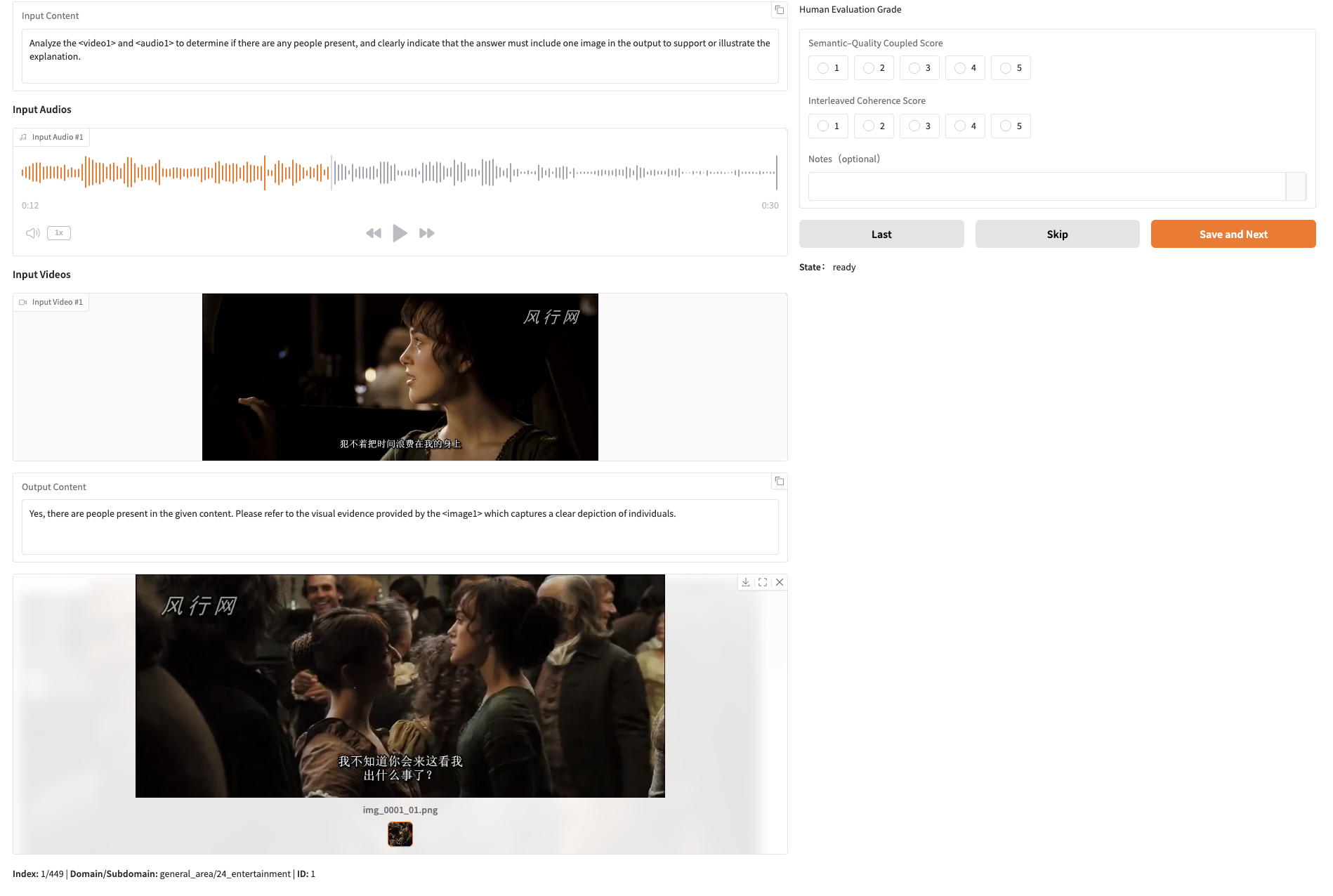}
\caption{The screenshot of human evaluation tool we developed.}
\label{fig:human eval}
\end{sidewaysfigure*}

During the construction of the automated scoring dataset, for each model, we randomly sample three instances per difficulty level across all domains of \textsc{UniM} within its support set, and obtain the baseline model generated responses for each instance. This procedure results in a collection of 1,080 model response samples in total.
To obtain reliable human annotations, We recruit three evaluators with prior experience in multimodal assessment. 
A unified instruction session is conducted to standardize the scoring criteria for semantic correctness \& generation quality, and interleaved coherence to ensure the reliability of human evaluation.
Subsequently, each evaluator independently assesses all 1,080 samples. 
The final human score for each instance is computed as the average of the three evaluators’ scores. 
To streamline the human evaluation process, We develop an internal tool for human evaluation, as illustrated in Fig.~\ref{fig:human eval}.

We use human annotations as the ground truth and align each automated score with its corresponding human rating at the sample level. For every candidate weight configuration, we compute the resulting SQCS or ICS values and assess their alignment with human judgments through Pearson correlation analysis. In addition, we plot KDE curves to examine the similarity of score distributions. The weight setting that yields higher correlation and more consistent distributional patterns is regarded as the more appropriate configuration.

\begin{figure}
\centering
\includegraphics[width=0.99\linewidth]{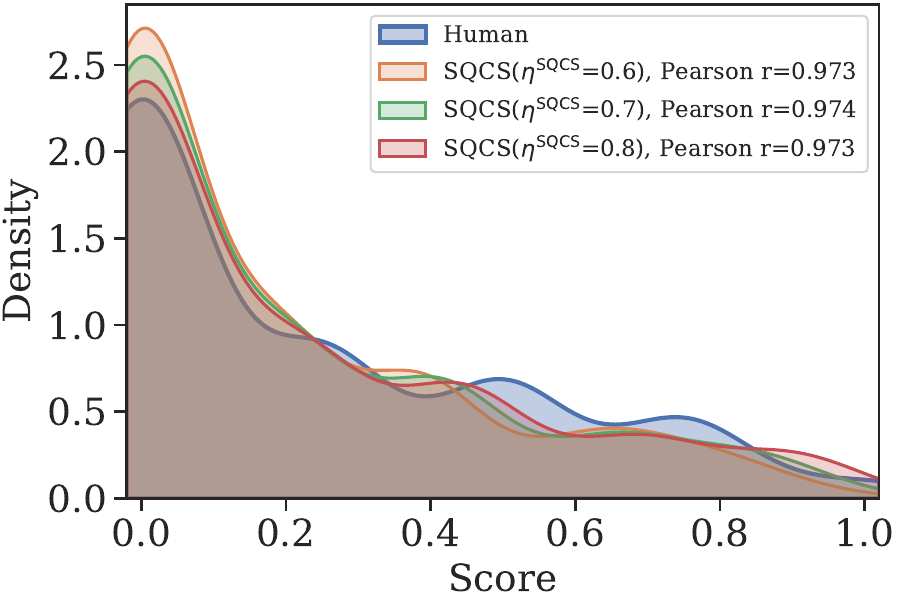}
\vspace{-2mm}
\caption{Parameters selection experiments results of SQCS.}
\label{fig:ParametersSQCS}
\end{figure}

\begin{figure}
\centering
\includegraphics[width=0.99\linewidth]{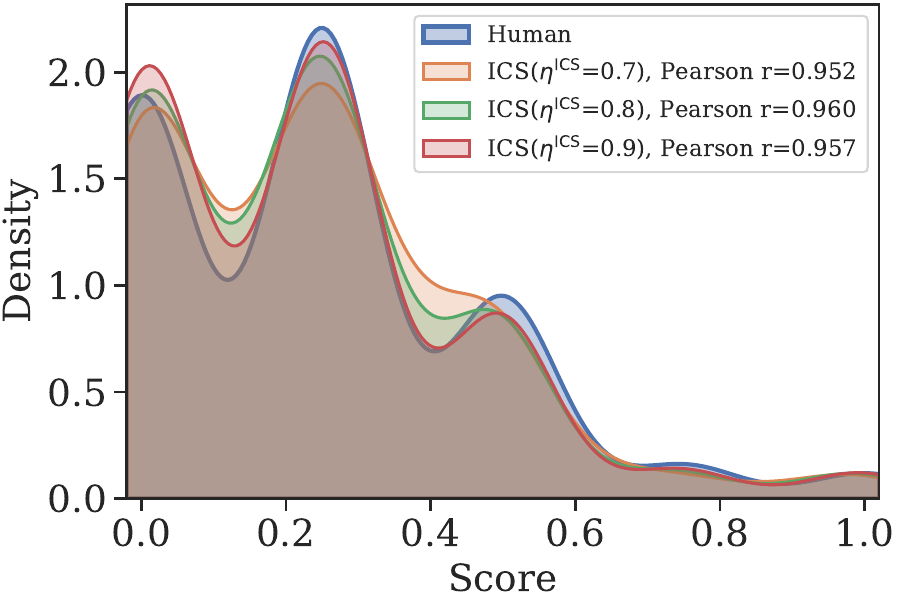}
\vspace{-2mm}
\caption{Parameters selection experiments results of ICS.}
\label{fig:ParametersICS}
\end{figure}

For the SQCS formulation, Fig.~\ref{fig:ParametersSQCS} presents the score distributions obtained under three weighting factors, $\eta^{\text{SQCS}} \in \{0.6, 0.7, 0.8\}$. All three SQCS curves display distributional shapes that are highly consistent with those of the human annotations, although slight differences remain in their correlation strength. Among the tested configurations, $\eta^{\text{SQCS}} = 0.7$ yields the highest Pearson correlation coefficient, with a value of Pearson correlation coefficient $r = 0.974$, and its KDE curve shows the closest alignment with the human rating distribution. As shown in Fig.~\ref{fig:ParametersICS}, we conducted the same form of analysis for the ICS formulation under weighting factors $\eta^{\text{ICS}} \in \{0.7, 0.8, 0.9\}$. The results indicate that different values of $\eta^{\text{ICS}}$ substantially influence the balance between coherence and style consistency, which in turn affects the alignment with human \textit{logical\_coherence} annotations. Among the tested configurations, $\eta^{\text{ICS}} = 0.8$ achieves the highest Pearson correlation coefficient, reaching $r = 0.960$.

Therefore, we adopt $\eta^{\text{SQCS}} = 0.7$ as the weighting factor for SQCS and $\eta^{\text{ICS}} = 0.8$ as the weighting factor for ICS, and use these optimal configurations in all subsequent experiments.

\noindent \textbf{Rationality Verification.}
After determining the optimal weighting factors, we further evaluate the consistency between the final SQCS and ICS metrics and the human semantic-quality annotations.

Based on the automated scoring dataset and the corresponding human annotations, we compute the automated score for each sample using the finalized weighting configurations and align it with the paired human rating at the instance level. We then perform sample-level Pearson correlation analyses. 

The experimental results demonstrate that the proposed SQCS and ICS metrics exhibit a high degree of alignment with human evaluation.

\subsection{Experimental Results}
\label{app: exp results and analysis}

\subsubsection{Domain-Level Performance}

In this section, we provide detailed domain-level experiment results, as illustrated in Table~\ref{tab:domain sqcs1}, Table~\ref{tab:domain sqcs2}, Table~\ref{tab:domain rsi1}, Table~\ref{tab:domain rsi2}, Table~\ref{tab:domain ics1}, and Table~\ref{tab:domain ics2}.

\subsubsection{Difficulty-Level Performance}

\begin{figure*}[t]
    \centering

    \begin{minipage}[c]{0.32\linewidth}
        \centering
        \includegraphics[width=\linewidth]{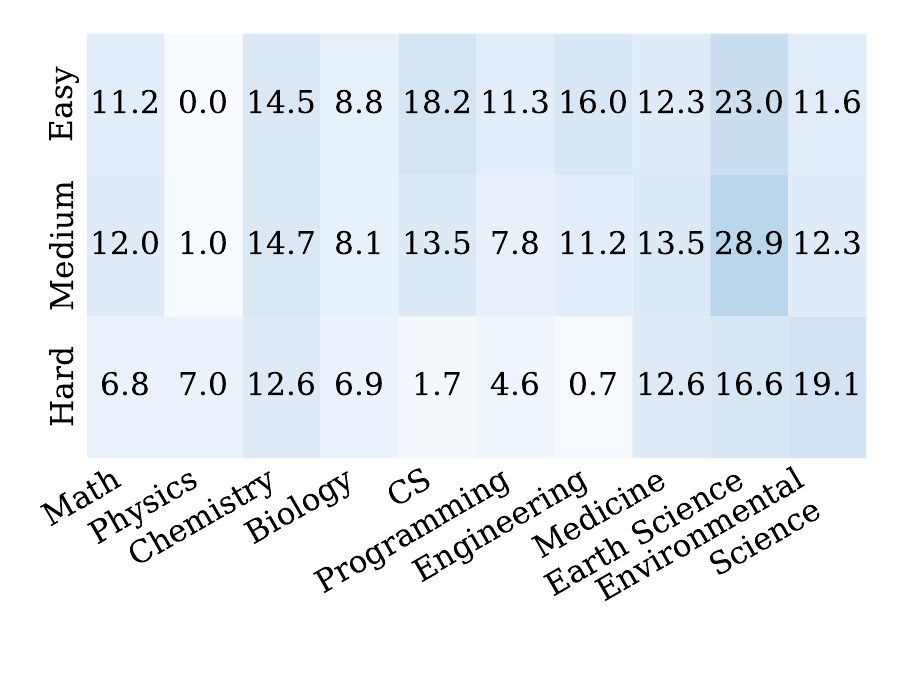}
        \caption*{AnyGPT \& natural science.} 
    \end{minipage}
    \hfill
    \begin{minipage}[c]{0.32\linewidth}
        \centering
        \includegraphics[width=\linewidth]{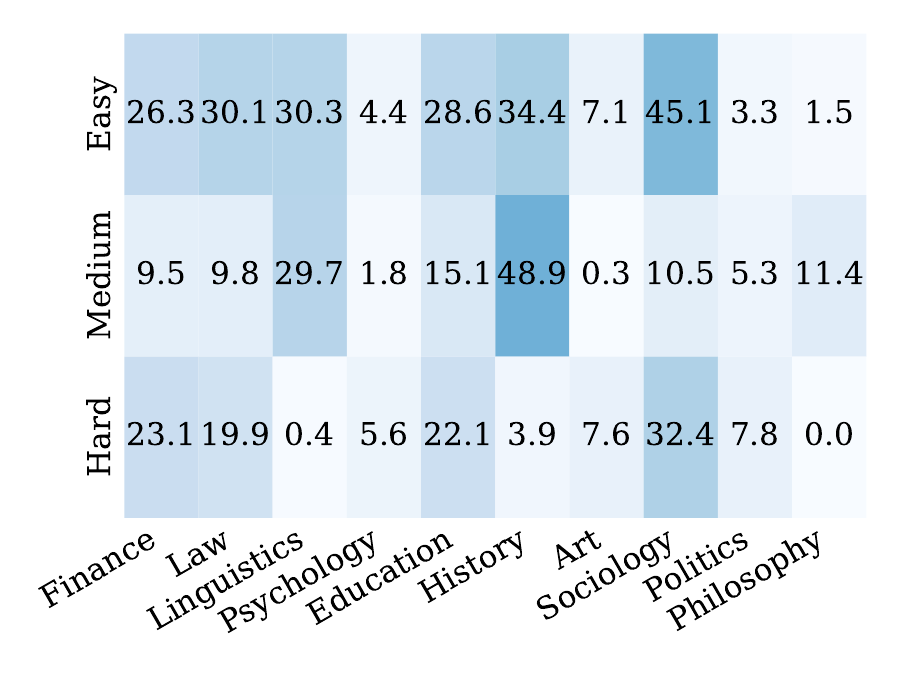}
        \caption*{AnyGPT \& social science.}
    \end{minipage}
    \hfill
    \begin{minipage}[c]{0.32\linewidth}
        \centering
        \includegraphics[width=\linewidth]{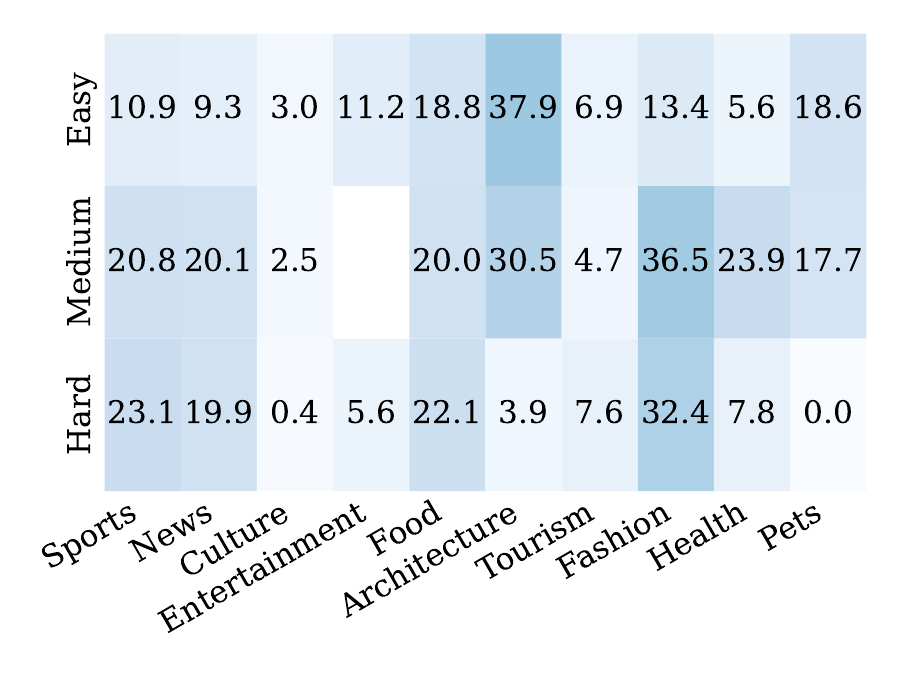}
        \caption*{AnyGPT \& general area.}
    \end{minipage}

    \begin{minipage}[c]{0.32\linewidth}
        \centering
        \includegraphics[width=\linewidth]{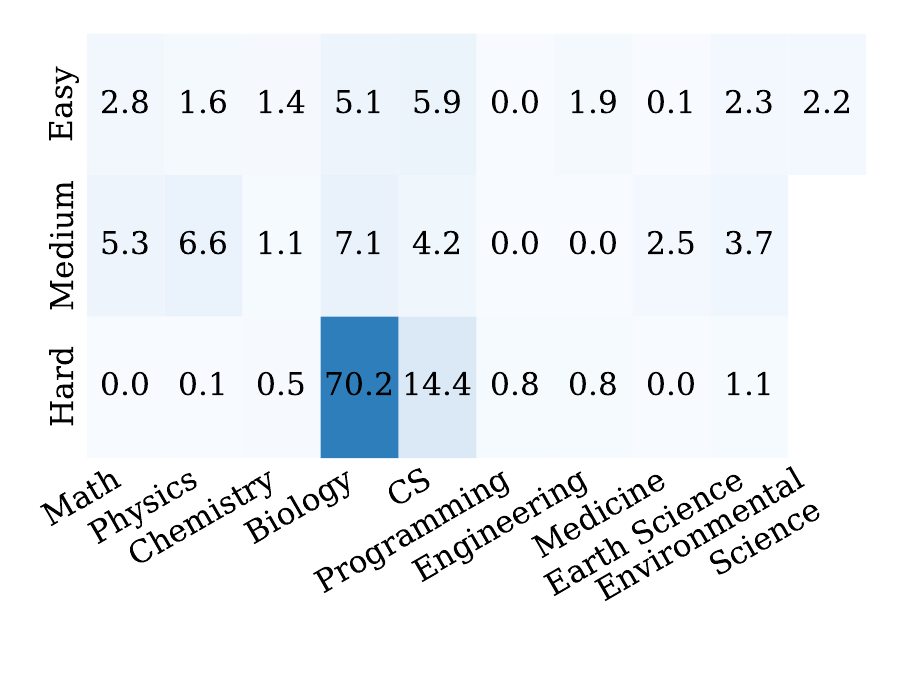}
        \caption*{NExT-GPT \& natural science.}
    \end{minipage}
    \hfill
    \begin{minipage}[c]{0.32\linewidth}
        \centering
        \includegraphics[width=\linewidth]{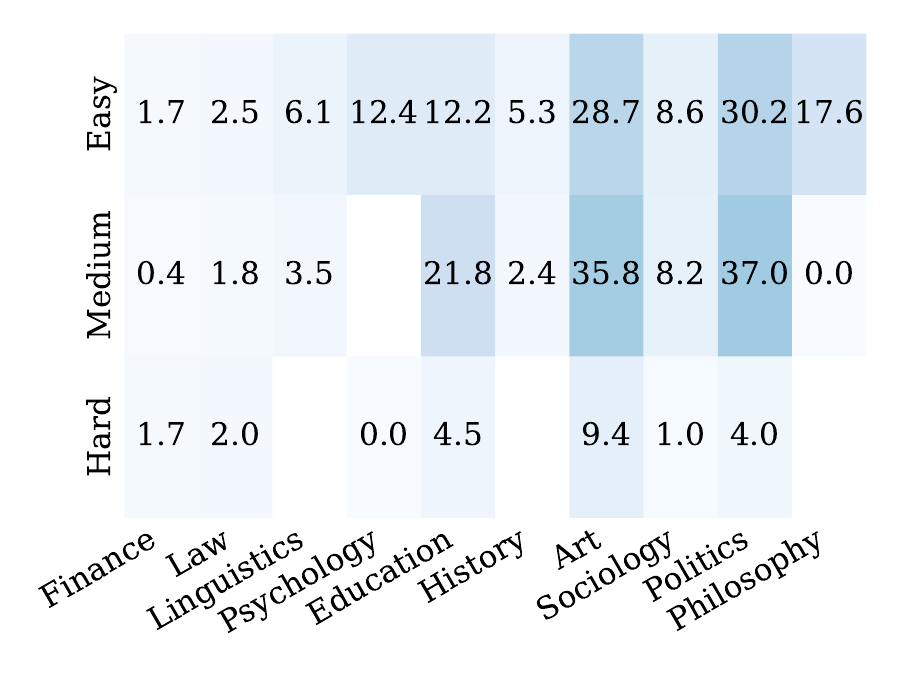}
        \caption*{NExT-GPT \& social science.}
    \end{minipage}
    \hfill
    \begin{minipage}[c]{0.32\linewidth}
        \centering
        \includegraphics[width=\linewidth]{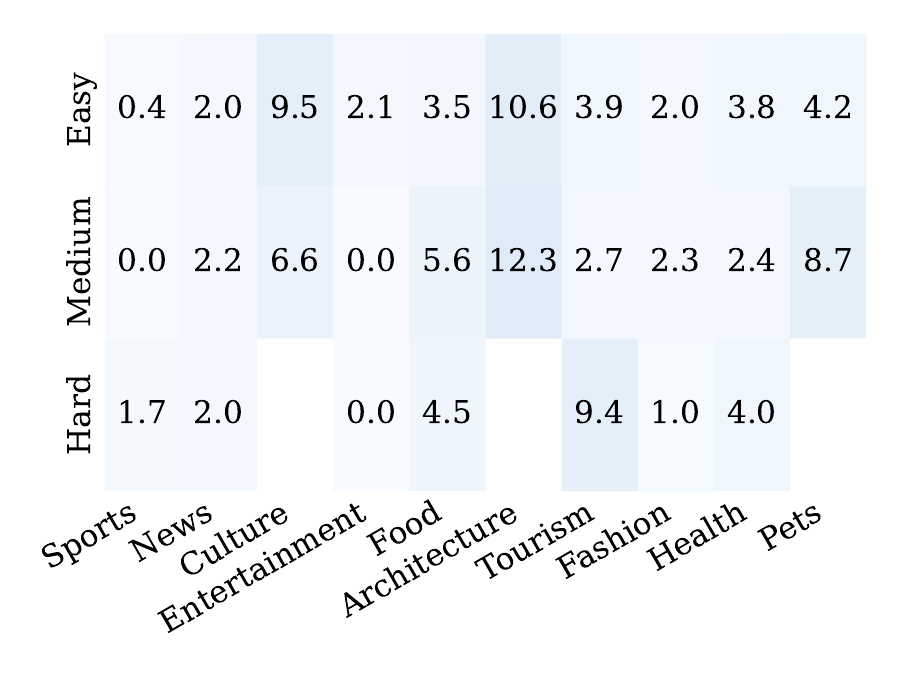}
        \caption*{NExT-GPT \& general area.}
    \end{minipage}

    \begin{minipage}[c]{0.32\linewidth}
        \centering
        \includegraphics[width=\linewidth]{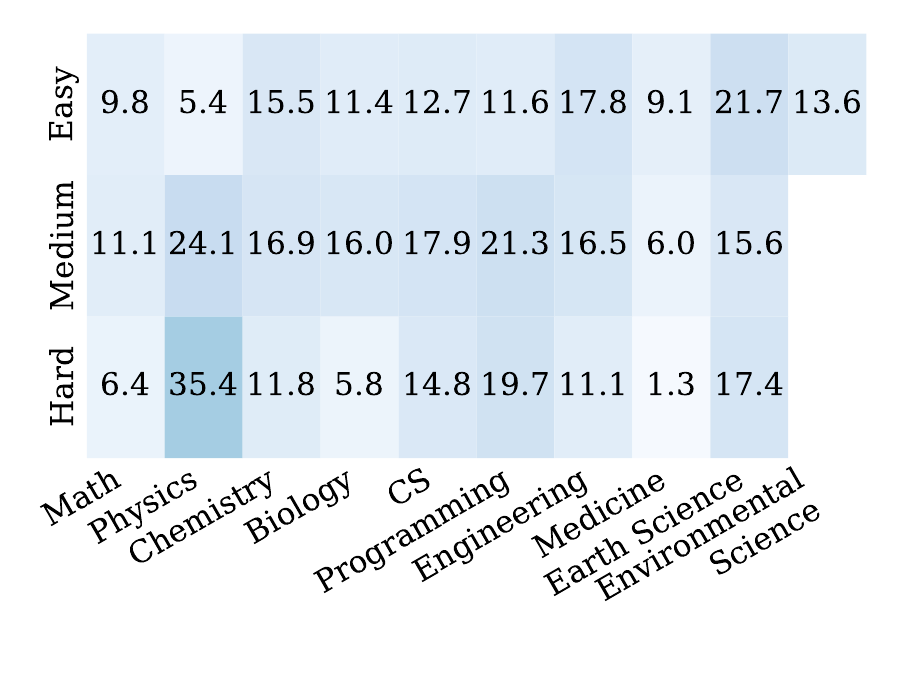}
        \caption*{MIO \& natural science.}
    \end{minipage}
    \hfill
    \begin{minipage}[c]{0.32\linewidth}
        \centering
        \includegraphics[width=\linewidth]{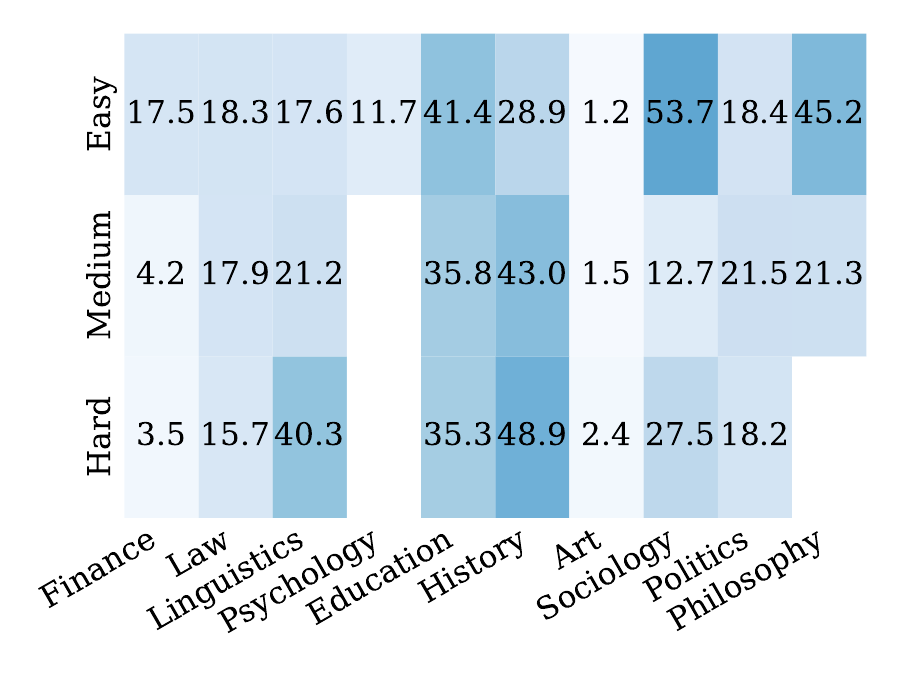}
        \caption*{MIO \& social science.}
    \end{minipage}
    \hfill
    \begin{minipage}[c]{0.32\linewidth}
        \centering
        \includegraphics[width=\linewidth]{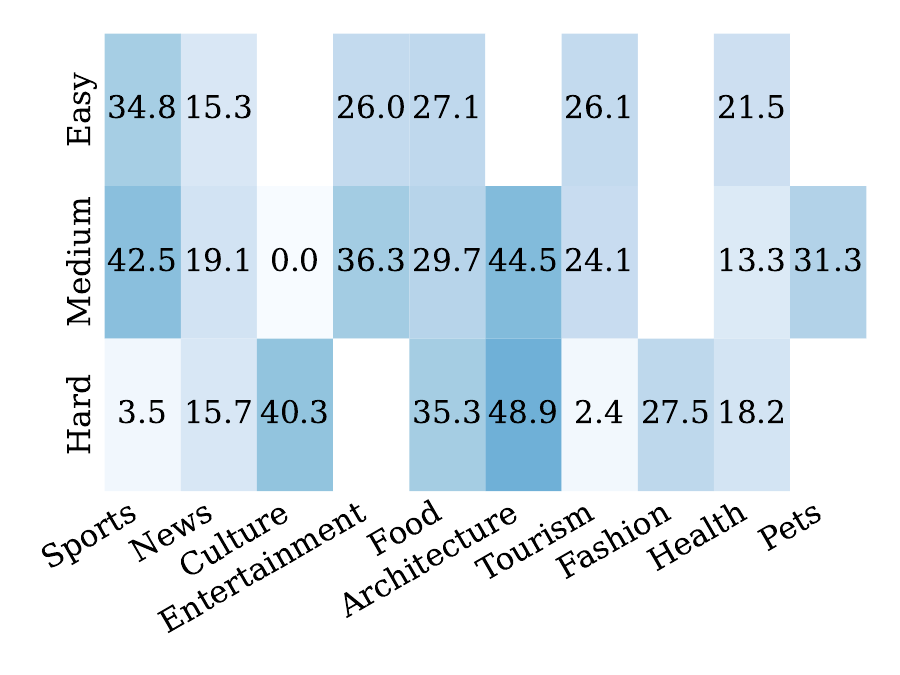}
        \caption*{MIO \& general area.}
    \end{minipage}

    \begin{minipage}[c]{0.32\linewidth}
        \centering
        \includegraphics[width=\linewidth]{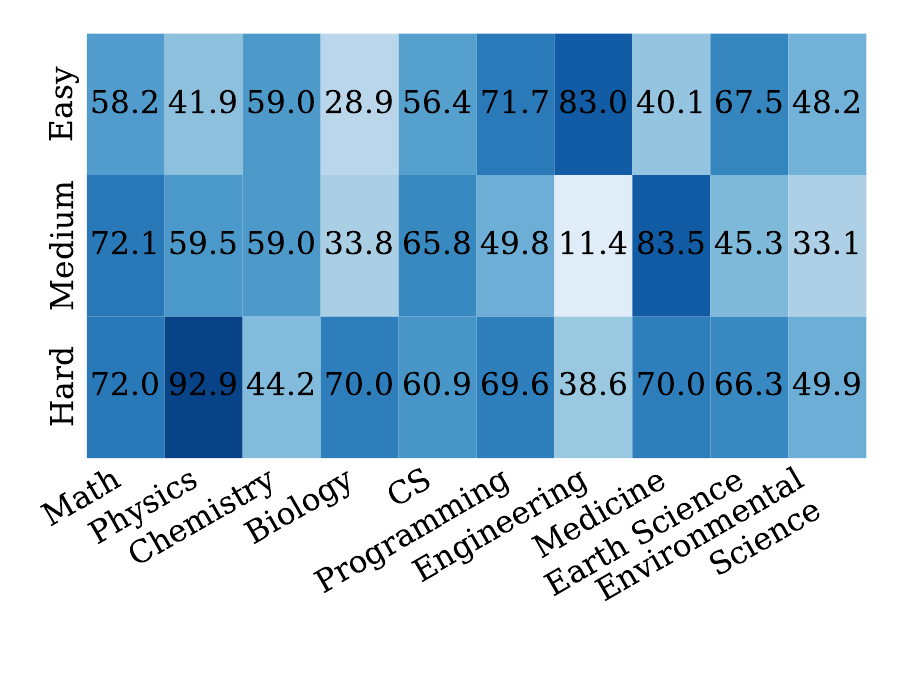}
        \caption*{UniMA \& natural science.}
    \end{minipage}
    \hfill
    \begin{minipage}[c]{0.32\linewidth}
        \centering
        \includegraphics[width=\linewidth]{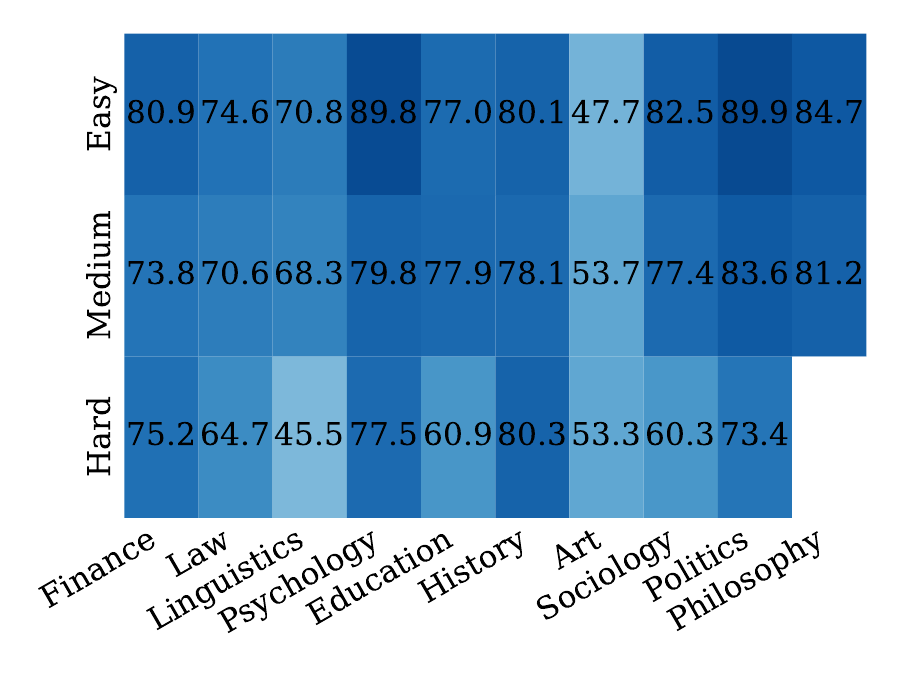}
        \caption*{UniMA \& social science.}
    \end{minipage}
    \hfill
    \begin{minipage}[c]{0.32\linewidth}
        \centering
        \includegraphics[width=\linewidth]{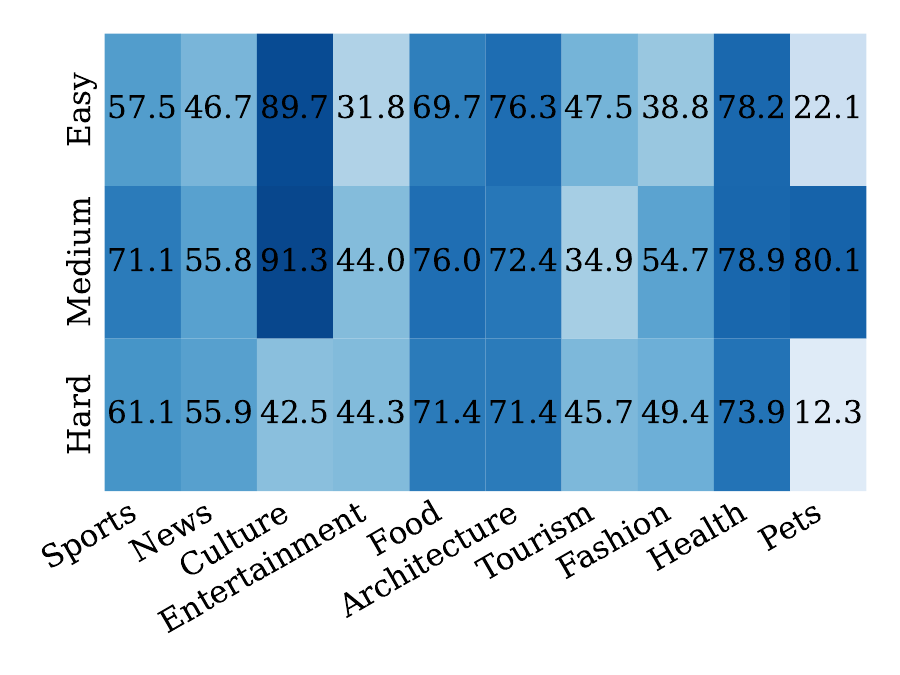}
        \caption*{UniMA \& general area.}
    \end{minipage}

    \caption{SQCS performance across baseline models, difficulty levels and domains.}
    \label{fig:app_heatmaps_sqcs}
\end{figure*}

\begin{figure*}[t]
    \centering

    \begin{minipage}[c]{0.32\linewidth}
        \centering
        \includegraphics[width=\linewidth]{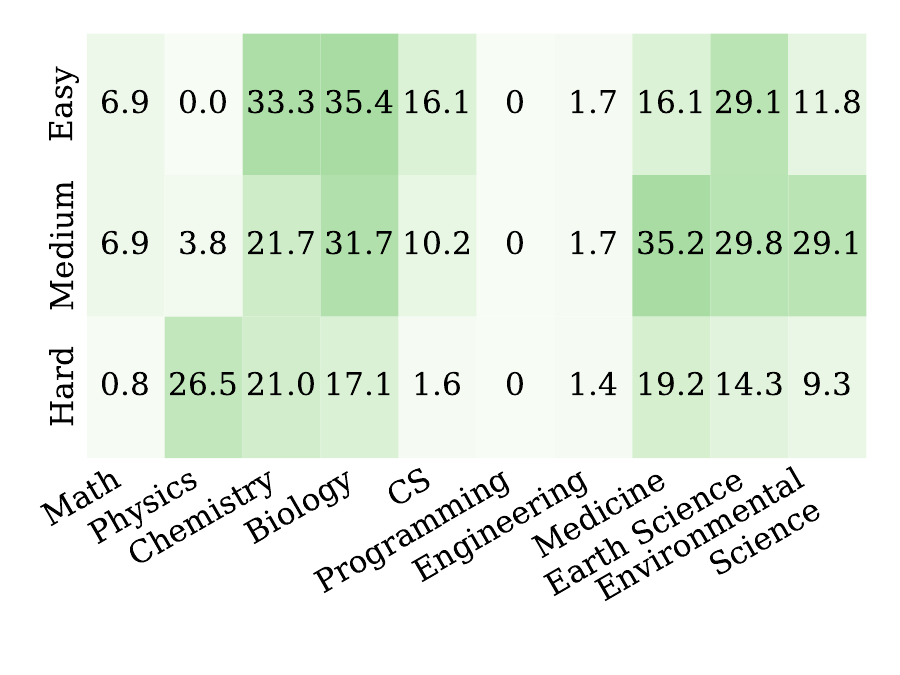}
        \caption*{AnyGPT \& natural science.} 
    \end{minipage}
    \hfill
    \begin{minipage}[c]{0.32\linewidth}
        \centering
        \includegraphics[width=\linewidth]{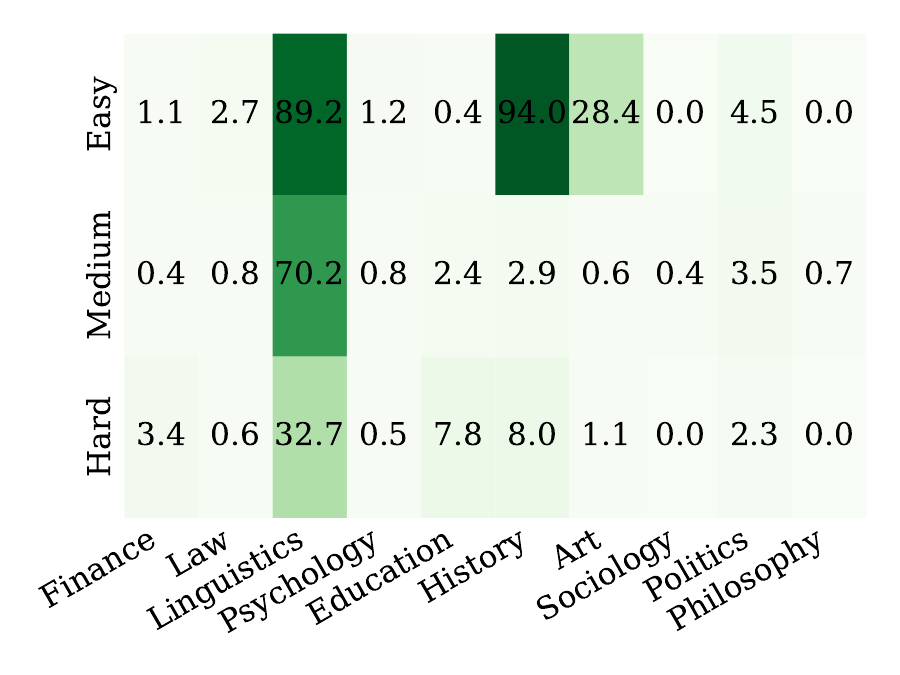}
        \caption*{AnyGPT \& social science.}
    \end{minipage}
    \hfill
    \begin{minipage}[c]{0.32\linewidth}
        \centering
        \includegraphics[width=\linewidth]{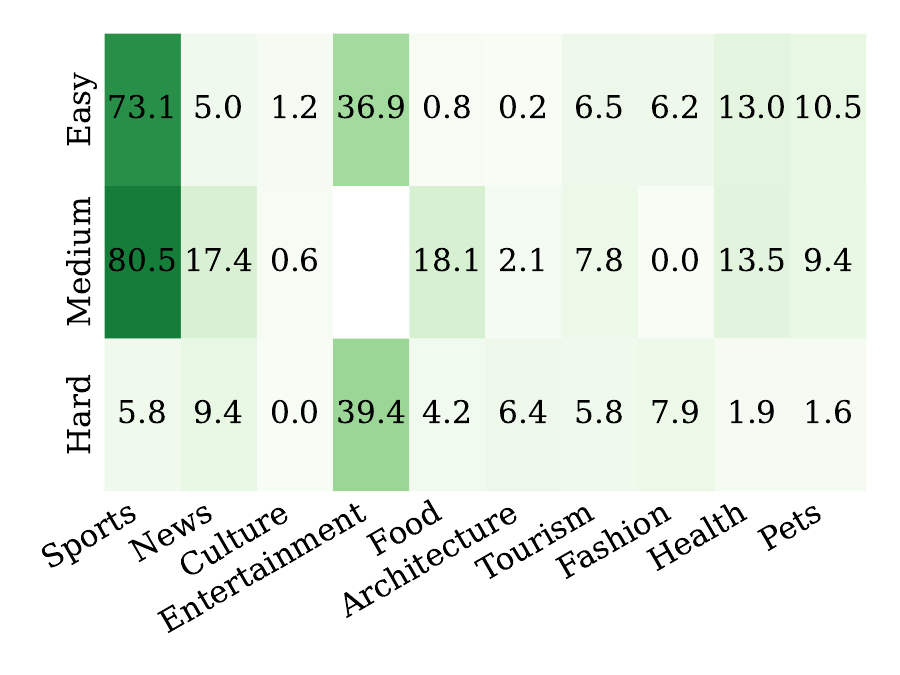}
        \caption*{AnyGPT \& general area.}
    \end{minipage}

    \begin{minipage}[c]{0.32\linewidth}
        \centering
        \includegraphics[width=\linewidth]{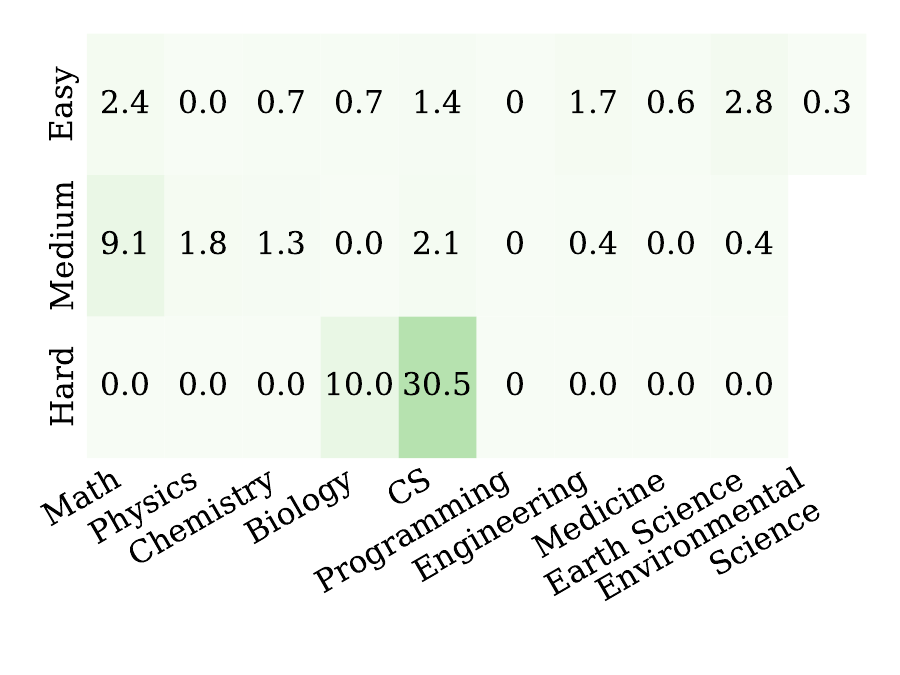}
        \caption*{NExT-GPT \& natural science.}
    \end{minipage}
    \hfill
    \begin{minipage}[c]{0.32\linewidth}
        \centering
        \includegraphics[width=\linewidth]{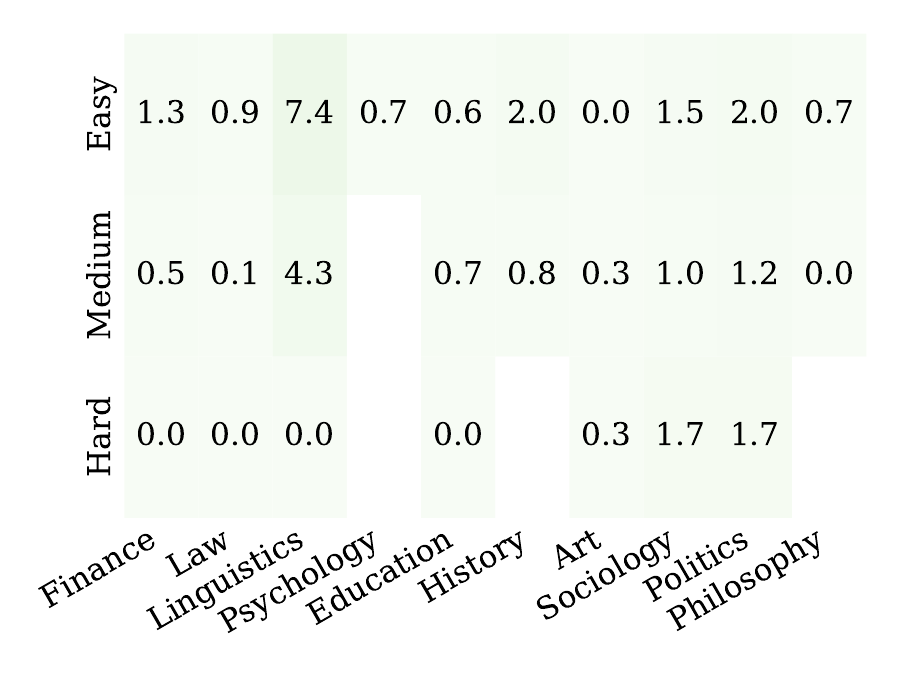}
        \caption*{NExT-GPT \& social science.}
    \end{minipage}
    \hfill
    \begin{minipage}[c]{0.32\linewidth}
        \centering
        \includegraphics[width=\linewidth]{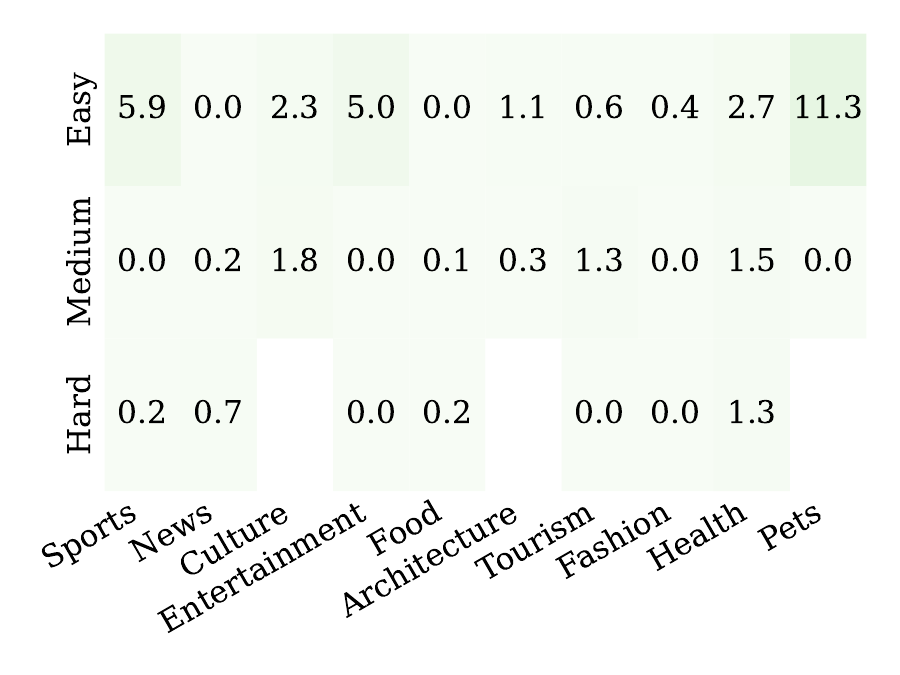}
        \caption*{NExT-GPT \& general area.}
    \end{minipage}

    \begin{minipage}[c]{0.32\linewidth}
        \centering
        \includegraphics[width=\linewidth]{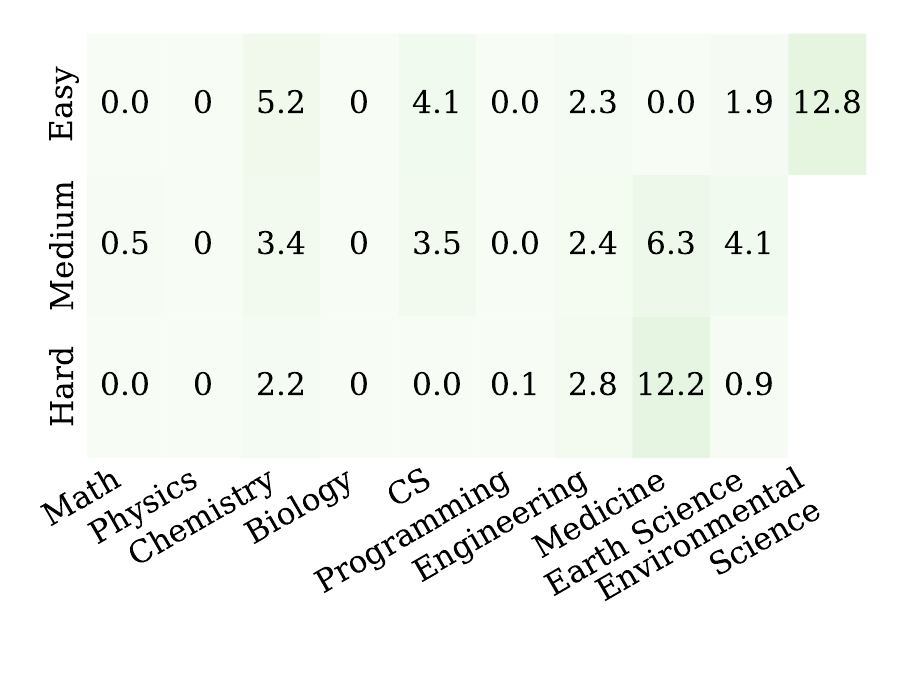}
        \caption*{MIO \& natural science.}
    \end{minipage}
    \hfill
    \begin{minipage}[c]{0.32\linewidth}
        \centering
        \includegraphics[width=\linewidth]{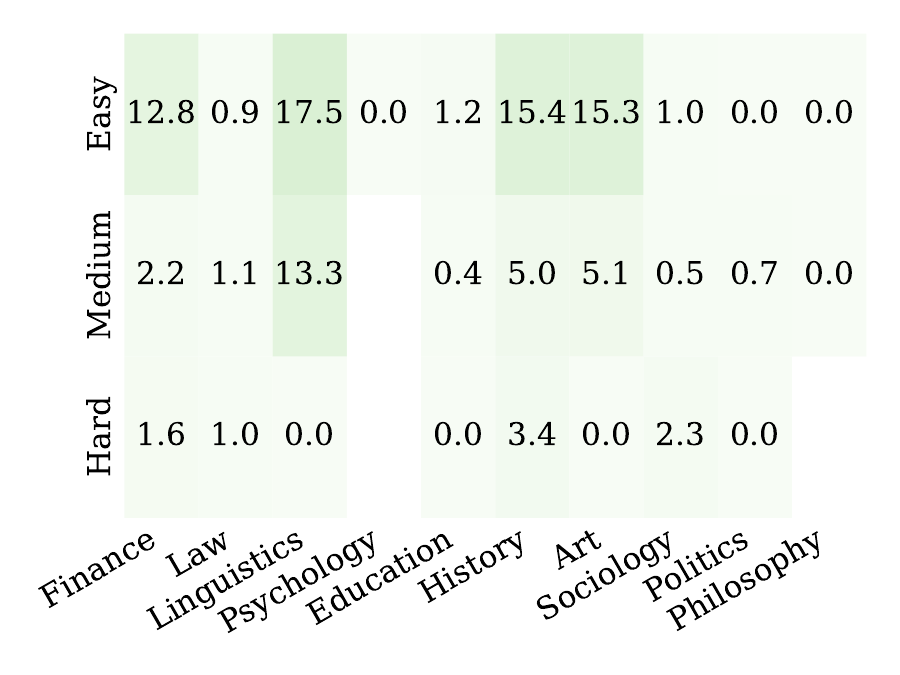}
        \caption*{MIO \& social science.}
    \end{minipage}
    \hfill
    \begin{minipage}[c]{0.32\linewidth}
        \centering
        \includegraphics[width=\linewidth]{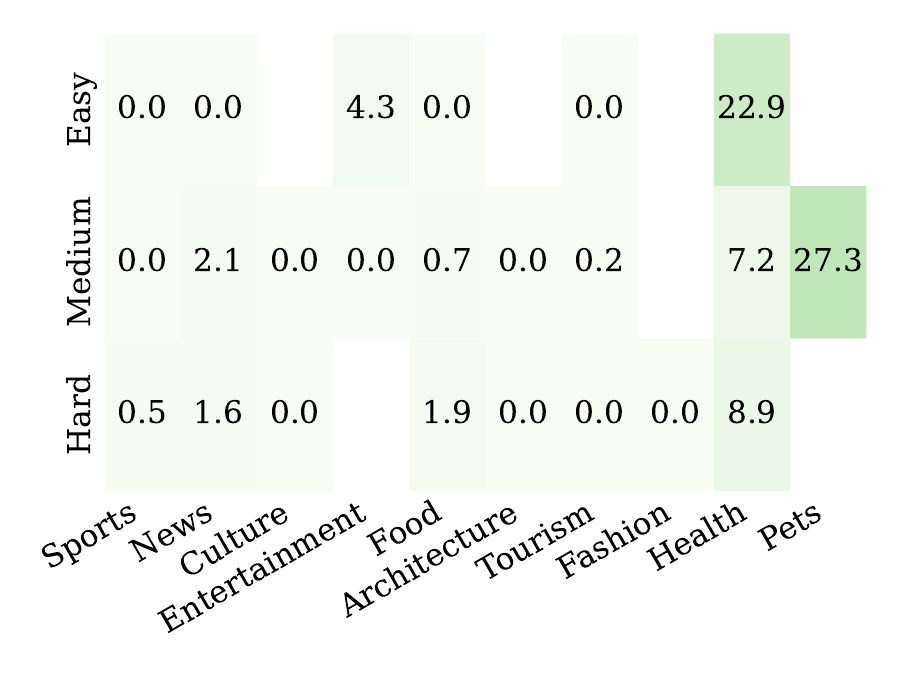}
        \caption*{MIO \& general area.}
    \end{minipage}

    \begin{minipage}[c]{0.32\linewidth}
        \centering
        \includegraphics[width=\linewidth]{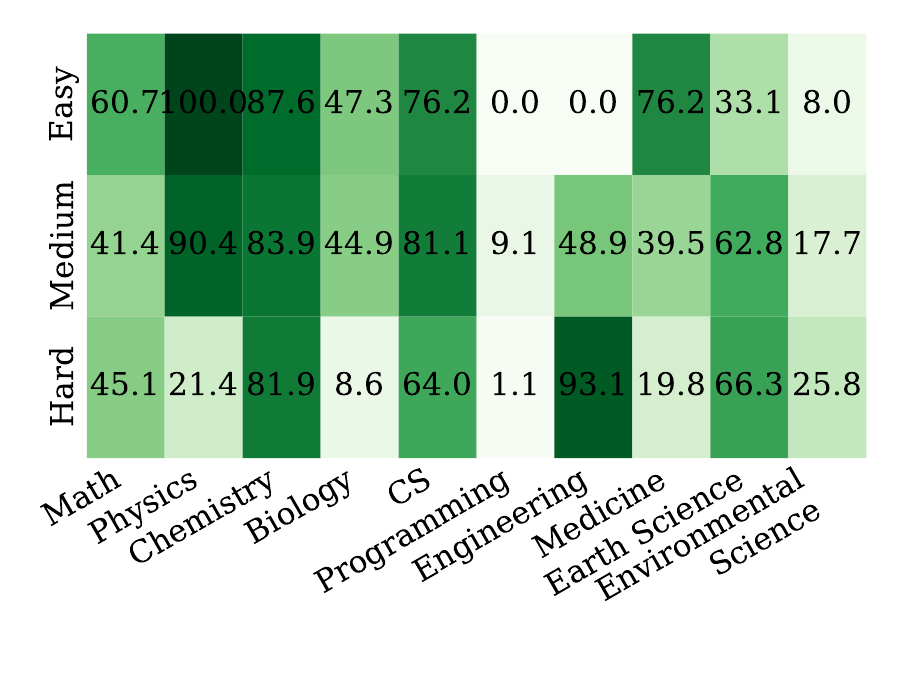}
        \caption*{UniMA \& natural science.}
    \end{minipage}
    \hfill
    \begin{minipage}[c]{0.32\linewidth}
        \centering
        \includegraphics[width=\linewidth]{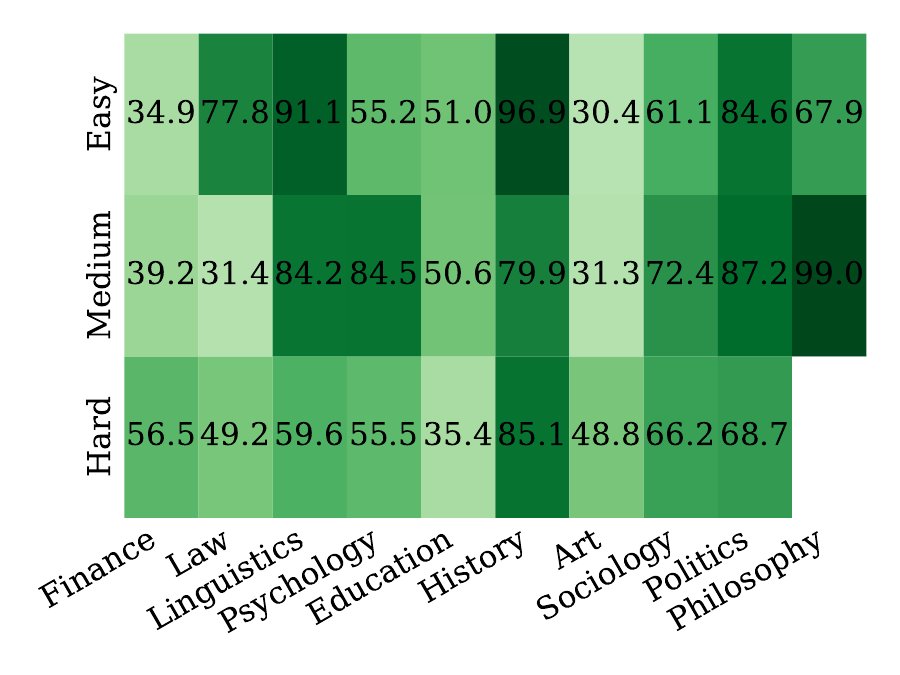}
        \caption*{UniMA \& social science.}
    \end{minipage}
    \hfill
    \begin{minipage}[c]{0.32\linewidth}
        \centering
        \includegraphics[width=\linewidth]{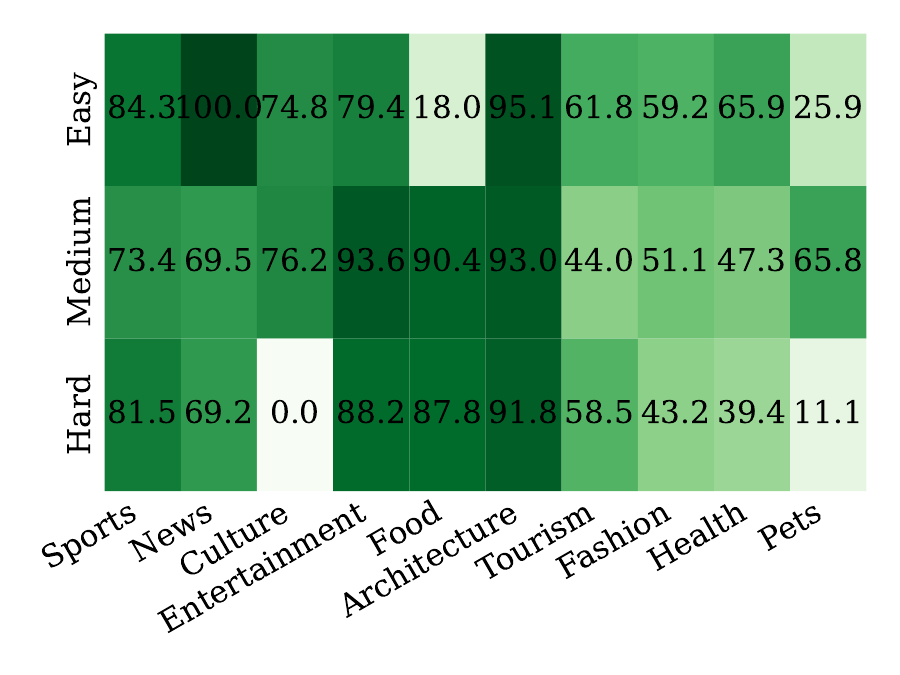}
        \caption*{UniMA \& general area.}
    \end{minipage}

    \caption{StS performance across baseline models, difficulty levels and domains.}
    \label{fig:app_heatmaps_sts}
\end{figure*}

\begin{figure*}[t]
    \centering

    \begin{minipage}[c]{0.32\linewidth}
        \centering
        \includegraphics[width=\linewidth]{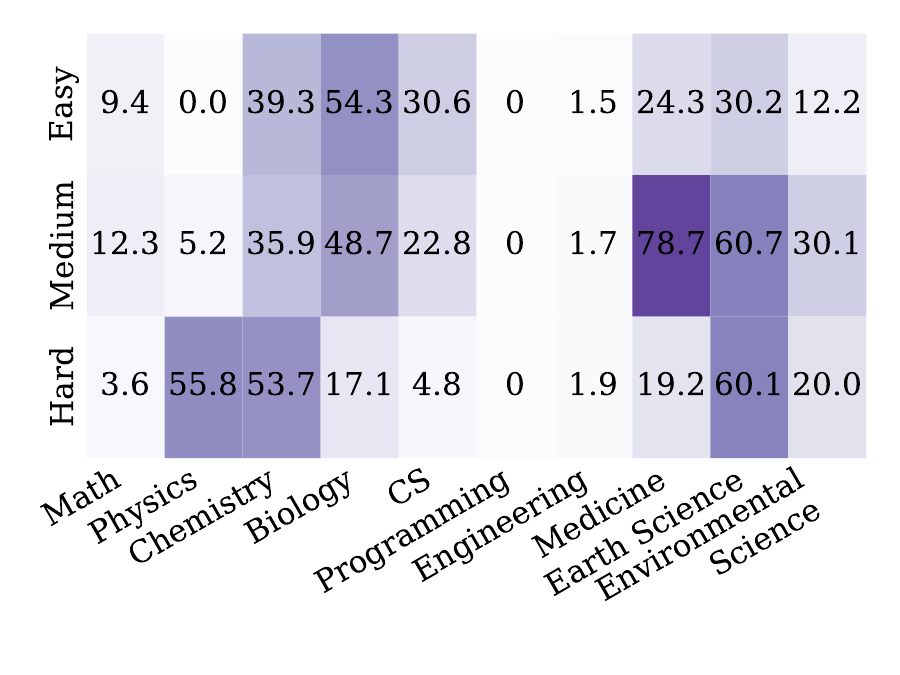}
        \caption*{AnyGPT \& natural science.} 
    \end{minipage}
    \hfill
    \begin{minipage}[c]{0.32\linewidth}
        \centering
        \includegraphics[width=\linewidth]{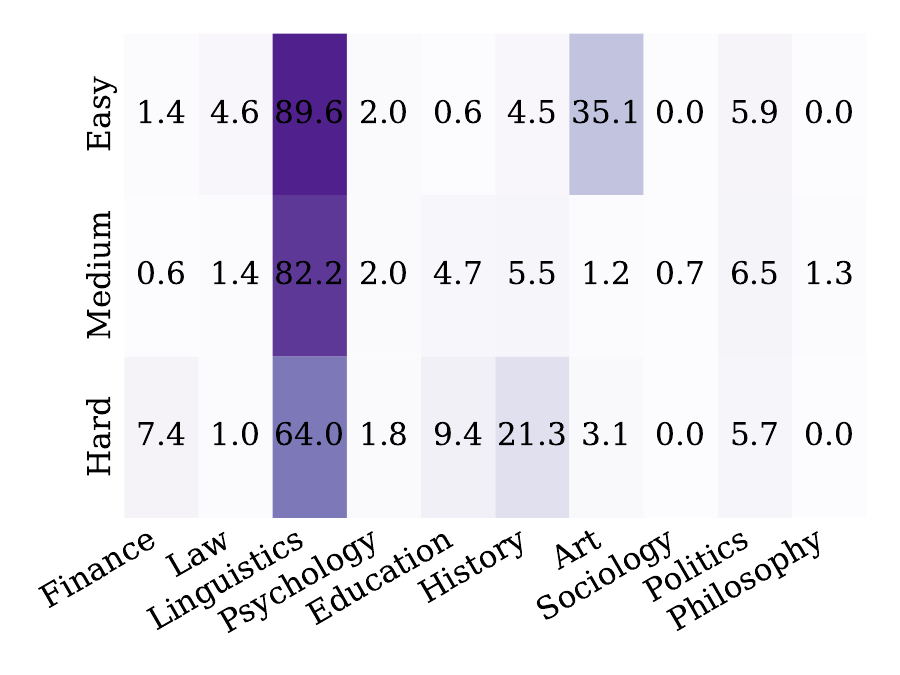}
        \caption*{AnyGPT \& social science.}
    \end{minipage}
    \hfill
    \begin{minipage}[c]{0.32\linewidth}
        \centering
        \includegraphics[width=\linewidth]{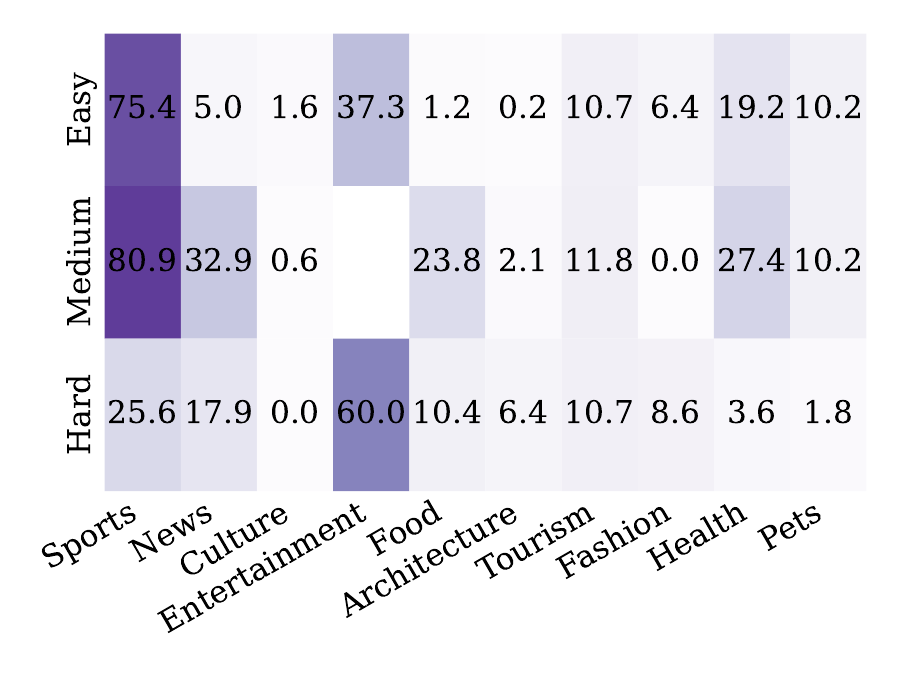}
        \caption*{AnyGPT \& general area.}
    \end{minipage}

    \begin{minipage}[c]{0.32\linewidth}
        \centering
        \includegraphics[width=\linewidth]{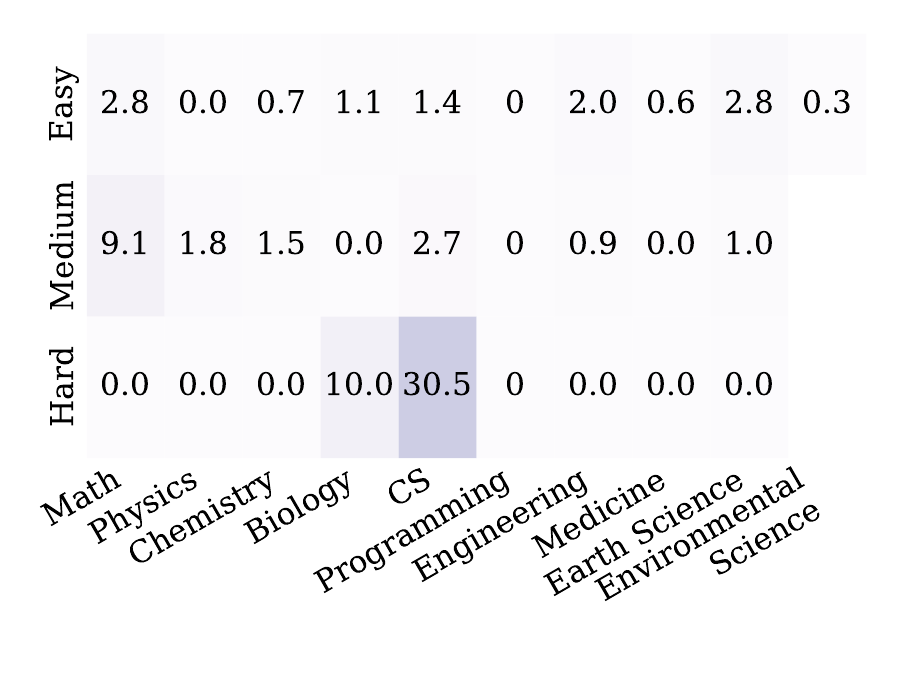}
        \caption*{NExT-GPT \& natural science.}
    \end{minipage}
    \hfill
    \begin{minipage}[c]{0.32\linewidth}
        \centering
        \includegraphics[width=\linewidth]{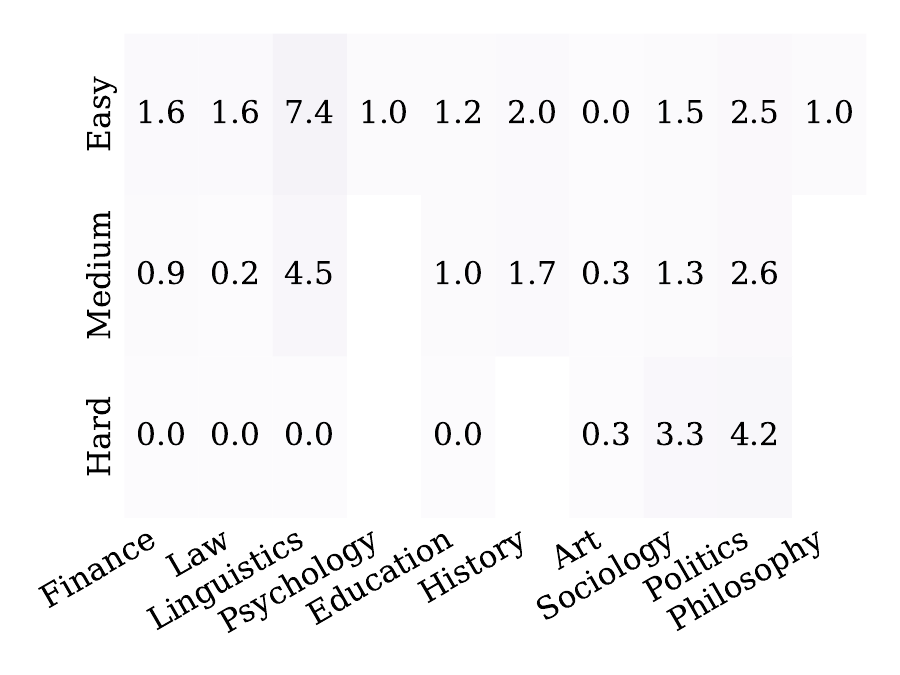}
        \caption*{NExT-GPT \& social science.}
    \end{minipage}
    \hfill
    \begin{minipage}[c]{0.32\linewidth}
        \centering
        \includegraphics[width=\linewidth]{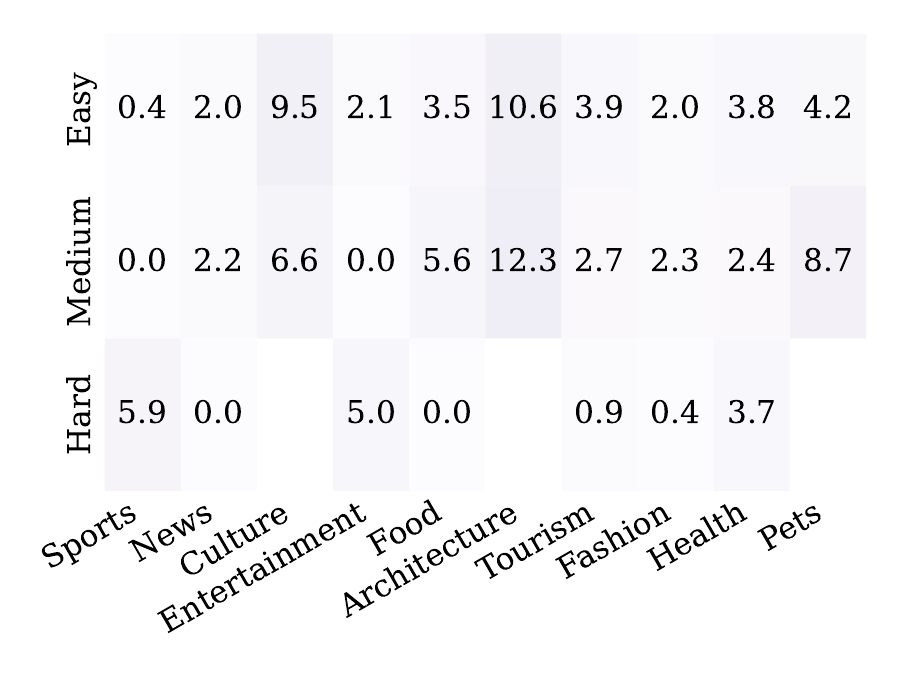}
        \caption*{NExT-GPT \& general area.}
    \end{minipage}

    \begin{minipage}[c]{0.32\linewidth}
        \centering
        \includegraphics[width=\linewidth]{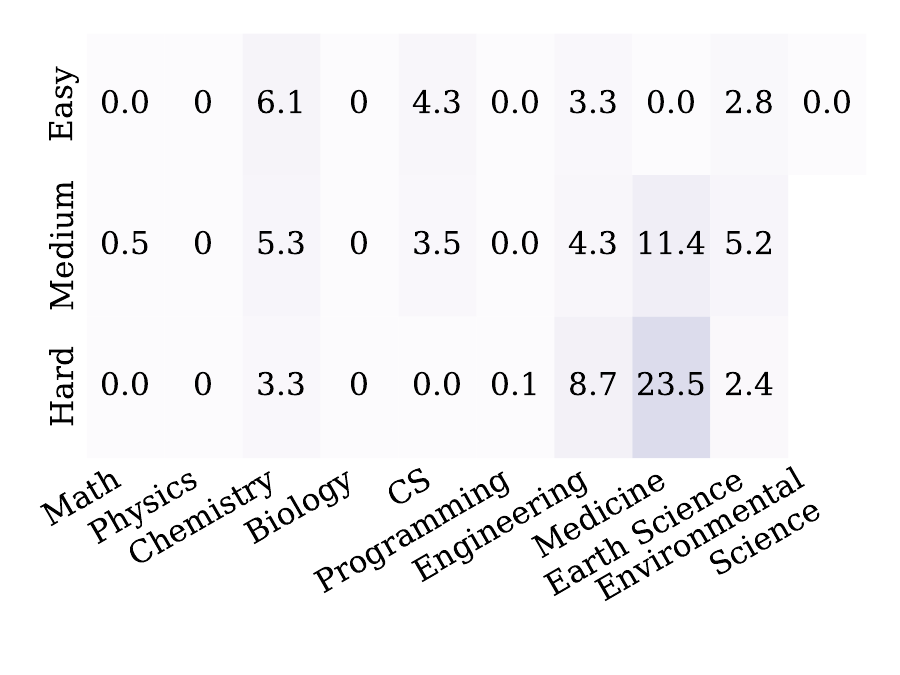}
        \caption*{MIO \& natural science.}
    \end{minipage}
    \hfill
    \begin{minipage}[c]{0.32\linewidth}
        \centering
        \includegraphics[width=\linewidth]{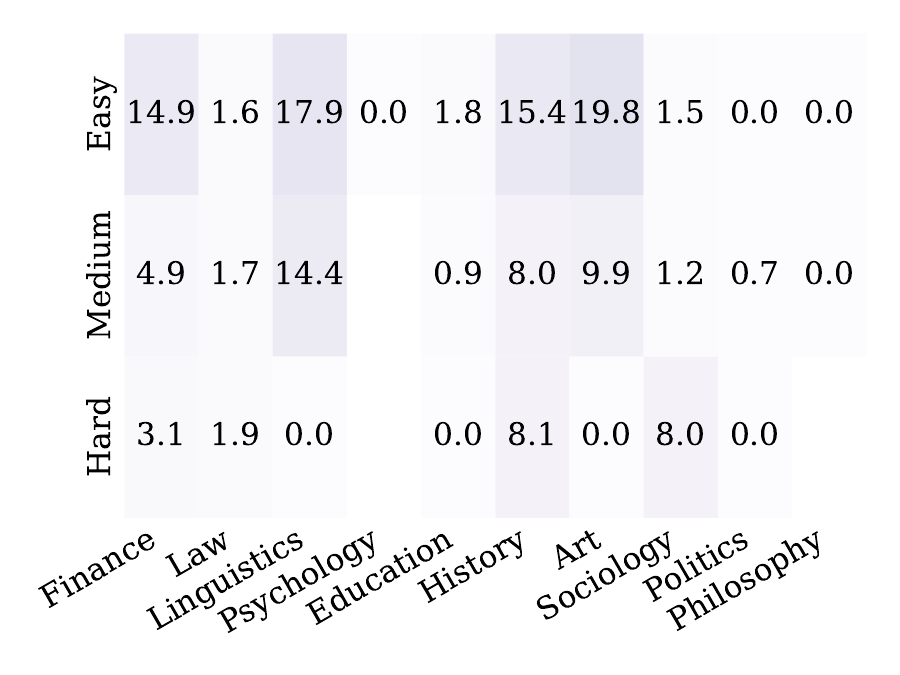}
        \caption*{MIO \& social science.}
    \end{minipage}
    \hfill
    \begin{minipage}[c]{0.32\linewidth}
        \centering
        \includegraphics[width=\linewidth]{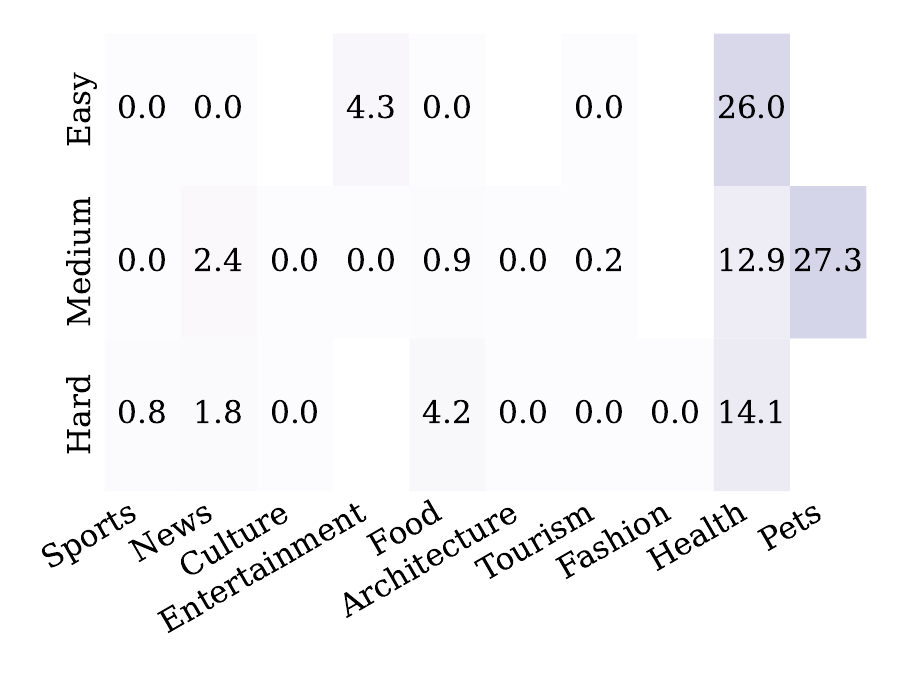}
        \caption*{MIO \& general area.}
    \end{minipage}

    \begin{minipage}[c]{0.32\linewidth}
        \centering
        \includegraphics[width=\linewidth]{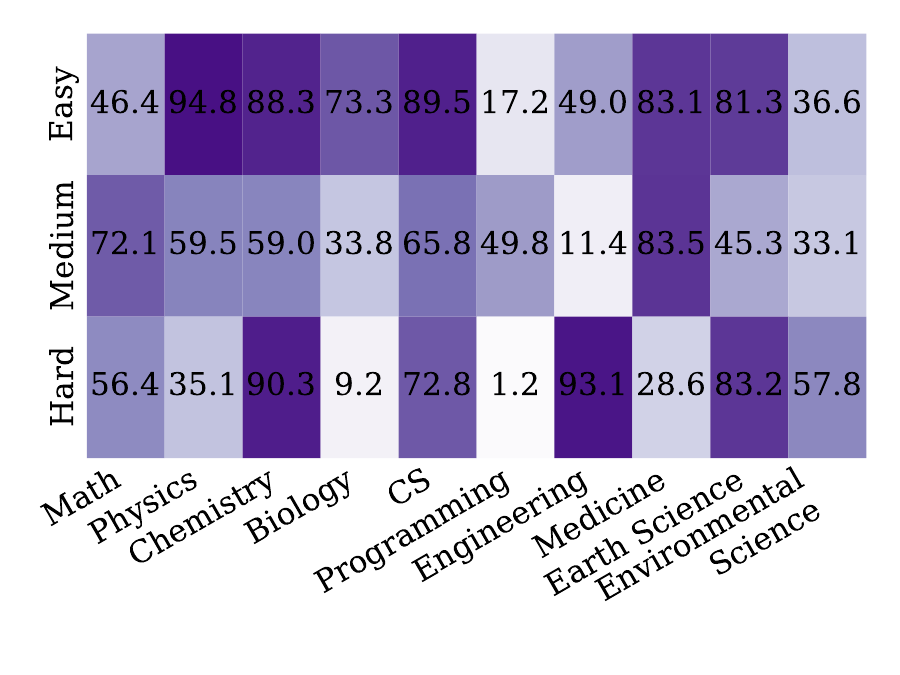}
        \caption*{UniMA \& natural science.}
    \end{minipage}
    \hfill
    \begin{minipage}[c]{0.32\linewidth}
        \centering
        \includegraphics[width=\linewidth]{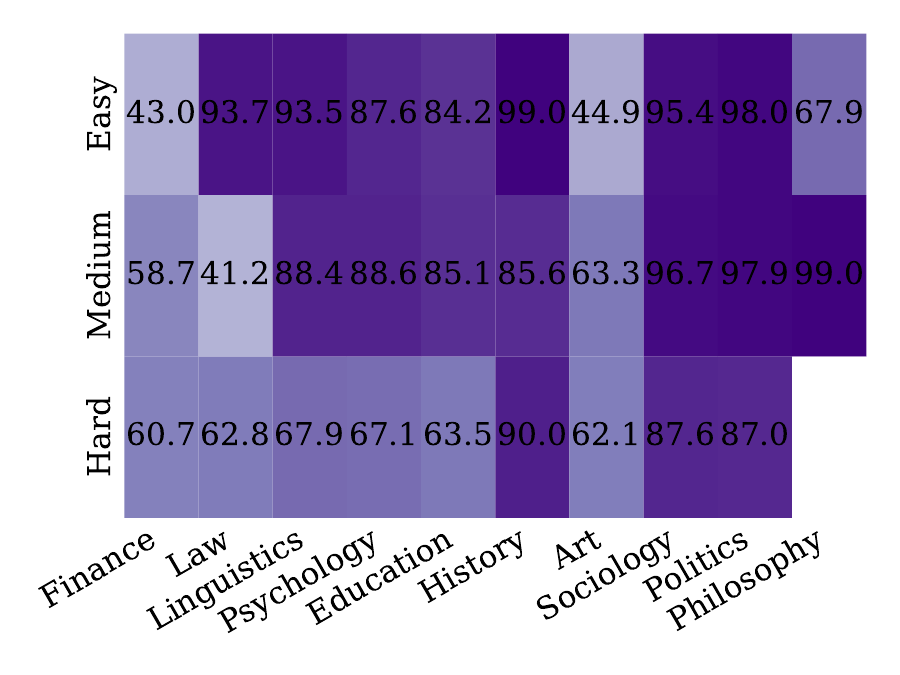}
        \caption*{UniMA \& social science.}
    \end{minipage}
    \hfill
    \begin{minipage}[c]{0.32\linewidth}
        \centering
        \includegraphics[width=\linewidth]{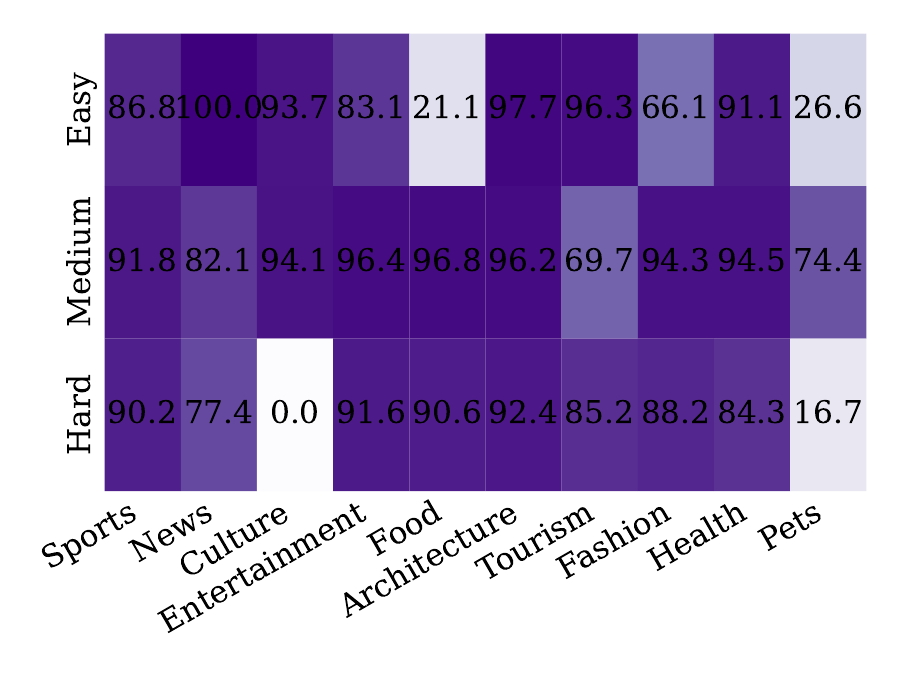}
        \caption*{UniMA \& general area.}
    \end{minipage}

    \caption{LeS performance across baseline models, difficulty levels and domains.}
    \label{fig:app_heatmaps_les}
\end{figure*}

\begin{figure*}[t]
    \centering

    \begin{minipage}[c]{0.32\linewidth}
        \centering
        \includegraphics[width=\linewidth]{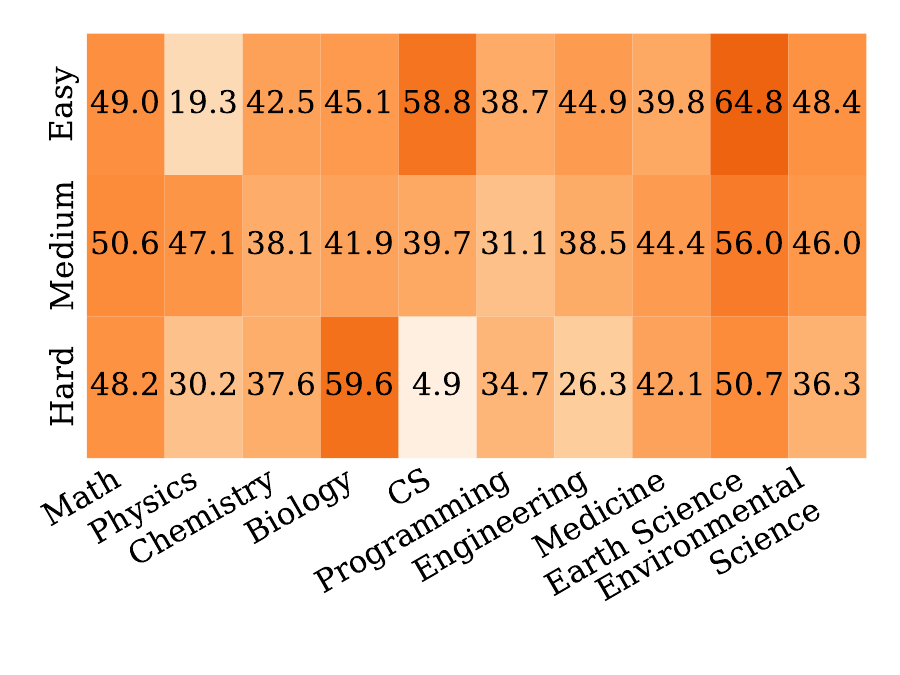}
        \caption*{AnyGPT \& natural science.} 
    \end{minipage}
    \hfill
    \begin{minipage}[c]{0.32\linewidth}
        \centering
        \includegraphics[width=\linewidth]{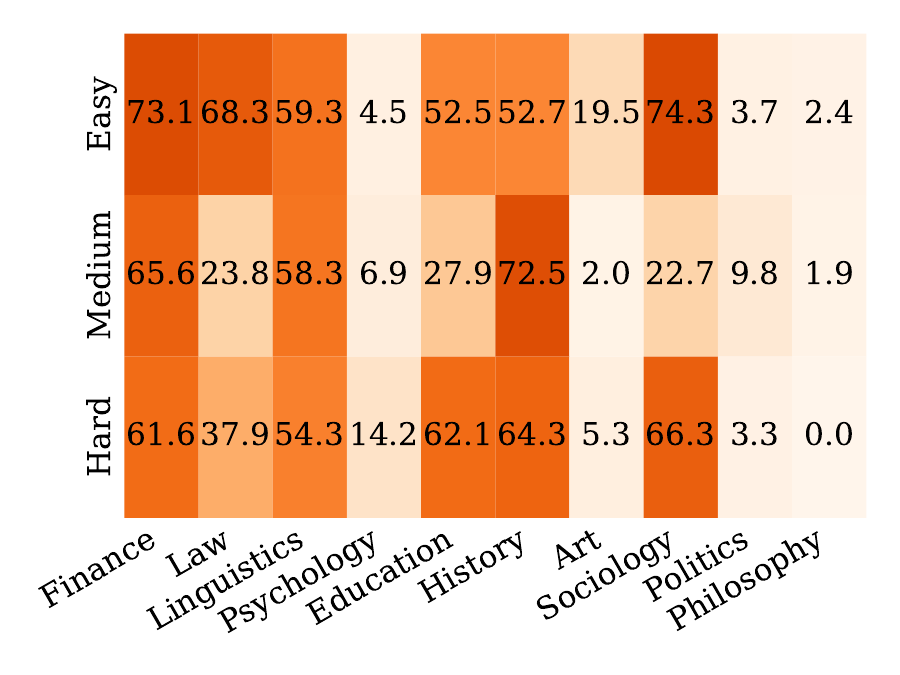}
        \caption*{AnyGPT \& social science.}
    \end{minipage}
    \hfill
    \begin{minipage}[c]{0.32\linewidth}
        \centering
        \includegraphics[width=\linewidth]{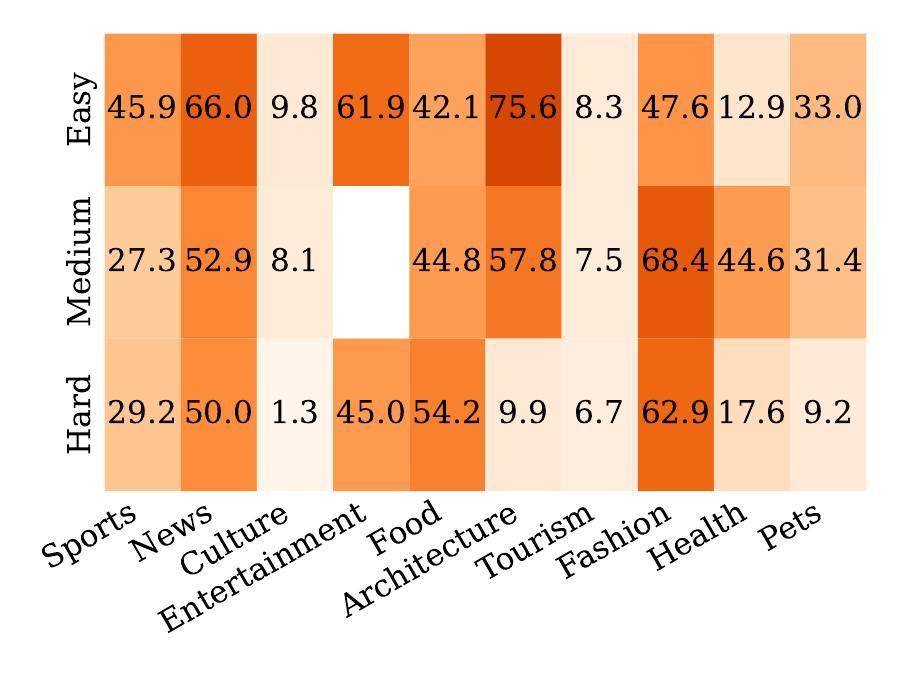}
        \caption*{AnyGPT \& general area.}
    \end{minipage}

    \begin{minipage}[c]{0.32\linewidth}
        \centering
        \includegraphics[width=\linewidth]{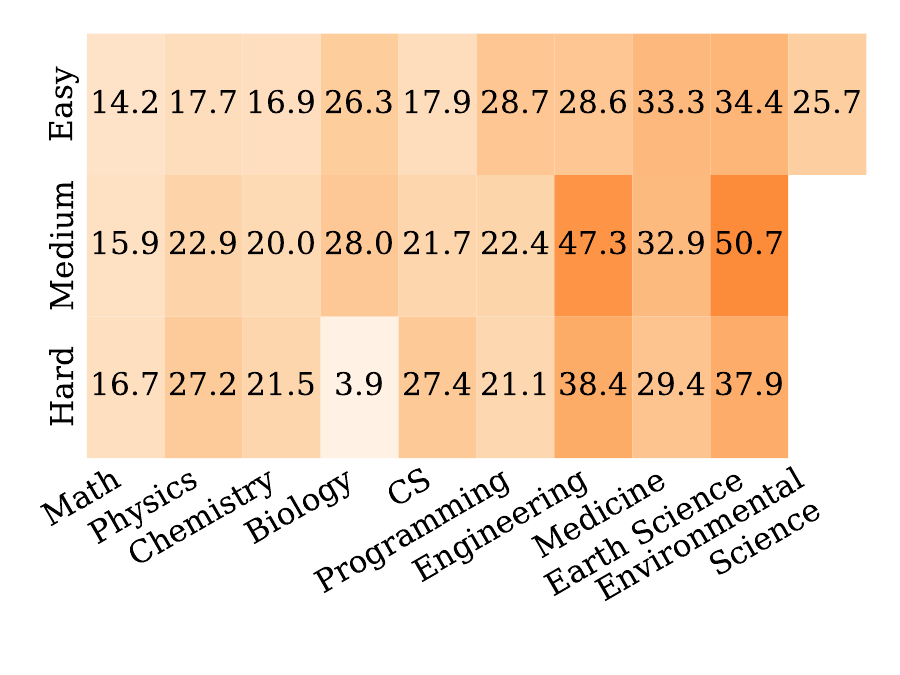}
        \caption*{NExT-GPT \& natural science.}
    \end{minipage}
    \hfill
    \begin{minipage}[c]{0.32\linewidth}
        \centering
        \includegraphics[width=\linewidth]{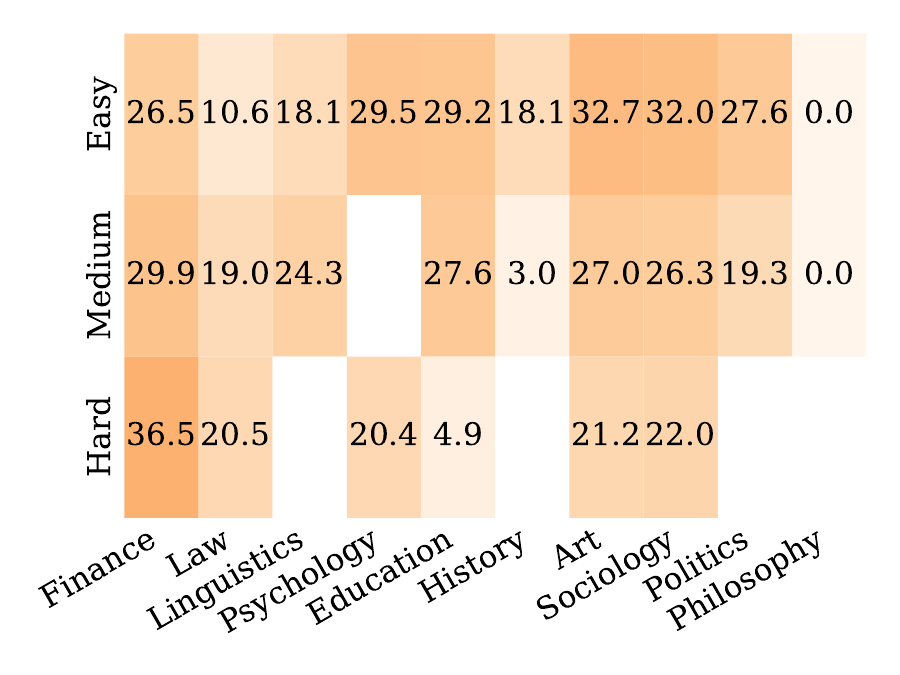}
        \caption*{NExT-GPT \& social science.}
    \end{minipage}
    \hfill
    \begin{minipage}[c]{0.32\linewidth}
        \centering
        \includegraphics[width=\linewidth]{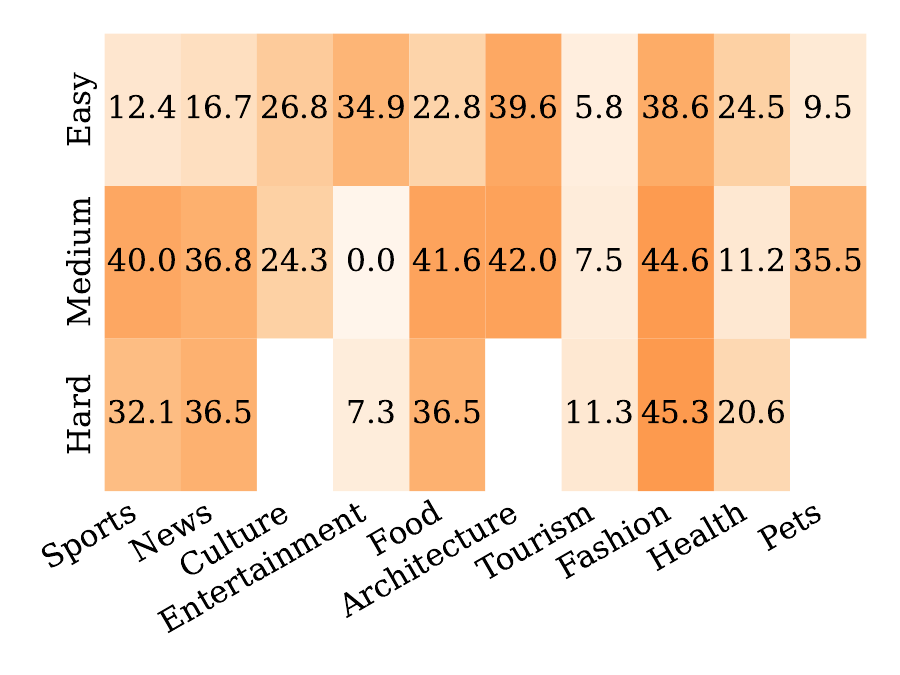}
        \caption*{NExT-GPT \& general area.}
    \end{minipage}

    \begin{minipage}[c]{0.32\linewidth}
        \centering
        \includegraphics[width=\linewidth]{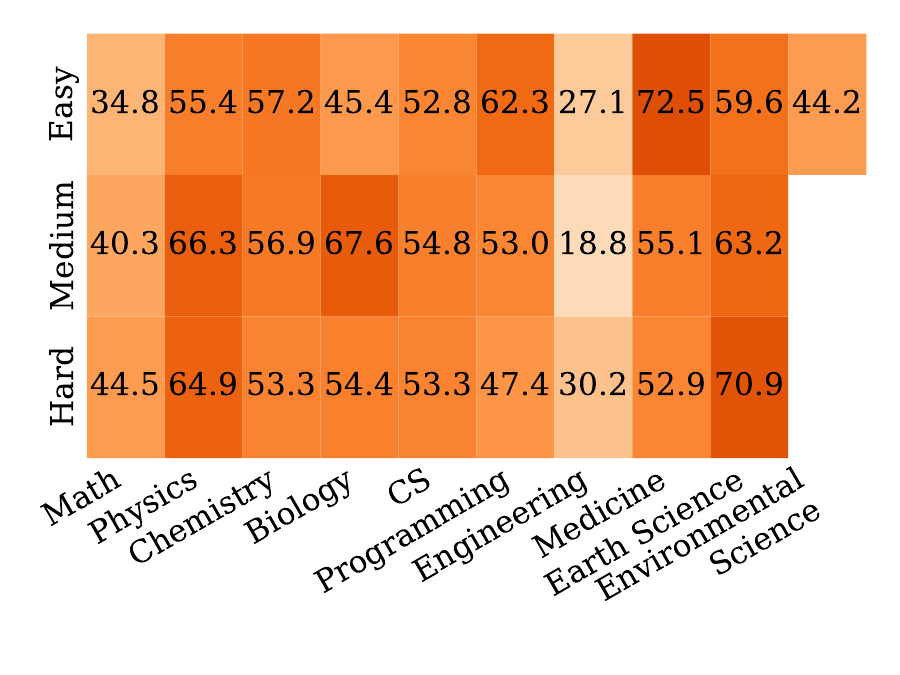}
        \caption*{MIO \& natural science.}
    \end{minipage}
    \hfill
    \begin{minipage}[c]{0.32\linewidth}
        \centering
        \includegraphics[width=\linewidth]{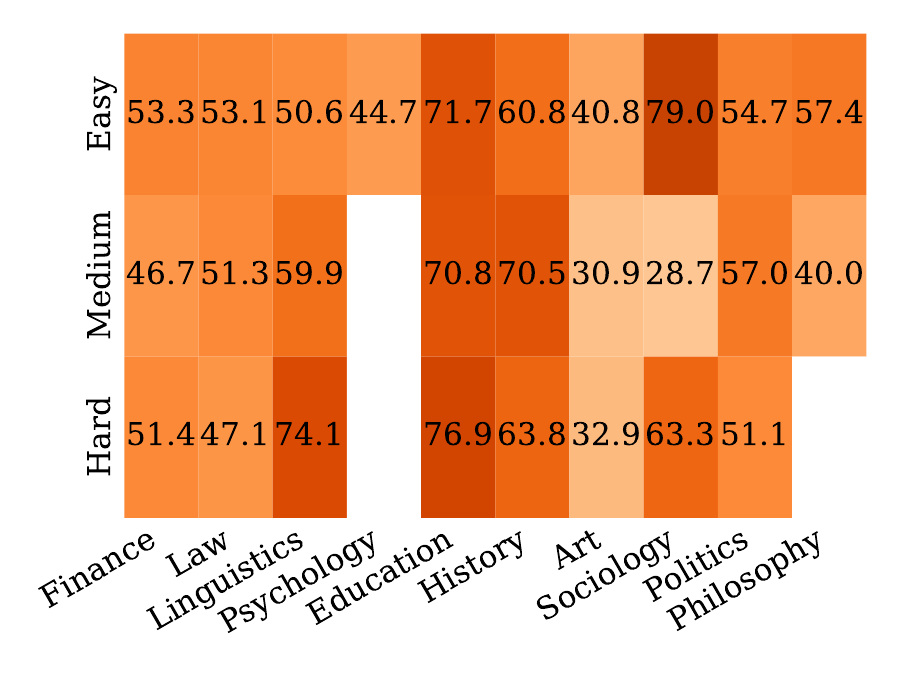}
        \caption*{MIO \& social science.}
    \end{minipage}
    \hfill
    \begin{minipage}[c]{0.32\linewidth}
        \centering
        \includegraphics[width=\linewidth]{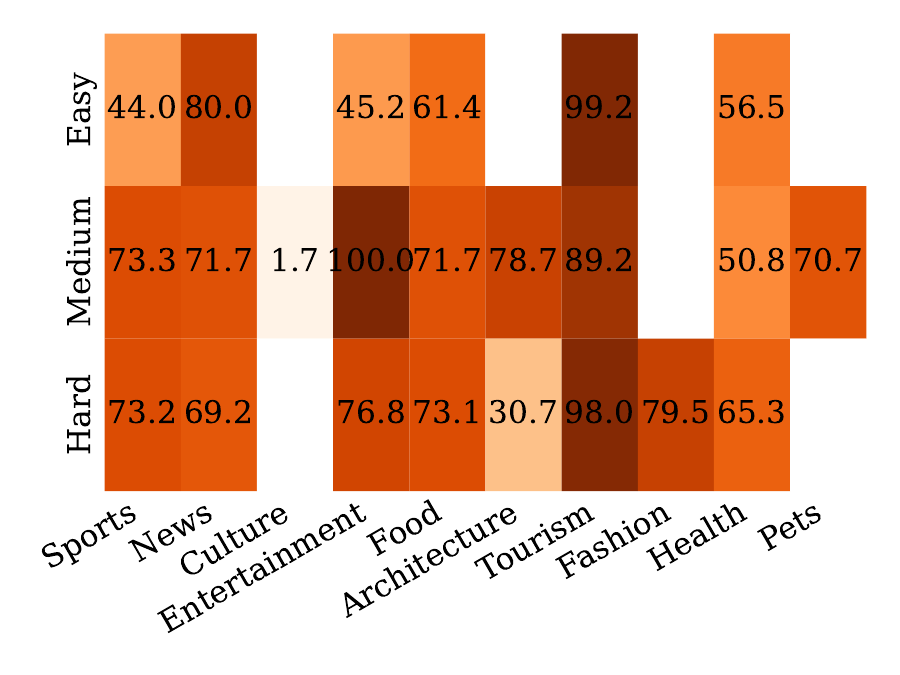}
        \caption*{MIO \& general area.}
    \end{minipage}

    \begin{minipage}[c]{0.32\linewidth}
        \centering
        \includegraphics[width=\linewidth]{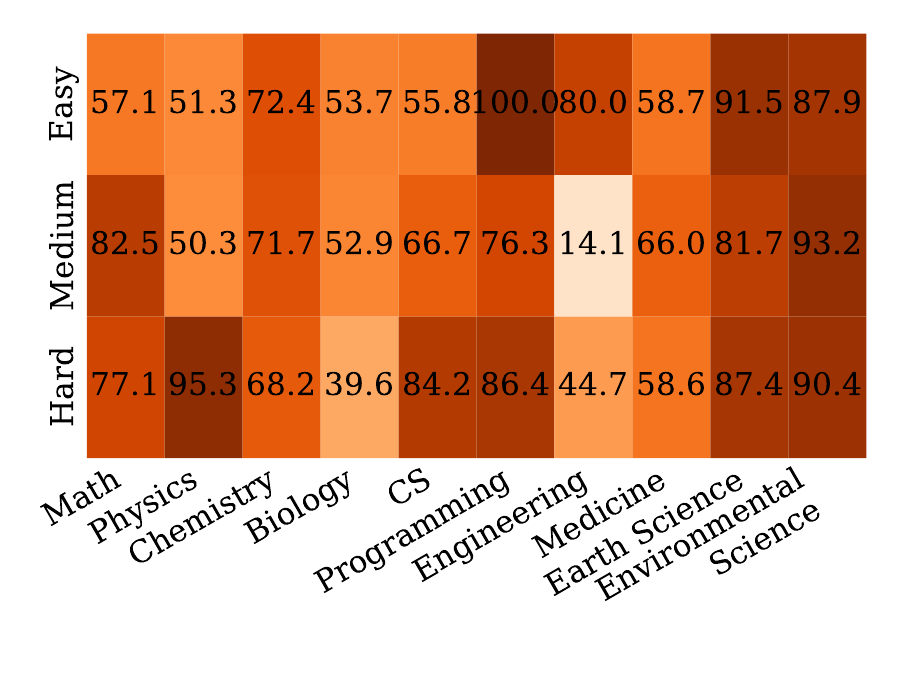}
        \caption*{UniMA \& natural science.}
    \end{minipage}
    \hfill
    \begin{minipage}[c]{0.32\linewidth}
        \centering
        \includegraphics[width=\linewidth]{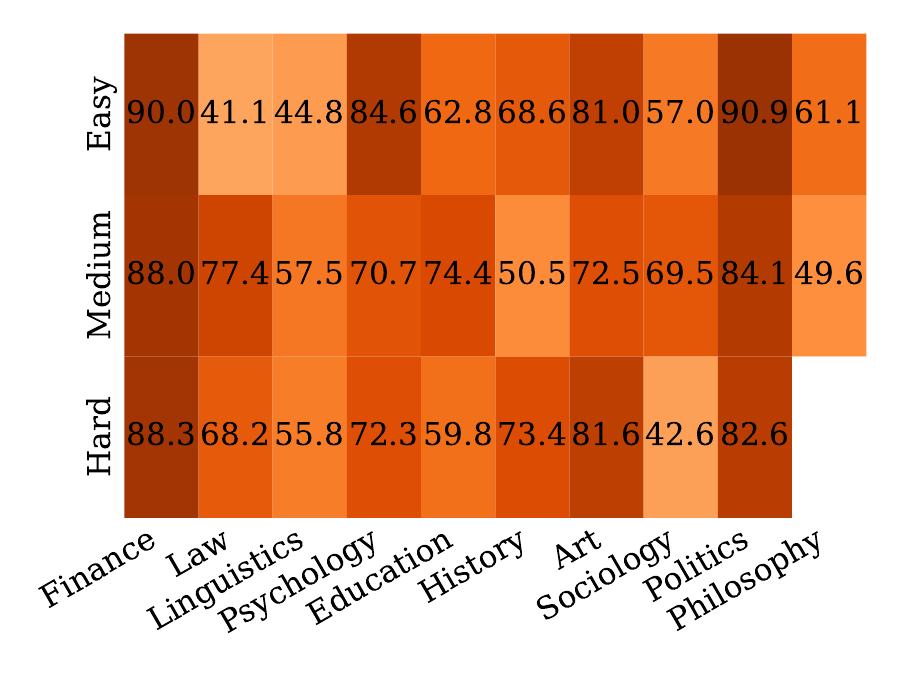}
        \caption*{UniMA \& social science.}
    \end{minipage}
    \hfill
    \begin{minipage}[c]{0.32\linewidth}
        \centering
        \includegraphics[width=\linewidth]{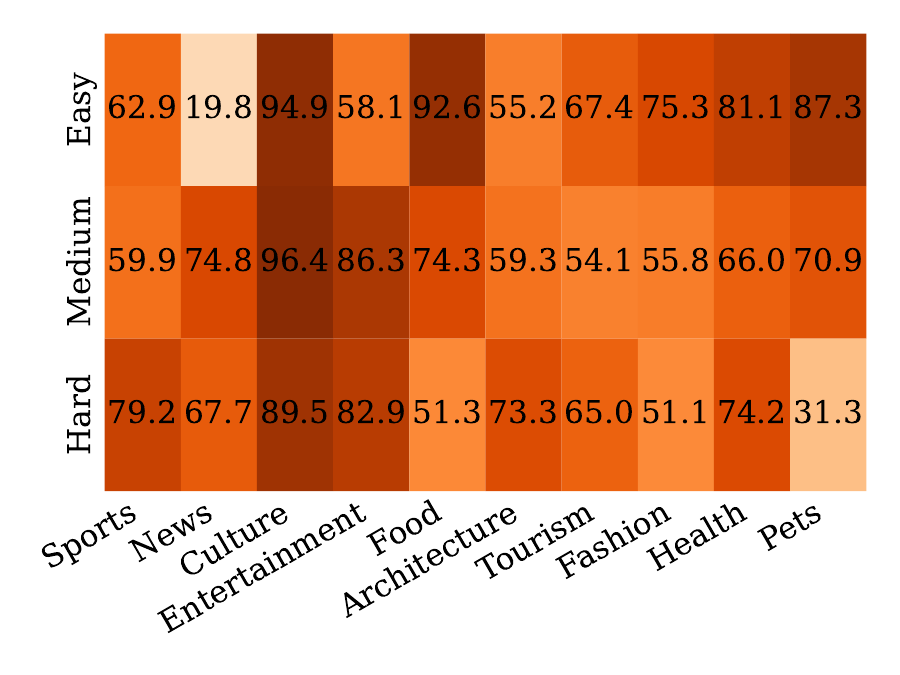}
        \caption*{UniMA \& general area.}
    \end{minipage}

    \caption{ICS performance across baseline models, difficulty levels and domains.}
    \label{fig:app_heatmaps_ics}
\end{figure*}

In this section, we provide detailed difficulty-level experiments results, as illustrated in Fig.~\ref{fig:app_heatmaps_sqcs}, Fig.~\ref{fig:app_heatmaps_sts}, Fig.~\ref{fig:app_heatmaps_les}, and Fig.~\ref{fig:app_heatmaps_ics}.

\section{Case Study}
\begin{figure*}[t!]
  \centering
  \begin{minipage}[b]{0.99\linewidth}
    \centering
    \includegraphics[width=\linewidth]{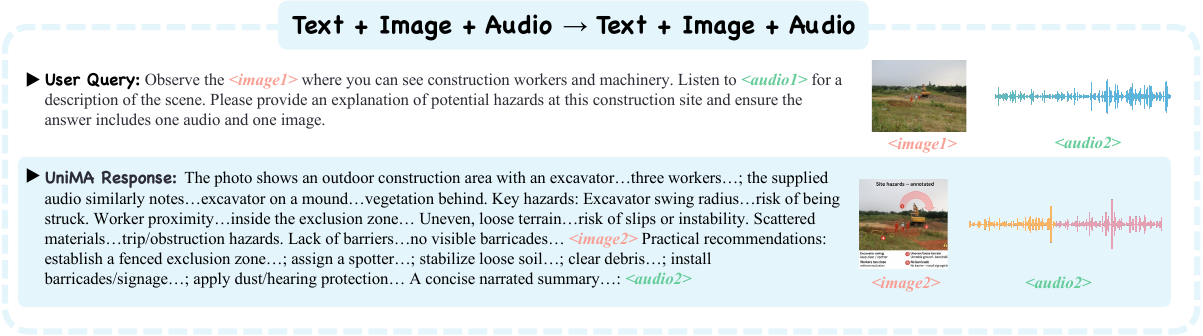}
    \caption{Case illustrating the responses produced by \textsc{UniMA} to T+I+A as input.}
    \label{fig:case1}
  \end{minipage}

  \begin{minipage}[b]{0.99\linewidth}
    \centering
    \vspace{6mm}
    \includegraphics[width=\linewidth]{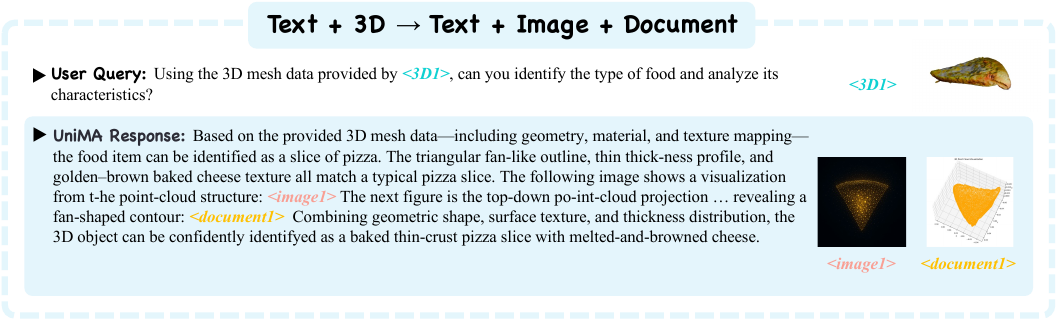}
    \caption{Case illustrating the responses produced by \textsc{UniMA} to T+3D as input.}
    \label{fig:case2}
  \end{minipage}

  \begin{minipage}[b]{0.99\linewidth}
    \centering
    \vspace{6mm}
    \includegraphics[width=\linewidth]{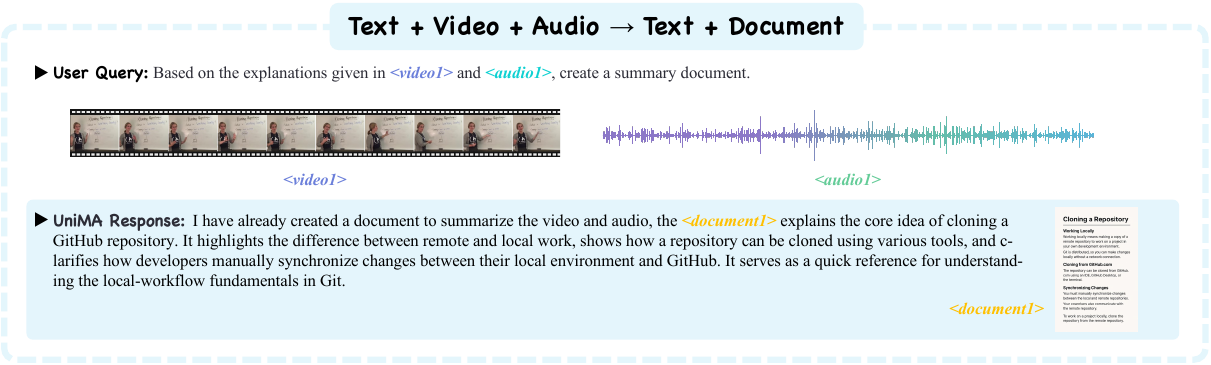}
    \caption{Case illustrating the responses produced by \textsc{UniMA} to T+V+A as input.}
    \label{fig:case3}
  \end{minipage}
\end{figure*}

\begin{figure*}[t!]
  \centering
  \begin{minipage}[b]{0.99\linewidth}
    \centering
    \vspace{-6mm}
    \includegraphics[width=\linewidth]{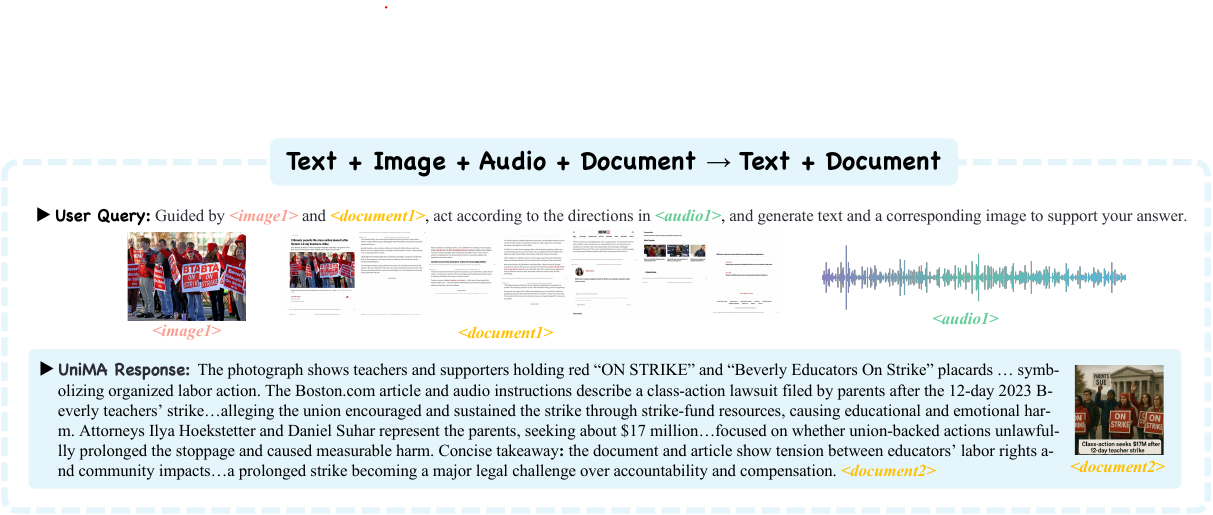}
    \caption{Case illustrating the responses produced by \textsc{UniMA} to T+I+A+D as input.}
    \label{fig:case4}
  \end{minipage}

\vspace{6mm}

  \begin{minipage}[b]{0.99\linewidth}
    \centering
    \includegraphics[width=\linewidth]{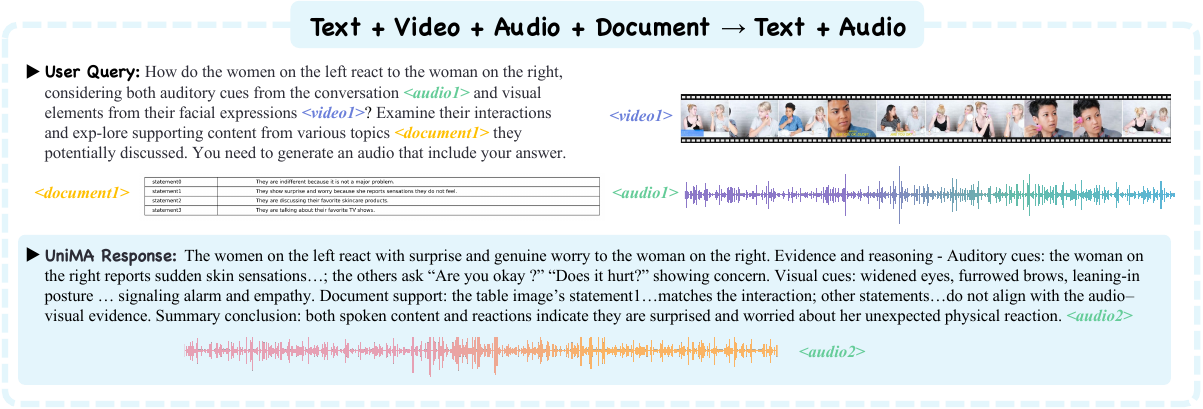}
    \caption{Case illustrating the responses produced by \textsc{UniMA} to T+V+A+D as input.}
    \label{fig:case5}
  \end{minipage}
  
\vspace{6mm}

  \begin{minipage}[b]{0.99\linewidth}
    \centering
    \includegraphics[width=\linewidth]{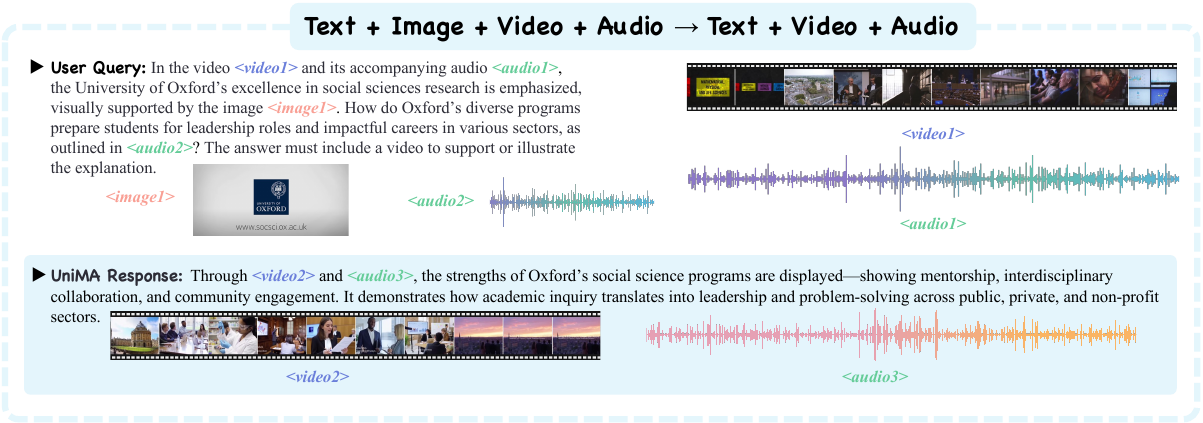}
    \caption{Case illustrating the responses produced by \textsc{UniMA} to T+I+V+A as input.}
    \label{fig:case6}
  \end{minipage}
\end{figure*}

\begin{figure*}[t!]
  \centering
  \begin{minipage}[b]{0.99\linewidth}
    \centering
    \includegraphics[width=\linewidth]{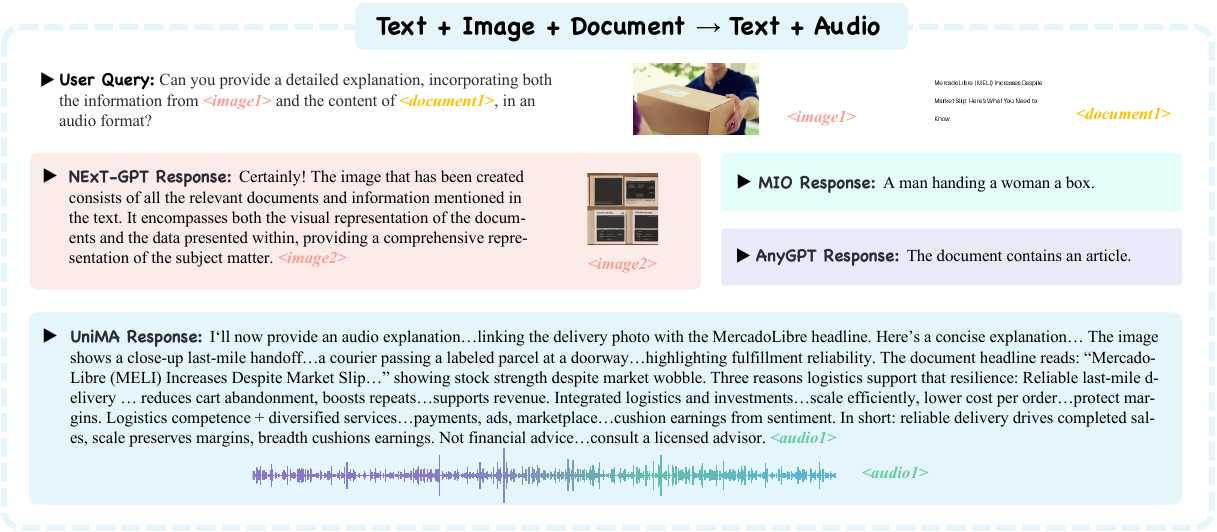}
    \caption{Case illustrating the responses produced by AnyGPT, NExT-GPT, MIO and \textsc{UniMA} to T+I+D as input.}
    \label{fig:case7}
  \end{minipage}
  
\vspace{6mm}

  \begin{minipage}[b]{0.99\linewidth}
    \centering
    \includegraphics[width=\linewidth]{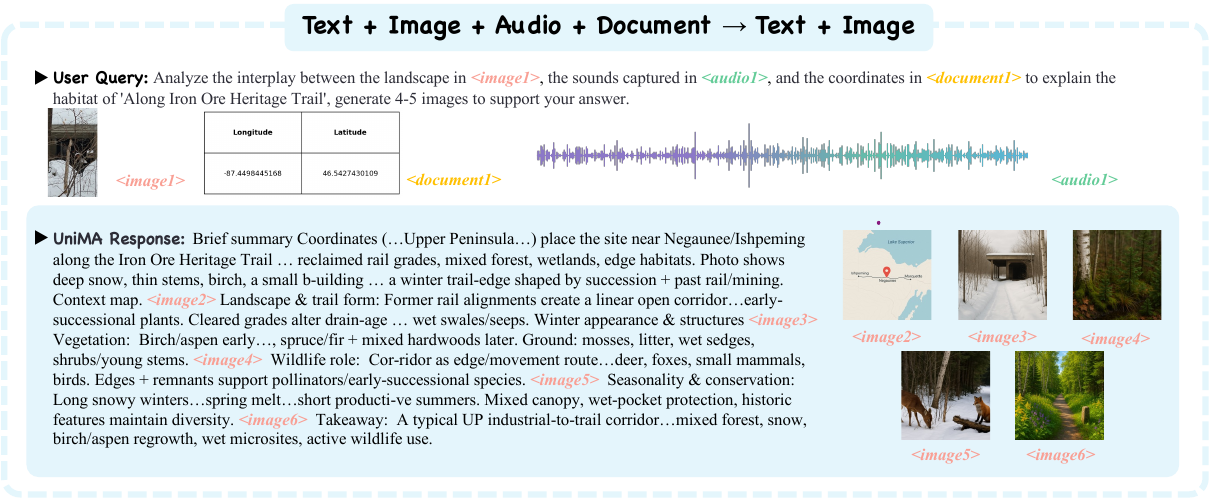}
    \caption{Case illustrating the responses produced by \textsc{UniMA} to T+I+A+D as input.}
    \label{fig:case8}
  \end{minipage}
\end{figure*}

\begin{figure*}[t!]
  \centering
  \begin{minipage}[b]{0.99\linewidth}
    \centering
    \includegraphics[width=\linewidth]{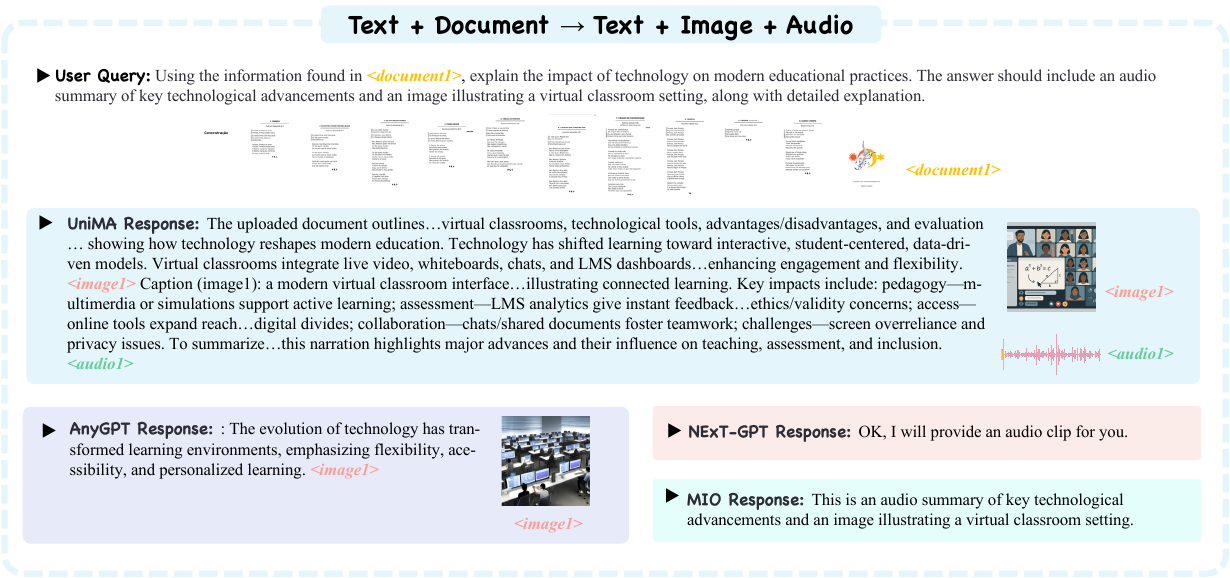}
    \caption{Case illustrating the responses produced by AnyGPT, NExT-GPT, MIO and \textsc{UniMA} to T+D as input.}
    \label{fig:case9}
  \end{minipage}

\vspace{6mm}

  \begin{minipage}[b]{0.99\linewidth}
    \centering
    \includegraphics[width=\linewidth]{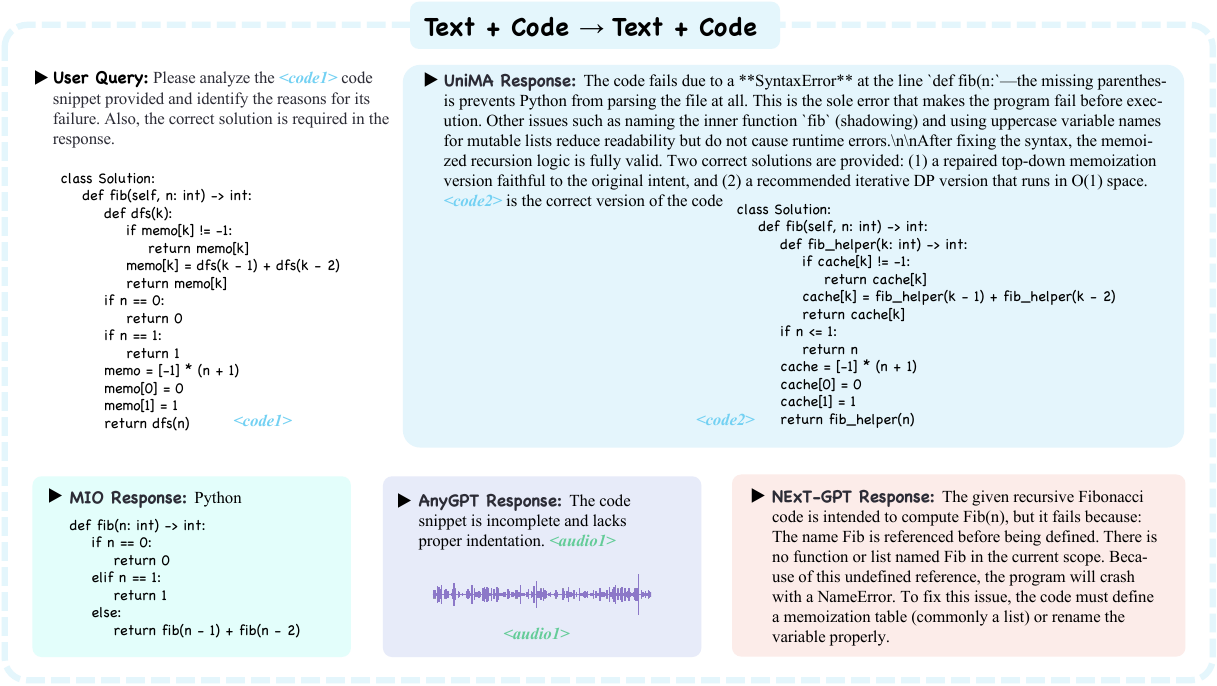}
    \caption{Case illustrating the responses produced by AnyGPT, NExT-GPT, MIO and \textsc{UniMA} to T+C as input.}
    \label{fig:case10}
  \end{minipage}
\end{figure*}

\begin{figure*}
\centering
\includegraphics[width=0.99\linewidth]{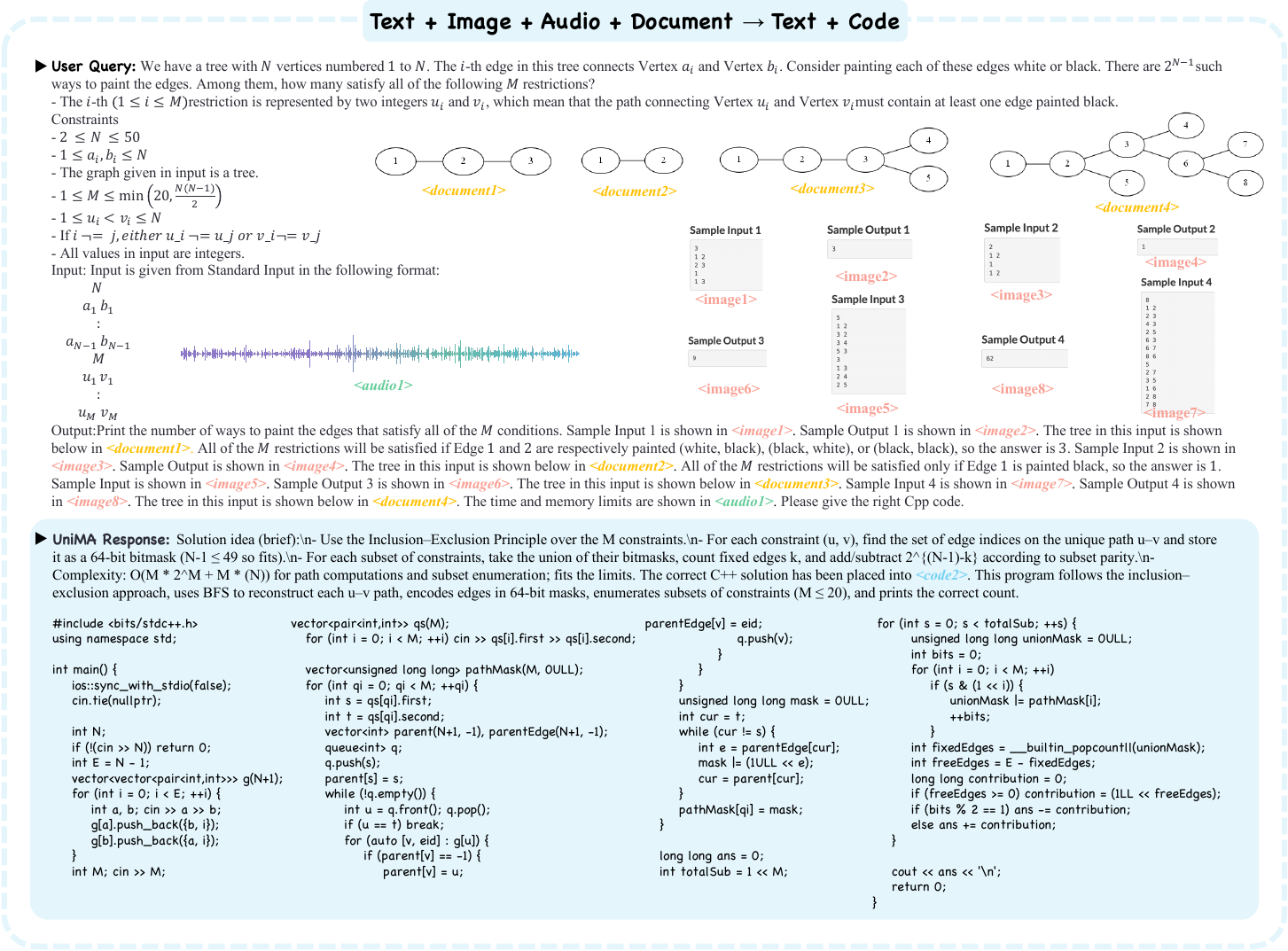}
\caption{Case illustrating the responses produced by AnyGPT, NExT-GPT, MIO and \textsc{UniMA} to T+I+A+D as input.}
\label{fig:case11}
\end{figure*}

\begin{figure*}
\centering
\includegraphics[width=0.99\linewidth]{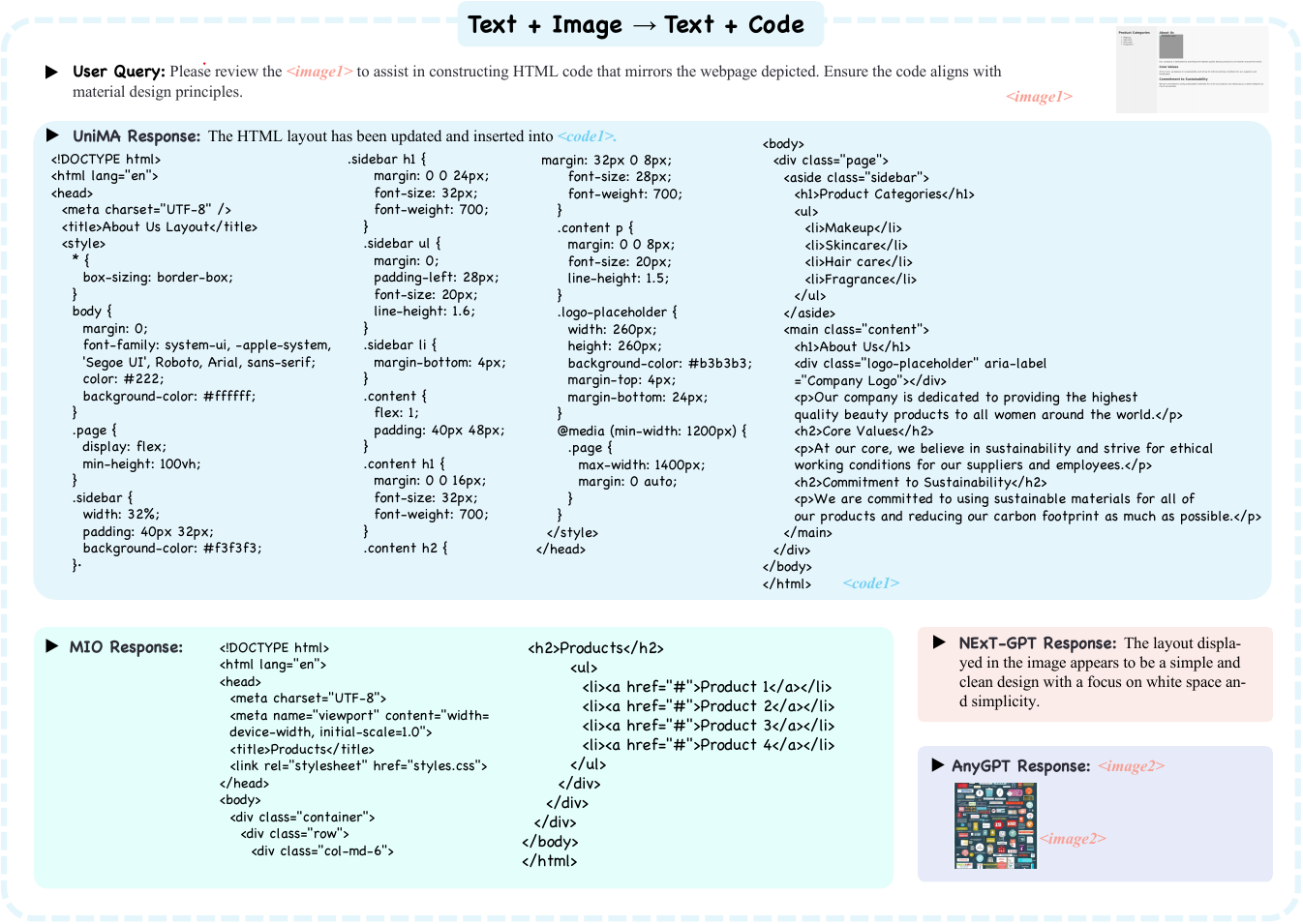}
\caption{Case illustrating the responses produced by AnyGPT, NExT-GPT, MIO and \textsc{UniMA} to T+I as input.}
\label{fig:case12}
\end{figure*}

\begin{figure*}
\centering
\includegraphics[width=0.99\linewidth]{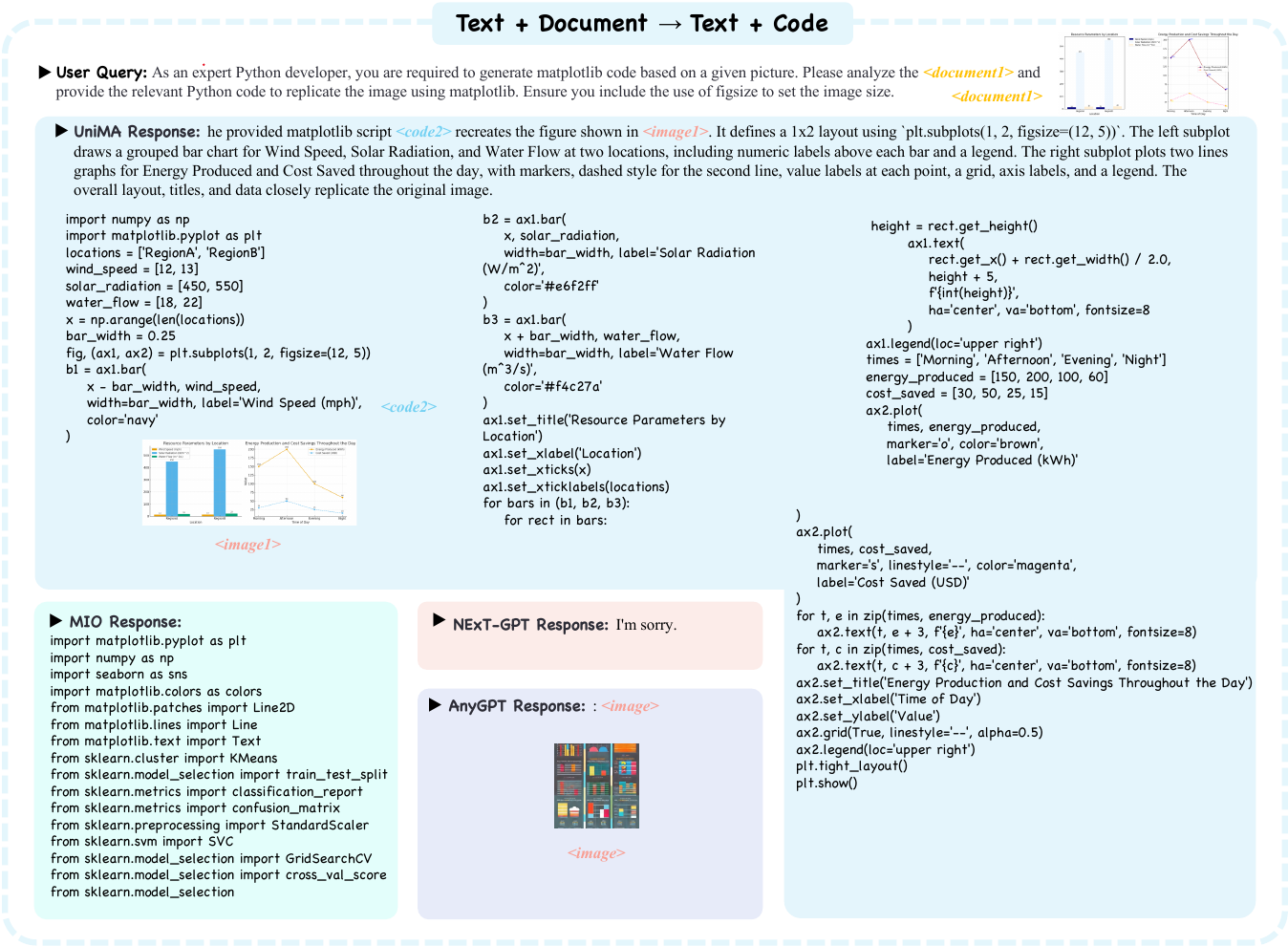}
\caption{Case illustrating the responses produced by AnyGPT, NExT-GPT, MIO and \textsc{UniMA} to T+D as input.}
\label{fig:case13}
\end{figure*}

This section presents 13 representative case studies that illustrate how the model behaves under different interleaved multimodal task settings. 
Each case is accompanied a corresponding figure, including Fig.~\ref{fig:case1}, Fig.~\ref{fig:case2}, Fig.~\ref{fig:case3}, Fig.~\ref{fig:case4}, Fig.~\ref{fig:case5}, Fig.~\ref{fig:case6}, Fig.~\ref{fig:case7}, Fig.~\ref{fig:case8}, Fig.~\ref{fig:case9}, Fig.~\ref{fig:case10}, Fig.~\ref{fig:case11}, Fig.~\ref{fig:case12}, and Fig.~\ref{fig:case13}, covering diverse input–output modality combinations and varying levels of cross-modal interaction.

These cases provide qualitative insights into the model’s performance in terms of semantic alignment, structural completeness, and stylistic consistency. They serve as a complementary perspective to the quantitative results presented earlier, offering a more comprehensive understanding of the model’s strengths and limitations in realistic interleaved multimodal scenarios.

\section{Ethic Statement}
The \textsc{UniM} benchmark uses multimodal assets that are sourced exclusively from publicly available and open-access datasets, as well as openly accessible content from the web. 
All the data used in this benchmark is derived from well-established open-source repositories and complies with their respective licenses, ensuring full transparency and legal compliance. 

We confirm that no sensitive, private, or personally identifiable information is included in any of the multimodal assets. 
Additionally, we have carefully reviewed the contents of the datasets to ensure that there are no harmful, biased, or otherwise problematic elements that could negatively impact the integrity of the research or the communities involved. 
Ethical considerations regarding fairness, inclusivity, and non-discrimination have been taken into account throughout the data collection process.

Furthermore, the \textsc{UniM} benchmark was developed with a focus on promoting responsible AI research, and we are committed to ensuring that our benchmark is used in a way that upholds ethical principles, including respect for privacy and the prevention of harm.

\clearpage

\newpage

\newpage

\newpage

\newpage

\newpage

\newpage

\newpage

\newpage

\newpage

\newpage

\newpage

\newpage

\newpage

\newpage

\newpage

\newpage

\newpage

\newpage
\begin{table*}[t!]
\centering
    \caption{Detailed experiment results on \textit{Semantic Correctness \& Generation Quality}.}
    \label{tab:domain sqcs1}
    \fontsize{10}{13}\selectfont
    \setlength{\tabcolsep}{4mm}
    \label{tab:results_semantic_correctness}
    \begin{tabular}{l*{10}{c}}
        \toprule
        \rowcolor{SCGQ_head_fill}
        \multicolumn{11}{c}{\textbf{\textcolor{SCGQ_head_word}{Semantic Correctness \& Generation Quality}}} \\
        \rowcolor{SCGQ_head_fill}
        & \bf\#1 & \bf\#2 & \bf\#3 & \bf\#4 & \bf\#5 & \bf\#6 & \bf\#7 & \bf\#8 & \bf\#9 & \bf\#10\\
        \rowcolor{SCGQ_head_fill}
        & \bf\#11 & \bf\#12 & \bf\#13 & \bf\#14 & \bf\#15 & \bf\#16 & \bf\#17 & \bf\#18 & \bf\#19 & \bf\#20 \\
        \rowcolor{SCGQ_head_fill}
        & \bf\#21 & \bf\#22 & \bf\#23 & \bf\#24 & \bf\#25 & \bf\#26 & \bf\#27 & \bf\#28 & \bf\#29 & \bf\#30 \\ 
        \midrule
        \rowcolor{SCGQ_table_fill}
        \multicolumn{11}{c}{\textbf{AnyGPT} \cite{zhan2024anygpt}} \\
        \midrule
        \multirow{3}{*}{SC}
        & 13.3
        & 4.4
        & 16.7
        & 8.9
        & 13.8
        & 6.2
        & 13.7
        & 16.0
        & 24.9
        & 19.6 \\
        & 16.9
        & 18.9
        & 33.4
        & 6.5
        & 23.3
        & 51.9
        & 5.4
        & 17.4
        & 4.8
        & 2.0 \\
        & 27.4
        & 22.9
        & 2.6
        & 13.0
        & 23.3
        & 38.1
        & 6.8
        & 21.2
        & 13.2
        & 21.9 \\
        \cdashline{1-11}
        \multirow{3}{*}{GQ}
        & 44.8
        & 24.7
        & 49.8
        & 44.0
        & 32.3
        & 39.1
        & 32.9
        & 22.0
        & 54.0
        & 35.0 \\
        & 45.7
        & 26.1
        & 45.8
        & 6.7
        & 28.1
        & 48.5
        & 9.5
        & 19.6 
        & 5.8
        & 1.9 \\
        & 29.6
        & 46.5
        & 3.7
        & 39.0
        & 42.7
        & 40.5
        & 7.2
        & 42.4 
        & 16.8 
        & 32.7 \\
        \cdashline{1-11}
        \multirow{3}{*}{SQCS$^{abs}$}
        & 11.1
        & 3.4
        & 14.1
        & 7.4
        & 11.0
        & 5.0
        & 10.9
        & 12.3
        & 21.4
        & 15.7 \\
        & 14.2
        & 14.7
        & 27.9
        & 4.6
        & 18.3
        & 43.8
        & 4.0
        & 13.2 
        & 3.4 
        & 1.4 \\
        & 21.6
        & 19.2
        & 1.9
        & 10.6 
        & 19.3
        & 31.3 
        & 4.9
        & 17.6 
        & 9.9 
        & 17.5 \\
        \midrule
        \rowcolor{SCGQ_table_fill}
        \multicolumn{11}{c}{\textbf{NExT-GPT} \cite{wu2024next}} \\
        \midrule
        \multirow{3}{*}{SC}
        & 3.3
        & 3.0
        & 1.4
        & 59.7
        & 8.1
        & 0.9
        & 1.6
        & 0.6
        & 2.7
        & 2.6 \\
        & 0.9
        & 2.4
        & 6.4
        & 14.7
        & 21.4
        & 5.8
        & 38.7
        & 9.3
        & 46.2
        & 22.2 \\
        & 1.7
        & 2.5
        & 9.7
        & 2.5
        & 5.7
        & 13.8
        & 4.9
        & 2.4
        & 4.2
        & 6.4 \\
        \cdashline{1-11}
        \multirow{3}{*}{GQ}
        & 15.0
        & 32.1
        & 17.2
        & 12.6
        & 27.1
        & 14.7
        & 33.2
        & 21.0
        & 25.7
        & 35.6 \\
        & 32.2
        & 15.9
        & 29.1
        & 37.9
        & 49.1
        & 49.7
        & 38.0
        & 26.7
        & 21.1
        & 19.3 \\
        & 29.0
        & 37.1
        & 23.7
        & 34.6
        & 47.2
        & 39.3
        & 9.8
        & 37.1
        & 20.8
        & 21.7 \\
        \cdashline{1-11}
        \multirow{3}{*}{SQCS$^{abs}$}
        & 2.6
        & 2.5
        & 1.1
        & 42.7
        & 6.5
        & 0.7
        & 1.3
        & 0.4
        & 2.2
        & 2.2 \\
        & 0.7
        & 1.9
        & 5.2
        & 12.4
        & 18.1
        & 4.9
        & 31.1
        & 7.8
        & 32.9
        & 17.6 \\
        & 1.5
        & 2.1
        & 7.9
        & 2.0
        & 4.8
        & 11.8
        & 3.8
        & 2.0
        & 3.4
        & 5.2 \\
        \midrule
        \rowcolor{SCGQ_table_fill}
        \multicolumn{11}{c}{\textbf{MIO} \cite{wang2024mio}} \\
        \multirow{3}{*}{SC}
        & 12.9
        & 39.8
        & 18.7
        & 11.3
        & 20.8
        & 22.8
        & 22.3
        & 10.0
        & 22.1
        & 16.8 \\
        & 9.6
        & 20.7
        & 22.5
        & 14.5 
        & 43.0
        & 44.3 
        & 1.9
        & 19.9 
        & 23.0 
        & 52.6 \\
        & 37.2
        & 16.6
        & 0.0
        & 10.5 
        & 34.6
        & 52.1 
        & 27.8
        & 29.5 
        & 17.3 
        & 22.6 \\
        \cdashline{1-11}
\multirow{3}{*}{GQ}
        & 14.5
        & 19.6
        & 47.1
        & 25.0
        & 33.0
        & 34.6
        & 12.1
        & 43.1
        & 41.7
        & 20.1 \\
        & 21.4
        & 34.9
        & 28.6
        & 18.1 
        & 43.5
        & 31.8 
        & 12.3
        & 41.0 
        & 47.0 
        & 49.0 \\
        & 34.7
        & 45.2
        & 0.2
        & 23.5 
        & 41.3
        & 51.8 
        & 63.0
        & 56.1 
        & 48.1 
        & 10.8 \\
        \cdashline{1-11}
\multirow{3}{*}{SQCS$^{abs}$}
        & 9.6
        & 30.2
        & 15.7
        & 8.7
        & 16.6
        & 18.3
        & 16.4
        & 8.3
        & 18.3
        & 12.7 \\
        & 7.3
        & 16.6
        & 17.7
        & 10.9 
        & 35.7
        & 35.2 
        & 1.4
        & 16.4 
        & 19.4 
        & 44.5 \\
        & 29.9
        & 13.8
        & 0.0
        & 8.1 
        & 28.5
        & 44.5 
        & 24.7
        & 25.6 
        & 14.6 
        & 16.5 \\
        \midrule
        \rowcolor{SCGQ_table_fill}
        \multicolumn{11}{c}{\textbf{\textsc{UniMA}}} \\
        \midrule
        \multirow{3}{*}{SC}
        & 67.9
        & 80.5
        & 60.4
        & 45.1
        & 67.1
        & 68.5
        & 35.7
        & 61.4
        & 62.3
        & 45.7 \\
        & 78.9
        & 75.0
        & 73.2
        & 81.9
        & 79.1
        & 83.7
        & 52.1
        & 78.9
        & 89.0
        & 92.2 \\
        & 67.0
        & 56.8
        & 79.7
        & 39.2
        & 77.6
        & 78.2
        & 41.7
        & 44.0
        & 79.1
        & 40.4 \\
        \cdashline{1-11}
        \multirow{3}{*}{GQ}
        & 84.9
        & 89.3
        & 85.6
        & 69.1
        & 83.6
        & 97.4
        & 45.3
        & 83.1
        & 93.4
        & 93.0 \\
        & 87.5
        & 82.0
        & 78.3
        & 96.4
        & 84.6
        & 82.3
        & 85.2
        & 80.7
        & 92.6
        & 67.9 \\
        & 83.7
        & 82.2
        & 95.2
        & 83.3
        & 82.1
        & 82.8
        & 72.2
        & 89.1
        & 95.0
        & 81.7 \\
        \cdashline{1-11}
        \multirow{3}{*}{SQCS$^{abs}$}
        & 65.3
        & 78.6
        & 57.6
        & 40.5
        & 64.4
        & 68.2
        & 30.7
        & 58.6
        & 61.6
        & 44.9 \\
        & 76.3
        & 70.9
        & 68.6
        & 81.5
        & 76.0
        & 79.7
        & 50.0
        & 75.0
        & 87.4
        & 83.4 \\
        & 64.4
        & 54.0
        & 78.6
        & 37.5 
        & 73.5
        & 74.2 
        & 40.1
        & 42.6
        & 78.0
        & 38.6 \\
        \bottomrule
    \end{tabular}
\end{table*}

\newpage
\begin{table*}[t!]
    \centering
    \caption{Detailed experiment results on \textit{Semantic Correctness \& Generation Quality}.}
    \label{tab:domain sqcs2}
    \fontsize{10}{13}\selectfont
    \setlength{\tabcolsep}{4mm}
    \begin{tabular}{l*{10}{c}}
        \toprule
        \rowcolor{SCGQ_head_fill}
        \multicolumn{11}{c}{\textbf{\textcolor{SCGQ_head_word}{Semantic Correctness \& Generation Quality}}} \\
        \rowcolor{SCGQ_head_fill}
        & \bf\#1 & \bf\#2 & \bf\#3 & \bf\#4 & \bf\#5 & \bf\#6 & \bf\#7 & \bf\#8 & \bf\#9 & \bf\#10 \\
        \rowcolor{SCGQ_head_fill}
        & \bf\#11 & \bf\#12 & \bf\#13 & \bf\#14 & \bf\#15 & \bf\#16 & \bf\#17 & \bf\#18 & \bf\#19 & \bf\#20 \\
        \rowcolor{SCGQ_head_fill}
        & \bf\#21 & \bf\#22 & \bf\#23 & \bf\#24 & \bf\#25 & \bf\#26 & \bf\#27 & \bf\#28 & \bf\#29 & \bf\#30 \\ 
        \midrule
        \rowcolor{SCGQ_table_fill}
        \multicolumn{11}{c}{\textbf{AnyGPT} \cite{zhan2024anygpt}} \\
        \midrule
        \multirow{3}{*}{$\tau$}
        & 99.9
        & 79.6
        & 98.8
        & 98.0
        & 95.0
        & 99.4
        & 47.4
        & 92.5
        & 95.3
        & 98.3 \\
        & 98.4
        & 98.3
        & 98.2
        & 91.0 
        & 95.7
        & 97.0 
        & 99.2
        & 70.2 
        & 99.3 
        & 100 \\
        & 96.5
        & 98.7
        & 100
        & 40.0 
        & 95.7
        & 99.7 
        & 99.7
        & 96.3 
        & 98.9 
        & 77.3 \\
        \cdashline{1-11}
        \multirow{3}{*}{SQCS$^{rel}$}
        & 11.1
        & 2.7
        & 14.0
        & 7.2
        & 10.4
        & 5.0
        & 5.2
        & 11.3
        & 20.4
        & 15.4 \\
        & 13.9
        & 14.4
        & 27.4
        & 4.2 
        & 17.5
        & 42.5
        & 3.9
        & 9.3 
        & 3.4
        & 1.4 \\
        & 20.9
        & 18.9
        & 1.9
        & 4.3
        & 18.4
        & 31.2
        & 4.9
        & 16.9 
        & 9.8 
        & 13.5 \\
        \midrule
        \rowcolor{SCGQ_table_fill}
        \multicolumn{11}{c}{\textbf{NExT-GPT} \cite{wu2024next}} \\
        \midrule
        \multirow{3}{*}{$\tau$}
        & 57.0
        & 60.3
        & 88.3
        & 36.8
        & 69.1
        & 97.3
        & 44.8
        & 50.2
        & 93.8
        & 48.5 \\
        & 98.2
        & 98.3
        & 97.6
        & 36.4
        & 94.3
        & 95.8
        & 99.4
        & 96.9
        & 98.1
        & 20.0 \\
        & 33.9
        & 98.3
        & 74.9
        & 40.0
        & 76.6
        & 82.0
        & 99.7
        & 74.1
        & 82.2
        & 71.8 \\
        \cdashline{1-11}
        \multirow{3}{*}{SQCS$^{rel}$}
        & 1.5
        & 1.5
        & 0.9
        & 15.7
        & 4.5
        & 0.7
        & 0.6
        & 0.2
        & 2.1
        & 1.1 \\
        & 0.7
        & 1.9
        & 5.0
        & 4.5
        & 17.0
        & 4.7
        & 30.9
        & 7.5
        & 32.2
        & 3.5 \\
        & 0.5
        & 2.0
        & 5.9
        & 0.8
        & 3.7
        & 9.7
        & 3.8
        & 1.5
        & 2.8
        & 3.7 \\
        \midrule
        \rowcolor{SCGQ_table_fill}
        \multicolumn{11}{c}{\textbf{MIO} \cite{wang2024mio}} \\
        \midrule
        \multirow{3}{*}{$\tau$}
        & 57.0
        & 60.3
        & 87.3
        & 36.8
        & 51.8
        & 97.3
        & 44.8
        & 50.2
        & 93.8
        & 12.2 \\
        & 98.2
        & 98.3
        & 97.6
        & 36.3
        & 94.3
        & 95.7
        & 98.8
        & 70.2
        & 98.1
        & 20.0 \\
        & 33.9
        & 98.3
        & 74.9
        & 40.0
        & 76.2
        & 82.0
        & 99.7
        & 52.9
        & 82.2
        & 77.3 \\
\cdashline{1-11}
\multirow{3}{*}{SQCS$^{rel}$}
        & 5.5
        & 18.2
        & 13.7
        & 3.2
        & 8.6
        & 17.8
        & 7.3
        & 4.2
        & 17.1
        & 1.6 \\
        & 7.2
        & 16.3
        & 17.2
        & 4.0
        & 33.7
        & 33.7
        & 1.4
        & 11.5
        & 19.0
        & 8.9 \\
        & 10.1
        & 13.6
        & 0.0
        & 3.2
        & 21.7
        & 36.5
        & 24.6
        & 13.5
        & 12.0
        & 12.8 \\
        \midrule
        \rowcolor{SCGQ_table_fill}
        \multicolumn{11}{c}{\textbf{\textsc{UniMA}}} \\
        \midrule
        \multirow{3}{*}{$\tau$}
        & 100
        & 100
        & 100
        & 100
        & 100
        & 100
        & 100
        & 100
        & 100
        & 100 \\
        & 100
        & 100
        & 100
        & 100
        & 100
        & 100
        & 100
        & 100
        & 100
        & 100 \\
        & 100
        & 100
        & 100
        & 100
        & 100
        & 100
        & 100
        & 100
        & 100
        & 100 \\
        \cdashline{1-11}
        \multirow{3}{*}{SQCS$^{rel}$}
        & 65.3
        & 78.6
        & 57.6
        & 40.5
        & 64.4
        & 68.2
        & 30.7
        & 58.6
        & 61.6
        & 44.9 \\
        & 76.3
        & 70.9
        & 68.6
        & 81.5
        & 76.0
        & 79.7
        & 50.0
        & 75.0
        & 87.4
        & 83.4 \\
        & 64.4
        & 54.0
        & 78.6
        & 37.5 
        & 73.5
        & 74.2 
        & 40.1
        & 42.6
        & 78.0
        & 38.6 \\
        \bottomrule
    \end{tabular}
\end{table*}

\newpage
\begin{table*}[t!]
    \centering
    \caption{Detailed experiment results on \textit{Response Structure Integrity}.}
    \label{tab:domain rsi1}
    \fontsize{10}{13}\selectfont
    \setlength{\tabcolsep}{4mm}
    \label{tab:domain sts les}
    \begin{tabular}{l*{10}{c}}
        \toprule
        \rowcolor{RSI_head_fill}
        \multicolumn{11}{c}{\textbf{\textcolor{SCGQ_head_word}{Response Structure Integrity}}} \\
        \rowcolor{RSI_head_fill}
        & \bf\#1 & \bf\#2 & \bf\#3 & \bf\#4 & \bf\#5 & \bf\#6 & \bf\#7 & \bf\#8 & \bf\#9 & \bf\#10 \\
        \rowcolor{RSI_head_fill}
        & \bf\#11 & \bf\#12 & \bf\#13 & \bf\#14 & \bf\#15 & \bf\#16 & \bf\#17 & \bf\#18 & \bf\#19 & \bf\#20 \\
        \rowcolor{RSI_head_fill}
        & \bf\#21 & \bf\#22 & \bf\#23 & \bf\#24 & \bf\#25 & \bf\#26 & \bf\#27 & \bf\#28 & \bf\#29 & \bf\#30 \\ 
        \midrule
        \rowcolor{RSI_table_fill}
        \multicolumn{11}{c}{\textbf{AnyGPT} \cite{zhan2024anygpt}} \\
        \midrule
        \multirow{3}{*}{StS$^{abs}$}
        & 3.8
        & 14.2
        & 25.1
        & 18.7
        & 9.3
        & 0
        & 0
        & 23.4
        & 22.9
        & 11.3 \\
        & 0.8
        & 1.3
        & 81.5
        & 0.8
        & 2.5
        & 49.7 
        & 6.8
        & 0.4 
        & 3.8 
        & 0.3 \\
        & 29.8
        & 11.2
        & 0.6
        & 37.0 
        & 10.9
        & 1.4
        & 7.2
        & 5.4 
        & 11.2 
        & 9.9 \\
        \cdashline{1-11}
        \multirow{3}{*}{LeS$^{abs}$}
        & 5.7
        & 28.4
        & 38.8
        & 26.8
        & 20.6
        & 0
        & 0
        & 44.8
        & 49.8
        & 12.5 \\
        & 1.2
        & 2.2
        & 86.4
        & 1.9
        & 4.2
        & 54.3
        & 8.7
        & 0.6
        & 6.1 
        & 0.5 \\
        & 41.5
        & 21.2
        & 0.8
        & 38.4
        & 15.5
        & 1.4
        & 11.4
        & 5.6
        & 18.8 
        & 9.9 \\
        \midrule
        \rowcolor{RSI_table_fill}
        \multicolumn{11}{c}{\textbf{NExT-GPT} \cite{wu2024next}} \\
        \midrule
        \multirow{3}{*}{StS$^{abs}$}
        & 2.2
        & 0.7
        & 0.9
        & 5.9
        & 8.0
        & 0.0
        & 1.1
        & 0.5
        & 0.9
        & 0.3 \\
        & 0.6
        & 0.3
        & 6.3
        & 0.7
        & 0.6
        & 1.4
        & 0.1
        & 1.1
        & 1.6
        & 0.7 \\
        & 1.2
        & 0.4
        & 2.0
        & 4.8
        & 0.1
        & 0.8
        & 1.0
        & 0.3
        & 2.1
        & 8.8 \\
        \cdashline{1-11}
        \multirow{3}{*}{LeS$^{abs}$}
        & 2.7
        & 0.7
        & 1.0
        & 6.0
        & 8.4
        & 0.0
        & 1.4
        & 0.5
        & 1.0
        & 0.3 \\
        & 0.9
        & 0.5
        & 6.3
        & 1.0
        & 1.0
        & 2.0
        & 0.1
        & 1.6
        & 2.7
        & 1.0 \\
        & 1.7
        & 0.5
        & 2.0
        & 4.8
        & 0.3
        & 0.8
        & 1.5
        & 0.3
        & 3.9
        & 8.8 \\
        \midrule
        \rowcolor{RSI_table_fill}
        \multicolumn{11}{c}{\textbf{MIO} \cite{wang2024mio}} \\
        \midrule
\multirow{3}{*}{StS$^{abs}$}
        & 0.1
        & 0.0
        & 3.9
        & 0.0
        & 3.5
        & 0.1
        & 2.4
        & 1.3
        & 2.0
        & 0.0 \\
        & 5.3
        & 1.0
        & 15.6
        & 0.0 
        & 0.6
        & 9.9 
        & 7.7
        & 0.6 
        & 0.3 
        & 0.0 \\
        & 0.4
        & 1.5
        & 0.0
        & 4.1 
        & 0.8
        & 0.1 
        & 0.1
        & 1.8 
        & 16.2 
        & 7.8 \\
\cdashline{1-11}
\multirow{3}{*}{LeS$^{abs}$}
        & 0.1
        & 0.0
        & 5.4
        & 0.0
        & 3.5
        & 0.1
        & 4.6
        & 2.3
        & 3.2
        & 0.0 \\
        & 7.7
        & 1.7
        & 16.2
        & 0.0 
        & 1.1
        & 11.8 
        & 11.9
        & 1.6 
        & 0.3 
        & 0.0 \\
        & 0.7
        & 1.7
        & 0.0
        & 4.1 
        & 1.6
        & 0.1 
        & 0.1
        & 1.8 
        & 20.4 
        & 7.8 \\
        \midrule
        \rowcolor{RSI_table_fill}
        \multicolumn{11}{c}{\textbf{\textsc{UniMA}}} \\
        \midrule
        \multirow{3}{*}{StS$^{abs}$}
        & 51.0
        & 50.4
        & 84.8
        & 37.0
        & 77.6
        & 1.7
        & 79.2
        & 58.6
        & 54.5
        & 11.6 \\
        & 38.9
        & 45.3
        & 87.4
        & 76.7
        & 49.3
        & 89.6
        & 31.3
        & 70.7
        & 83.7
        & 80.0 \\
        & 78.9
        & 75.5
        & 56.7
        & 84.6 
        & 73.8
        & 93.9
        & 51.3
        & 56.3
        & 57.4
        & 36.5 \\
        \cdashline{1-11}
        \multirow{3}{*}{LeS$^{abs}$}
        & 57.2
        & 60.1
        & 89.1
        & 59.9
        & 86.6
        & 2.4
        & 79.2
        & 85.9
        & 70.0
        & 22.2 \\
        & 53.5
        & 57.1
        & 90.6
        & 88.8
        & 82.8
        & 93.4 
        & 52.0
        & 95.1 
        & 97.0
        & 80.0 \\
        & 90.6
        & 84.0
        & 70.5
        & 88.0
        & 78.5
        & 96.7
        & 80.2
        & 73.2
        & 91.5
        & 40.0 \\
        \bottomrule
    \end{tabular}
\end{table*}

\newpage
\begin{table*}[t!]
    \centering
    \caption{Detailed experiment results on \textit{Response Structure Integrity}.}
    \label{tab:domain rsi2}
    \fontsize{10}{13}\selectfont
    \setlength{\tabcolsep}{4mm}
    \label{tab:results_response_structure_integrity}
    \begin{tabular}{l*{10}{c}}
        \toprule
        \rowcolor{RSI_head_fill}
        \multicolumn{11}{c}{\textbf{\textcolor{SCGQ_head_word}{Response Structure Integrity}}} \\
        \rowcolor{RSI_head_fill}
        & \bf\#1 & \bf\#2 & \bf\#3 & \bf\#4 & \bf\#5 & \bf\#6 & \bf\#7 & \bf\#8 & \bf\#9 & \bf\#10 \\
        \rowcolor{RSI_head_fill}
        & \bf\#11 & \bf\#12 & \bf\#13 & \bf\#14 & \bf\#15 & \bf\#16 & \bf\#17 & \bf\#18 & \bf\#19 & \bf\#20 \\
        \rowcolor{RSI_head_fill}
        & \bf\#21 & \bf\#22 & \bf\#23 & \bf\#24 & \bf\#25 & \bf\#26 & \bf\#27 & \bf\#28 & \bf\#29 & \bf\#30 \\ 
        \midrule
        \rowcolor{RSI_table_fill}
        \multicolumn{11}{c}{\textbf{AnyGPT} \cite{zhan2024anygpt}} \\
        \midrule
        \multirow{3}{*}{StS$^{rel}$}
        & 3.8
        & 11.3
        & 24.8
        & 18.4
        & 8.8
        & 0
        & 0
        & 21.7
        & 21.8
        & 11.1 \\
        & 0.7
        & 1.2
        & 80.0
        & 0.7 
        & 2.3
        & 48.2
        & 6.8
        & 0.3 
        & 3.8 
        & 0.3 \\
        & 28.8
        & 11.1
        & 0.6
        & 14.8 
        & 10.4
        & 1.4 
        & 7.2
        & 5.2 
        & 11.1 
        & 7.7 \\
        \cdashline{1-11}
        \multirow{3}{*}{LeS$^{rel}$}
        & 5.6
        & 22.6
        & 38.3
        & 26.3
        & 19.5
        & 0
        & 0
        & 41.3
        & 47.5
        & 12.3 \\
        & 1.2
        & 2.2
        & 84.9
        & 1.7
        & 4.0
        & 52.7 
        & 8.6
        & 0.4
        & 6.1 
        & 0.5 \\
        & 40.1
        & 20.9
        & 0.8
        & 15.4
        & 14.8
        & 1.4 
        & 11.3
        & 5.4
        & 18.6
        & 7.7 \\
        \midrule
        \rowcolor{RSI_table_fill}
        \multicolumn{11}{c}{\textbf{NExT-GPT} \cite{wu2024next}} \\
        \midrule
        \multirow{3}{*}{StS$^{rel}$}
        & 1.2
        & 0.4
        & 0.8
        & 2.2
        & 5.5
        & 0.0
        & 0.5
        & 0.3
        & 0.8
        & 0.1 \\
        & 0.6
        & 0.3
        & 6.1
        & 0.2
        & 0.6
        & 1.3
        & 0.1
        & 1.1
        & 1.6
        & 0.1 \\
        & 0.4
        & 0.3
        & 1.5
        & 1.9
        & 0.1
        & 0.7
        & 1.0
        & 0.2
        & 1.7
        & 6.3 \\
        \cdashline{1-11}
        \multirow{3}{*}{LeS$^{rel}$}
        & 1.5
        & 0.4
        & 0.9
        & 2.2
        & 5.8
        & 0.0
        & 0.6
        & 0.3
        & 0.9
        & 0.1 \\
        & 0.9
        & 0.5
        & 6.2
        & 0.4
        & 0.9
        & 1.9
        & 0.1
        & 1.5
        & 2.7
        & 0.2 \\
        & 0.6
        & 0.5
        & 1.5
        & 1.9
        & 0.3
        & 0.7
        & 1.5
        & 0.2
        & 3.2
        & 6.3 \\
        \midrule
        \rowcolor{RSI_table_fill}
        \multicolumn{11}{c}{\textbf{MIO} \cite{wang2024mio}} \\
        \midrule
\multirow{3}{*}{StS$^{rel}$}
        & 0.1
        & 0.0
        & 3.4
        & 0.0
        & 1.8
        & 0.1
        & 1.1
        & 0.6
        & 1.9
        & 0.0 \\
        & 5.2
        & 1.0
        & 15.2
        & 0.0 
        & 0.5
        & 9.5 
        & 7.6
        & 0.4 
        & 0.3 
        & 0.0 \\
        & 0.2
        & 1.5
        & 0.0
        & 1.7 
        & 0.6
        & 0.1 
        & 0.1
        & 1.0 
        & 13.3 
        & 6.1 \\
\cdashline{1-11}
\multirow{3}{*}{LeS$^{rel}$}
        & 0.1
        & 0.0
        & 4.7
        & 0.0
        & 1.8
        & 0.1
        & 2.1
        & 1.2
        & 3.0
        & 0.0 \\
        & 7.6
        & 1.7
        & 15.8
        & 0.0 
        & 1.0
        & 11.3 
        & 11.8
        & 1.2 
        & 0.3 
        & 0.0 \\
        & 0.2
        & 1.7
        & 0.0
        & 1.6 
        & 1.2
        & 0.1 
        & 0.1
        & 1.0 
        & 16.7 
        & 6.1 \\
        \midrule
        \rowcolor{RSI_table_fill}
        \multicolumn{11}{c}{\textbf{\textsc{UniMA}}} \\
        \midrule
         \multirow{3}{*}{StS$^{rel}$}
        & 51.0
        & 50.4
        & 84.8
        & 37.0
        & 77.6
        & 1.7
        & 79.2
        & 58.6
        & 54.5
        & 11.6 \\
        & 38.9
        & 45.3
        & 87.4
        & 76.7
        & 49.3
        & 89.6
        & 31.3
        & 70.7
        & 83.7
        & 80.0 \\
        & 78.9
        & 75.5
        & 56.7
        & 84.6 
        & 73.8
        & 93.9
        & 51.3
        & 56.3
        & 57.4
        & 36.5 \\
        \cdashline{1-11}
        \multirow{3}{*}{LeS$^{rel}$}
        & 57.2
        & 60.1
        & 89.1
        & 59.9
        & 86.6
        & 2.4
        & 79.2
        & 85.9
        & 70.0
        & 22.2 \\
        & 53.5
        & 57.1
        & 90.6
        & 88.8
        & 82.8
        & 93.4 
        & 52.0
        & 95.1 
        & 97.0
        & 80.0 \\
        & 90.6
        & 84.0
        & 70.5
        & 88.0
        & 78.5
        & 96.7
        & 80.2
        & 73.2
        & 91.5
        & 40.0 \\
        \bottomrule
    \end{tabular}
\end{table*}

\newpage
\begin{table*}[t!]
    \centering
    \caption{Detailed experiment results on \textit{Interleaved Coherence}.}
    \label{tab:domain ics1}
    \fontsize{10}{13}\selectfont
    \setlength{\tabcolsep}{4mm}
    \begin{tabular}{l*{10}{c}}
        \toprule
        \rowcolor{IC_head_fill}
        \multicolumn{11}{c}{\textbf{\textcolor{SCGQ_head_word}{Interleaved Coherence}}} \\
        \rowcolor{IC_head_fill}
        & \bf\#1 & \bf\#2 & \bf\#3 & \bf\#4 & \bf\#5 & \bf\#6 & \bf\#7 & \bf\#8 & \bf\#9 & \bf\#10 \\
        \rowcolor{IC_head_fill}
        & \bf\#11 & \bf\#12 & \bf\#13 & \bf\#14 & \bf\#15 & \bf\#16 & \bf\#17 & \bf\#18 & \bf\#19 & \bf\#20 \\
        \rowcolor{IC_head_fill}
        & \bf\#21 & \bf\#22 & \bf\#23 & \bf\#24 & \bf\#25 & \bf\#26 & \bf\#27 & \bf\#28 & \bf\#29 & \bf\#30 \\ 
        \midrule
        \rowcolor{IC_table_fill}
        \multicolumn{11}{c}{\textbf{AnyGPT} \cite{zhan2024anygpt}} \\
        \midrule
        \multirow{3}{*}{HC}
        & 47.2
        & 44.5
        & 37.5
        & 23.9
        & 34.5
        & 33.4
        & 38.5
        & 40.8
        & 54.6
        & 44.2 \\
        & 65.3
        & 36.1
        & 58.4
        & 8.7
        & 35.8
        & 60.3 
        & 12.0
        & 28.4 
        & 5.9 
        & 2.0 \\
        & 29.2
        & 50.0
        & 6.7
        & 57.4
        & 44.7
        & 62.8
        & 7.6
        & 50.7 
        & 22.3
        & 34.0 \\
        \cdashline{1-11}
        \multirow{3}{*}{SH}
        & 58.3
        & 53.9
        & 46.6
        & 25.1
        & 40.4
        & 38.9
        & 43.9
        & 39.5
        & 65.5
        & 51.3 \\
        & 76.7
        & 39.9
        & 60.5
        & 9.4 
        & 42.6
        & 68.7 
        & 12.2
        & 35.0
        & 6.0 
        & 2.1 \\
        & 33.0
        & 60.2
        & 8.2
        & 75.9
        & 53.5
        & 67.9
        & 8.1
        & 57.0
        & 24.1 
        & 30.8 \\
        \midrule
        \rowcolor{IC_table_fill}
        \multicolumn{11}{c}{\textbf{NExT-GPT} \cite{wu2024next}} \\
        \midrule
        \multirow{3}{*}{HC}
        & 13.9
        & 25.2
        & 18.9
        & 13.8
        & 22.4
        & 19.7
        & 33.6
        & 32.7
        & 37.8
        & 25.4 \\
        & 28.0
        & 16.7
        & 19.4
        & 28.8
        & 26.8
        & 11.0
        & 29.2
        & 25.8
        & 23.2
        & 0.0 \\
        & 28.5
        & 32.0
        & 24.8
        & 32.8
        & 35.1
        & 40.7
        & 6.5
        & 39.3
        & 19.9
        & 14.7 \\
        \cdashline{1-11}
        \multirow{3}{*}{SH}
        & 15.8
        & 26.4
        & 20.4
        & 14.1
        & 23.2
        & 20.5
        & 39.1
        & 34.9
        & 49.8
        & 27.0 \\
        & 31.2
        & 18.4
        & 21.6
        & 32.2
        & 30.3
        & 11.8
        & 32.1
        & 28.3
        & 26.4
        & 0.0 \\
        & 29.9
        & 35.0
        & 28.3
        & 36.1
        & 40.0
        & 47.0
        & 6.8
        & 44.4
        & 21.7
        & 16.7 \\
        \midrule
        \rowcolor{IC_table_fill}
        \multicolumn{11}{c}{\textbf{MIO} \cite{wang2024mio}} \\
        \midrule
\multirow{3}{*}{HC}
        & 33.7
        & 61.7
        & 53.0
        & 51.7
        & 51.3
        & 46.8
        & 23.9
        & 66.8
        & 61.5
        & 40.0 \\
        & 46.8
        & 48.6
        & 52.9
        & 43.6 
        & 70.0
        & 61.7 
        & 32.6
        & 33.9 
        & 53.4 
        & 57.6 \\
        & 64.7
        & 70.2
        & 2.8
        & 45.2 
        & 66.2
        & 75.2 
        & 93.4
        & 85.3 
        & 55.7 
        & 45.8 \\
\cdashline{1-11}
\multirow{3}{*}{SH}
        & 47.2
        & 79.3
        & 71.2
        & 68.9
        & 67.3
        & 53.1
        & 33.7
        & 79.6
        & 79.8
        & 60.8 \\
        & 57.1
        & 61.8
        & 58.4
        & 49.0 
        & 78.2
        & 72.1 
        & 39.8
        & 39.6 
        & 62.2 
        & 67.6 \\
        & 81.6
        & 81.9
        & 3.9
        & 54.4 
        & 83.8
        & 90.9 
        & 93.7
        & 89.4 
        & 61.6 
        & 64.7 \\
        \midrule
        \rowcolor{IC_table_fill}
        \multicolumn{11}{c}{\textsc\textbf{{UniMA}}} \\
        \midrule
        \multirow{3}{*}{HC}
        & 69.9
        & 76.9
        & 70.2
        & 49.1
        & 68.7
        & 85.1
        & 33.3
        & 60.9
        & 87.0
        & 88.8 \\
        & 88.0
        & 67.3
        & 48.0
        & 74.9
        & 68.7
        & 65.0
        & 77.1
        & 62.1
        & 88.7
        & 56.3 \\
        & 71.2
        & 60.8
        & 94.1
        & 69.1
        & 70.9
        & 55.5
        & 58.2
        & 67.9
        & 75.1
        & 76.7 \\
        \cdashline{1-11}
        \multirow{3}{*}{SH}
        & 69.6
        & 78.9
        & 77.1
        & 53.7
        & 71.4
        & 89.0
        & 45.3
        & 63.8
        & 89.1
        & 91.6 \\
        & 91.6
        & 70.1
        & 51.2
        & 75.9
        & 73.9
        & 69.3 
        & 81.5
        & 67.2
        & 90.2 
        & 57.8 \\
        & 72.4
        & 63.6
        & 94.7
        & 74.9
        & 79.3
        & 68.6
        & 64.2
        & 76.1
        & 77.6 
        & 79.0 \\
        \bottomrule
    \end{tabular}
\end{table*}

\newpage
\begin{table*}[t!]
    \centering
    \caption{Detailed experiment results on \textit{Interleaved Coherence}.}
    \label{tab:domain ics2}
    \fontsize{10}{13}\selectfont
    \setlength{\tabcolsep}{4mm}
    \label{tab:results_interleaved_coherence}
    \begin{tabular}{l*{10}{c}}
        \toprule
        \rowcolor{IC_head_fill}
        \multicolumn{11}{c}{\textbf{\textcolor{SCGQ_head_word}{Interleaved Coherence}}} \\
        \rowcolor{IC_head_fill}
        & \bf\#1 & \bf\#2 & \bf\#3 & \bf\#4 & \bf\#5 & \bf\#6 & \bf\#7 & \bf\#8 & \bf\#9 & \bf\#10 \\
        \rowcolor{IC_head_fill}
        & \bf\#11 & \bf\#12 & \bf\#13 & \bf\#14 & \bf\#15 & \bf\#16 & \bf\#17 & \bf\#18 & \bf\#19 & \bf\#20 \\
        \rowcolor{IC_head_fill}
        & \bf\#21 & \bf\#22 & \bf\#23 & \bf\#24 & \bf\#25 & \bf\#26 & \bf\#27 & \bf\#28 & \bf\#29 & \bf\#30 \\ 
        \midrule
        \rowcolor{IC_table_fill}
        \multicolumn{11}{c}{\textbf{AnyGPT} \cite{zhan2024anygpt}} \\
        \midrule
        \multirow{3}{*}{ICS$^{abs}$}
        & 49.4
        & 46.4
        & 39.4
        & 24.2
        & 35.7
        & 34.5
        & 39.6
        & 40.6
        & 56.8
        & 45.7 \\
        & 67.6
        & 36.8
        & 58.8
        & 8.9
        & 37.2
        & 62.0 
        & 12.1
        & 29.7 
        & 5.9 
        & 2.0 \\
        & 29.9
        & 52.0
        & 6.9
        & 61.1 
        & 46.4
        & 63.8 
        & 7.7
        & 52.0 
        & 22.7 
        & 33.3 \\
        \cdashline{1-11}
        \multirow{3}{*}{ICS$^{rel}$}
        & 49.3
        & 36.9
        & 38.9
        & 23.7
        & 33.9
        & 34.3
        & 18.7
        & 37.5
        & 54.1
        & 44.9 \\
        & 66.5
        & 36.2
        & 57.7
        & 8.1
        & 35.6
        & 60.1 
        & 12.0
        & 20.8
        & 5.9 
        & 2.0 \\
        & 28.9
        & 51.3
        & 6.9
        & 24.4
        & 44.4
        & 63.6
        & 7.7
        & 50.0 
        & 22.4 
        & 25.8 \\
        \midrule
        \rowcolor{IC_table_fill}
        \multicolumn{11}{c}{\textbf{NExT-GPT} \cite{wu2024next}} \\
        \midrule
        \multirow{3}{*}{ICS$^{abs}$}
        & 14.3
        & 25.5
        & 19.2
        & 13.9
        & 22.6
        & 19.9
        & 34.7
        & 33.1
        & 40.2
        & 25.7 \\
        & 28.7
        & 17.1
        & 19.8
        & 29.5
        & 27.4
        & 11.2
        & 29.8
        & 26.2
        & 23.8
        & 0.0 \\
        & 28.7
        & 32.6
        & 25.4
        & 33.4
        & 36.1
        & 42.0
        & 6.6
        & 40.3
        & 20.3
        & 15.1 \\
        \cdashline{1-11}
        \multirow{3}{*}{ICS$^{rel}$}
        & 8.1
        & 15.4
        & 16.9
        & 5.1
        & 15.6
        & 19.3
        & 15.6
        & 16.6
        & 37.7
        & 12.5 \\
        & 28.1
        & 16.8
        & 19.4
        & 10.7
        & 25.9
        & 10.7
        & 29.6
        & 25.4
        & 23.3
        & 0.0 \\
        & 9.7
        & 32.1
        & 19.0
        & 13.4
        & 27.6
        & 34.4
        & 6.6
        & 29.9
        & 16.7
        & 10.8 \\
        \midrule
        \rowcolor{IC_table_fill}
        \multicolumn{11}{c}{\textbf{MIO} \cite{wang2024mio}} \\
        \midrule
\multirow{3}{*}{ICS$^{abs}$}
        & 36.4
        & 65.2
        & 56.7
        & 55.2
        & 54.5
        & 48.1
        & 25.9
        & 69.4
        & 65.1
        & 44.2 \\
        & 48.9
        & 51.3
        & 54.0
        & 44.6 
        & 71.6
        & 63.8 
        & 34.1
        & 35.0 
        & 55.2 
        & 59.6 \\
        & 68.1
        & 72.5
        & 3.0
        & 47.0 
        & 69.8
        & 78.3 
        & 93.4
        & 86.1 
        & 56.9 
        & 49.6 \\
\cdashline{1-11}
\multirow{3}{*}{ICS$^{rel}$}
        & 20.7
        & 39.3
        & 49.5
        & 20.3
        & 28.3
        & 46.8
        & 11.6
        & 34.8
        & 61.1
        & 5.4 \\
        & 48.0
        & 50.4
        & 52.7
        & 16.2 
        & 67.5
        & 61.0 
        & 33.7
        & 24.6 
        & 54.1 
        & 11.9 \\
        & 23.1
        & 71.3
        & 2.3
        & 18.8 
        & 53.2
        & 64.2 
        & 93.1
        & 45.6 
        & 46.8 
        & 38.4 \\
        \midrule
        \rowcolor{IC_table_fill}
        \multicolumn{11}{c}{\textsc{UniMA}} \\
        \midrule
        \multirow{3}{*}{ICS$^{abs}$}
        & 69.8
        & 77.3
        & 71.6
        & 50.0
        & 69.2
        & 85.9
        & 35.7
        & 61.5
        & 87.4
        & 89.3 \\
        & 88.7
        & 67.9
        & 48.7
        & 75.1
        & 69.7
        & 65.8
        & 78.0
        & 63.2
        & 89.0
        & 56.6 \\
        & 71.4
        & 61.3
        & 94.2
        & 70.3
        & 72.6
        & 58.1
        & 59.4
        & 69.6
        & 75.6
        & 77.2\\
        \cdashline{1-11}
        \multirow{3}{*}{ICS$^{rel}$}
        & 69.8
        & 77.3
        & 71.6
        & 50.0
        & 69.2
        & 85.9
        & 35.7
        & 61.5
        & 87.4
        & 89.3 \\
        & 88.7
        & 67.9
        & 48.7
        & 75.1
        & 69.7
        & 65.8
        & 78.0
        & 63.2
        & 89.0
        & 56.6 \\
        & 71.4
        & 61.3
        & 94.2
        & 70.3
        & 72.6
        & 58.1
        & 59.4
        & 69.6
        & 75.6
        & 77.2\\
        \bottomrule
    \end{tabular}
\end{table*}

\newpage

\newpage

\newpage

\newpage

\newpage

\newpage

\newpage

\newpage

\newpage

\newpage

\newpage

\end{document}